\documentclass[10pt,twocolumn,letterpaper]{article}

\usepackage[pagenumbers]{cvpr} % To force page numbers, e.g. for an arXiv version

\definecolor{cvprblue}{rgb}{0.21,0.49,0.74}
\usepackage[pagebackref,breaklinks,colorlinks,allcolors=cvprblue]{hyperref}
\usepackage{times}
\usepackage{epsfig}
\usepackage{graphicx}
\usepackage{amssymb}
\usepackage{booktabs}
\usepackage{multirow}
\usepackage{float}
\usepackage{multicol}
\usepackage{tikz}
\usepackage{comment}
\usepackage{wrapfig}
\usepackage{graphicx}
\usepackage[nopar]{lipsum}
\usepackage{stfloats}
\usepackage{amsmath,amssymb} % define this before the line numbering.
\usepackage{amsfonts}
\usepackage{siunitx}

\usepackage{color}
\usepackage[hypcap=true]{subcaption}
\usepackage[misc]{ifsym}

\usepackage{array}     % For tabular formatting and \arraystretch
\usepackage{stfloats}
\usepackage{etoolbox}
\usepackage{tocloft}
\usepackage{setspace}

\usepackage{tabularx}  % Optional, for advanced table control

\usepackage{xcolor}
\usepackage{paralist}
\usepackage[capitalize]{cleveref}
\usepackage{afterpage}
\usepackage{stfloats}
\usepackage{lineno}

\crefname{section}{Sec.}{Secs.}
\Crefname{section}{Section}{Sections}
\Crefname{table}{Table}{Tables}
\crefname{table}{Tab.}{Tabs.}

\newcommand{\proj}[1]{\operatorname{proj}(#1)}
\newcommand{\homo}[1]{\operatorname{homo}(#1)}
\newcommand{\interp}[1]{\operatorname{interp}(#1)}
\newcommand{\fpp}[2]{\frac{\partial #1}{\partial #2}}
\usepackage{dsfont}

\def\dD{\mathcal{D}}

\def\lL{\mathcal{L}}

\def\pP{\mathcal{P}}

\def\rR{\mathcal{R}}

\newcommand\paren[1]{\left(#1\right)}

\DeclareMathSymbol{@}{\mathord}{letters}{"3B}

\def\latex/{\LaTeX}
\def\bibtex/{\hologo{BibTeX}}

\newcommand{\RN}[1]{%
  \textup{\uppercase\expandafter{\romannumeral#1}}%
}

\usepackage{stfloats}
\usepackage[hypcap=true]{caption}
\usepackage{amssymb}% http://ctan.org/pkg/amssymb
\usepackage{pifont}% http://ctan.org/pkg/pifont
\newcommand{\cmark}{\ding{51}}%
\newcommand{\xmark}{\ding{55}}%
\def\paperID{6818} % *** Enter the Paper ID here
\def\confName{ICCV}
\def\confYear{2025}

\title{Self-Calibrating Gaussian Splatting for Large Field-of-View Reconstruction}

\author{
    Youming Deng\textsuperscript{1}%\footnotemark[1] 
    \hspace{1cm} 
    Wenqi Xian\textsuperscript{2}%\footnotemark[1]
    \hspace{1cm} 
    Guandao Yang\textsuperscript{3}  \and 
    Leonidas Guibas\textsuperscript{3} \hspace{1cm}
    Gordon Wetzstein\textsuperscript{3} \hspace{1cm} 
    Steve Marschner\textsuperscript{1} \hspace{1cm} 
    Paul Debevec\textsuperscript{2} \and
    \textsuperscript{1}Cornell University \hspace{1cm} 
    \textsuperscript{2}Netflix Eyeline Studios \hspace{1cm} 
    \textsuperscript{3}Stanford University
}

\begin{document}

\twocolumn[{%
\renewcommand\twocolumn[1][]{#1}%
\maketitle
\begin{center}
    % \vspace{-5mm}
    \setlength{\tabcolsep}{0.1em}
    \renewcommand{\arraystretch}{0.75}
    % \begin{tabular}{ccccc}
    %     \includegraphics[height=0.22\textwidth]{images/teaser/gt.jpg} &  
    %     \includegraphics[height=0.22\textwidth]{images/teaser/vanilla.jpg} &  
    %     \includegraphics[height=0.22\textwidth]{images/teaser/fisheye-gs.jpg} &  
    %     \includegraphics[height=0.22\textwidth]{images/teaser/ours.jpg} &  
    %     \includegraphics[height=0.22\textwidth]{images/teaser/gt_.jpg} \\
    %     (a) Ground Truth Full-image & 
    %     (b) Vanilla GS &
    %     (c) Fisheƒye-GS &
    %     (d) Ours &
    %     (e) Ground Truth
    % \end{tabular}
    \begin{tabular}{ccc @{\hskip 10pt} c}
        \includegraphics[height=0.35\textwidth]{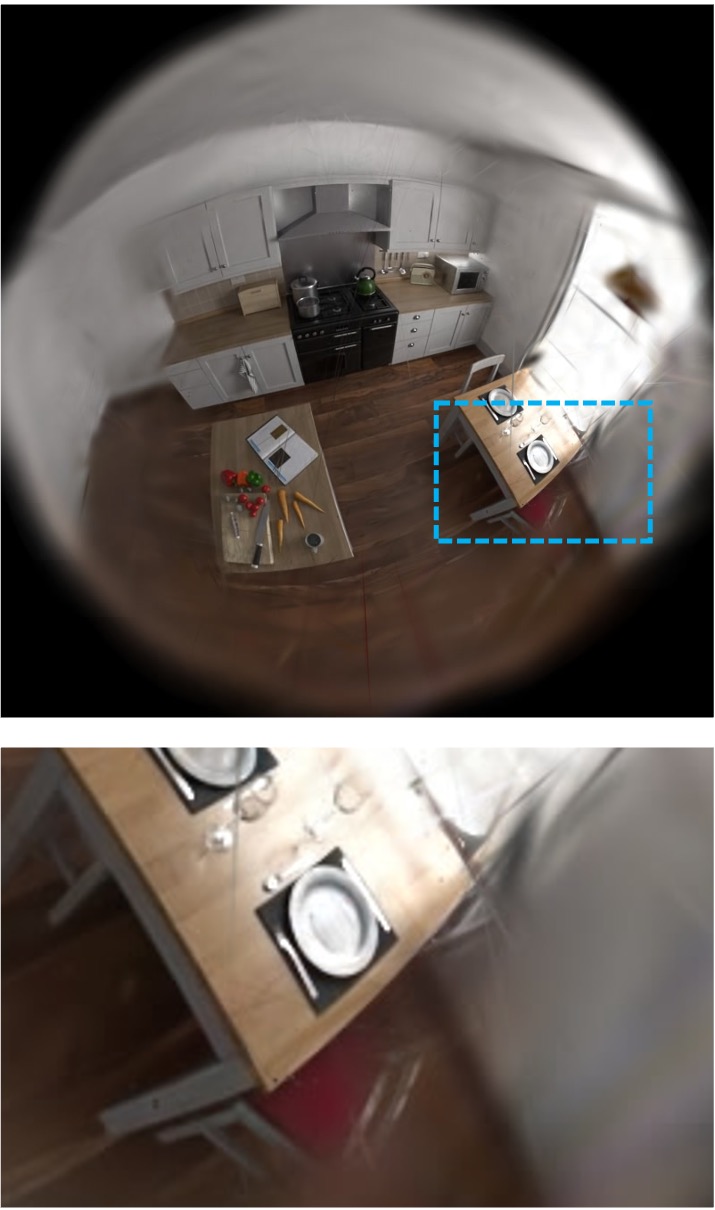} &  
        \includegraphics[height=0.35\textwidth]{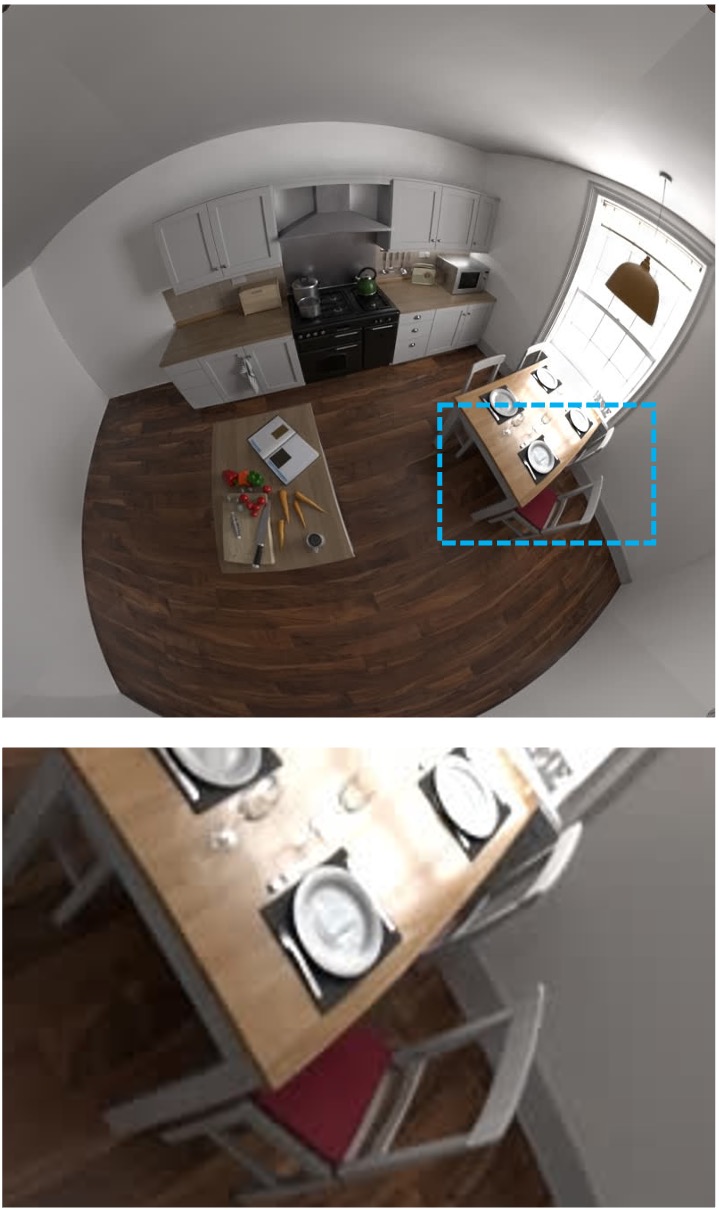} &  
        \includegraphics[height=0.35\textwidth]{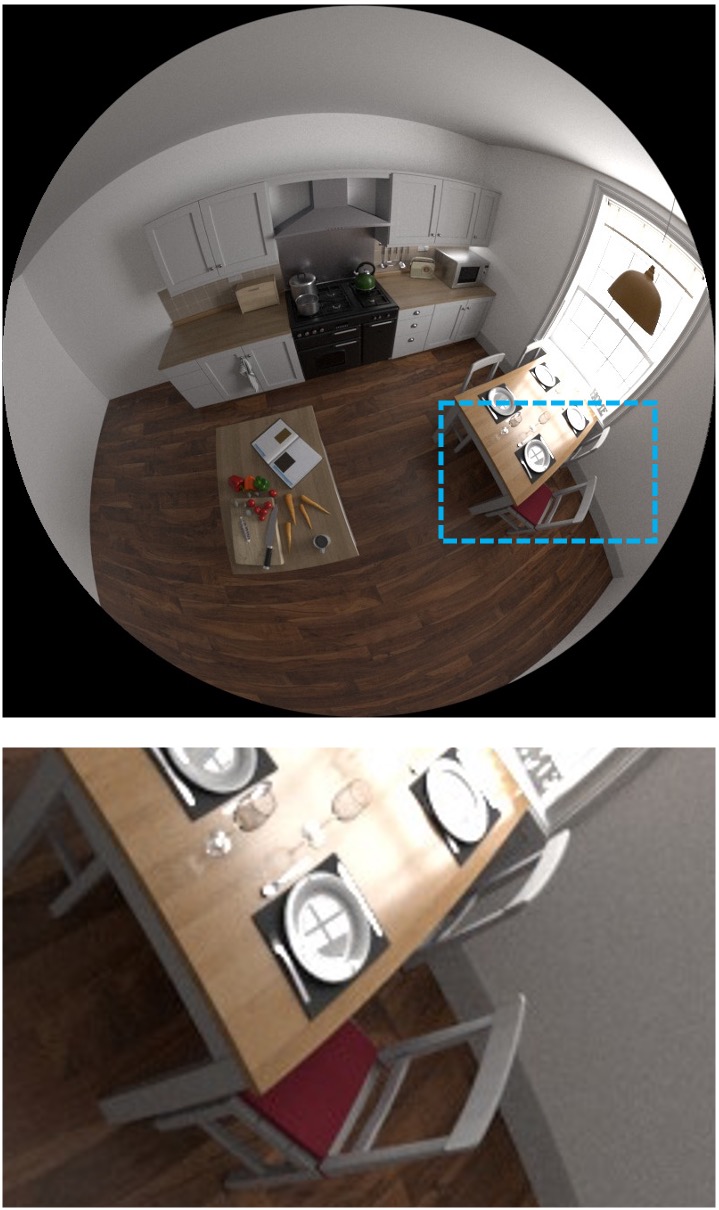} & 
        \includegraphics[height=0.35\textwidth]{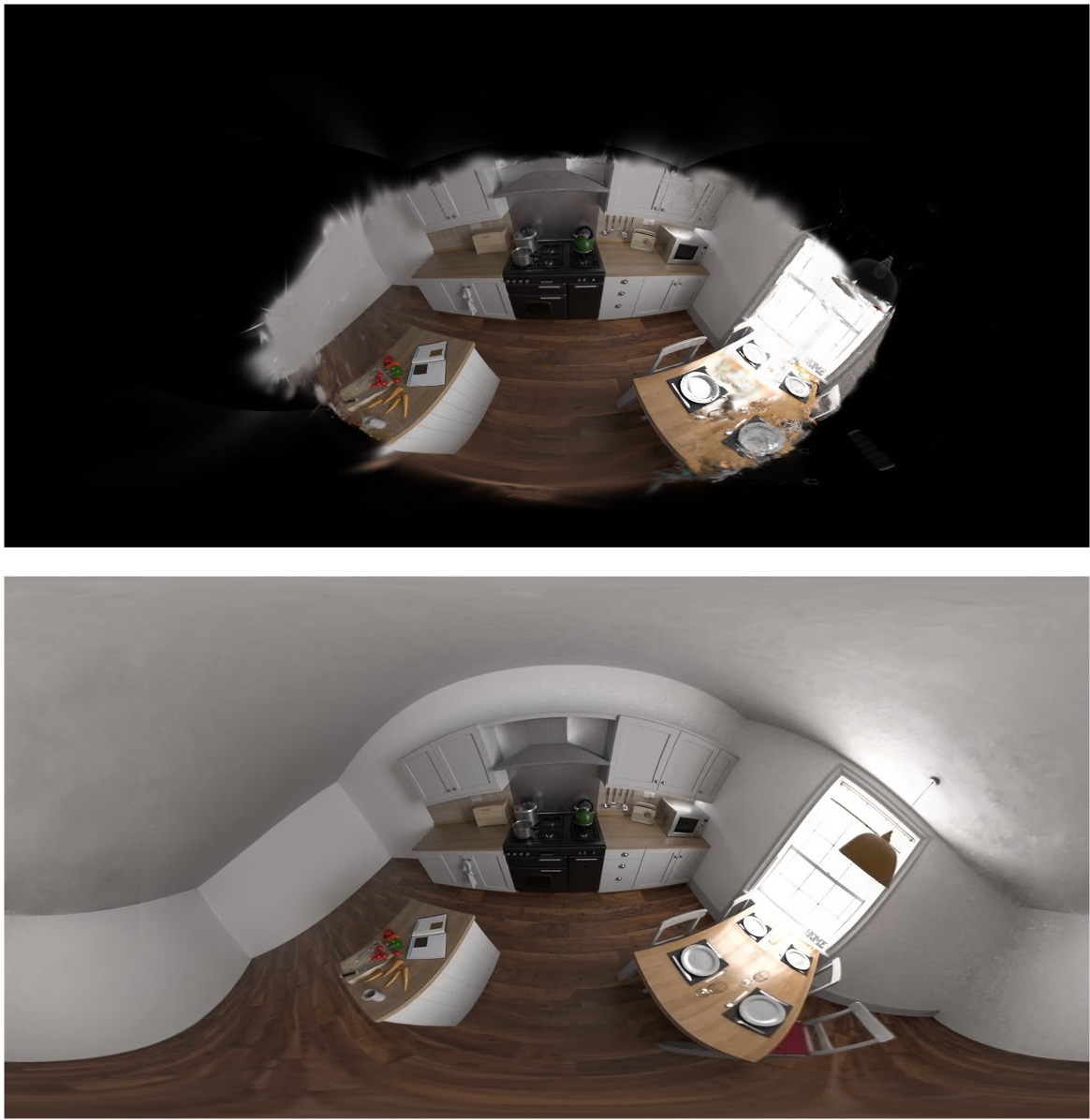} \\
        
        (a) Fisheye-GS~\cite{liao2024fisheye} &
        (b) Ours &
        (c) Ground Truth &
        (d) Rendered Panorama
    \end{tabular}
    % \vspace{-1em}
    \captionof{figure}{
        We introduce Self-Calibrating Gaussian Splatting, a differentiable rasterization pipeline with a hybrid lens distortion field that can produce high-quality novel view synthesis results from uncalibrated wide-angle photographs. (a) Existing methods such as Fisheye-GS~\cite{liao2024fisheye} fail to accurately handle complex lens distortions due to the traditional parametric distortion model. (b) Our method accurately models large distortions, especially in the peripheral regions, utilizing the entire highly distorted raw images for reconstruction. (d) Our method (bottom) provides extensive coverage, whereas conventional pipelines (top) can only recover the center.
    }
    \label{fig:teaser}
\end{center}%
}]
\renewcommand{\thefootnote}{\fnsymbol{footnote}}

% \footnotetext[1]{Equal Contribution.}

\begin{abstract}
Large field-of-view (FOV) cameras can simplify and accelerate scene capture because they provide complete coverage with fewer views. However, existing reconstruction pipelines fail to take full advantage of large-FOV input data because they convert input views to perspective images, resulting in stretching that prevents the use of the full image. Additionally, they calibrate lenses using models that do not accurately fit real fisheye lenses in the periphery.
We present a new reconstruction pipeline based on Gaussian Splatting that uses a flexible lens model and supports fields of view approaching 180 degrees. We represent lens distortion with a hybrid neural field based on an Invertible ResNet and use a cubemap to render wide-FOV images while retaining the efficiency of the Gaussian Splatting pipeline. Our system jointly optimizes lens distortion, camera intrinsics, camera poses, and scene representations using a loss measured directly against the original input pixels.
We present extensive experiments on both synthetic and real-world scenes, demonstrating that our model accurately fits real-world fisheye lenses and that our end-to-end self-calibration approach provides higher-quality reconstructions than existing methods.
More details and videos can be found at the project page: \url{https://denghilbert.github.io/self-cali/}.

\end{abstract}

\section{Introduction}
\label{sec:intro}
Large field-of-view (FOV) lenses, such as fisheye lenses, are widely used in robotics~\cite{de2018robust}, virtual reality~\cite{klanvcar2004wide}, and autonomous driving~\cite{cui2019real} because they capture scenes with fewer images, enabling efficient data acquisition for reconstruction and novel view synthesis (NVS)\cite{ma20153d}. However, most modern 3D reconstruction systems based on Neural Radiance Fields (NeRFs)\cite{mildenhall2021nerf} and Gaussian Splatting (3DGS)~\cite{kerbl20233d} cannot be directly applied to this type of input because they rely on perspective projection. As a result, fisheye images are typically re-projected into linear perspective images, and the resulting non-uniform sampling and stretching pose challenges for accurate camera calibration and scene reconstruction from wide-angle imagery.

Traditionally, imaging systems are pre-calibrated using specialized setups, such as calibration checkerboards. During calibration, polynomial distortion models~\cite{opencv_library, schoenberger2016sfm} are estimated and then used to re-project raw fisheye images into perspective ones for scene reconstruction. There are three important problems with this practice. First, resampling wide-field images leads to severe stretching and non-uniform sampling, as shown in~\cref{fig:hilbert_180}, where perspective images exhibit far less uniform sampling of directions than fisheyes. Second, traditional parametric distortion models lack sufficient expressiveness to accurately model large FOV (\textit{e.g.,} 180\si{\degree}) lenses. Third, pre-calibration prevents lens parameters from being optimized end-to-end along with reconstruction, meaning the final reconstruction does not fully minimize the reconstruction loss.

To avoid the separate pre-calibration stage, many recent works~\cite{moenne20243d, jeong2021self, liao2024fisheye} integrate distortion modeling~\cite{zeller1996camera, pollefeys1999stratified, chandraker2007autocalibration, chandraker2010globally} and optimization into NeRF and 3DGS frameworks. However, even when using photometric loss to refine distortion parameters alongside reconstruction, these approaches still exhibit significant misalignment in peripheral regions due to their reliance on traditional distortion models and single-plane projection. Both issues constrain the ability of current reconstruction frameworks to represent wide-FOV optical systems.

In this work, we introduce \textit{Self-Calibrating Gaussian Splatting}, a differentiable rasterization pipeline that jointly optimizes lens distortion, camera intrinsics, camera poses, and scene representations using 3D Gaussians. Our approach effectively addresses the three aforementioned challenges, achieving high-quality reconstruction from large-FOV images without requiring resampling, pre-calibration, or polynomial distortion models.

To improve lens modeling, we replace the conventional parametric distortion model with a novel hybrid neural field that balances expressiveness and computational efficiency, as illustrated in~\cref{fig:pipeline}. Our method uses invertible residual networks~\cite{behrmann2019invertible} to predict displacements on a normalized sparse grid, followed by bilinear interpolation to generate a continuous distortion field.

To mitigate the stretching caused by resampling to a single-plane perspective, we use cubemap sampling as in \cite{Cohen:1985:hemicube} (\cref{fig:hilbert_cubemap}), which significantly reduces stretching and yields a more uniform pixel density, even in peripheral areas. Gaussians project into each limited-FOV face of the cubemap with bounded distortion.

Finally, our method combines lens calibration with bundle adjustment and scene reconstruction into an integrated end-to-end self-calibration process that minimizes the rendering loss over all unknowns jointly, leading to lower reconstruction errors compared to pre-calibration of distortion and/or camera pose.

To validate our method, we conduct extensive experiments on both synthetic datasets and real-world scenes, including the FisheyeNeRF datasets~\cite{jeong2021self}, our own real-world captured scenes, and a synthetic dataset. Our system effectively calibrates extrinsics, intrinsics, and lens distortion, achieving better reconstruction performance compared to existing methods using uncalibrated fisheye cameras. Importantly, our system is not constrained to a single fisheye camera model; rather, it is designed to be flexible and adaptable, accommodating a wide range of cameras, from perspective to extreme fisheye, without requiring pre-calibration. This flexibility enables our method to fully leverage the unique capabilities of available lenses, ensuring comprehensive scene coverage and high-quality reconstructions.

{
\begin{figure}[t]
    \centering
    \setlength{\tabcolsep}{1pt} % Adjust space between columns if needed
    \begin{tabular}{cc} % 4 columns
        \subcaptionbox{Conventional Paradigm\label{fig:hilbert_180}}{
            \includegraphics[width=0.28\textwidth]{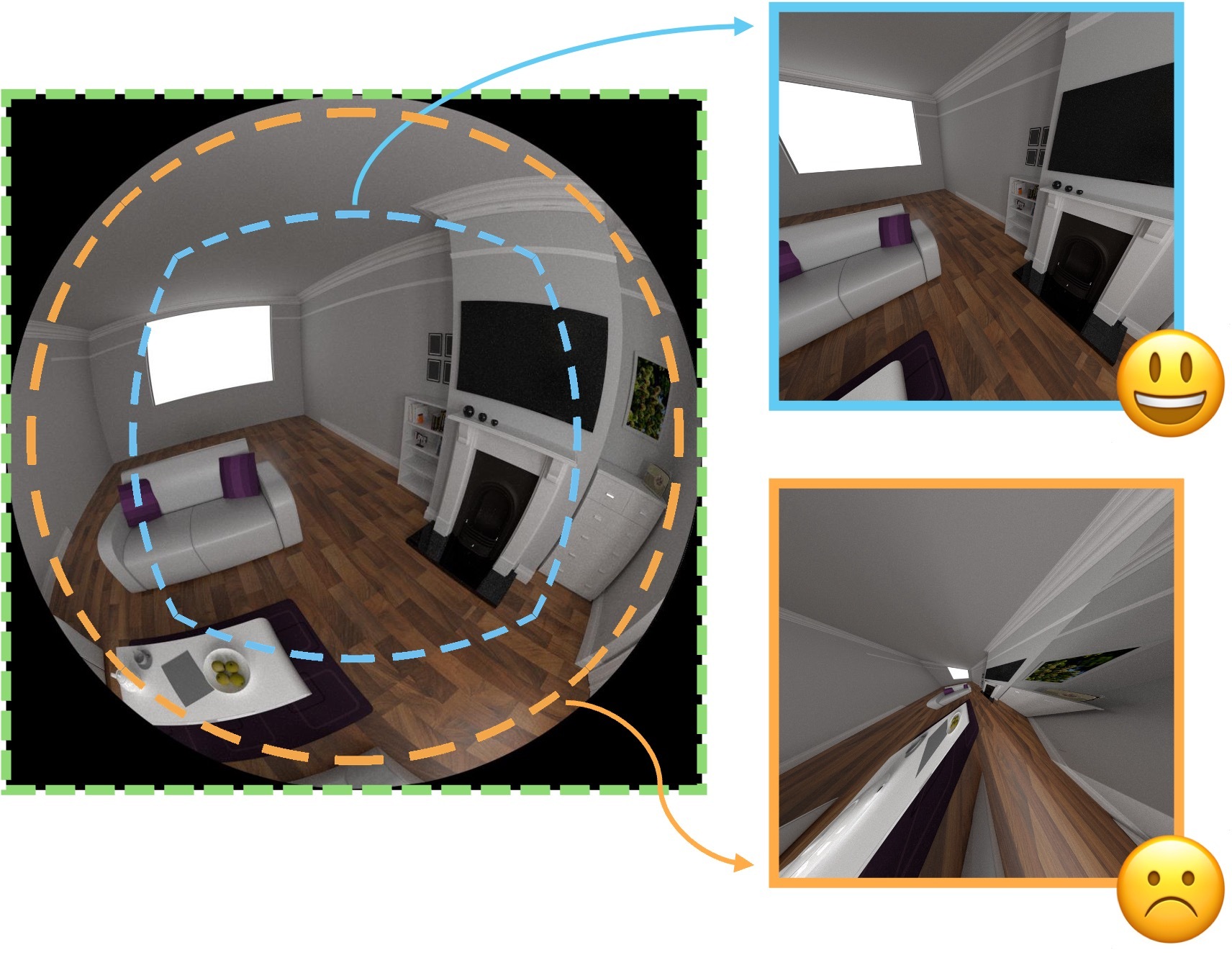}
            \label{fig:conventional}
        } &

        % Fourth image with label (c)
        \subcaptionbox{Ours\label{fig:hilbert_cubemap}}{
            \includegraphics[width=0.16\textwidth]{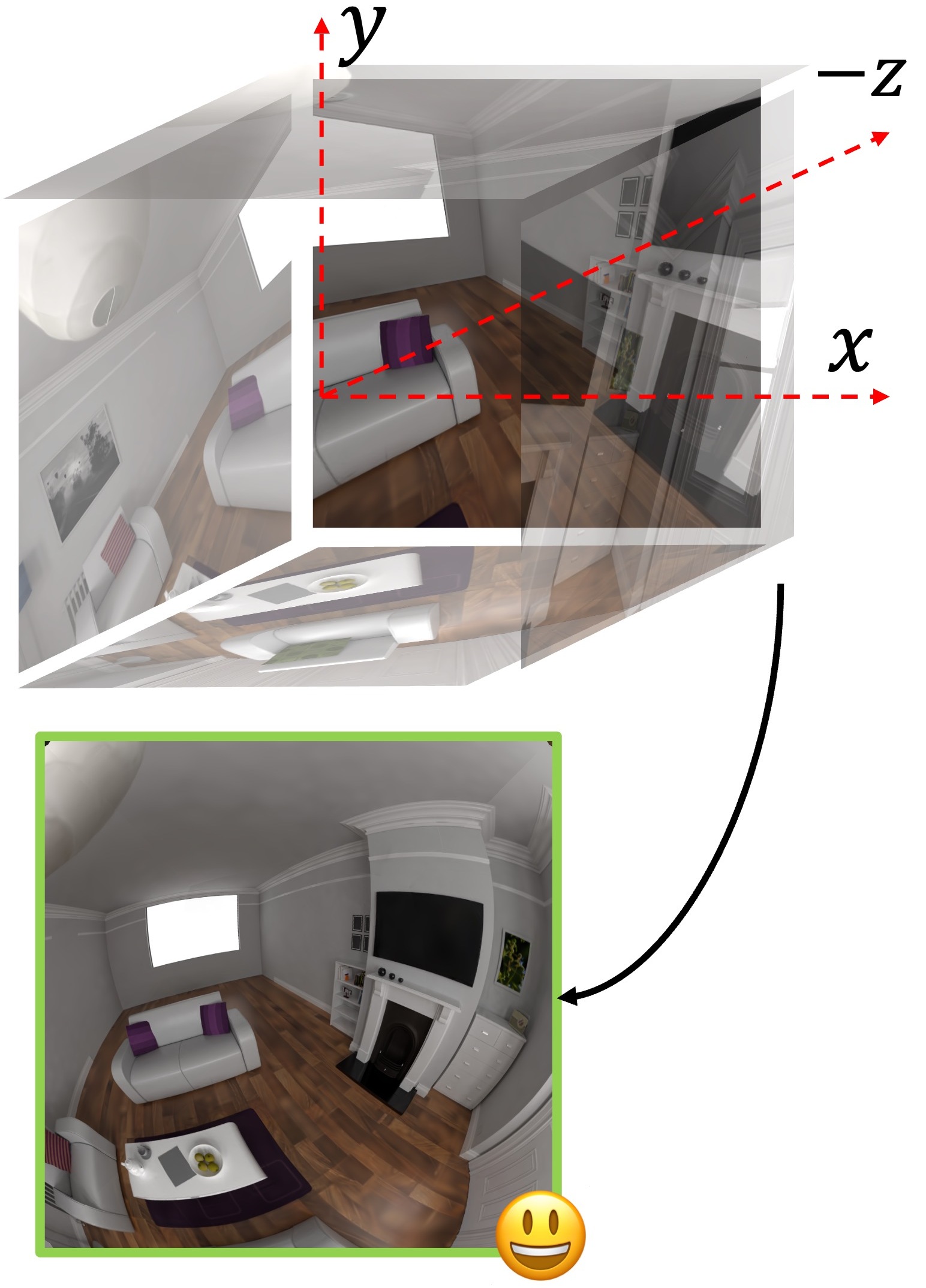}
        }
    \end{tabular}
    \vspace{-0.6em}
    \caption{\textbf{Conventional Paradigm vs.\ Our Method}. (a) Conventional approaches require reprojecting the image into perspective views compatible with 3DGS rasterization. As the field of view increases, pixel stretching becomes progressively severe, significantly compromising the quality of the reconstruction. (b) In contrast, our cubemap resampling strategy maintains a consistent pixel density across the entire field of view. This approach, combined with our hybrid distortion field, utilizes the peripheral regions (the annular area outside the blue box) without severe distortion or pixel stretching. Moreover, our method can handle fields of view up to 180°, as demonstrated by the green box, allowing for comprehensive and accurate reconstructions.}
    \label{fig:limitation_single_perspective}
    \vspace{-0.5em}
\end{figure}
}

\section{Related Work}
\label{sec:related_work}
% \vspace{-0.5em}
\paragraph{Camera Modeling and Lens Distortion.}
Lens distortion is an inherent property of all cameras. In general, nonlinear distortion can be formulated as:
\begin{equation}
\textbf{x}_d = \textbf{K} \cdot D\left(\pi\left(\textbf{R}\cdot\textbf{X} + \textbf{t}\right)\right),
\end{equation}
where $\textbf{K}$ and $[\textbf{R}|\textbf{t}]$ represent the intrinsic and extrinsic parameters, respectively. $\textbf{X}$ is a 3D point in world coordinates. $\pi(\cdot)$ denotes the pinhole projection, including dehomogenization to obtain 2D points $\textbf{x}_n$ on the image plane. The distortion model $D(r(\textbf{x}_n))$ is parameterized as a polynomial function of the radial distance:
\begin{equation} \label{eq:radial}
D(r(\textbf{x}_n)) = 1 + k_1r^2 + k_2r^4 + k_3r^6 + \ldots
\end{equation}
where $k_1$, $k_2$, $k_3$, $\ldots$ are the parameters of the Brown–Conrady model~\cite{conrady1919decentred,brown1996decentering}, derived from calibration, and $r = \sqrt{x_n^2 + y_n^2}$.
Scaramuzza \textit{et al.}\cite{scaramuzza2006toolbox} first proposed a unified model for large-FOV fisheye lenses, which has been adopted in several works\cite{bujnak2010new,kukelova2015efficient}. The most widely used fisheye model~\cite{itseez2015opencv} describes distortion as a function of the angular distance from the projection center:
\begin{equation} \label{eq:fisheye}
D(r(\textbf{x}_n)) = \frac{\theta}{r} \left(1 + k_1\theta^2 + k_2\theta^4 + k_3\theta^6 + \ldots\right),
\end{equation}
where $\theta = \arctan\left(\frac{r}{1}\right)$, and the distances of projected points to the image plane are normalized to 1.
The 3D Gaussian Splatting method~\cite{kerbl20233d} assumes a standard perfect pinhole camera model and typically relies on COLMAP~\cite{schoenberger2016sfm} to undistort images before reconstruction. To remove this constraint, recent methods~\cite{liao2024fisheye,moenne20243d} adopt parametric models like~\cref{eq:fisheye} to extend 3D Gaussian Splatting techniques to fisheye images. However, these methods still depend heavily on camera calibration for accurate estimation and fix the projection in rasterization, limiting their generalizability to various camera types.
Some works~\cite{moenne20243d} introduce ray tracing into the rasterization pipeline and approximate Gaussian bounding using an icosahedron, which can potentially compromise rasterization efficiency. Other approaches~\cite{li2024omnigs,bai2024360,huang2025sc} explore reconstruction from omnidirectional 360° panoramas, but the key difference is that panoramas require calibration to stitch two fisheye images together, which does not preserve raw geometric consistency at the stitching boundary.
We address these limitations by introducing a hybrid distortion field that is compatible with the 3D Gaussian Splatting pipeline. Our experiments demonstrate that existing methods, such as Fisheye-GS~\cite{liao2024fisheye} and ADOP~\cite{ruckert2022adop}, which incorporate traditional camera distortion models from~\cref{eq:radial,eq:fisheye} into the rasterization process, are not expressive enough to handle the severe distortions present in large-FOV cameras.

{
\begin{figure}[t]
    \centering
    {\includegraphics[width=0.46\textwidth]{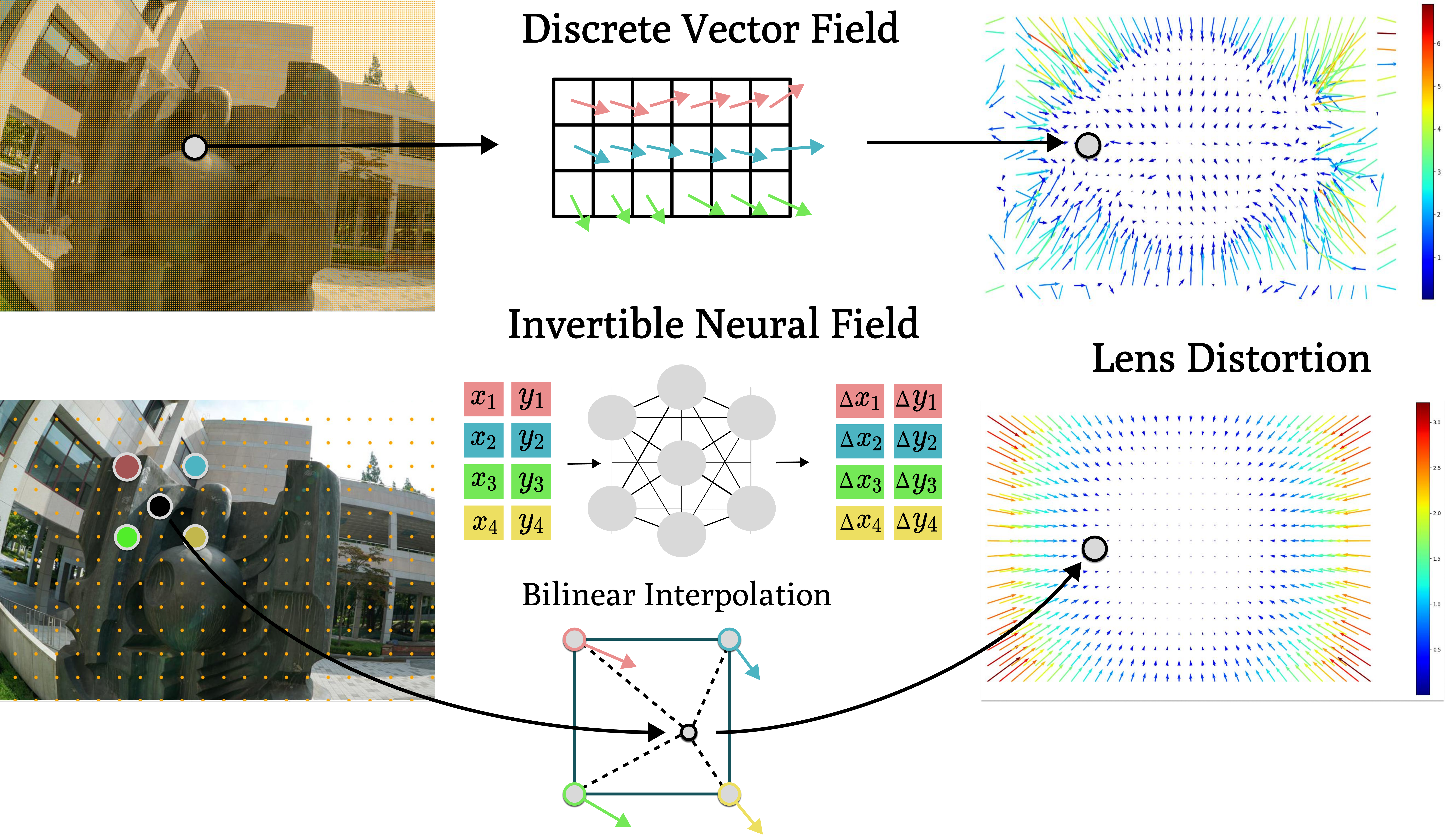}}
    % \vspace{-0.75em}
    \caption{
    \textbf{Overview of Our Method}. In contrast to the explicit distortion vector field illustrated in the upper row, our hybrid approach maintains computational efficiency by leveraging explicit control points. Additionally, the regularization provided by the invertible neural field effectively balances the trade-off between the expressiveness and smoothness of the distortion field.
    }
    \label{fig:pipeline}
    \vspace{-1.em}
\end{figure}
}
% \vspace{-1.2em}
\paragraph{Self-Calibrating Reconstruction.}
The bundle adjustment process can be extended to optimize camera lens parameters alongside poses, a process known as self-calibration~\cite{zhang1999flexible,pollefeys1999self,devernay2001straight,hartley2003multiple}. Camera calibration without a known calibration target is particularly challenging, as it relies on strong assumptions about scene structure and geometric priors to establish reliable correspondences~\cite{barreto2005geometric,carroll2009optimizing,aleman2014automatic}. Camera auto-calibration methods~\cite{zhang1999flexible,pollefeys1999self,devernay2001straight} extend this idea by deriving camera intrinsics from multi-view observations of unstructured scenes, an approach further advanced in recent studies~\cite{fang2022self,engel2016photometrically,ha2016high,deng2024physics}.
Several non-parametric models have been developed to ensure broad applicability across different camera and lens combinations~\cite{grossberg2001general,camposeco2015non,hartley2007parameter,li2006plane}, while additional regularization is often required to maintain smooth underlying distortion~\cite{pan2022camera}. With advances in differentiable rendering and rasterization pipelines~\cite{kopanas2021point,yifan2019differentiable}, recent works~\cite{ruckert2022adop,xian2023neural,jeong2021self} have demonstrated that camera lens distortion can be optimized jointly with other parameters through a differentiable projection module.
Prior works have also adapted NeRF for panoramic and fisheye-distorted inputs~\cite{gu2022omni,huang2022360roam,xu2023vr,kulkarni2023360fusionnerf}. These solutions typically rely on parametric models tailored for specific lenses, limiting their generalizability to a broader range of lens types. SCNeRF~\cite{jeong2021self} models a residual projection matrix and residual raxel parameters~\cite{grossberg2005raxel}, which are interpolated on a sub-sampled pixel grid. NeuroLens~\cite{xian2023neural} optimizes lens parameters through an invertible neural network, while SC-OmniGS~\cite{huang2025sc} optimizes camera parameters jointly with reconstruction but relies on calibrated captures.
Building on insights from prior self-calibration methods, this work introduces a novel and efficient approach to modeling lens distortion, fully integrated with 3DGS~\cite{kerbl20233d}.

% \vspace{-0.8em}
\section{Method}
\label{sec:method}
Given uncalibrated wide-angle captures, we aim to develop an algorithm that produces high-quality reconstructions using 3D Gaussians. Our method is designed to be robust against severe distortion in the peripheral regions of images and various wide-angle lens effects.
First, we extend Gaussian Splatting to support a broader range of camera models, including fisheye lenses, as discussed in~\cref{sec:iresnet}. To model lens distortion, we introduce a hybrid distortion field, enabling our approach to generalize across diverse real-world scenarios involving cameras with varying distortions.
Second, we replace the traditional single-plane projection in 3DGS with a cubemap representation and introduce a distance-sorting strategy accordingly, as detailed in~\cref{sec:cubemap}.
Finally, we derive an efficient optimization for camera parameters, supporting self-calibration with distortion, as described in~\cref{sec:opt_cam}.

\subsection{Background of Gaussian Splatting} 
A 3D Gaussian $G(x) = e^{-(x-\mu_i)^T\Sigma_i^{-1} (x-\mu_i)}$ is parameterized by center $\mu_i\in\mathbb{R}^3$, covariance $\Sigma_i$, opacity $\sigma_i$, and color $C_i$, represented using spherical harmonics. The 3D Gaussians are projected~\cite{zwicker2001ewa} into $\mu_i^{2D} = \pP(\mu_i, \Theta)$ and $\Sigma^{2D}=J_\pP^T\Sigma J_\pP$, where $J_\pP\in\mathbb{R}^{3\times 2}$ is the affine approximation of the projection $\pP$ at point $\mu_i$, parameterized by $\Theta$. 
For a pixel location $u$, the RGB color is produced with alpha blending~\cite{kopanas2021point,yifan2019differentiable}:
\begin{align}
\label{eq:alpha_blending}
    \hat{I}(u) = \sum_{i=1}^{|G|}C_i\alpha_i\prod_{j\in \mathcal{N}_{<i}(G)}(1-\alpha_j),
\end{align}
where $\mathcal{N}(G)$ is an ordered index of Gaussians sorted by depth, and $\alpha_i$ depends on $\sigma_i$, $\Sigma_i^{2D}$, and $\mu_i^{2D}$. Finally, this set of 3D Gaussians is optimized using the L1 loss and D-SSIM~\cite{baker2022dssim} loss, along with adaptive control~\cite{kerbl20233d}.

\subsection{Lens Distortion Modeling}
\label{sec:iresnet}
In this section, we extend the Gaussian Splatting technique to accommodate a broader class of camera lenses, including fisheye and wide-angle cameras, by modeling lens distortion.
Lens distortion is typically captured by a distortion function defined in camera coordinates.
A distortion function $\dD_\theta:\mathbb{R}^2 \to \mathbb{R}^2$ parameterized by $\theta$ maps pixel locations from a rectified image to locations in a distorted image.
Ideally, the mapping $\dD_\theta$ should be:
1) expressive enough to model various lens distortions,
2) well-regularized so that it can be optimized together with the 3D scene, and
3) efficient, ensuring that it does not add significant computational overhead.
While existing methods have explored using parametric camera models, grid-based methods, and deep-learning methods, none of these approaches perfectly satisfy all three criteria.

% \vspace{-1.em}
\paragraph{Grid-based method.}
The simplest way to implement a generic camera model is to explicitly optimize for the distortion in a grid of pixel coordinates and apply bilinear interpolation to extract a continuous distortion field:
\begin{align}
    \dD_\theta(\textbf{x}) = \textbf{x} + \interp{\textbf{x}, \theta}, 
\end{align}
where the optimizable parameters $\theta\in \mathbb{R}^{H\times W\times 2}$ define an $H\times W$ grid storing 2D vectors representing the distortion. The bilinear interpolation function is given by $\interp{\textbf{x}, \theta} = W(\textbf{x}, \theta)\cdot\theta$, where $W(\textbf{x}, \theta)\in \mathbb{R}^{H\times W}$ represents the bilinear interpolation weights at location $\textbf{x}$.
Such a grid-based method is both expressive and efficient, as $W(\textbf{x}, \theta)$ is sparse, and increasing the grid resolution allows for modeling more complex functions.
However, the grid-based method lacks the smoothness required to model lens distortion properly, leading to overfitting and suboptimal solutions (\cref{fig:pipeline} Top).

% \vspace{-1.em}
\paragraph{Invertible Residual Networks.} An alternative way to model distortion is by using a neural network with an appropriate inductive bias.
NeuroLens~\cite{xian2023neural} proposes using an invertible ResNet~\cite{behrmann2019invertible} to represent non-linear lens distortions as a diffeomorphism.
Specifically, the deformation mapping is modeled by a residual network:
\begin{align}
\label{eq:invert}
    \dD_\theta(\textbf{x}) = F_L \circ \cdots \circ F_1(\textbf{x}), \quad F_i(z) = z + f^{(i)}_{\theta_i}(z), 
\end{align}
where $f^{(i)}_{\theta_i}$ is a neural block parameterized by $\theta_i$ with a Lipschitz constant bounded by 1 (\ie, $|f^{(i)}_\theta(x) - f^{(i)}_\theta(y)| < |x-y|$ for all $x$, $y$, and $\theta$). $f^{(i)}_{\theta_i}$ represents a residual block with four linear layers. $L$ denotes the total number of blocks, which is 5 in our case, and the circle denotes function composition.
Such constraints make the network invertible, and its inverse can be obtained using a fixed-point algorithm~\cite{behrmann2019invertible}. In the supplementary, we also provide additional comparisons and insights between iResNet and a regular ResNet.

While an invertible ResNet offers both expressiveness (\ie, the ability to model various lenses) and regularization, it is computationally prohibitive to apply it directly to 3DGS. 
To backpropagate gradients to the alpha-blending weights $\alpha_i$ when rendering an image in \cref{eq:alpha_blending}, the computational graph for the backward passes of $\dD_\theta$ must be maintained for each Gaussian. This is infeasible due to the large number of 3D Gaussians in a single scene, often reaching millions and leading to out-of-memory errors.
This limitation motivates us to develop a more efficient solution that leverages the inherent inductive bias of the invertible ResNet while maintaining computational efficiency.

% \vspace{-1.em}
\paragraph{Hybrid Distortion Field.}
Given that the grid-based method is efficient yet tends to overfit, while the invertible ResNet has an appropriate inductive bias but is not efficient, we propose a hybrid method that combines the advantages of both.
Specifically, we use the invertible ResNet to predict the flow field on a sparse grid and apply bilinear interpolation to each projected 2D Gaussian:
\begin{align}
\label{eq:hybrid_field}
    \dD_\theta(\textbf{x}) = \textbf{x} + \interp{\textbf{x}, \rR_\theta(\textbf{P}_c) - \textbf{P}_c},
\end{align}
where $\textbf{P}_c\in \mathbb{R}^{H\times W\times 2}$ is a sparse grid of fixed control points (pixel locations, where $H\times W$ represents the resolution of control points rather than image resolution), and $\rR_\theta$ is an invertible ResNet parameterized by $\theta$. 

Unlike existing hybrid neural fields~\cite{muller2022instant}, where networks are applied after grid interpolation, our approach uses iResNet to predict displacement vectors on a sparse grid, with bilinear interpolation applied to produce a continuous displacement field, as shown at the bottom of~\cref{fig:pipeline}. 
The advantage of this architecture is that we only need to compute the expensive forward and backward ResNet mappings for locations at $\textbf{P}_c$, which scales with the grid resolution and is independent of the number of Gaussians in the scene.
The additional operation required for each Gaussian is $\interp{\cdot}$, which is computationally affordable and parallelizable.

\subsection{Cubemap for Large FOV}
\label{sec:cubemap}
In order to apply our method to cameras with larger FOV, we extend the single-planar perspective projection to a cubemap projection~\cite{wan2007isocube,jiang2021cubemap}. Mathematically, single-planar projection requires upsampling in the peripheral regions, and the sampling rate increases drastically as the FOV approaches 180\si{\degree}. In contrast, rendering with a cubemap maintains a relatively uniform pixel density from the image center to the edges, making it ideal for wide-angle rendering.

% \vspace{-1.em}
\paragraph{Single-Planar Projection.} Given the parameters estimated from SfM~\cite{schoenberger2016sfm}, the parametric model is then applied to reproject the raw image into perspective images, as shown in the blue box in~\cref{fig:hilbert_180}. These reprojected images are then used for reconstruction through perspective-based pipelines like NeRF~\cite{mildenhall2021nerf} or 3D Gaussian Splatting (3DGS)~\cite{kerbl20233d}. However, this process stretches the pixels in the peripheral regions, and the effect becomes more pronounced when the images are reprojected to larger FOV perspectives, as shown in the orange box.
More specifically, the stretching rate of each pixel is defined by the inverse of~\cref{eq:fisheye}, which exhibits a trend similar to $\tan(r)$, where $r$ is the FOV angle from the center of the raw image. When the FOV of the reprojected image is 110\si{\degree} (as in the blue example), the upsampling rate from the blue circle to the box is approximately 1.4. However, when the FOV increases to 170\si{\degree}, as in the orange example, this rate increases to 11.4, inevitably sacrificing a significant amount of high-frequency information for reconstruction.

Moreover, to preserve central details, the resolution of undistorted images needs to be higher, as the pixel density at the center of the raw image should ideally match that of perspective ones. For example, when undistorting a fisheye image in~\cref{fig:hilbert_180}, the resulting perspective image in the orange region would have an extremely high resolution, making it computationally expensive to render. A common solution is to crop away the periphery, following COLMAP's solution~\cite{schoenberger2016sfm}, but this strategy contradicts our intention of using a fisheye camera to capture wide-angle information.

% \vspace{-1.em}
\paragraph{Multi-Planar Projections.} Inspired by the representation of environment maps using cubemaps~\cite{greene1986environment} and hemi-cube~\cite{Cohen:1985:hemicube} in computer graphics, we propose representing extreme wide-angle renderings with cubemap projections, each covering 90\si{\degree} FOV and oriented orthogonally to one another, as illustrated in~\cref{fig:hilbert_cubemap}. By resampling across the cubemap faces, we can render perspective or distorted images with FOVs even larger than 180\si{\degree}. Fast rasterization is first applied to obtain each face of the cubemap. For each rendered pixel, we look up its corresponding position in the constructed cubemap. Our hybrid distortion field then resamples from the lookup table to achieve the distorted rendering. In practice, it is not necessary to render all faces of the cubemap at once, as the number of cubemap faces may vary for different FOV camera lenses.

The resampling step involves only a simple coordinate transformation along with our hybrid field distortion. The distance from the shared camera center to each plane is normalized to 1. To render a pixel outside the 90\si{\degree} FOV at $(x, y, -1)$ where $x>1$ and $|y|<1$, for instance, the pixel on the right face can be obtained as $(\frac{-1}{x}, \frac{y}{x}, -1)$ in camera coordinates, looking toward the right side. When considering lens distortion, the sampling mapping is distorted according to~\cref{eq:hybrid_field}, altering the lookup on the right face to $(\frac{-1}{x'}, \frac{y'}{x'}, -1)$, where $(x', y') = \dD_\theta(x, y)$. By doing so, the entire distortion field can be directly applied to the cubemap for large FOV rendering. The entire resampling process is fully differentiable, making it directly applicable in our hybrid distortion field as a plug-and-play module.

% \vspace{-1.em}
\paragraph{Gaussian Sorting.} 
3DGS~\cite{kerbl20233d} constructs an ordered set $\mathcal{N}(G)$ of Gaussians before alpha blending~\cite{kerbl20233d}. Gaussians are sorted based on their orthogonal projection distance to the image plane. This approach is valid as long as a single Gaussian is not projected onto multiple faces. However, with the cubemap representation, Gaussians can span the boundary between two faces, leading to multiple projections with inconsistent ordering across faces. This discrepancy introduces intensity discontinuities at the boundaries.
To address this issue, we replace the original sorting strategy with a distance-based approach, ordering Gaussians by their distance from the camera center. This ensures that the rasterization order remains consistent across all cubemap faces, thereby alleviating intensity discontinuities. 
Due to the affine approximation used in~\cite{kerbl20233d,zwicker2001ewa}, the 2D covariance of Gaussians near cubemap face boundaries still has a slight influence on the final rendering, which we further discuss in~\cref{sec:discussions} and the supplementary material.

\subsection{Optimization of Camera Parameters} 
\label{sec:opt_cam}
Our pipeline differentiates all camera parameters. We theoretically derive the gradient for all camera parameters, including extrinsics and intrinsics, making the camera module completely differentiable and capable of being optimized alongside distortion modeling. All gradient calculations are implemented with a native CUDA kernel. A comprehensive mathematical derivation and experimental results are provided in the supplementary.

% \subsection{Other Lens Effects}
% \TODO{TBD}
% entrance pupil shift
% \label{sec:shift}
{\begin{table*}[ht]

  \centering
  \scalebox{0.64}{
    \begin{tabular}{ccccccccccccccccccc}
    \toprule
    \multicolumn{1}{c}{\multirow{2}[2]{*}{Method}} & \multicolumn{3}{c}{Chairs} & \multicolumn{3}{c}{Cube} & \multicolumn{3}{c}{Flowers} & \multicolumn{3}{c}{Globe} & \multicolumn{3}{c}{Heart} & \multicolumn{3}{c}{Rock}\\
    \cmidrule(lr){2-4}
    \cmidrule(lr){5-7}
    \cmidrule(lr){8-10}
    \cmidrule(lr){11-13}
    \cmidrule(lr){14-16}
    \cmidrule(lr){17-19}
    \multicolumn{1}{c}{} & SSIM  & PSNR  & LPIPS & SSIM  & PSNR & LPIPS & SSIM  & PSNR & LPIPS & SSIM  & PSNR & LPIPS & SSIM  & PSNR  & LPIPS & SSIM  & PSNR  & LPIPS \\
    \midrule
    3DGS-perspective~\cite{kerbl20233d} &0.431&14.06&0.547&0.507&15.21&0.533&0.281&12.91&0.609&0.502&15.09& 0.530&0.505&15.19&0.549&0.297&12.70&0.595\\
    3DGS-COLMAP~\cite{kerbl20233d}   &0.583&18.28&0.290&0.637&21.64&0.296&0.443&18.09&0.379&0.580&19.63&0.327&0.660&20.87&0.282&0.511&20.24&0.280\\

    % Dense-Grid
    % &0.511&15.14&0.525&0.316&13.29&0.581&0.513&15.36&0.517&0.556&16.02&0.489&0.556&16.03&0.489&0.381&13.74&0.527\\
    % Sparse-Grid
    % &0.532&15.79&0.442&0.529&16.06&0.513&0.339&14.19&0.560&0.535&16.29&0.502&0.576&16.58&0.475&0.419&15.10&0.482\\

    % COLMAP+GS                       &0.583&18.28&0.290&0.637&21.64&0.296&0.443&18.09&0.379&0.580&19.63&0.327&0.660&20.87&0.282&0.511&20.24&0.280\\
    Adop-GS~\cite{ruckert2022adop}    &0.829&22.59&0.200&0.755&22.12&0.289&0.646&19.96&0.314&0.758&21.35&0.294&0.741&21.37&0.306&0.726&22.48&0.254\\

    Fisheye-GS~\cite{liao2024fisheye} & 0.785 & 21.68 & 0.110 & 0.754 & 23.29 & 0.166 & 0.615 & 20.23 & 0.214 & 0.728 & 22.11 & 0.160 & 0.722 & 21.37 & 0.218 & 0.697 & 22.38 & 0.177 \\
    
    \midrule

    Ours &\textbf{0.832}&\textbf{23.45}&\textbf{0.106}&\textbf{0.786}&\textbf{24.63}&\textbf{0.162}&\textbf{0.693}&\textbf{22.01}&\textbf{0.172}&\textbf{0.790}&\textbf{23.63}&\textbf{0.126}&\textbf{0.775}&\textbf{23.42}&\textbf{0.195}&\textbf{0.787}&\textbf{24.88}&\textbf{0.145}\\
    \bottomrule
    \end{tabular}%
  } 
  % \vspace{-2mm}
  \caption{\textbf{Quantitative Evaluation on the FisheyeNeRF Dataset}~\cite{jeong2021self}. We evaluate our method on a challenging real-world benchmark. Our method consistently outperforms existing baselines. Additional qualitative results are shown in~\cref{fig:qualitative_fisheyenerf}.}
    % \vspace{-2mm}
  \label{tab:quantitative_fisheyenerf}%
\end{table*}%
}
{
\begin{figure*}[t]
    \centering
    \setlength{\tabcolsep}{1pt} % Adjust space between columns if needed
    \includegraphics[width=1\textwidth]{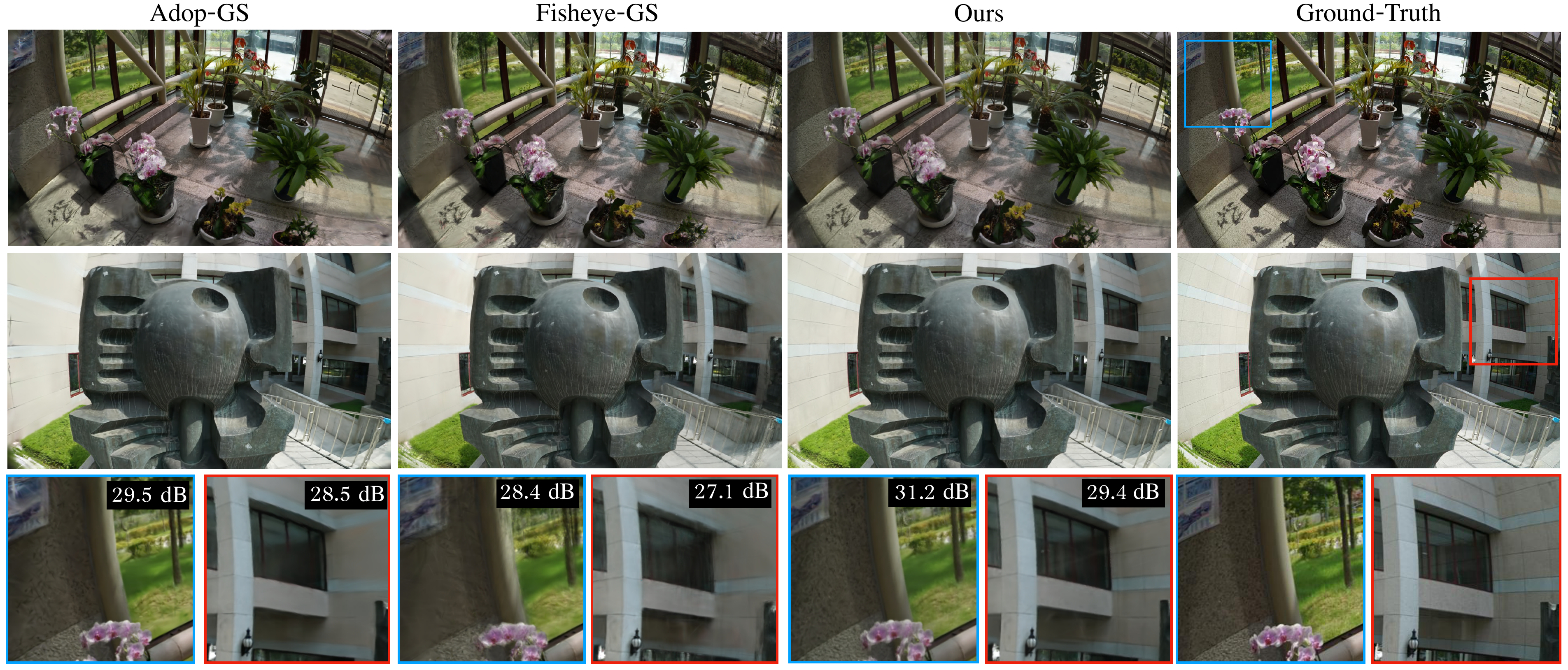}
    \caption{\textbf{Qualitative Comparisons with Baselines on the FisheyeNeRF Dataset}~\cite{jeong2021self}. The images show comparisons across different scenes using two baselines (\textit{e.g.,} ADOP-GS~\cite{ruckert2022adop} and Fisheye-GS~\cite{liao2024fisheye}) and our method. PSNRs are computed for each patch.}
    % \vspace{-1.em}
    \label{fig:qualitative_fisheyenerf}
\end{figure*}

}
% \vspace{-2mm}
\section{Experiments\label{sec:experiments}}
In this section, we first briefly introduce our data preparation pipeline in \cref{sec:data}.
We compare our method against various baselines for scene reconstruction with large-FOV input (\cref{sec:evaluation} and \cref{sec:eval_cubemap}), followed by ablation studies for both the neural distortion model and the cubemap projection in \cref{sec:ablation}.

\subsection{Data Acquisition}
\label{sec:data}
We customized a camera module in the Mitsuba ray tracer~\cite{jakob2010mitsuba} to incorporate fisheye camera parameters derived from the open-source Lensfun database~\cite{lensfun}. Using a 180\si{\degree} fisheye camera, we rendered three scenes from~\cite{resources16}. We generated a training set with both perspective and fisheye views for baselines and our method, respectively. Additionally, we captured several real-world datasets using different uncalibrated cameras to evaluate our method. These datasets consist of casual walk-around video footage captured using a camera with an approximate 150\si{\degree} FOV. Furthermore, we tested our method on the existing FisheyeNeRF benchmark dataset~\cite{jeong2021self}, which contains fisheye images with an FOV of approximately 120\si{\degree}.

{\begin{figure*}[ht]
    \centering
    \setlength{\tabcolsep}{1pt} % Adjust space between columns if needed
    \scalebox{0.9}{
    \begin{tabular}{cccccc} % 4 columns (Vertical Caption | Image Set 1 | Image Set 2 | Image Set 3)

        \includegraphics[height=0.142\textwidth]{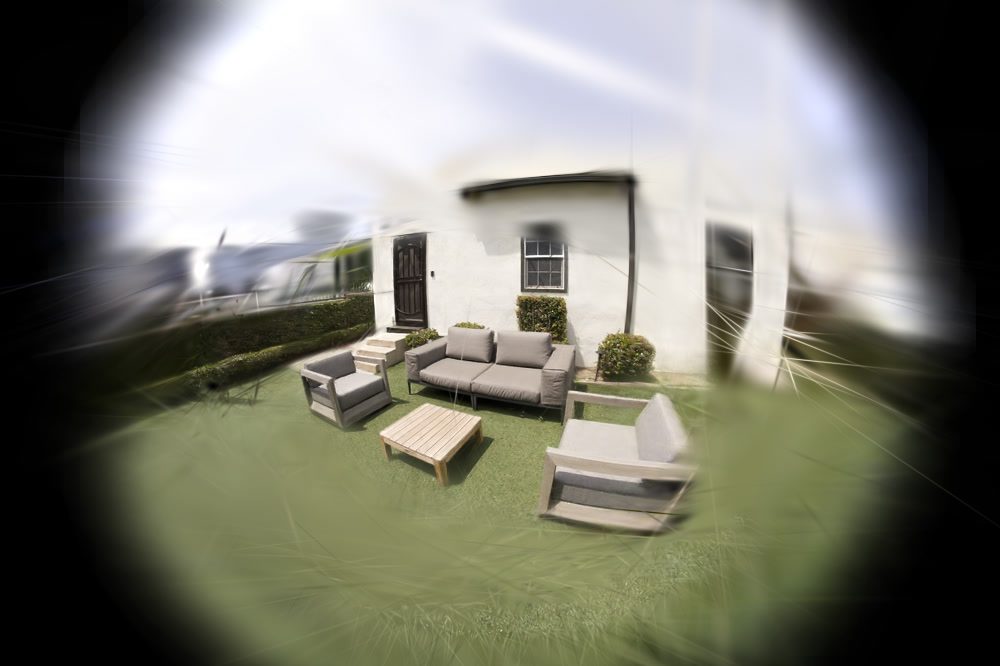} &
        \includegraphics[height=0.142\textwidth]{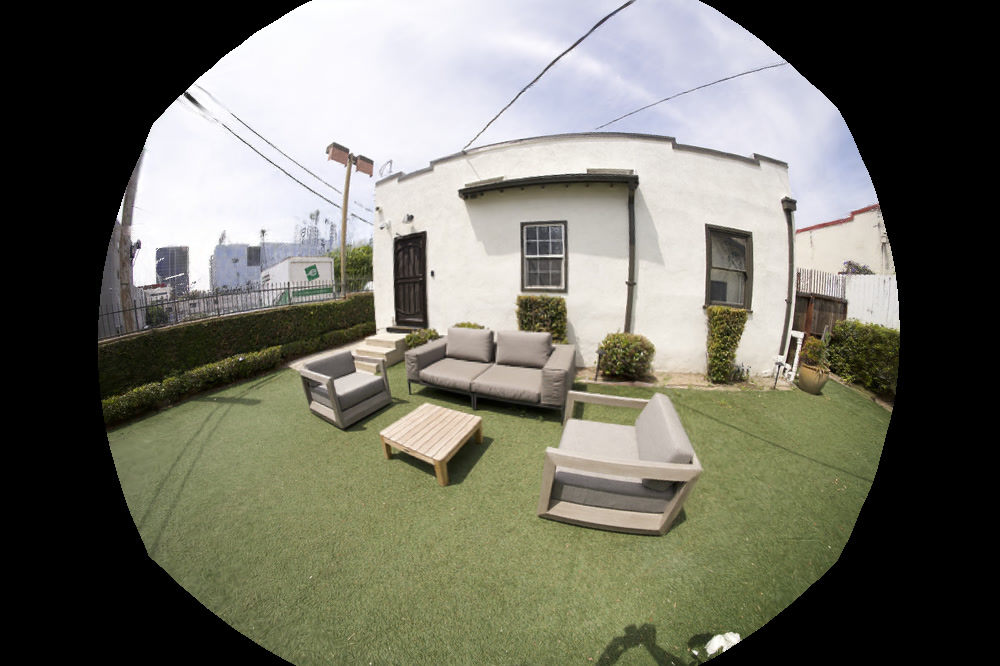} &
        \includegraphics[height=0.142\textwidth]{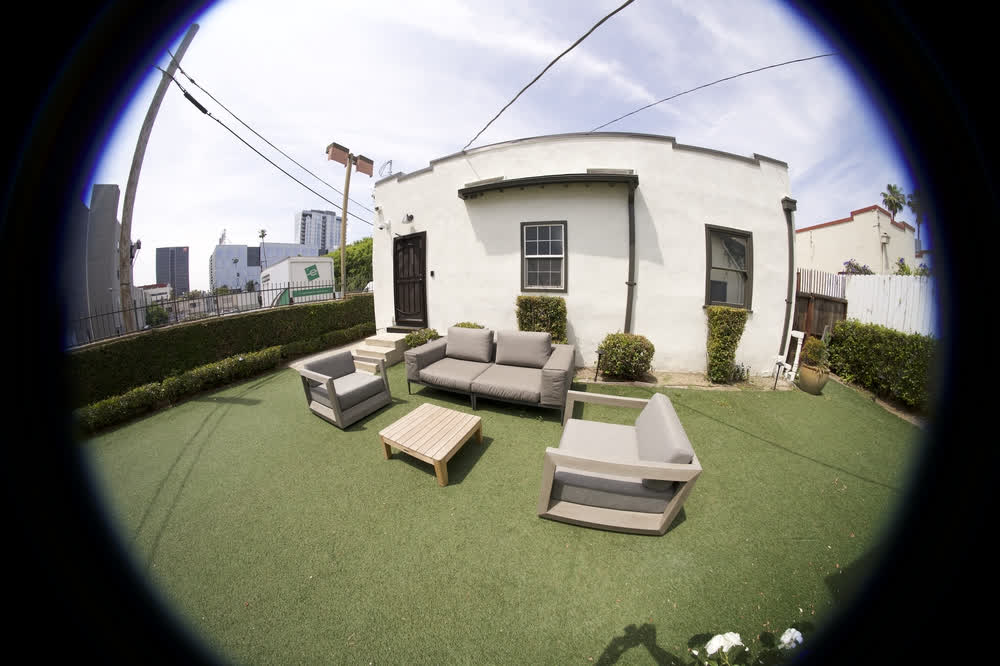} &
        \includegraphics[height=0.142\textwidth]{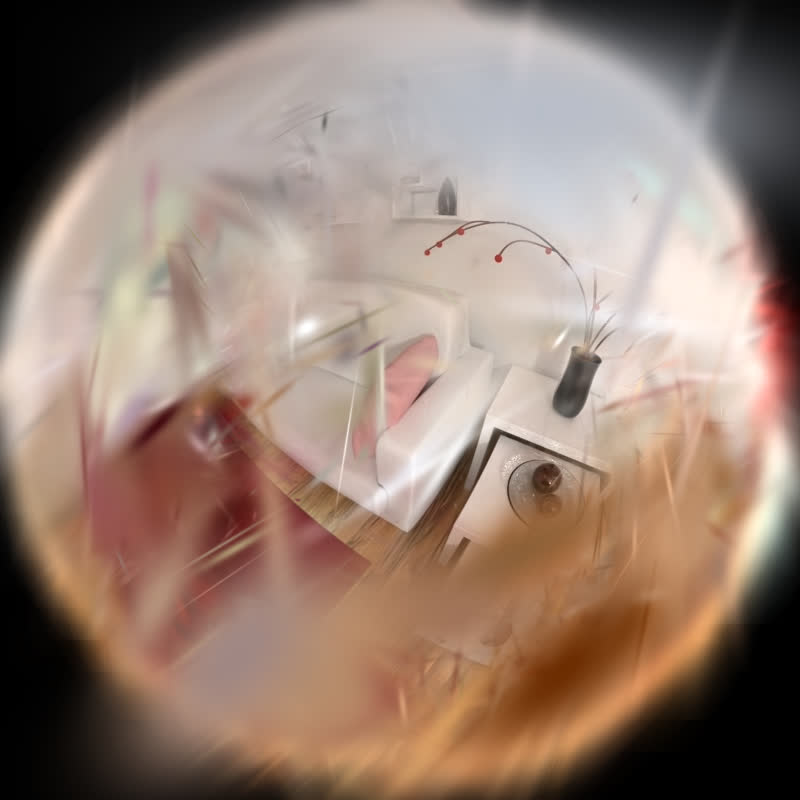} &
        \includegraphics[height=0.142\textwidth]{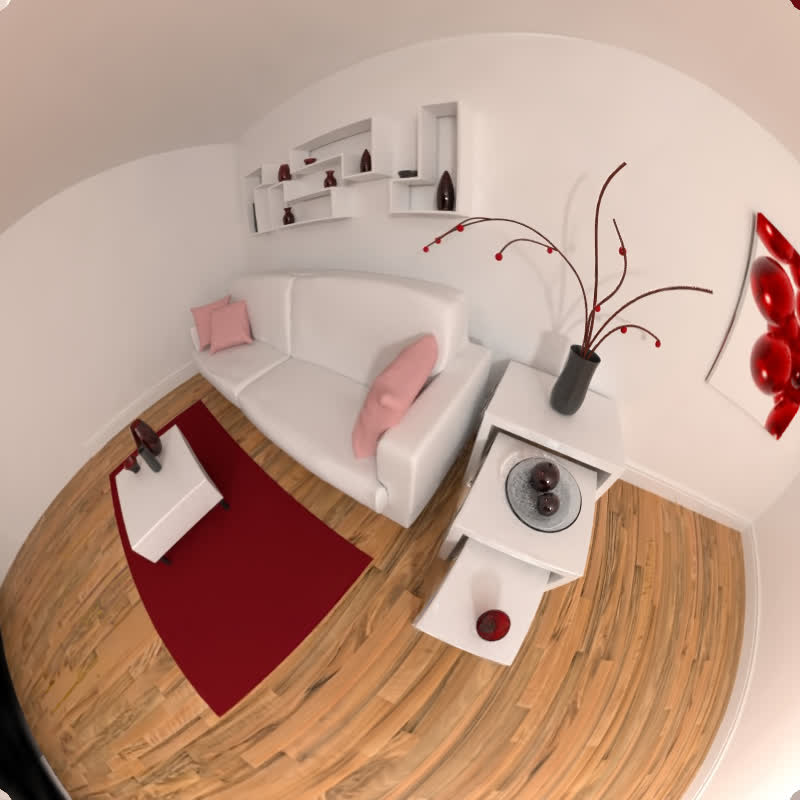} &
        \includegraphics[height=0.142\textwidth]{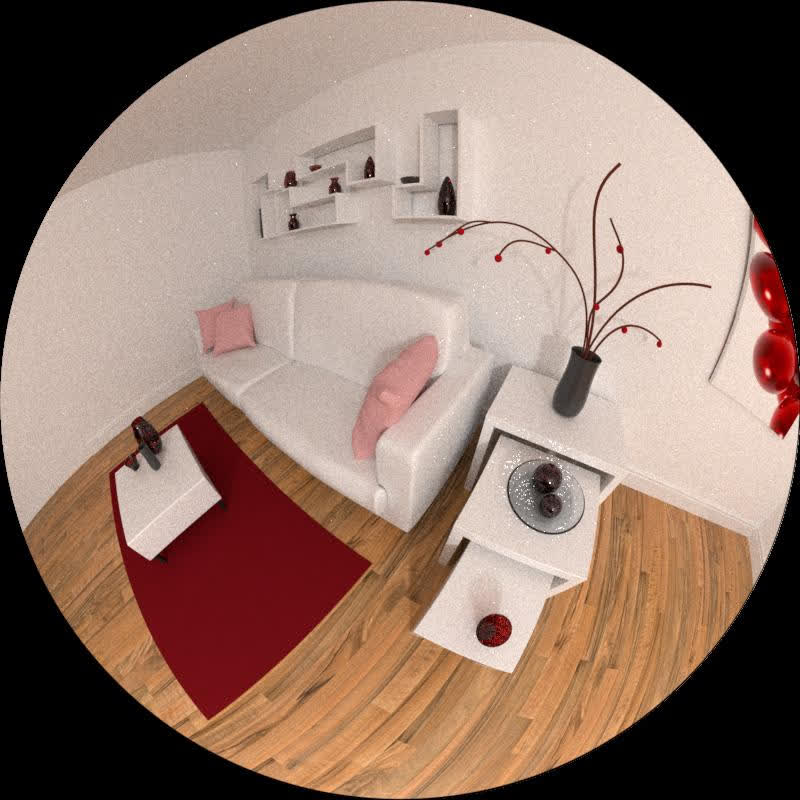}
        \\

        % \raisebox{0.5\height}{\hspace{-0.5cm}\rotatebox{90}{{Kitchen}}} &
        % \includegraphics[width=0.3\textwidth]{images/pano/kitchen_smallfov_pano.jpg} &
        % \includegraphics[width=0.3\textwidth]{images/pano/kitchen_ours_pano.jpg} &
        % \includegraphics[width=0.142\textwidth]{images/large_vs_small_fov_jpg/kitchen_fisheye-gs_fisheye.jpg} &
        % \includegraphics[width=0.142\textwidth]{images/large_vs_small_fov_jpg/kitchen_ours_fisheye.jpg} &
        % \includegraphics[width=0.142\textwidth]{images/large_vs_small_fov_jpg/kitchen_gt_fisheye.jpg}
        % \\

        \includegraphics[height=0.142\textwidth]{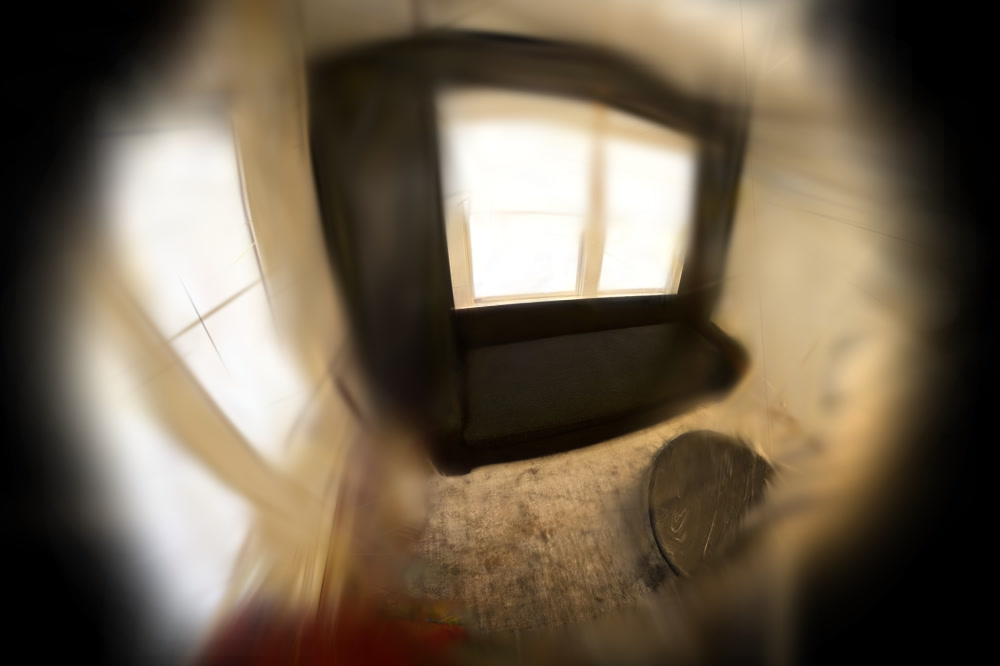} &
        \includegraphics[height=0.142\textwidth]{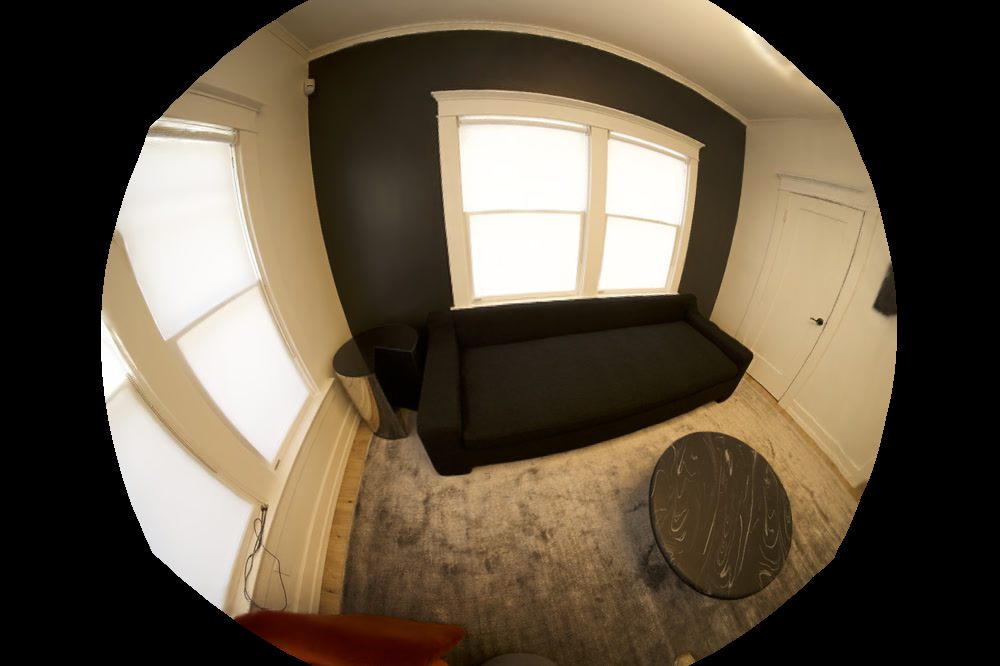} &
        \includegraphics[height=0.142\textwidth]{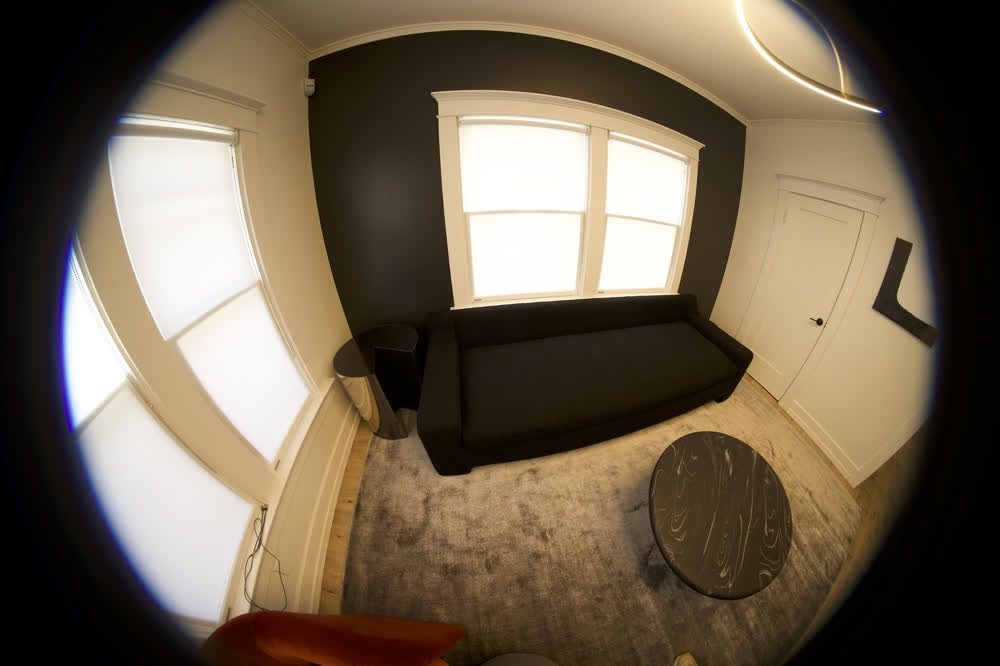} &
        \includegraphics[height=0.142\textwidth]{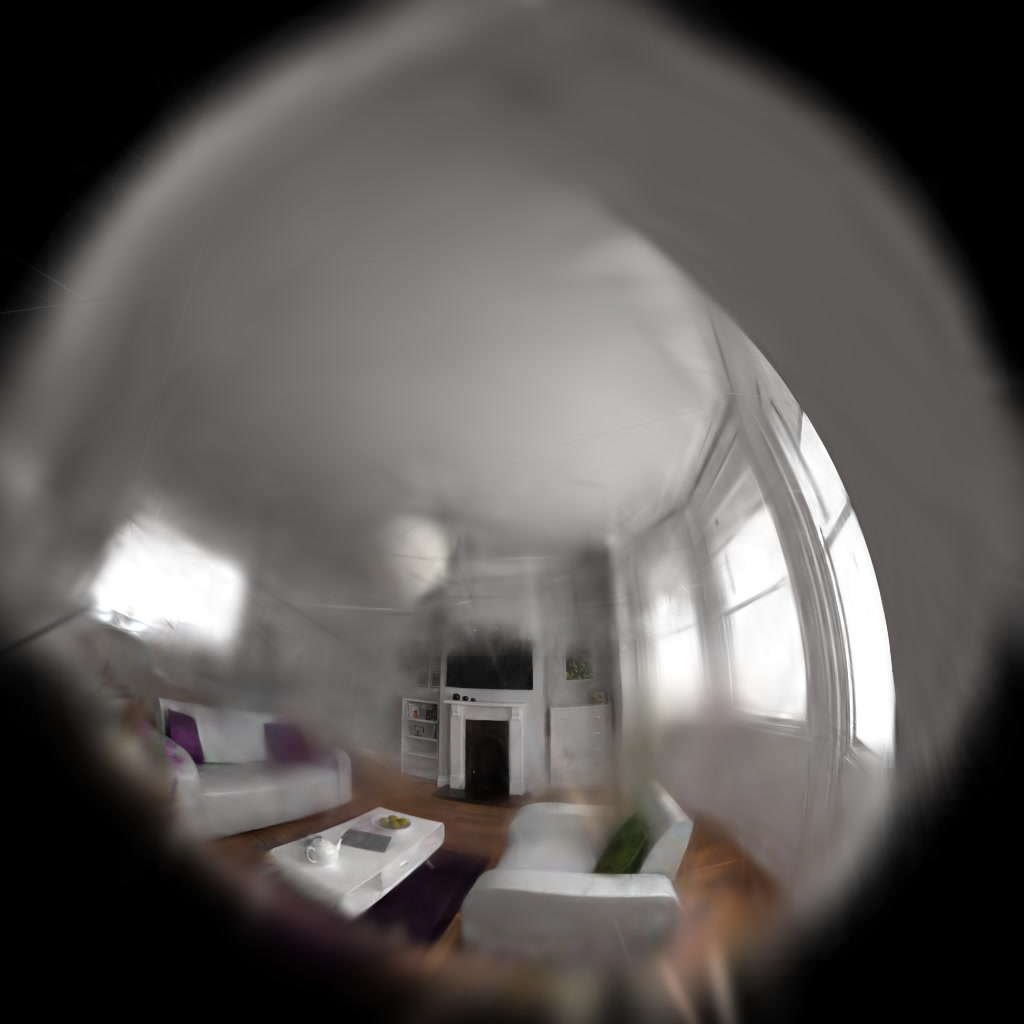} &
        \includegraphics[height=0.142\textwidth]{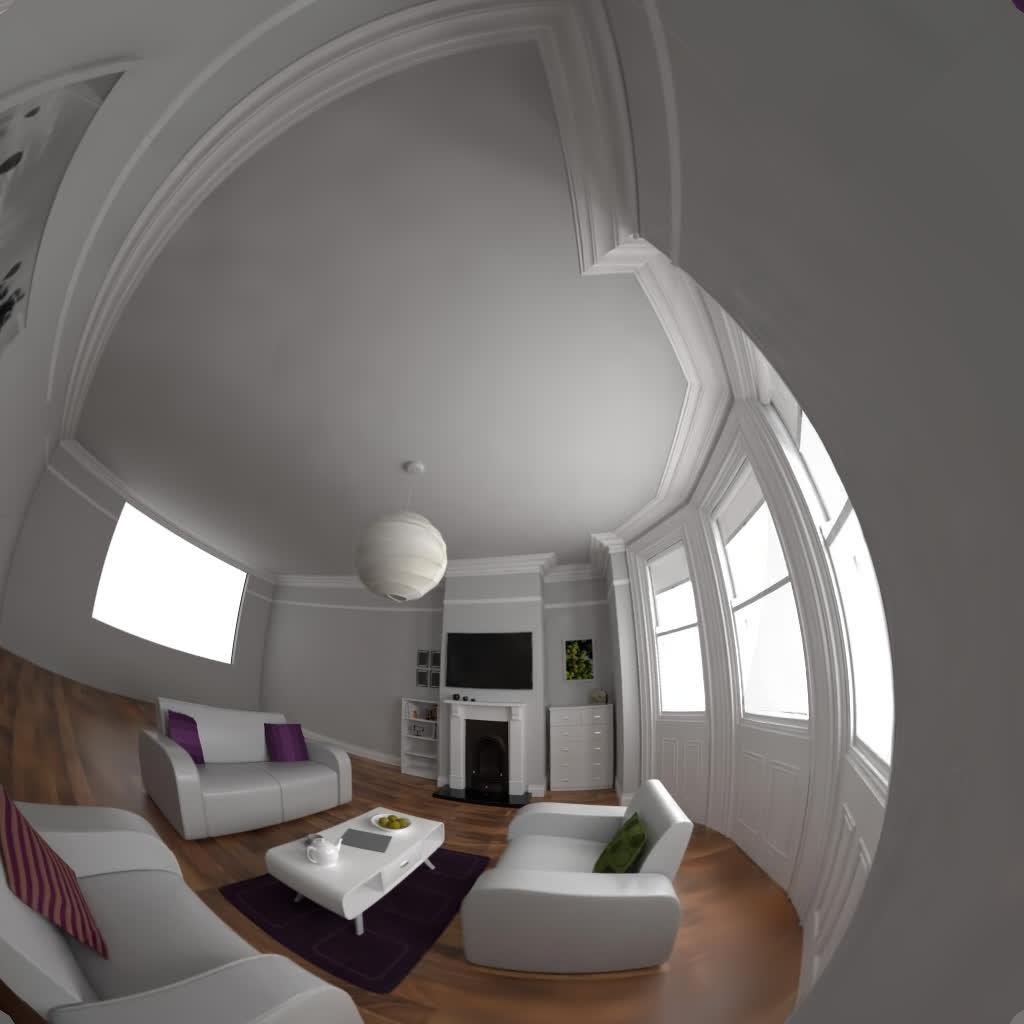} &
        \includegraphics[height=0.142\textwidth]{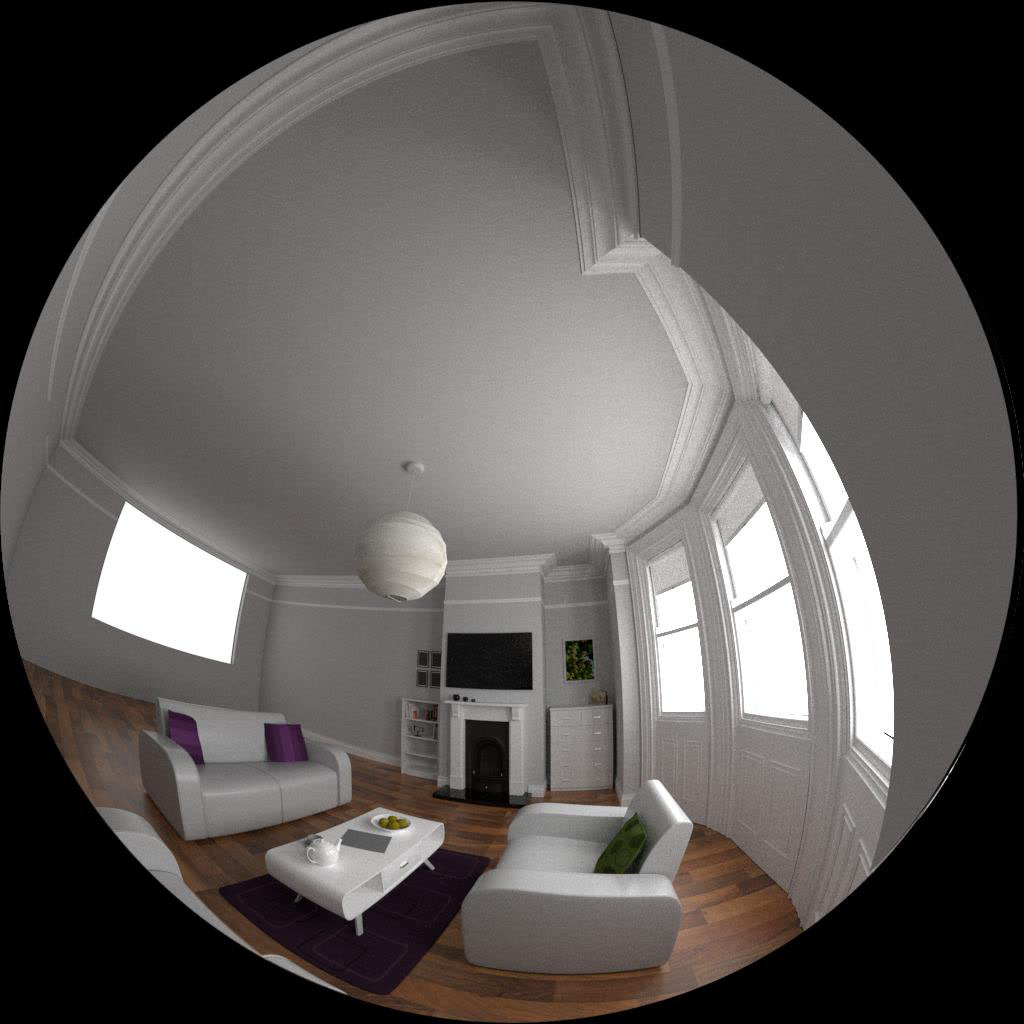}
        \\
        
        \multicolumn{1}{c}{(a) Fisheye-GS} & \multicolumn{1}{c}{(b) Ours} & \multicolumn{1}{c}{(c) GT} & \multicolumn{1}{c}{(d) Fisheye-GS} & \multicolumn{1}{c}{(e) Ours} & \multicolumn{1}{c}{(f) GT}
        \\
    \end{tabular}
    }
    % \vspace{-2mm}
    \caption{\textbf{Qualitative Comparisons with Fisheye-GS~\cite{liao2024fisheye}}. To validate our hybrid distortion modeling, we further compare our method with Fisheye-GS~\cite{liao2024fisheye} on larger FOV scenes, including real-world captures using 150\si{\degree} cameras (a–c) and simulations using a 180\si{\degree} camera (d–f) in Mitsuba~\cite{jakob2010mitsuba}. Our method successfully recovers details in peripheral regions, whereas Fisheye-GS~\cite{liao2024fisheye} struggles.}

    \label{fig:largefov_vs_smallfov_mitsuba_pano_fisheye}
    % \vspace{-2mm}
\end{figure*}
}

{
\begin{figure}[t]
    \centering
    % set the image length
    % \newlength 0.15\textwidth
    % set the space between columns
    \setlength{\tabcolsep}{0.1em} 
    \renewcommand{\arraystretch}{0.99}
    \scalebox{0.82}{
        \begin{tabular}{cccccc}
        \includegraphics[width=0.135\textwidth]{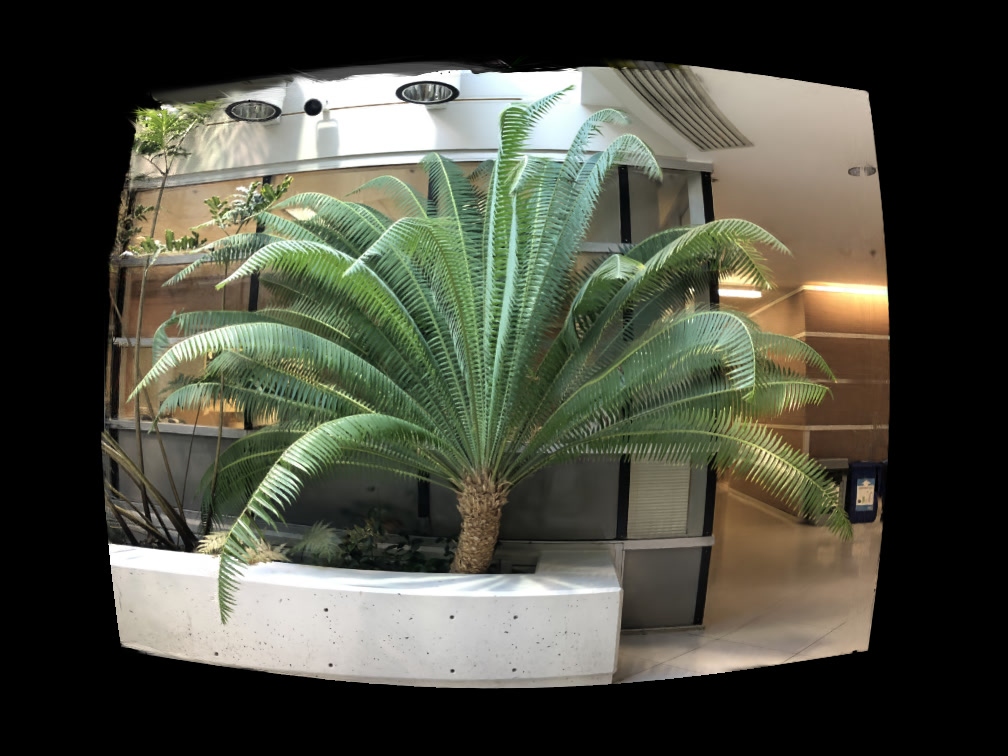} &  
        \includegraphics[width=0.135\textwidth]{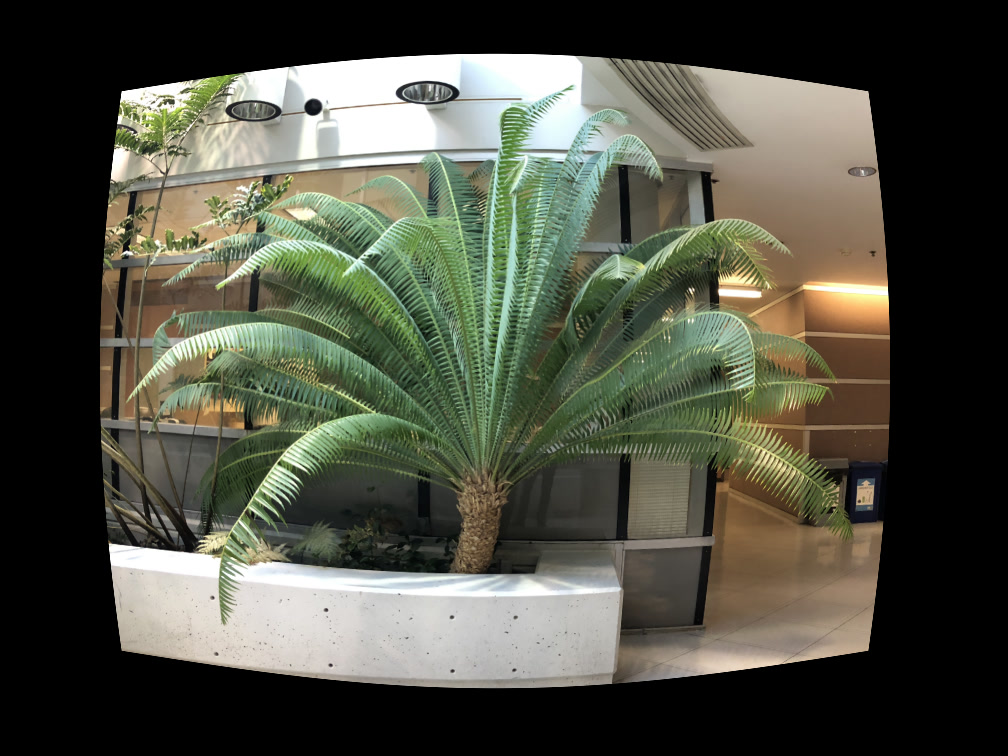} &  
        \includegraphics[width=0.144\textwidth]{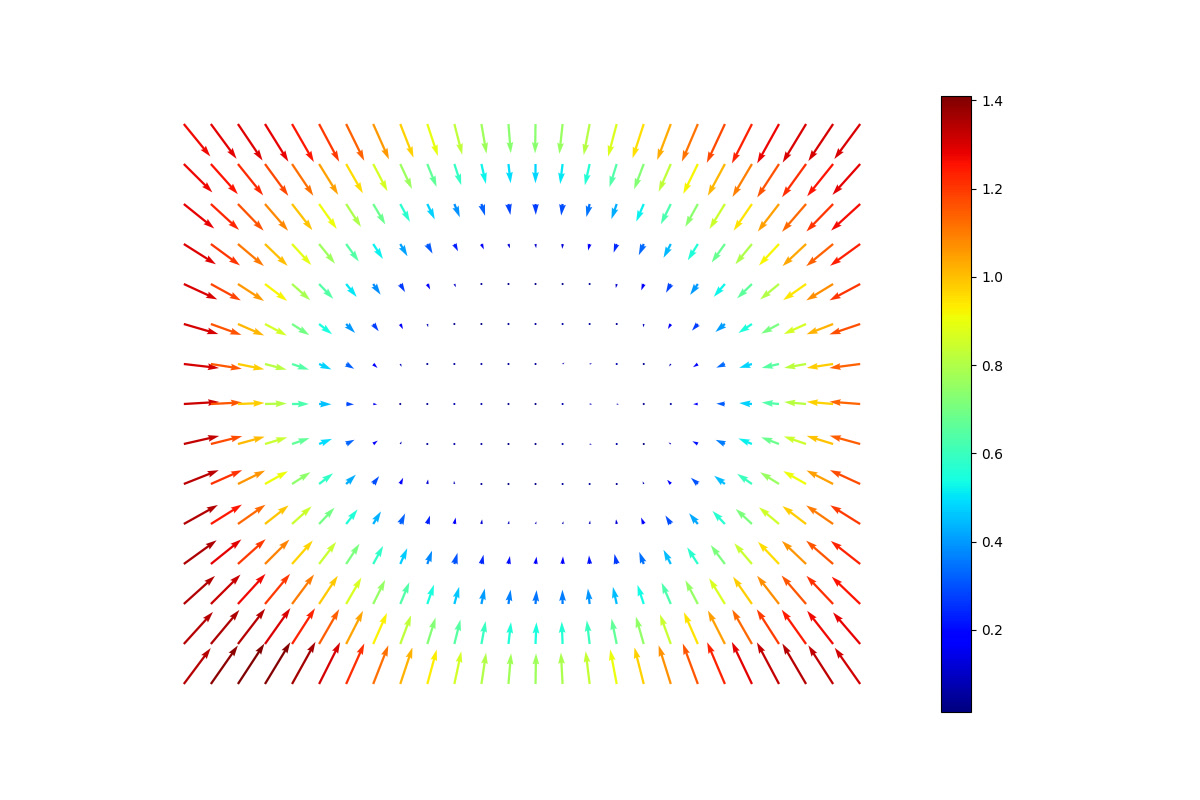} &  
        \includegraphics[width=0.135\textwidth]{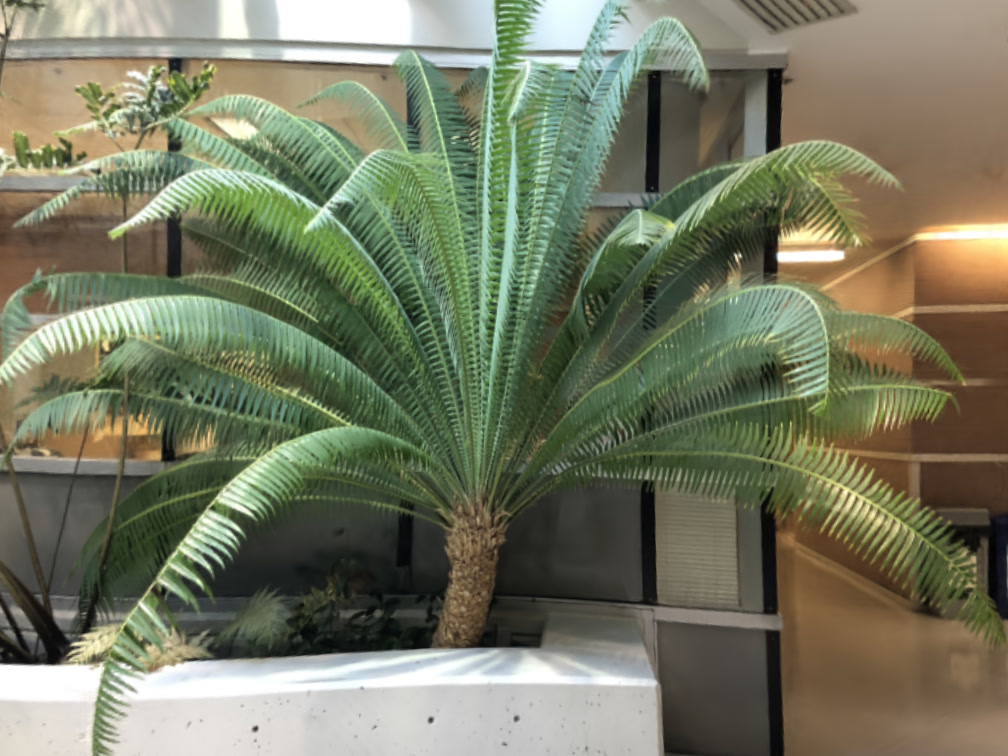}
        \\
        \includegraphics[width=0.135\textwidth]{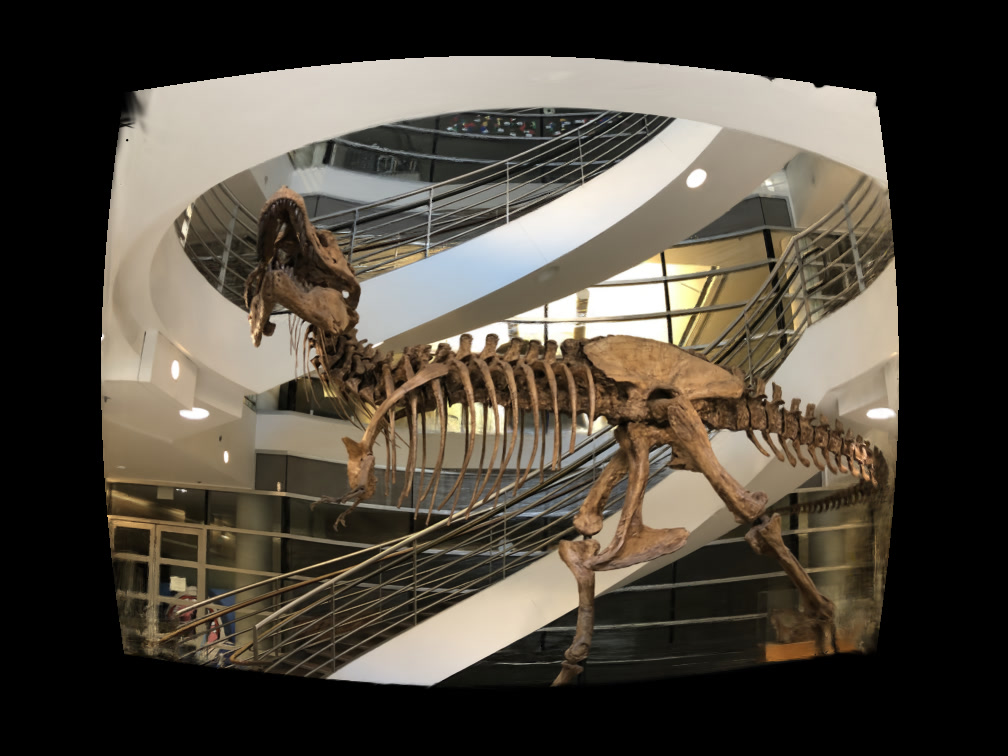} &  
        \includegraphics[width=0.135\textwidth]{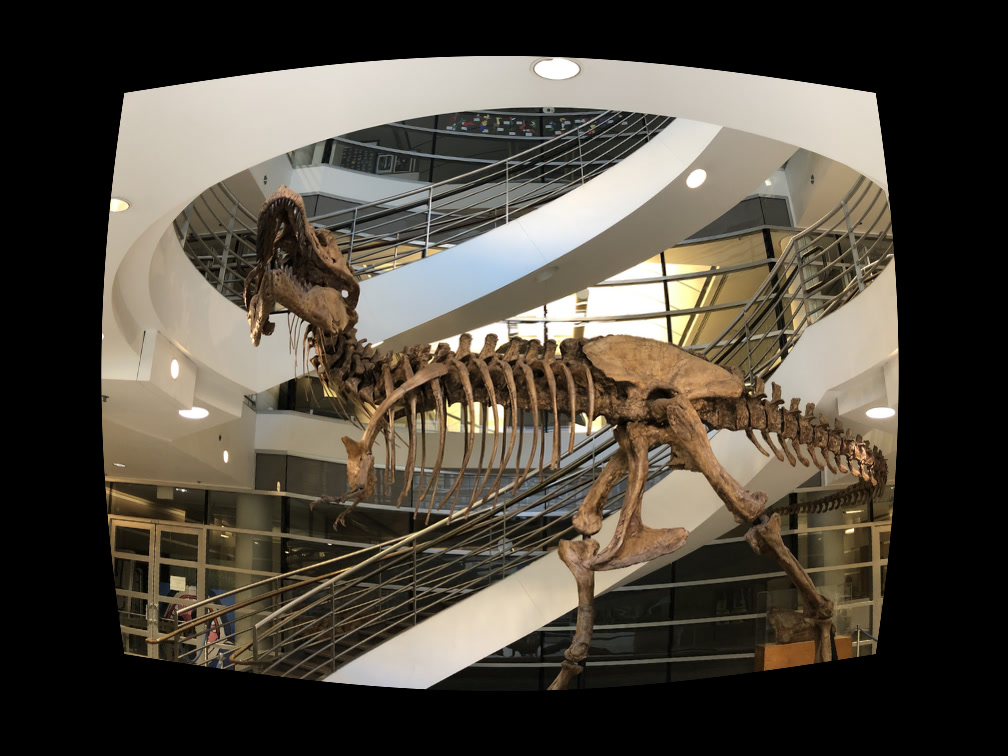} &  
        \includegraphics[width=0.144\textwidth]{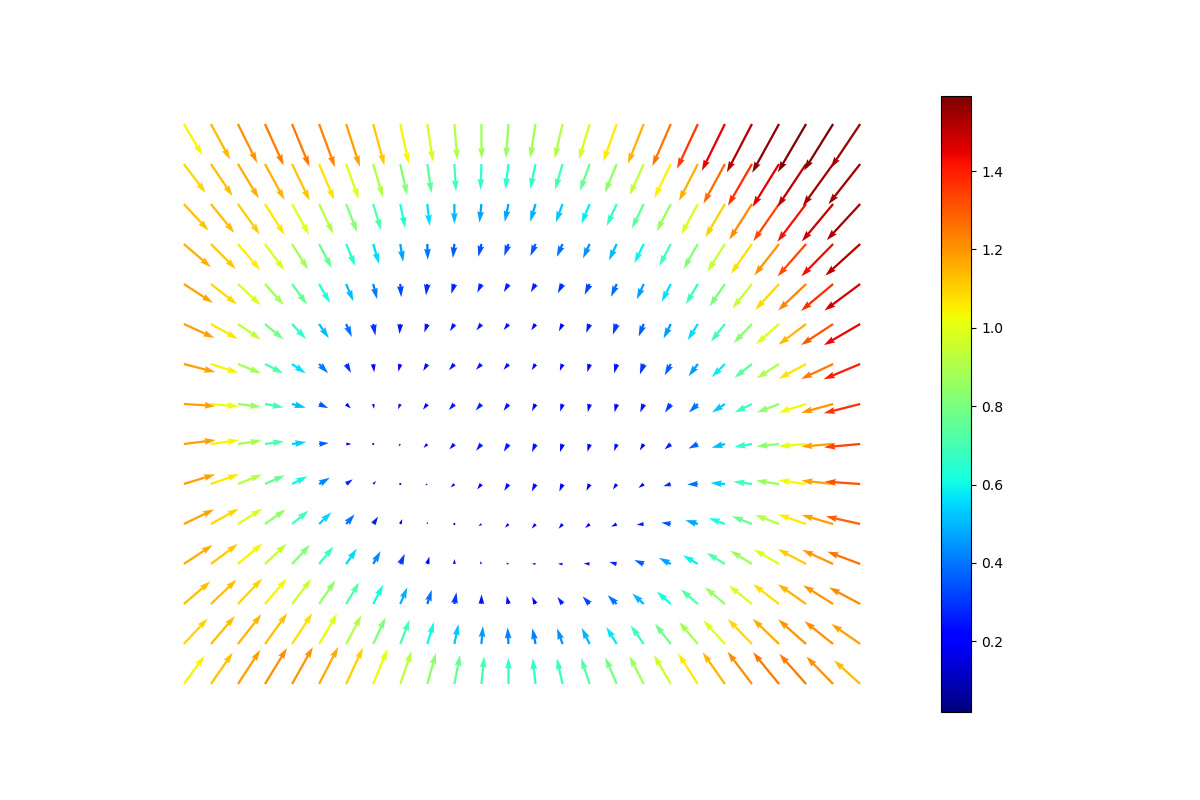} &  
        \includegraphics[width=0.135\textwidth]{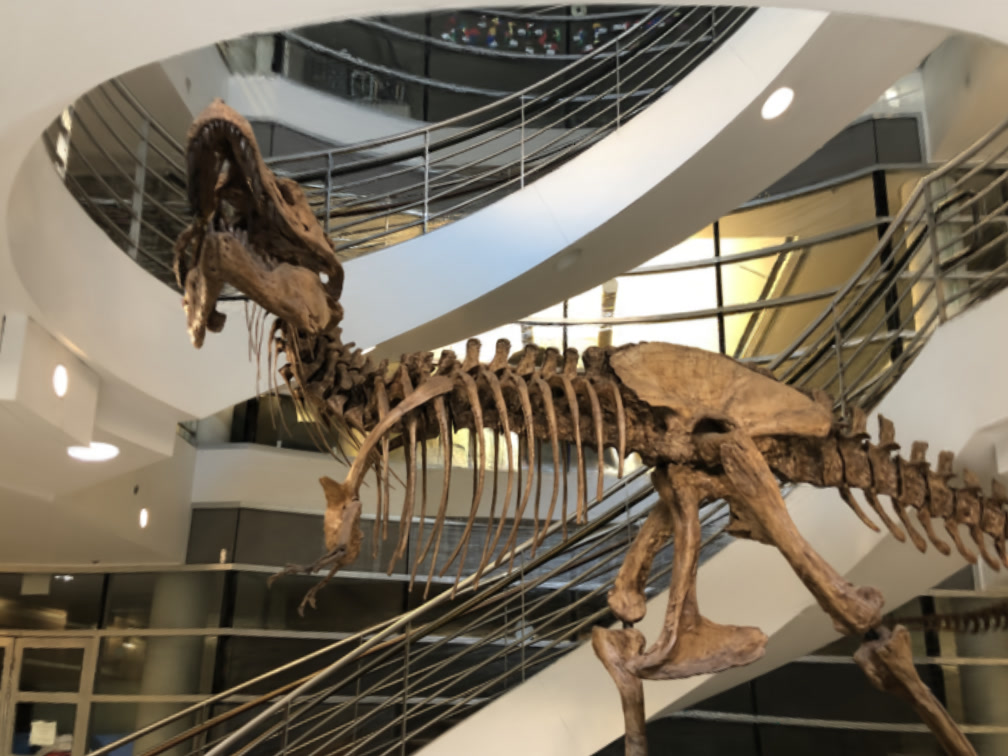} 
        \\
        (a) Ours & (b) GT & (c) Distortion & (d) Ours\\
    \end{tabular}
    }
    % \vspace{-2mm}
     \caption{\textbf{Qualitative Results of Radial and Tangential Distortion}. We apply synthetic distortion to real-world images and (a) show the distorted rendering, comparing it with the reference images in (b). After training, we use the distortion field in (c) to reproject the image to a perspective view. Notably, we apply a combination of radial and tangential distortion to T-Rex.}
    % \vspace{-5mm}
    \label{fig:radial_tangential}
\end{figure}
} 
\subsection{Comparisons to Traditional Lens Models} 
\label{sec:evaluation}
We first validate that our hybrid field representation models large-FOV cameras better than traditional polynomial distortion models~\cite{opencv_library, schoenberger2016sfm}. Using the FisheyeNeRF dataset~\cite{jeong2021self}, we compare our method against four baselines. The first baseline is Vanilla 3DGS~\cite{kerbl20233d}, which only uses a perspective camera model. Second, we directly apply the polynomial distortion model estimated by COLMAP~\cite{schoenberger2016sfm} to 2D Gaussians after the rasterization of 3DGS~\cite{kerbl20233d}. Third, we evaluate Fisheye-GS~\cite{liao2024fisheye}, which modifies the projection of Gaussians to a specific fisheye parametric model. Finally, we re-implemented ADOP~\cite{ruckert2022adop} using an omnidirectional camera model as the fourth baseline.

We report quantitative results in~\cref{tab:quantitative_fisheyenerf}, and our method consistently outperforms all baselines. We also show novel-viewpoint renderings in~\cref{fig:qualitative_fisheyenerf}. Baseline methods such as ADOP~\cite{ruckert2022adop} and Fisheye-GS~\cite{liao2024fisheye} tend to produce more artifacts in the corners of the test views, whereas our method reconstructs finer details.

To evaluate how our method performs with increasing FOV, we conducted an additional experiment on two datasets with different camera FOVs: real-world scenes captured with 150\si{\degree} cameras and a synthetic dataset with a 180\si{\degree} FOV. As the FOV approaches 180\si{\degree}, as shown in~\cref{fig:largefov_vs_smallfov_mitsuba_pano_fisheye}, our method successfully handles peripheral regions, whereas Fisheye-GS~\cite{liao2024fisheye} fails, producing spiky Gaussians in the surroundings and introducing artifacts at the center that blur the rendering. The quantitative evaluation of~\cref{fig:largefov_vs_smallfov_mitsuba_pano_fisheye} can be found in the supplementary material.

Our method can also be applied to different types of lens distortion. We introduce radial and tangential distortions to images in the LLFF dataset~\cite{mildenhall2019llff}, using camera parameters derived from the Lensfun database~\cite{lensfun}. 
By applying these parameters, we generate a combination of distortions, as demonstrated in the T-Rex scene. 
We optimize our model using these distorted images and obtain the distortion map learned by our lens model.
\cref{fig:radial_tangential} shows that our method accurately recovers various types of camera distortion without manual calibration or access to the physical camera.

{\begin{figure*}[t]
    \centering
    \setlength{\tabcolsep}{1pt} % Adjust space between columns if needed
    \begin{tabular}{cccccc} % 4 columns (Vertical Caption | Image Set 1 | Image Set 2 | Image Set 3)
                
        % \raisebox{0.4\height}{\hspace{-0.5cm}\rotatebox{90}{{Garden}}} &
        \includegraphics[height=0.14\textwidth]{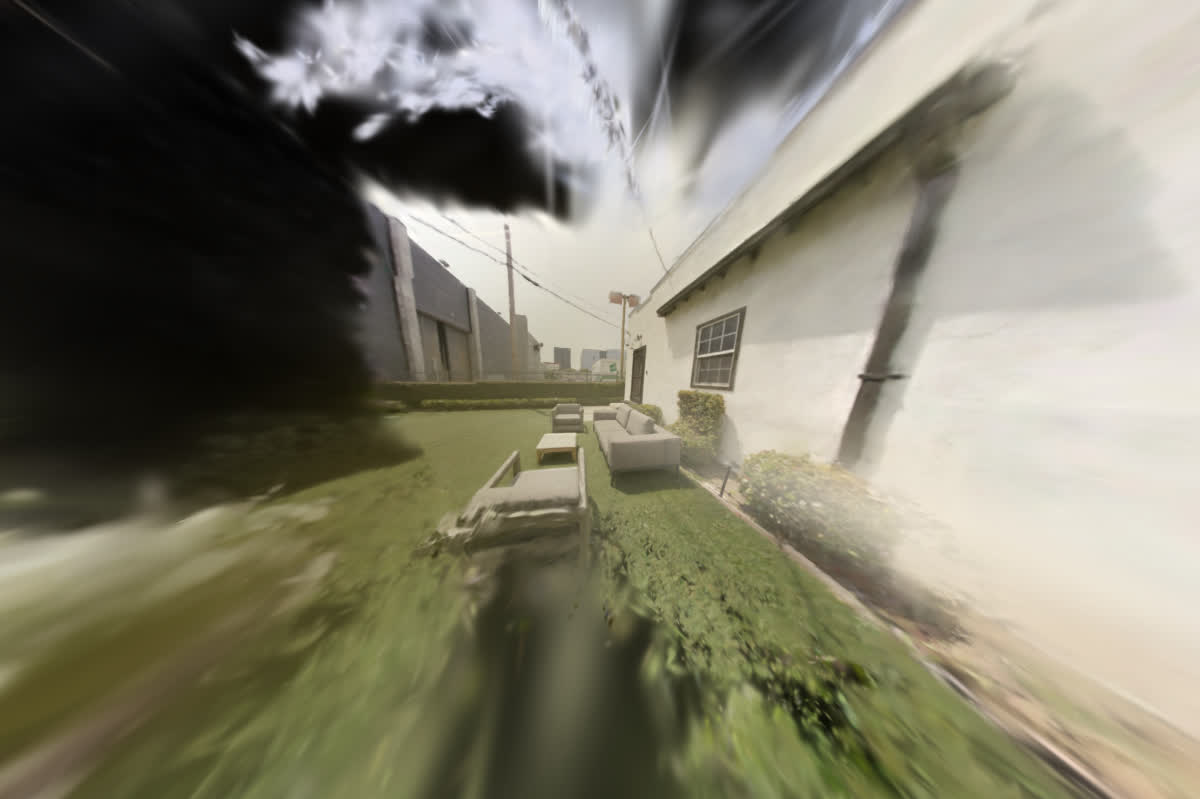} &
        \includegraphics[height=0.14\textwidth]{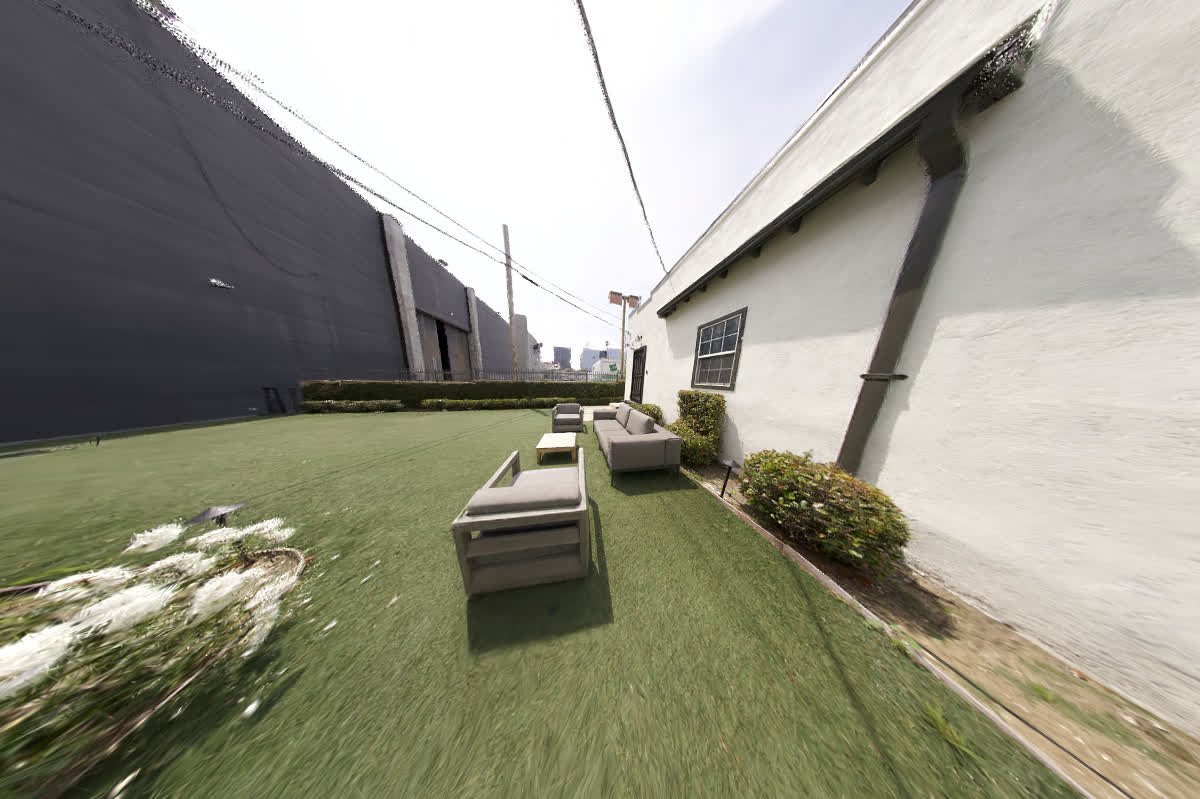} &
        % \raisebox{0.7\height}{\hspace{-0.5cm}\rotatebox{90}{{Room1}}} &
        \includegraphics[height=0.14\textwidth]{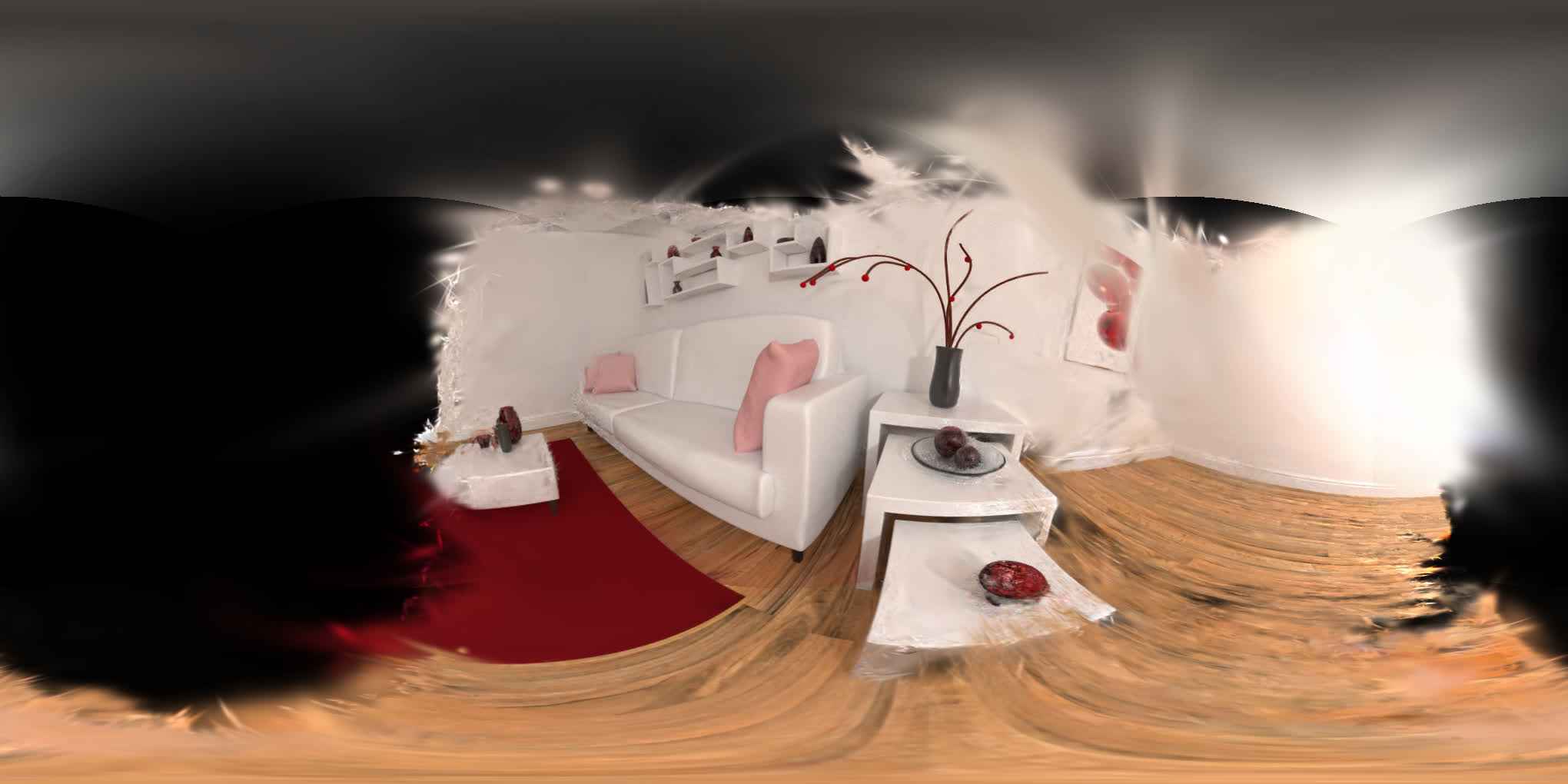} &
        \includegraphics[height=0.14\textwidth]{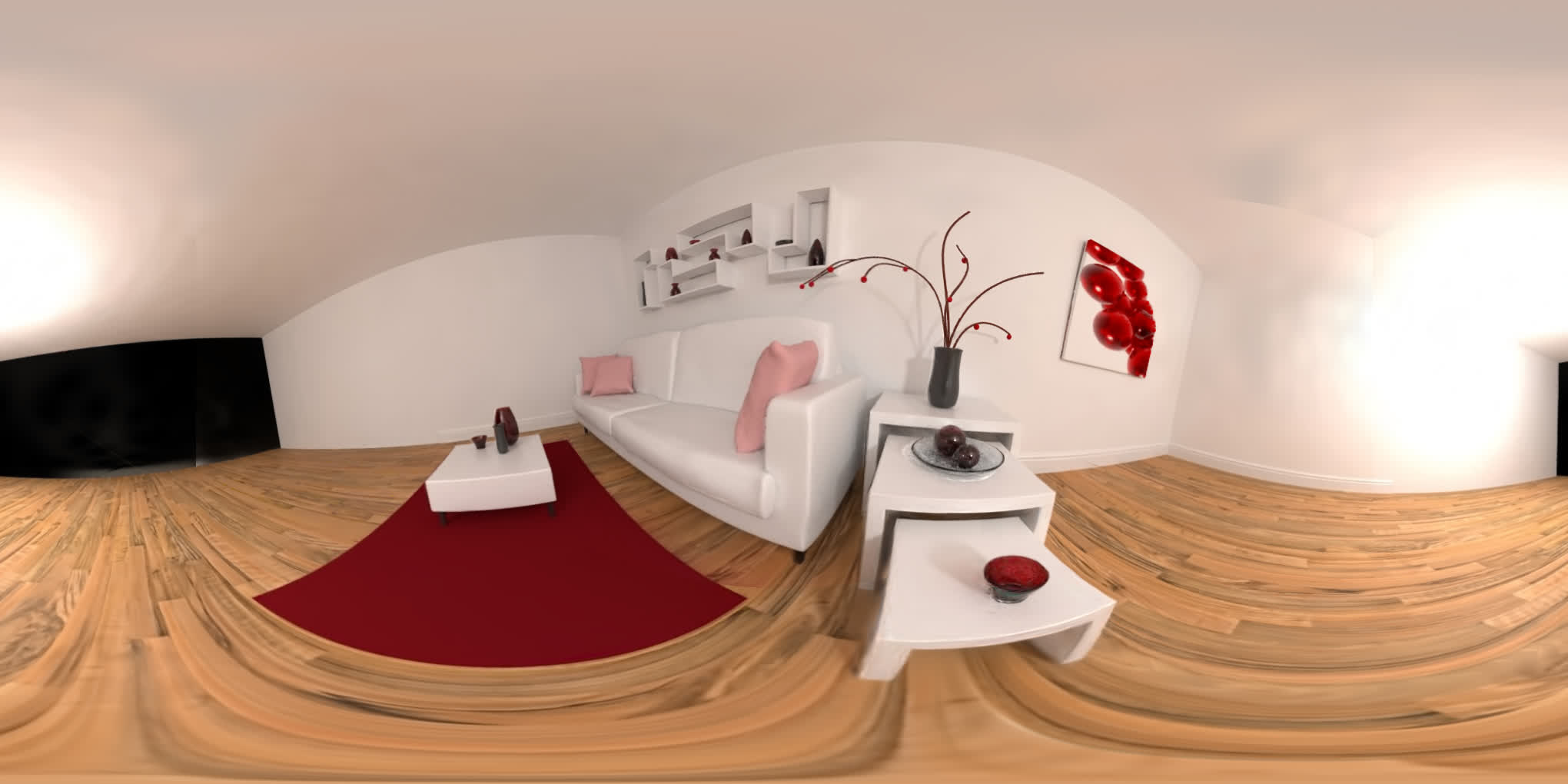} \\
                        
        % \raisebox{0.5\height}{\hspace{-0.5cm}\rotatebox{90}{{Studio}}} &
        \includegraphics[height=0.14\textwidth]{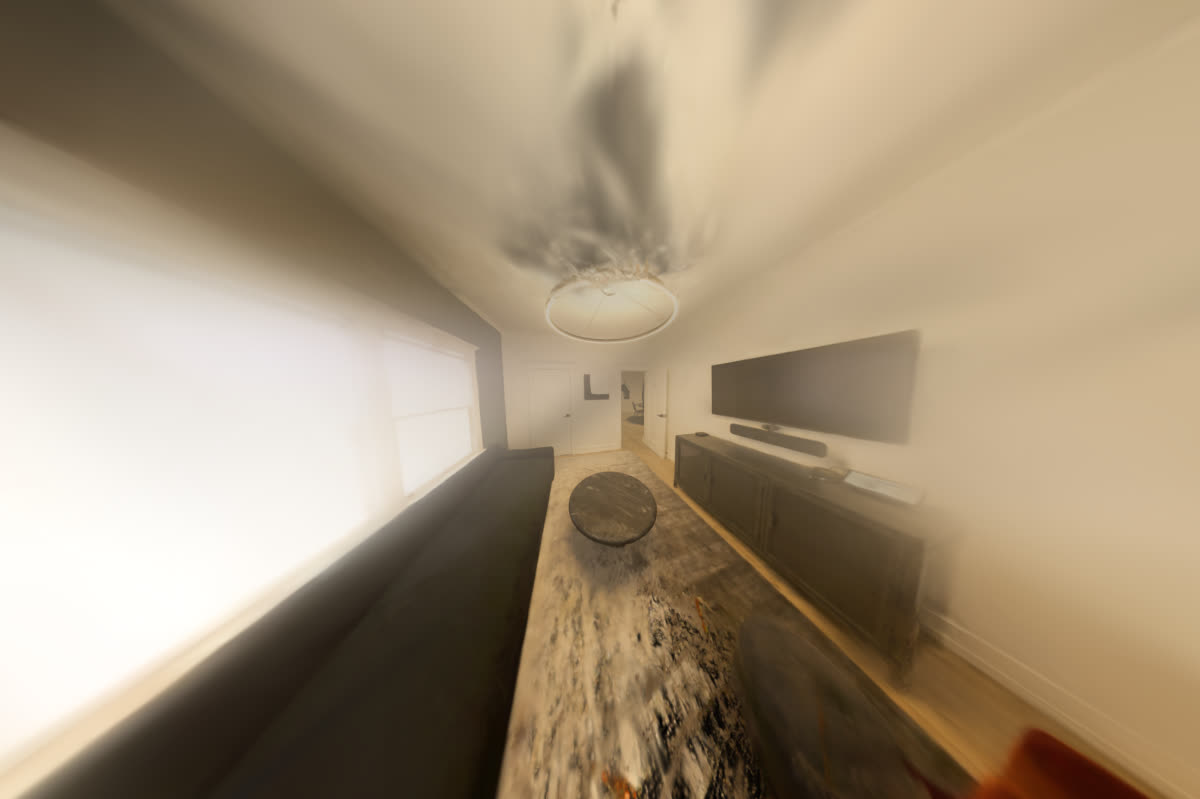} &
        \includegraphics[height=0.14\textwidth]{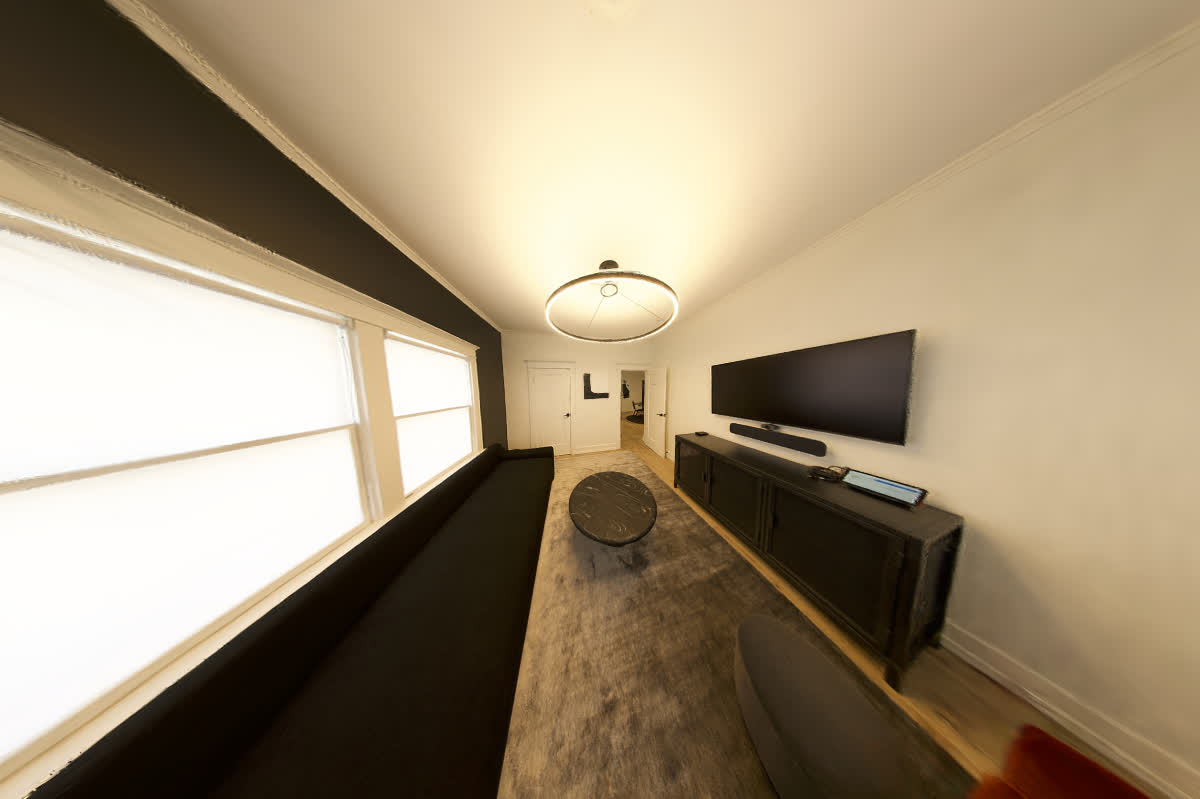} &
        % \raisebox{0.8\height}{\hspace{-0.5cm}\rotatebox{90}{{Room2}}} &
        \includegraphics[height=0.14\textwidth]{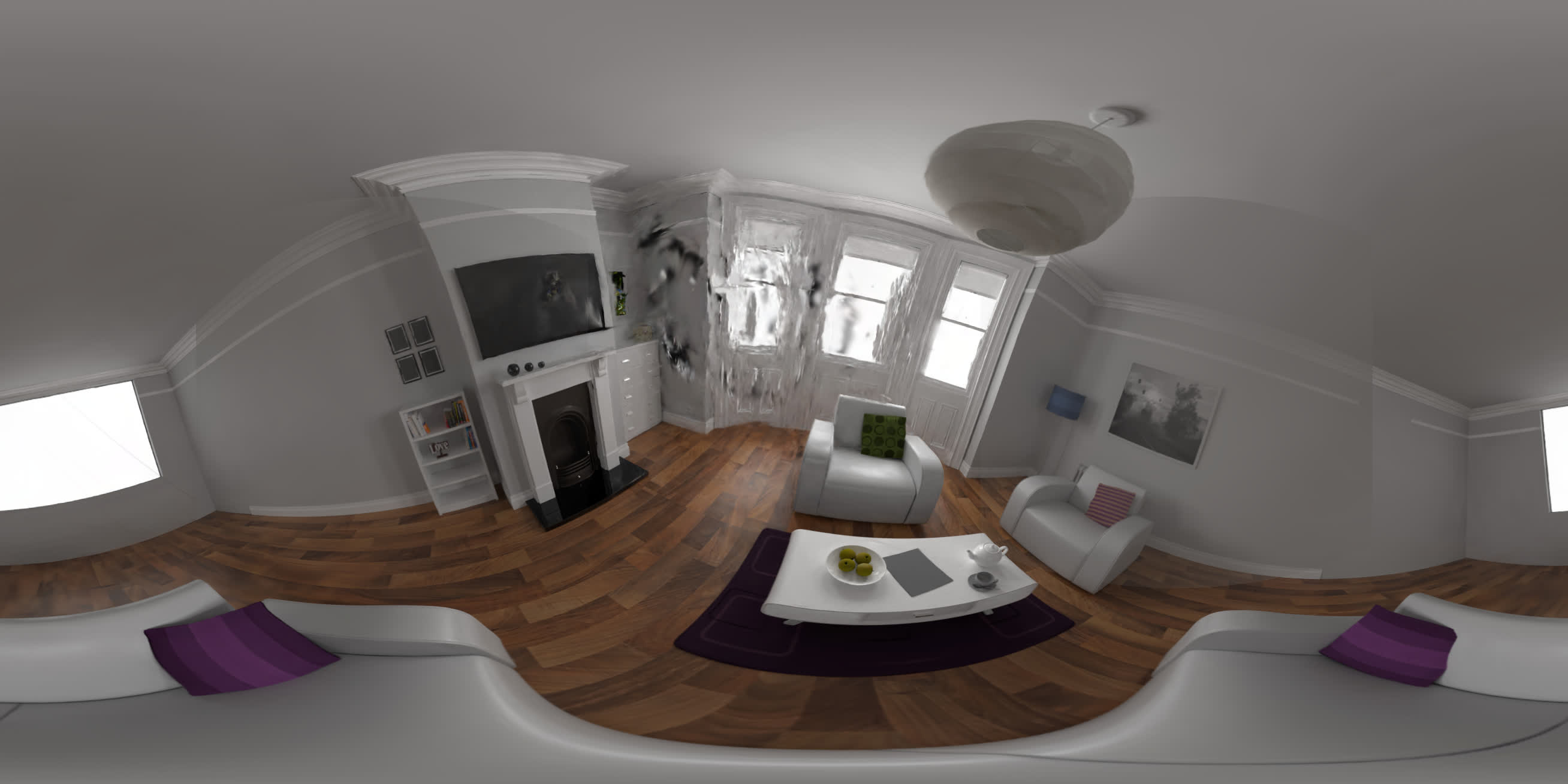} &
        \includegraphics[height=0.14\textwidth]{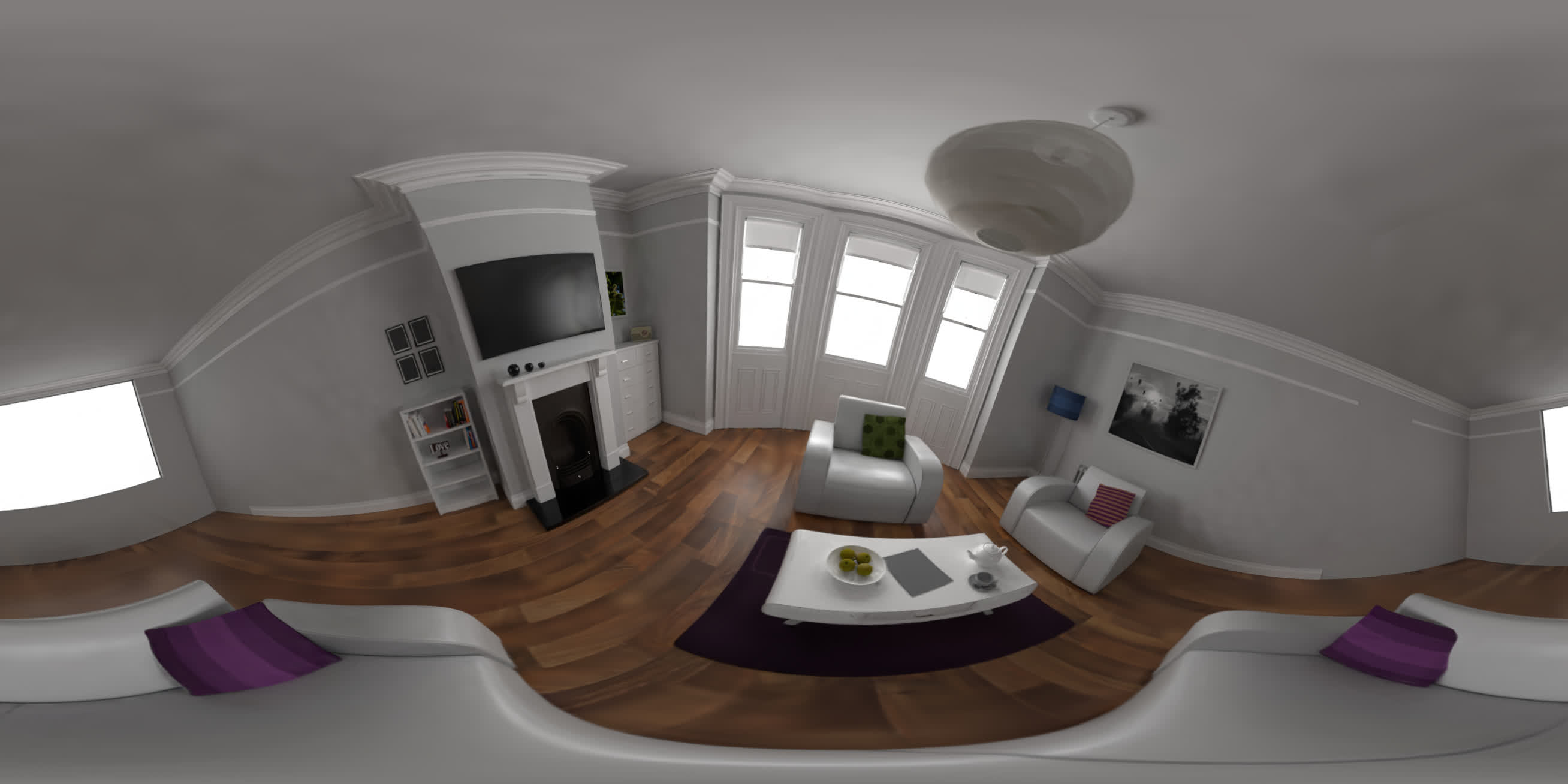} \\

        \multicolumn{1}{c}{(a) 3DGS w/ COLMAP} & \multicolumn{1}{c}{(b) Ours } & \multicolumn{1}{c}{(c) 3DGS w/ Regular Camera} & \multicolumn{1}{c}{(d) Ours}
        \\
    \end{tabular}
    % \vspace{-2mm}
    \caption{\textbf{Qualitative Comparison of Reconstruction Coverage}. We evaluate our method on real-world scenes where (a) 3DGS~\cite{kerbl20233d} relies on an SfM method like COLMAP~\cite{schoenberger2016sfm}, which crops the highly distorted periphery when the FOV is large. (b) Our method recovers a wider region with fewer artifacts at the center. We also visualize reconstructions using panorama views of both (c) 3DGS~\cite{kerbl20233d} and (d) our method. 3DGS~\cite{kerbl20233d} only supports perspective inputs captured with regular-FOV cameras, whereas our method can be directly applied to 180\si{\degree} inputs, demonstrating better coverage even with fewer training views, as shown in~\cref{tab:quantitative_large_small}.}
    \label{fig:largefov_vs_smallfov_netflix}
    % \vspace{-4mm}

\end{figure*}
}
\subsection{Large FOV Reconstruction with Few Captures} 
\label{sec:eval_cubemap}
By enabling the use of large-FOV views, our method can reconstruct scenes with fewer images than would be needed for narrower perspective views.
We evaluate our method on 150\si{\degree} real-world captures and 180\si{\degree} synthetic scenes. For real-world scenes, we follow the conventional paradigm of using COLMAP to first resample the images to perspective—including peripheral cropping performed by COLMAP~\cite{schoenberger2016sfm}—before reconstruction. For synthetic scenes, we can flexibly modify the camera model and render images with a regular 90\si{\degree} FOV as input for 3DGS~\cite{kerbl20233d}. Since our method supports perspective rendering after training, we compare it against the baseline using a set of perspective hold-out cameras for a fair evaluation. Quantitative evaluation is conducted only on synthetic data, as obtaining ground truth perspective images for real-world captures is not feasible.

{

\begin{table}[t]
  \centering
  \scalebox{1}{
    \begin{tabular}{ccccc}
        \toprule
        Method & Num & SSIM & PSNR & LPIPS \\
        \midrule
        3DGS~\cite{kerbl20233d} & 200 & 0.654 & 19.08 & 0.332 \\
        \midrule
        \multirow{4}{*}{Ours} 
        & 100 & \textbf{0.800} & \textbf{29.01} & \textbf{0.231} \\
        & 50 & 0.735 & 26.26 & 0.267 \\
        & 25 & 0.709 & 24.59 & 0.292 \\
        & 10 & 0.615 & 22.77 & 0.356 \\
        \bottomrule
    \end{tabular}
  }
  % \vspace{-2mm}
    \caption{\textbf{Evaluation on Mitsuba Scenes}. We compare our method with Vanilla 3DGS~\cite{kerbl20233d} to demonstrate the advantages of using wide-angle cameras over regular-FOV cameras. The testing views are rendered in perspective. Our method achieves better performance and coverage, even with fewer captures.}
    \label{tab:quantitative_large_small}%
  % \vspace{-5mm}
\end{table}
}

We report quantitative results in~\cref{tab:quantitative_large_small}, where our method outperforms the baseline even with far fewer input views, down to 10-15\%. 
The trajectories of Room 1 and Garden are similar, both focusing inward, while Room 2 and Studio involve a walk-around motion covering the entire space. As a result, we observe large incomplete regions in the first row of~\cref{fig:largefov_vs_smallfov_netflix} for the baseline, whereas our method achieves complete reconstruction. The walk-around scenes shown in the last row of~\cref{fig:largefov_vs_smallfov_netflix} exhibit noticeable artifacts and floating elements, primarily due to small-FOV captures or cropping in COLMAP.
These experiments demonstrate the efficiency of our approach compared to regular-FOV cameras, achieving high-quality reconstruction with significantly fewer captures, as shown in~\cref{tab:quantitative_large_small}.

\begin{table}[t]
  \centering
  \scalebox{1}{
    \begin{tabular}{ccccc}
    \toprule
    Explicit Grid & \multicolumn{1}{c}{iResNet} & PSNR & SSIM & LPIPS \\
    \midrule
    \xmark & \cmark & \multicolumn{3}{c}{Out-of-Memory}\\
    \xmark & \xmark & 14.19 & 0.421 & 0.561\\
    % & Explicit Grid & 15.67 & 0.488 & 0.496\\
    \cmark & \xmark & 19.79 & 0.569 & 0.309\\
    \cmark & \cmark & \textbf{23.67} & \textbf{0.777} & \textbf{0.151}\\
    \bottomrule
    \end{tabular}
  }
  % \vspace{-1mm}
  \caption{\textbf{Ablations of Hybrid Field.} We conduct ablation studies on our full method, including an explicit grid initialized with COLMAP's traditional polynomial distortion parameters, to demonstrate the necessity of adopting a hybrid neural distortion field representation in FisheyeNeRF~\cite{jeong2021self}. In the results, ``-'' indicates cases that are computationally infeasible to evaluate.}
  \label{tab:ablation_hybrid}
  % \vspace{-3mm}
\end{table}

{
\begin{table}[t]

  \centering
  \scalebox{1}{
    \begin{tabular}{cccc}
    \toprule
    Projection & PSNR & SSIM & LPIPS \\
    \midrule
    % \xmark & \cmark & \xmark & 19.08 & 0.654 & 0.332 \\
    Single Plane & 24.10 & 0.676 & 0.312 \\
    Cubemap & \textbf{29.01} & \textbf{0.792} & \textbf{0.253} \\
    \bottomrule
    \end{tabular}
  }
  % \vspace{-2mm}
  \caption{\textbf{Ablation of Cubemap}. We evaluate single-plane and cubemap projections using the same hybrid field on Mitsuba scenes, where the FOV is close to 180\si{\degree}, causing significant degradation in the single-plane projection.}
  \label{tab:ablation_cubemap_single}
  % \vspace{-5mm}
\end{table}
}

\subsection{Ablation Studies}
\label{sec:ablation}
% In this section, we will evaluate the effectiveness of hybrid fields and cubemap resampling.

\paragraph{Hybrid Field.} To verify the effectiveness of each module in our hybrid field, we isolate individual components (either iResNet or the explicit grid) and evaluate their performance separately. We report quantitative results in~\cref{tab:ablation_hybrid} on FisheyeNeRF~\cite{jeong2021self} scenes, showing that neither module alone achieves optimal performance compared to our full method. To test the grid alone, we use an explicit grid of learnable displacement vectors. This strategy provides only a slight improvement, primarily because the optimization of the explicit grid is prone to overfitting and can become trapped in local minima. We visualize the difference between the explicit grid and our hybrid field distortion flow in~\cref{fig:pipeline}. We cannot report results for iResNet without the grid, as explained in the last paragraph of \textbf{Invertible Residual Networks} in~\cref{sec:iresnet}. Our full method (iResNet + control grid) achieves the best performance by combining the expressiveness of iResNet with the computational efficiency of the explicit grid. This ablation verifies the necessity of our hybrid representation for modeling distortion.

% \vspace{-1.2em}
\paragraph{Cubemap Resampling.} To validate the necessity of cubemap resampling for wide-FOV reconstruction, we compare results obtained using our hybrid distortion field with single-plane projection versus cubemap resampling. 
We evaluate the performance of these two methods in the Mitsuba synthetic scenes.
As presented in~\cref{tab:ablation_cubemap_single}, using cubemap projections produces significantly better results when reconstructing images captured by 180\si{\degree} fisheye lenses.
Additional qualitative comparisons between single-plane and cubemap projections are provided in the supplementary.

\section{Discussion}
\label{sec:discussions}

This paper presents a method for optimizing 3D Gaussian representations while self-calibrating camera parameters and lens distortion. 
Our approach enables the use of large field-of-view captures to achieve efficient and high-quality reconstruction without cumbersome pre-calibration. Even with fewer input captures, our method maintains comprehensive scene coverage and reconstruction quality.
% We compare our method with state-of-the-art techniques and demonstrate superior performance on datasets containing large FoV captures.

\vspace{-1em}
\paragraph{Limitations and Future directions.} 
This work does not account for entrance pupil shift in fisheye lenses, which affects near-field scenes by causing splat misalignment but is negligible for distant scenes. Vignetting and chromatic aberration were also not modeled, leaving room for future extensions. Additionally, intensity discontinuities arise at cubemap face boundaries due to variations in 2D covariances. A simple fix involves switching from depth sorting to distance sorting, while a more refined solution could involve projecting covariances orthogonally to the viewing ray. More discussions can be found in the supplementary.

% \vspace{-1em}
\paragraph{Societal Impact.} 
This research can benefit industries that depend on 3D reconstruction, such as film production and virtual reality. 
However, a potential downside is the environmental impact associated with the increased computational resources required for model optimization.

\section{Acknowledgement}
This work was supported in part by the National Science Foundation under grant 2212084 and the Vannevar Bush Faculty Fellowship.
We want to express gratitude to Xichen Pan, Xiangzhi Tong, Julien Philip, Li Ma, Hansheng Chen, Jan Ackermann, and Eric Chen for their discussion and suggestions on this paper. 

\clearpage
% \input{sec/6_conclusion}

% \clearpage
{
    \small
    \bibliographystyle{ieeenat_fullname}
    \bibliography{main}
}

\clearpage
\maketitlesupplementary
\setcounter{section}{0}

{
% Format settings for TOC
\renewcommand{\cftsecafterpnum}{\vskip15pt}
\setlength{\cftsecindent}{1em}
\setlength{\cftbeforesecskip}{0.5em}

% First exclude all sections from TOC
\setcounter{tocdepth}{-1}

% Then add only supplementary sections to TOC
\begingroup
\renewcommand{\contentsname}{Content}
\tableofcontents
\endgroup

% After the TOC, manually add entries for just supplementary sections
\addtocontents{toc}{\protect\setcounter{tocdepth}{1}}
}

\section{Supplementary Video}
We provide a video, ``supp\_video.mp4," to better compare our method with baselines. Our video is organized into three parts.

The first part presents a comparison between our method and baselines on the FisheyeNeRF dataset~\cite{jeong2021self} across three scenes. 
Vanilla 3DGS~\cite{kerbl20233d} completely fails to reconstruct the scenes because lens distortion is not accounted for. 
Fisheye-GS~\cite{liao2024fisheye} adopts a parametric model, but the peripheral regions produce blurry results, as highlighted by the red box in the corners. 
In contrast, our method achieves clean and sharp reconstructions.

The second part of the video shows reconstruction results using our method on walk-around captures, including both synthetic and real-world scenes. 
Our approach achieves sharp and clean renderings upon completing the reconstruction. 

The third part of the video compares visualizations of our method with a conventional reconstruction pipeline that either uses narrow-FOV perspective images or undistorted images from COLMAP~\cite{schoenberger2016sfm} as input. 
Our method successfully reconstructs larger regions, particularly for scenes captured with large 180\si{\degree} cameras. Besides, we render videos in fisheye views for each scene.

Additionally, we provide ``opt\_pose.mp4," which illustrates the camera changes during optimization, and ``fisheye-gs\_failure.mp4," which demonstrates failure cases of Fisheye-GS~\cite{liao2024fisheye} in extremely large-FOV settings.

We strongly encourage all reviewers to watch the provided video for a more comprehensive visual understanding of our results.

\section{Optimization of Camera Parameters}
In this section, we first derive the gradients for all camera parameters during training in~\cref{sec:2.1}. We then demonstrate the effectiveness of the joint optimization of distortion alongside extrinsic and intrinsic parameters in~\cref{sec:2.2}. Finally, we evaluate our camera optimization on synthetic NeRF~\cite{mildenhall2021nerf} scenes with significant noise in~\cref{sec:2.3}.

\subsection{Derivation of Gradient}
\label{sec:2.1}
We first derive how to compute the gradient for a pinhole camera and then extend the derivation to account for distortion.

As defined above, the gradient of camera parameters can be written as:
\begin{align}
\label{eq:simpify-grad}
\fpp{\lL}{\Theta} = \sum_{j=1}^{|G|} \fpp{\lL}{C_j}\fpp{C_j}{\Theta} + \fpp{\lL}{\mu^{2D}_j}\fpp{\mu^{2D}_j}{\Theta} + \fpp{\lL}{\Sigma^{2D}_j}\fpp{\Sigma^{2D}_j}{\Theta}.
\end{align}

For the color term, we first define the input of the spherical harmonics stored by each Gaussian as the view direction. The 3D location of a Gaussian in world coordinates is $\textbf{X}_w=[x, y, z]^T$, and the camera center is $\textbf{C}=[C_x, C_y, C_z]^T$:
\begin{align}
    \omega_j = [x - C_x, y - C_y, z - C_z]^T.
\end{align}
Then, the derivative with respect to the camera center is:
\begin{align}
    \sum_{j=1}^{|G|} \fpp{\lL}{C_j}\fpp{C_j}{\Theta}&=\sum_{j=1}^{|G|} \fpp{\lL}{C_j}\fpp{C_j}{\omega_j}\fpp{\omega_j}{\textbf{C}}\\
    &=\sum_{j=1}^{|G|} \fpp{\lL}{C_j}\fpp{C_j}{\omega_j}(-\fpp{\omega_j}{\textbf{X}}).
\end{align}

For now, we only compute the derivative with respect to the camera center. However, the gradient can be easily propagated back since the camera center corresponds to the translation vector of the inverse of the world-to-camera matrix.

Next, we derive the gradient from the projected 2D Gaussian position $\mu_j^{2D}$. For simplicity, we integrate the intrinsic and extrinsic parameters into a projection matrix $\textbf{P}$ since we do not yet consider distortion. The projection from a 3D Gaussian in world coordinates is defined by:
\begin{align}
    \relax[\Tilde{x}, \Tilde{y}, \Tilde{z}, \Tilde{\omega}]^T=\textbf{P}[x,y,z,1]^T.
\end{align}

We apply a projective transformation:
\begin{equation}
\begin{aligned}
    f(x,y,z)=x'=\frac{\Tilde{x}}{\Tilde{\omega}}=\frac{p_0x+p_4y+p_8z+p_{12}}{p_3x+p_7y+p_{11}z+p_{15}}, \\ 
    g(x,y,z)=y'=\frac{\Tilde{y}}{\Tilde{\omega}}=\frac{p_1x+p_5y+p_9z+p_{13}}{p_3x+p_7y+p_{11}z+p_{15}},\\
    z'=\frac{\Tilde{z}}{\Tilde{\omega}}=\frac{p_2x+p_6y+p_{10}z+p_{14}}{p_3x+p_7y+p_{11}z+p_{15}},
\end{aligned}
\end{equation}
where
\begin{align}
    \textbf{P}=\begin{bmatrix}
    p_0 & p_4 & p_8 & p_{12} \\
    p_1 & p_5 & p_9 & p_{13} \\
    p_2 & p_6 & p_{10} & p_{14} \\
    p_3 & p_7 & p_{11} & p_{15} \\
    \end{bmatrix}.
\end{align}

Then, we project from NDC to screen space to obtain the 2D location $\mu^{2D}=(u, v)$:
\begin{align}
    u=\frac{(f(x,y,z)+1)W-1}{2},\\
    v=\frac{(g(x,y,z)+1)H-1}{2}.
\end{align}

The second term can be represented as:
\begin{align}
    \fpp{\lL}{\mu^{2D}_j}\fpp{\mu^{2D}_j}{\Theta}=\fpp{\lL}{\mu^{2D}_j}\fpp{\mu^{2D}_j}{[x', y']^T}\fpp{[x', y']^T}{\textbf{P}}.
\end{align}

We compute the gradients for each $p_i$ in the projection matrix $\textbf{P}$:
\begin{equation}
    \begin{aligned}
        \frac{\partial L}{\partial p_0} =& \frac{\partial L}{\partial u}\frac{W}{2}\frac{x}{\Tilde{\omega}},\\
        \frac{\partial L}{\partial p_4} =& \frac{\partial L}{\partial u}\frac{W}{2}\frac{y}{\Tilde{\omega}},\\
        \frac{\partial L}{\partial p_8} =& \frac{\partial L}{\partial u}\frac{W}{2}\frac{z}{\Tilde{\omega}},\\
        \frac{\partial L}{\partial p_{12}} =& \frac{\partial L}{\partial u}\frac{W}{2}\frac{1}{\Tilde{\omega}},\\        
        \frac{\partial L}{\partial p_1} =& \frac{\partial L}{\partial v}\frac{H}{2}\frac{x}{\Tilde{\omega}},\\
        \frac{\partial L}{\partial p_5} =& \frac{\partial L}{\partial v}\frac{H}{2}\frac{y}{\Tilde{\omega}},\\
        \frac{\partial L}{\partial p_9} =& \frac{\partial L}{\partial v}\frac{H}{2}\frac{z}{\Tilde{\omega}},\\
        \frac{\partial L}{\partial p_{13}} =& \frac{\partial L}{\partial v}\frac{H}{2}\frac{1}{\Tilde{\omega}},
    \end{aligned}
\end{equation}
and
\begin{equation}
    \begin{aligned}
        \frac{\partial L}{\partial p_3}=&\frac{\partial L}{\partial u}(-x')\left(\frac{W}{2}\frac{x}{\Tilde{\omega}}\right)+\frac{\partial L}{\partial v}(-y')\left(\frac{H}{2}\frac{x}{\Tilde{\omega}}\right),\\
        \frac{\partial L}{\partial p_7}=&\frac{\partial L}{\partial u}\frac{-\Tilde{x}}{\Tilde{\omega}}\left(\frac{W}{2}\frac{y}{\Tilde{\omega}}\right)+\frac{\partial L}{\partial v}\frac{-\Tilde{y}}{\Tilde{\omega}}\left(\frac{H}{2}\frac{y}{\Tilde{\omega}}\right),\\
        \frac{\partial L}{\partial p_{11}}=& \frac{\partial L}{\partial u}\frac{-\Tilde{x}}{\Tilde{\omega}}\left(\frac{W}{2}\frac{z}{\Tilde{\omega}}\right)+\frac{\partial L}{\partial v}\frac{-\Tilde{y}}{\Tilde{\omega}}\left(\frac{H}{2}\frac{z}{\Tilde{\omega}}\right),\\
        \frac{\partial L}{\partial p_{15}}=&\frac{\partial L}{\partial u}\frac{W}{2}\frac{-\Tilde{x}}{\Tilde{\omega}^2}+\frac{\partial L}{\partial v}\frac{H}{2}\frac{-\Tilde{y}}{\Tilde{\omega}^2}.
    \end{aligned}
\end{equation}

The gradient flow back to intrinsic and extrinsic parameters can be easily computed since $\textbf{P}=K[R|t]$.

Finally, we compute the last term. The 2D covariance $\Sigma^{2D}$ depends only on the view matrix, so instead of using $\textbf{P}$, we use only the view matrix (\ie, the world-to-camera matrix) $\textbf{V}$, and the transformed 3D location in camera coordinates is $\textbf{X}_c=[x_c, y_c, z_c]^T$:
\begin{equation}
    \begin{aligned}
        \relax[x_c, y_c, z_c]^T&=\textbf{V}[x,y,z,1]^T\\
        &=\begin{bmatrix}
        v_{0} & v_{4} & v_{8} & v_{12}\\
        v_{1} & v_{5} & v_{9} & v_{13}\\
        v_{2} & v_{6} & v_{10} & v_{14}\\
        \end{bmatrix}
        \cdot\begin{bmatrix}
        x\\
        y\\
        z\\
        1
        \end{bmatrix}.
    \end{aligned}
\end{equation}

We compute the 2D covariance as follows, given known focal lengths $f_x$ and $f_y$:

\begin{equation}
    \begin{aligned}
        J &= \begin{bmatrix}
        J_{00} & J_{01} & J_{02}\\
        J_{10} & J_{11} & J_{12}\\
        J_{20} & J_{21} & J_{22}
        \end{bmatrix} = \begin{bmatrix}
        \frac{f_x}{z_c} & 0 & \frac{-f_x\cdot x_c}{z_c^2}\\
        0 & \frac{f_y}{z_c} & \frac{-f_y\cdot y_c}{z_c^2}\\
        0 & 0 & 0
        \end{bmatrix},\\
         \textbf{W} &= \begin{bmatrix}
        v_{0} & v_{4} & v_{8} \\
        v_{1} & v_{5} & v_{9} \\
        v_{2} & v_{6} & v_{10}
        \end{bmatrix}= \begin{bmatrix}
        w_{00} & w_{01} & w_{02} \\
        w_{10} & w_{11} & w_{12} \\
        w_{20} & w_{21} & w_{22}
        \end{bmatrix},\\ 
        \textbf{T}&=\textbf{W}\cdot J=\begin{bmatrix}
        T_{00} & T_{10} & T_{20} \\
        T_{01} & T_{11} & T_{21} \\
        T_{02} & T_{12} & T_{22}
        \end{bmatrix},\\
        \Sigma^{2D}&=\textbf{T}^T\cdot\Sigma\cdot \textbf{T}=\textbf{T}^T\cdot\begin{bmatrix}
        c_{0} & c_{1} & c_{2} \\
        c_{1} & c_{3} & c_{4} \\
        c_{2} & c_{4} & c_{5}
        \end{bmatrix}\cdot \textbf{T},\\
        a&=\Sigma^{2D}[0][0],\quad b=\Sigma^{2D}[0][1],\quad c=\Sigma^{2D}[1][1],\\
        \textbf{Conic}&=\begin{bmatrix}
            a & b\\
            b & c
        \end{bmatrix}^{-1}=\begin{bmatrix}
            \frac{c}{ac-b^2} & \frac{-b}{ac-b^2}\\
            \frac{-b}{ac-b^2} & \frac{a}{ac-b^2}
        \end{bmatrix}.
    \end{aligned}
\end{equation}

Since we only extract the upper three elements of $\Sigma^{2D}$, we do not compute gradients for $T_{20}$, $T_{21}$, and $T_{22}$.

\begin{equation}
    \begin{aligned}
        \frac{\partial L}{\partial v_0}=&\frac{\partial L}{\partial T_{00}}\frac{f_x}{z_c},
        \quad \frac{\partial L}{\partial v_1}=\frac{\partial L}{\partial T_{01}}\frac{f_x}{z_c},\\
        \frac{\partial L}{\partial v_4}=&\frac{\partial L}{\partial T_{10}}\frac{f_y}{z_c},
        \quad \frac{\partial L}{\partial v_5}=\frac{\partial L}{\partial T_{11}}\frac{f_y}{z_c},\\
        \frac{\partial L}{\partial v_2}=&\frac{-(v_0\frac{\partial L}{\partial T_{00}}+v_1\frac{\partial L}{\partial T_{01}})\cdot x\cdot f_x}{z_c^2}\\
        &+\frac{-(v_4\frac{\partial L}{\partial T_{10}}+v_5\frac{\partial L}{\partial T_{11}}+v_6\frac{\partial L}{\partial T_{12}})\cdot x\cdot f_y}{z_c^2}\\
        &+\frac{(z_c-v_2x)f_x\frac{\partial L}{\partial T_{02}}}{z_c^2},\\
        \frac{\partial L}{\partial v_6}=&\frac{-(v_0\frac{\partial L}{\partial T_{00}}+v_1\frac{\partial L}{\partial T_{01}}+v_2\frac{\partial L}{\partial T_{02}})\cdot y\cdot f_x}{z_c^2}\\
        &+\frac{-(v_4\frac{\partial L}{\partial T_{10}}+v_5\frac{\partial L}{\partial T_{11}})\cdot y\cdot f_y}{z_c^2}\\
        &+\frac{(z_c-v_6y)f_y\frac{\partial L}{\partial T_{12}}}{z_c^2},\\ 
        \frac{\partial L}{\partial v_{10}}=&\frac{-((v_0\frac{\partial L}{\partial T_{00}}+v_1\frac{\partial L}{\partial T_{01}}+v_2\frac{\partial L}{\partial T_{02}})f_x}{z_c^2}\\
        &+\frac{(v_4\frac{\partial L}{\partial T_{10}}+v_5\frac{\partial L}{\partial T_{11}}+v_6\frac{\partial L}{\partial T_{12}})f_y)z}{z_c^2},\\
        \frac{\partial L}{\partial v_{14}}=&\frac{-((v_0\frac{\partial L}{\partial T_{00}}+v_1\frac{\partial L}{\partial T_{01}}+v_2\frac{\partial L}{\partial T_{02}})f_x}{z_c^2}\\
        &+\frac{(v_4\frac{\partial L}{\partial T_{10}}+v_5\frac{\partial L}{\partial T_{11}}+v_6\frac{\partial L}{\partial T_{12}})f_y}{z_c^2}.
    \end{aligned}
\end{equation}

To account for distortion, we decompose the projection matrix into two separate operations. Following the definitions above, we apply an invertible ResNet in between:

\begin{align}
    \relax[\Tilde{x}, \Tilde{y}, \Tilde{z}, \Tilde{\omega}]^T={K\homo{z_c\cdot\homo{\dD_\theta\paren{\proj{x_c, y_c, z_c}}}}}.
\end{align}

For the projection of the Gaussian covariance $\Sigma$, we compute a new Jacobian matrix $J$ using the distorted $\textbf{X}_c$:

\begin{align}
    [x_c, y_c, z_c]^T=z_c\cdot\homo{\dD_\theta\paren{\proj{x_c, y_c, z_c}}}.
\end{align}

Both processes are differentiable, allowing us to compute the intermediate Jacobian for $\dD_\theta$ to update the invertible ResNet.

\begin{figure}[t]
    \centering
    \setlength{\tabcolsep}{1pt} % Adjust space between columns if needed
    \begin{tabular}{cc} % 4 columns
        \subcaptionbox{w/o Cameras Optimization\label{fig:left}}{
            \includegraphics[width=0.22\textwidth]{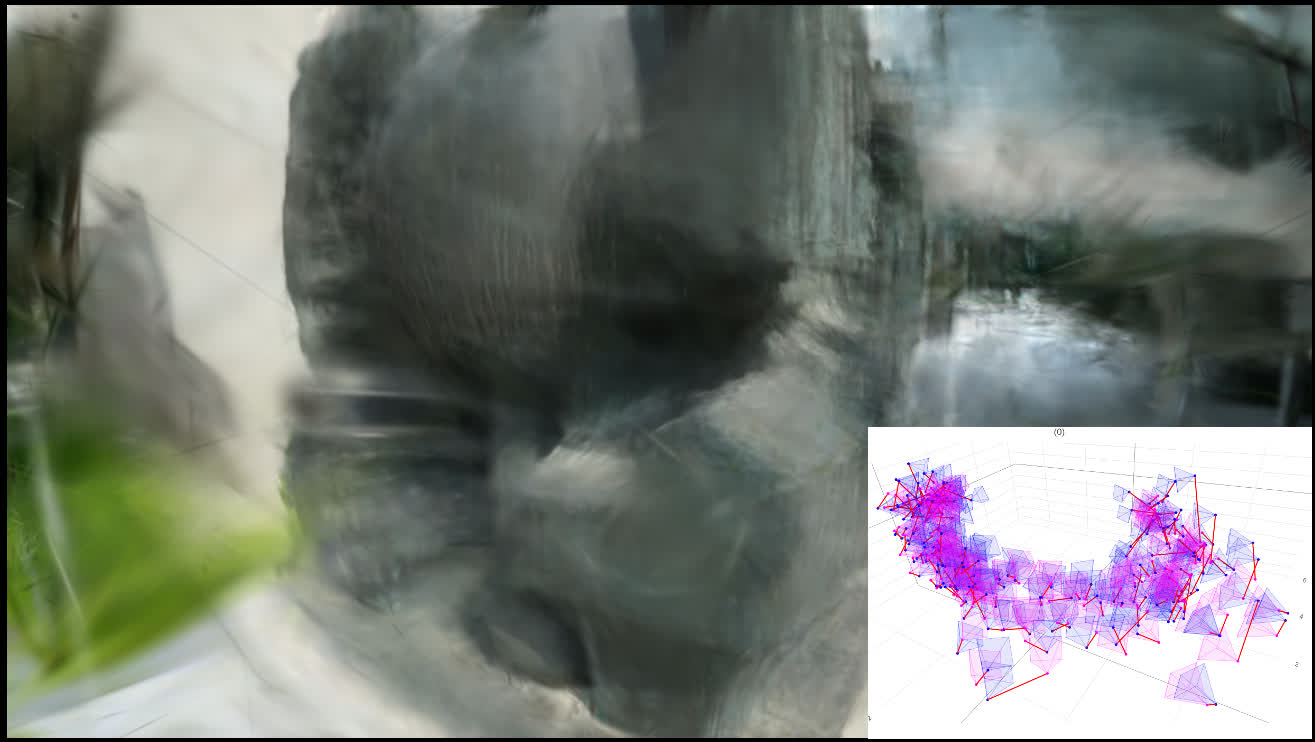}
        } &

        % Fourth image with label (c)
        \subcaptionbox{w/ Cameras Optimization\label{fig:right}}{
            \includegraphics[width=0.22\textwidth]{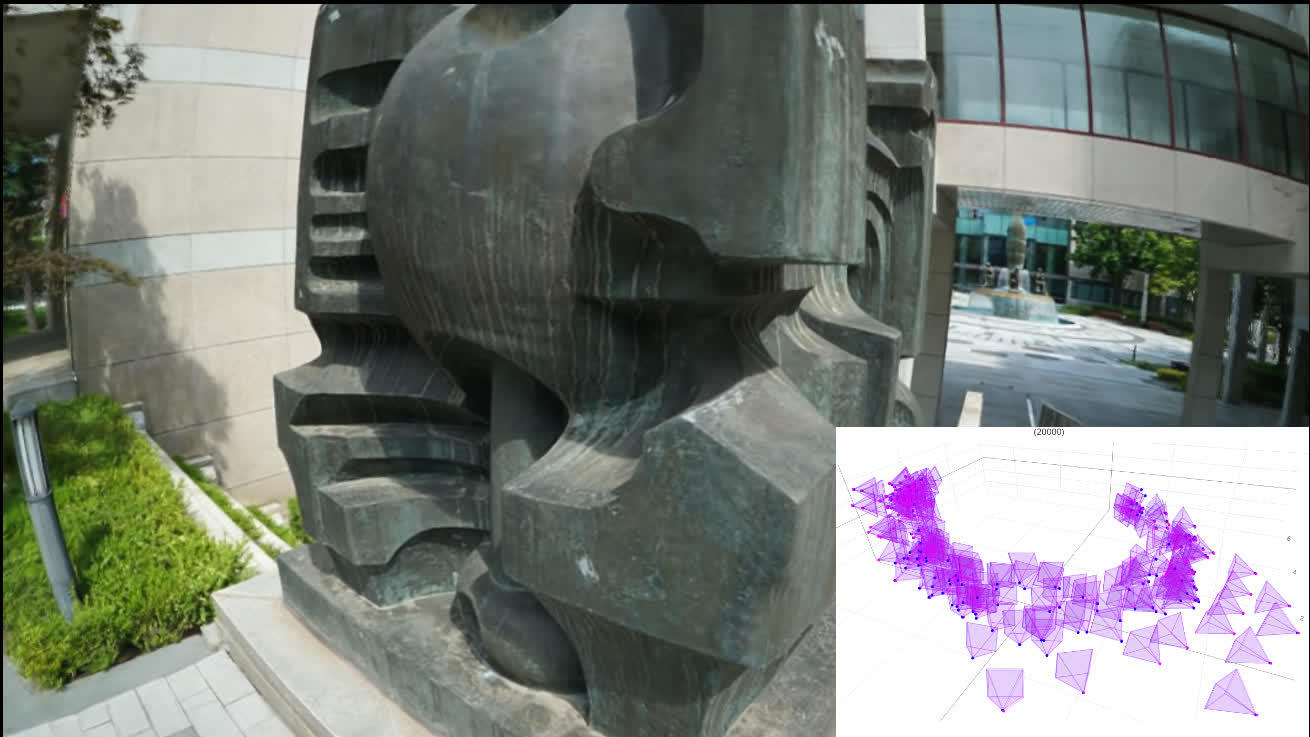}
        }
    \end{tabular}
    \vspace{-0.5em}
    \caption{\textbf{Camera Parameters Optimization}. We initialize noisy cameras from COLMAP~\cite{schoenberger2016sfm}. (a) Modeling fisheye distortion without optimizing camera parameters, while (b) jointly optimizing both in a self-calibration manner. Our full model can recover accurate lens distortion and camera parameters simultaneously.}
    \label{fig:pose_distortion}
    \vspace{-1.5em}
\end{figure}
\begin{figure}
    \centering
    \includegraphics[width=0.45\textwidth]{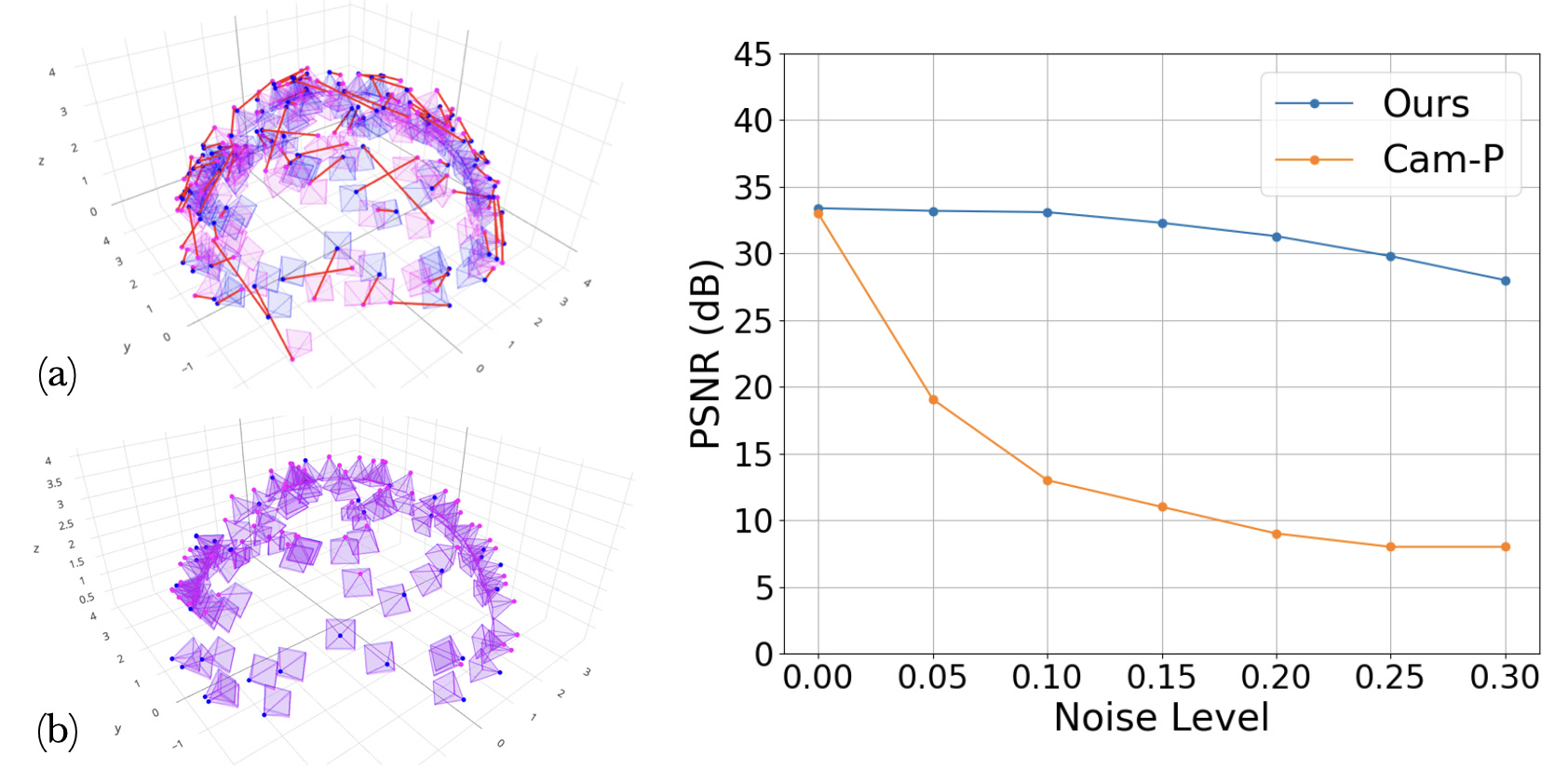}
     \caption{Visual comparison of (a) the initial perturbed ($s=0.15$) and GT poses and (b) optimized camera poses in the Lego scene. The chart demonstrates the different level of perturbations and the effectiveness of our optimization. Our method successfully recovers accurate camera frames. }
    \label{fig:robust_to_noise}   
\end{figure}

{
\begin{figure*}[th]
    \centering
    \setlength{\tabcolsep}{1pt} % Adjust space between columns if needed
    \begin{tabular}{cccc} % 4 columns (Vertical Caption | Image Set 1 | Image Set 2 | Image Set 3)

        \raisebox{1\height}{\hspace{-0.5cm}\rotatebox{90}{{Chairs}}} &
        \includegraphics[width=0.3\textwidth]{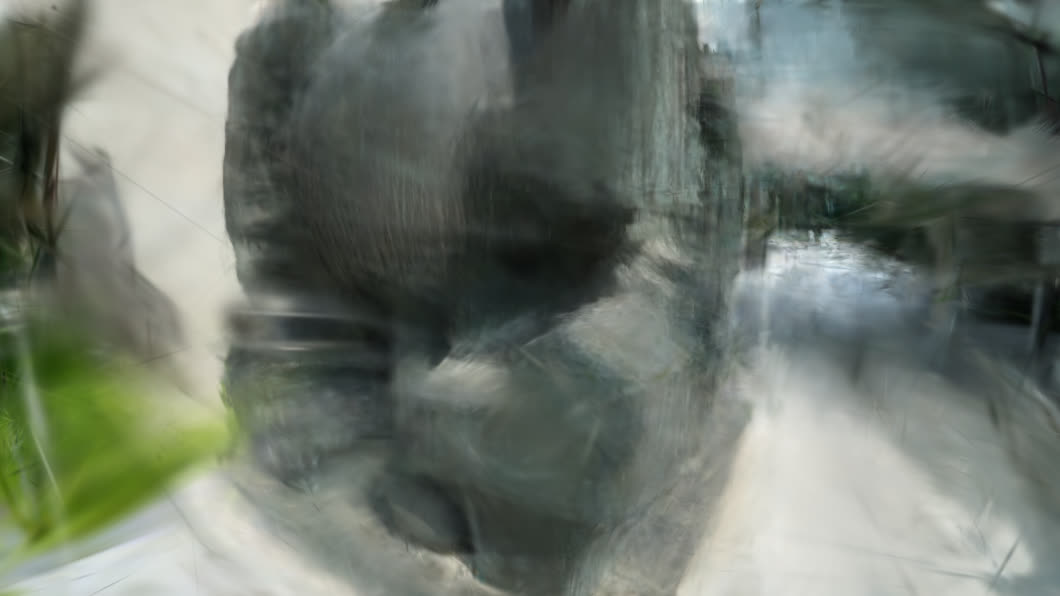} &
        \includegraphics[width=0.3\textwidth]{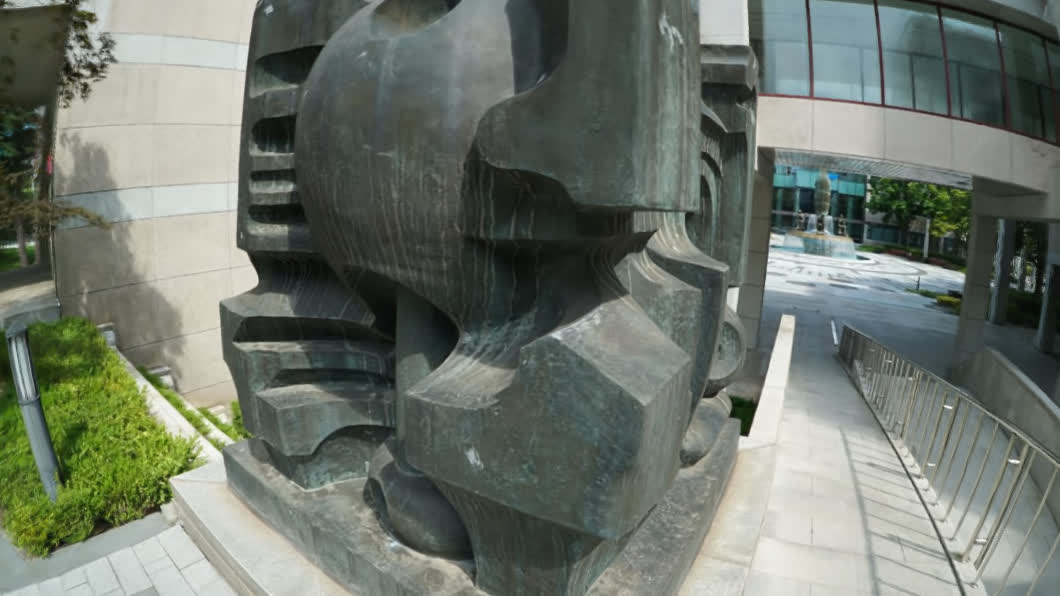} &
        \includegraphics[width=0.3\textwidth]{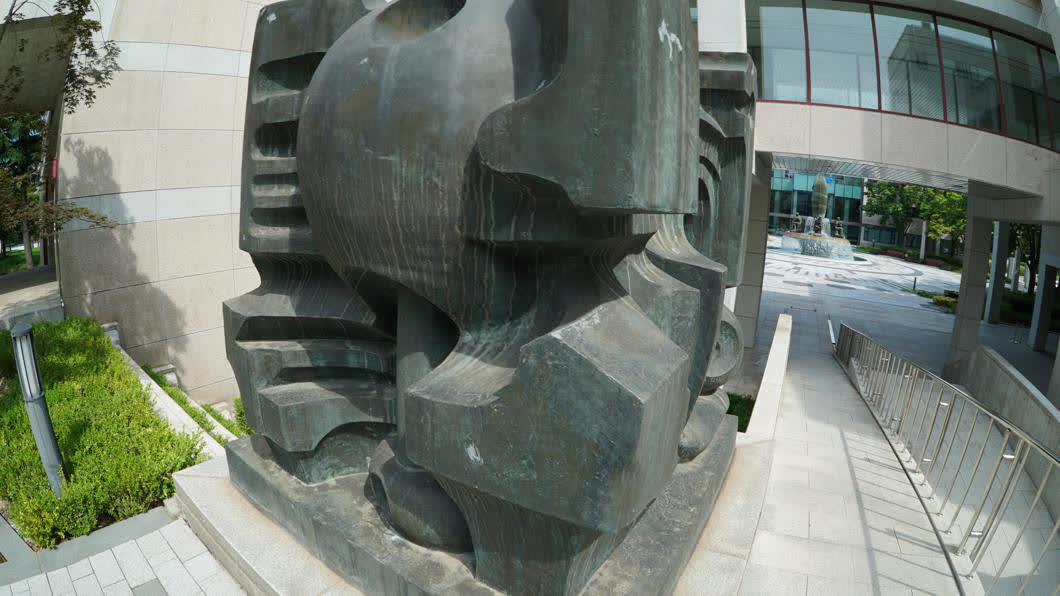} \\

        \raisebox{1.4\height}{\hspace{-0.5cm}\rotatebox{90}{Rock}} &
        \includegraphics[width=0.3\textwidth]{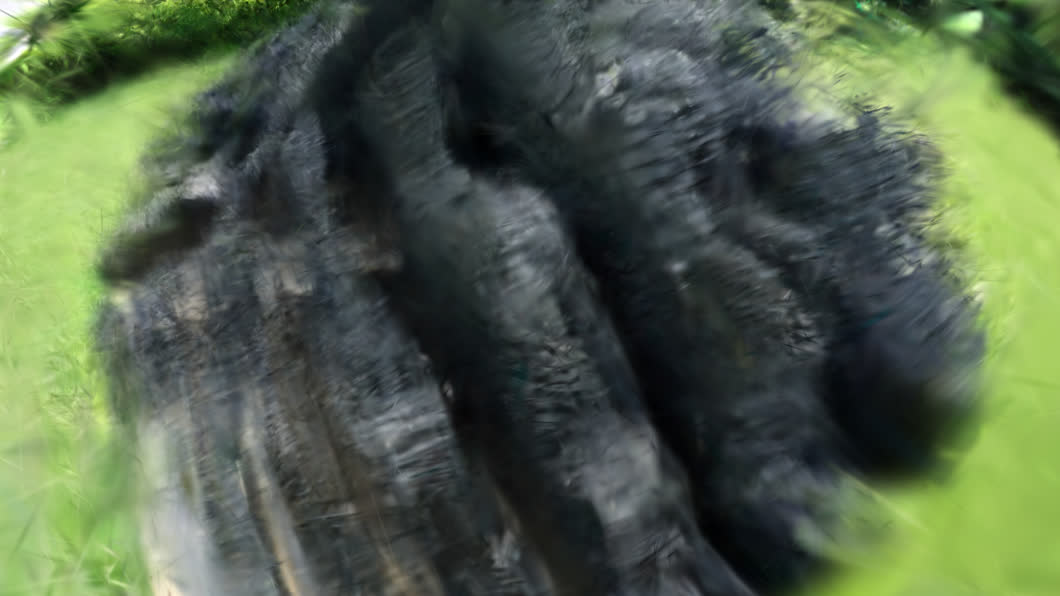} &
        \includegraphics[width=0.3\textwidth]{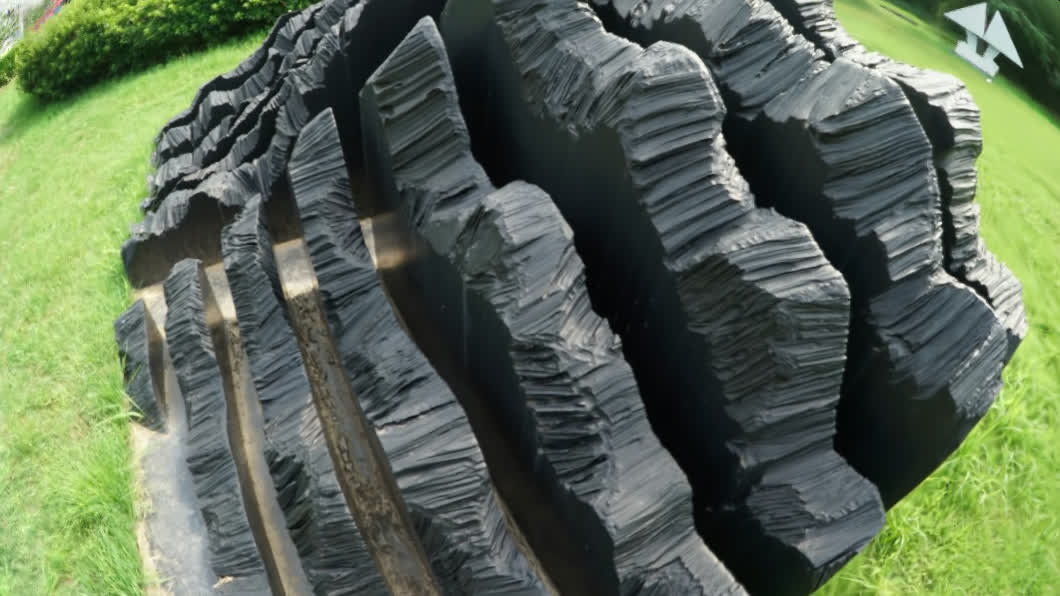} &
        \includegraphics[width=0.3\textwidth]{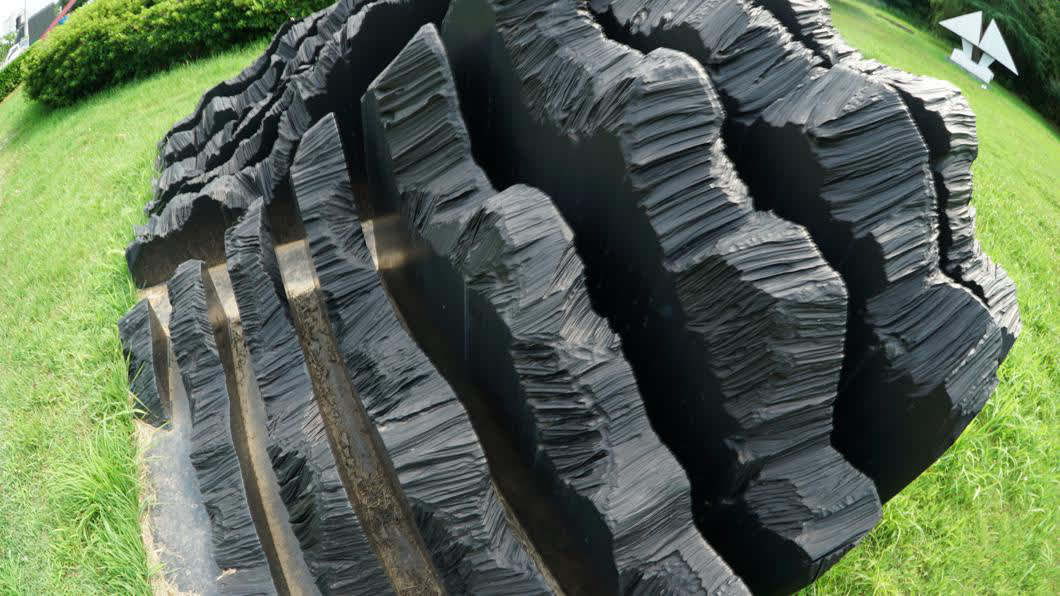} \\
        
        \raisebox{1\height}{\hspace{-0.5cm}\rotatebox{90}{{Chairs}}} &
        
        \includegraphics[width=0.3\textwidth]{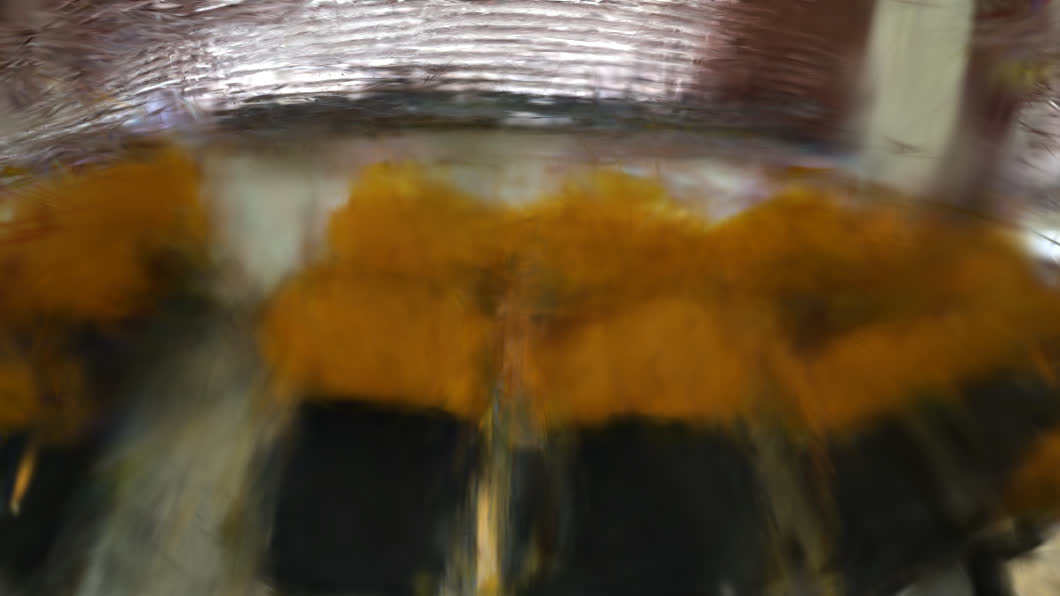} &
        \includegraphics[width=0.3\textwidth]{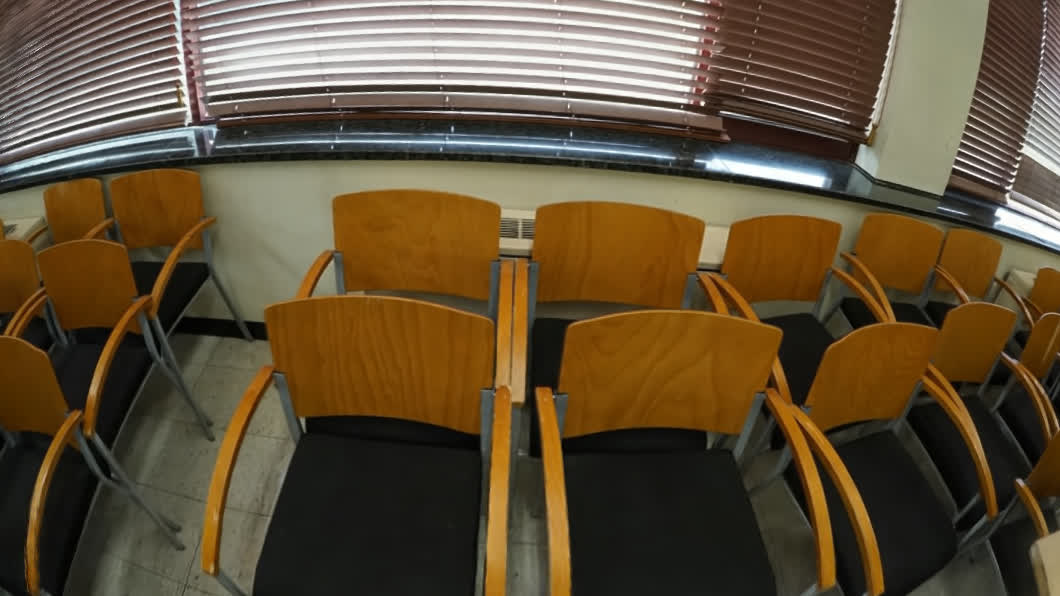} &
        \includegraphics[width=0.3\textwidth]{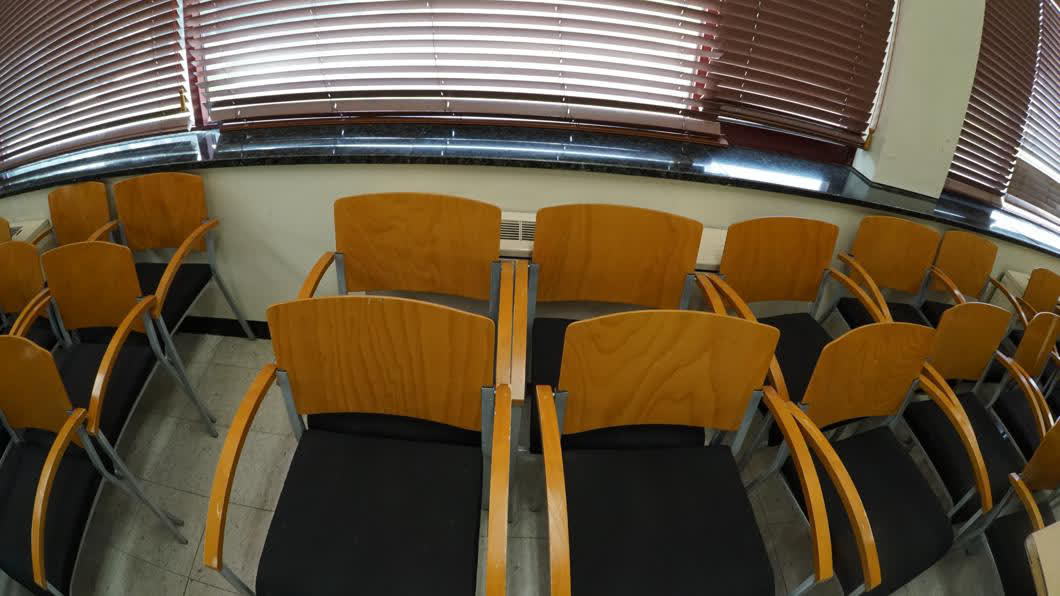} \\
                
        \raisebox{1\height}{\hspace{-0.5cm}\rotatebox{90}{{Globe}}} &
        
        \includegraphics[width=0.3\textwidth]{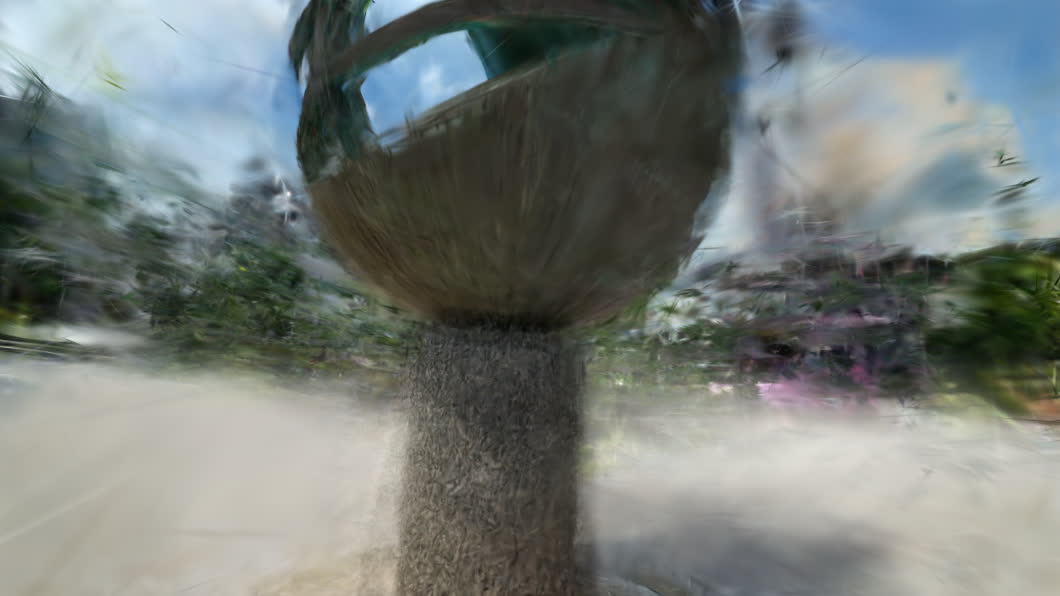} &
        \includegraphics[width=0.3\textwidth]{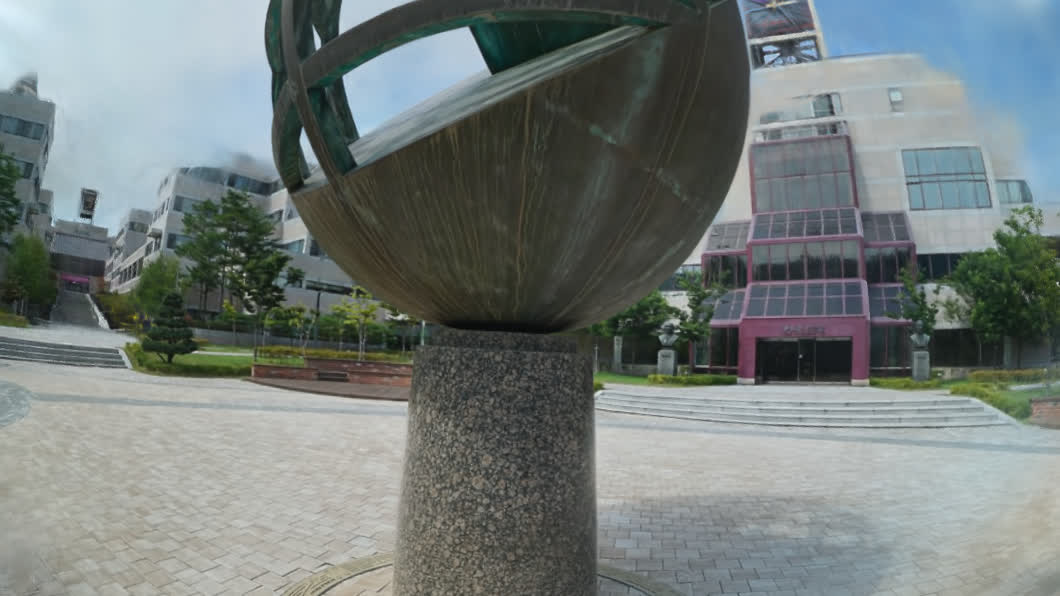} &
        \includegraphics[width=0.3\textwidth]{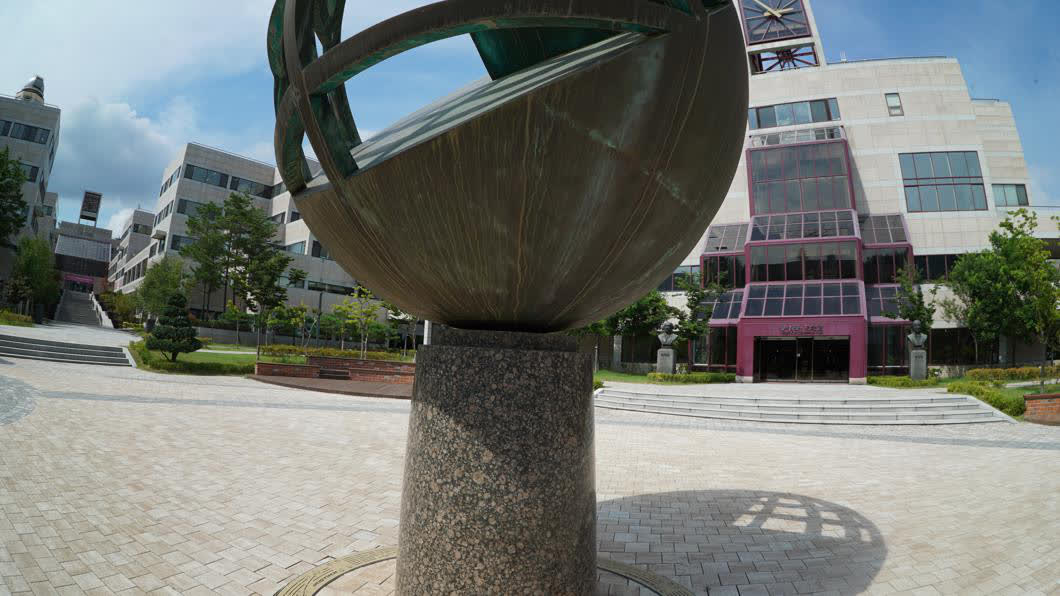} \\

        \raisebox{0.8\height}{\hspace{-0.5cm}\rotatebox{90}{{Flowers}}} &
        
        \includegraphics[width=0.3\textwidth]{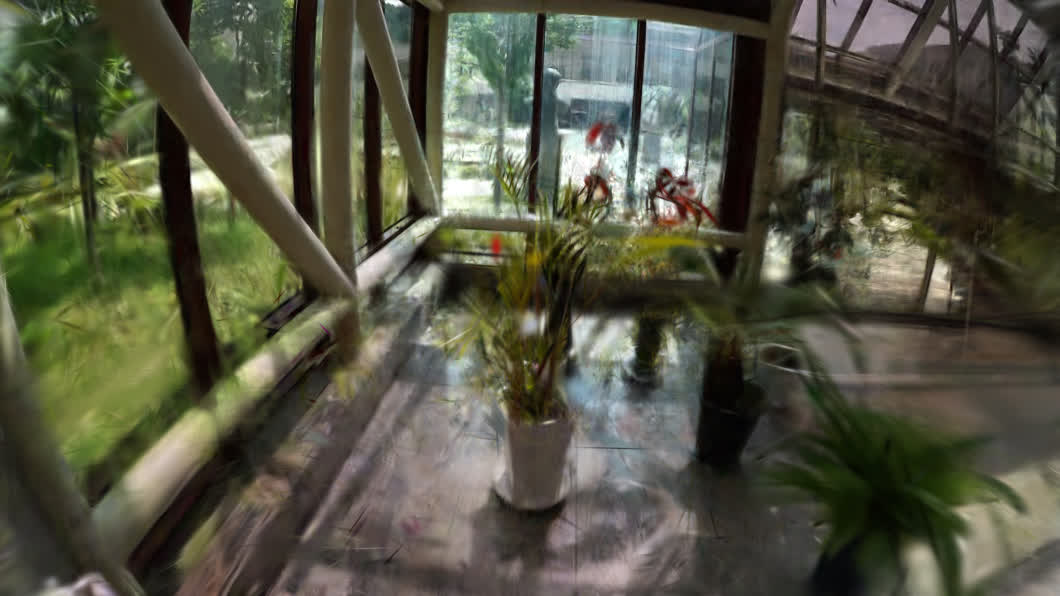} &
        \includegraphics[width=0.3\textwidth]{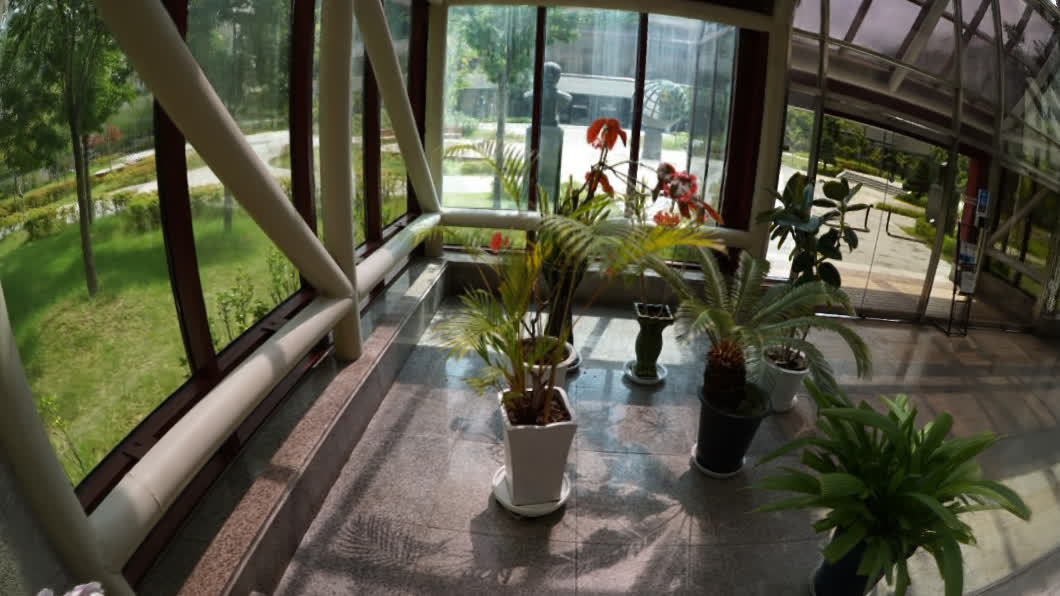} &
        \includegraphics[width=0.3\textwidth]{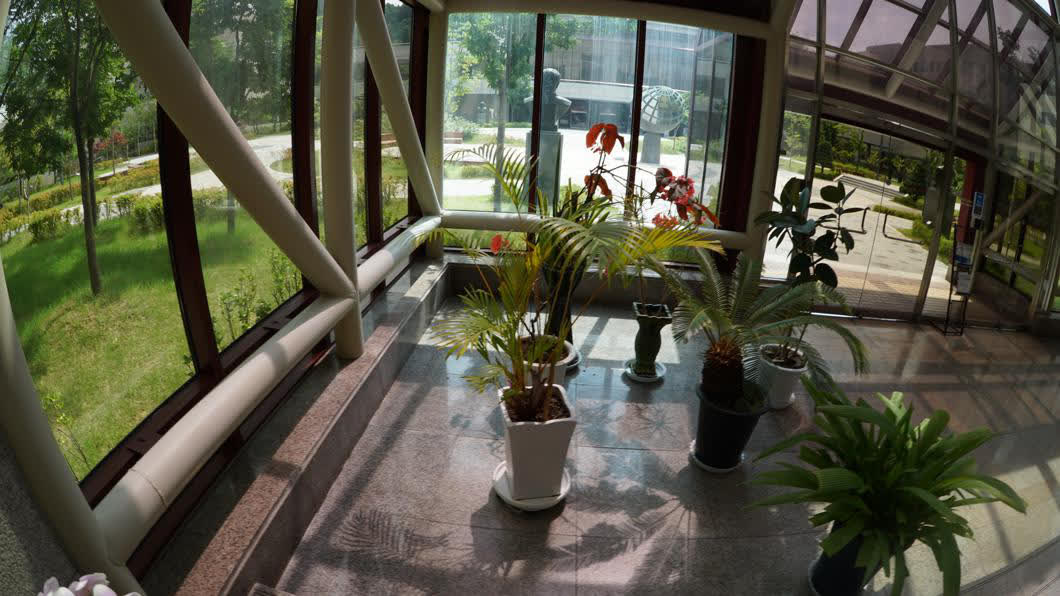} \\

        \raisebox{0.9\height}{\hspace{-0.5cm}\rotatebox{90}{{Flowers}}} &
        
        \includegraphics[width=0.3\textwidth]{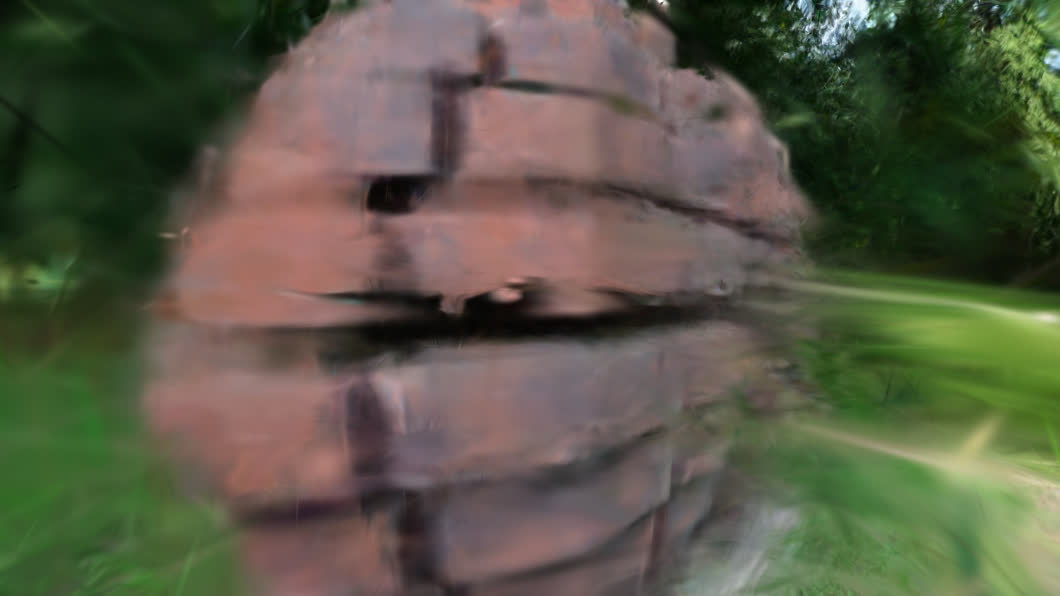} &
        \includegraphics[width=0.3\textwidth]{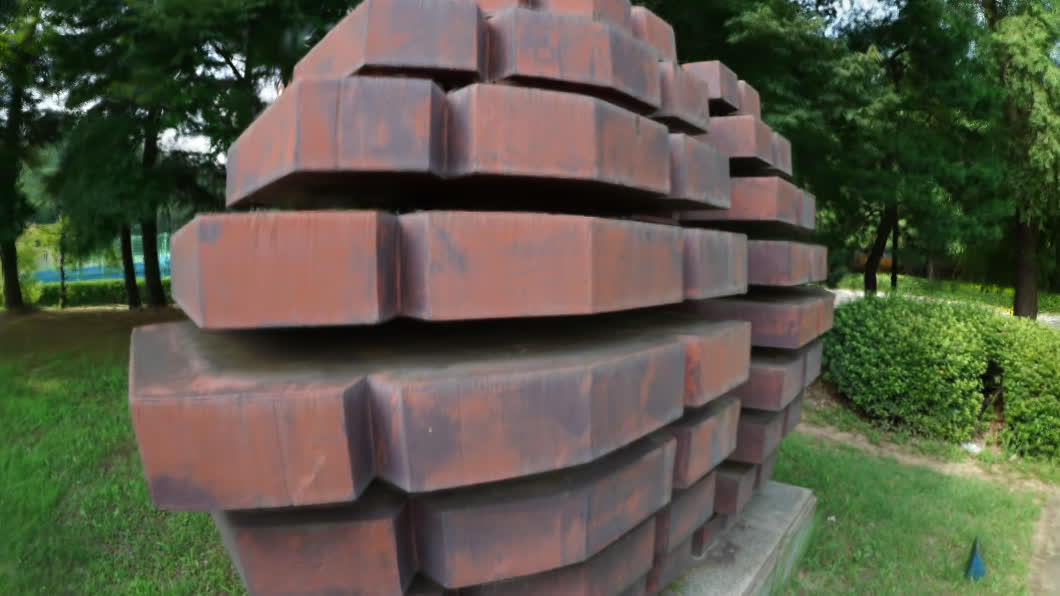} &
        \includegraphics[width=0.3\textwidth]{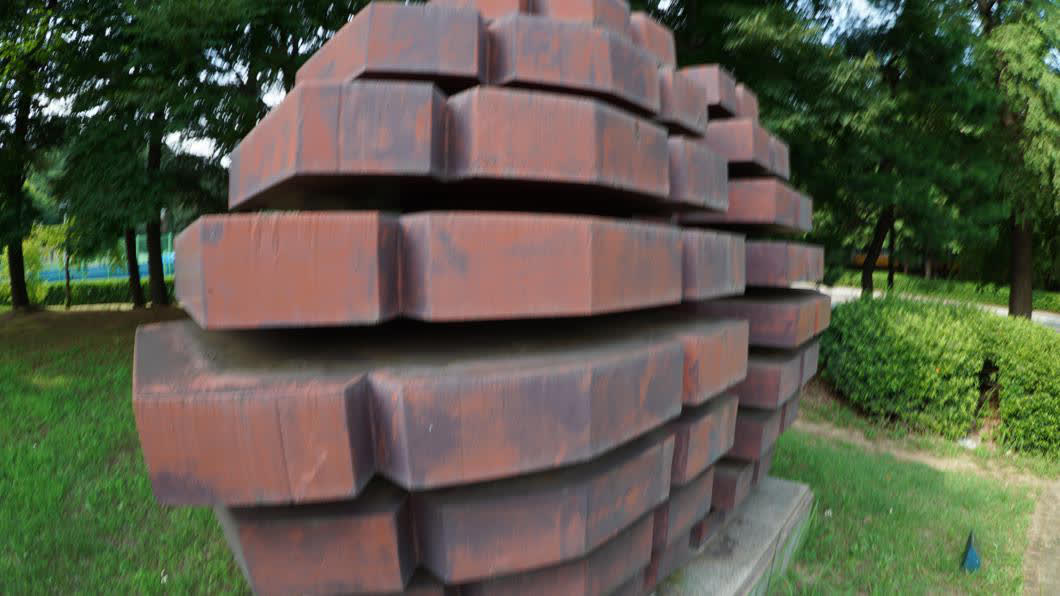} \\
        \multicolumn{1}{c}{} & \multicolumn{1}{c}{(a) w/o Optimize Cameras} & \multicolumn{1}{c}{(b) Optimize Cameras}& \multicolumn{1}{c}{(c) GT}
        \\
    \end{tabular}

    \caption{\textbf{Qualitative Comparison on Perturbed FisheyeNeRF dataset~\cite{jeong2021self}}. We show the novel view rendering with perturbed camera poses. We disable and enable camera optimization to illustrate the capability of our pipeline on recovering inaccurate poses along with distortion modeling.}
    \label{fig:fisheyenerf_Perturb}

\end{figure*}
}

\begin{figure*}[t]
    \centering
    \setlength{\tabcolsep}{0.1em} 
    \renewcommand{\arraystretch}{0.91}
    \begin{tabular}{cccccc}
        \includegraphics[height=0.157\textwidth]{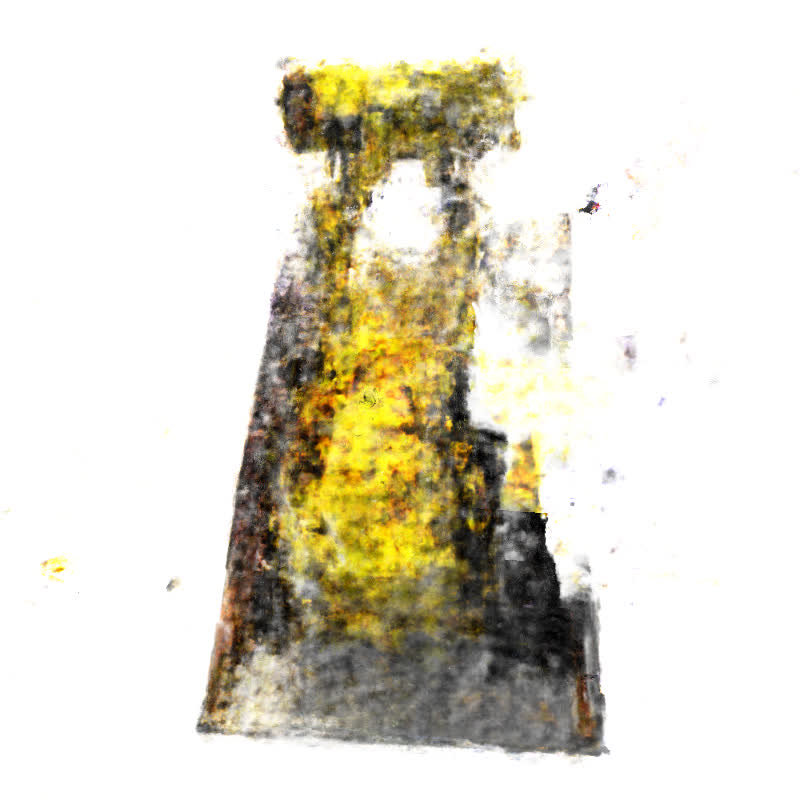} &  
        \includegraphics[height=0.157\textwidth]{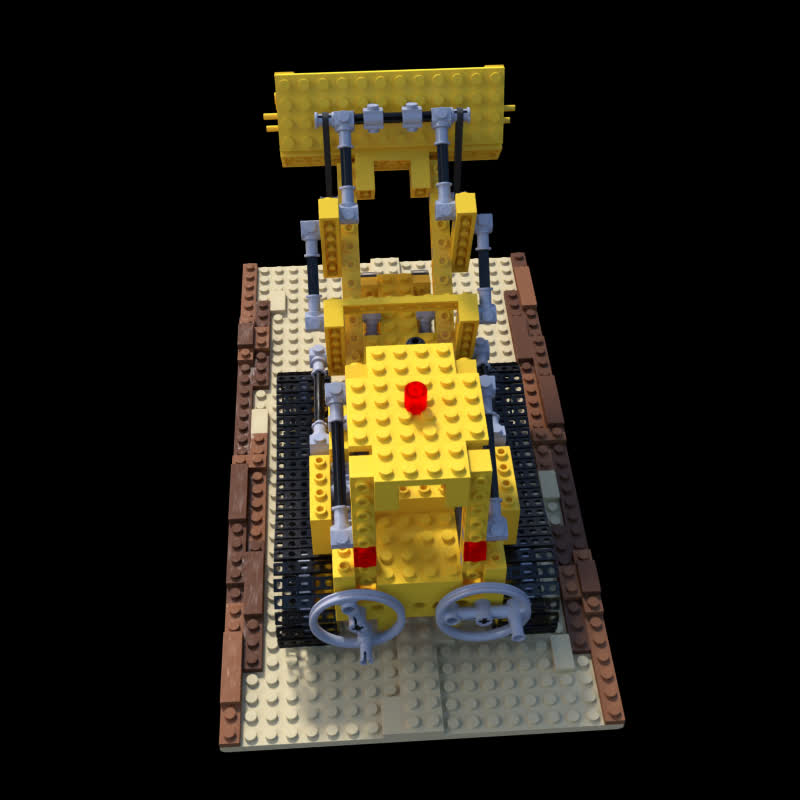} &  
        \includegraphics[height=0.157\textwidth]{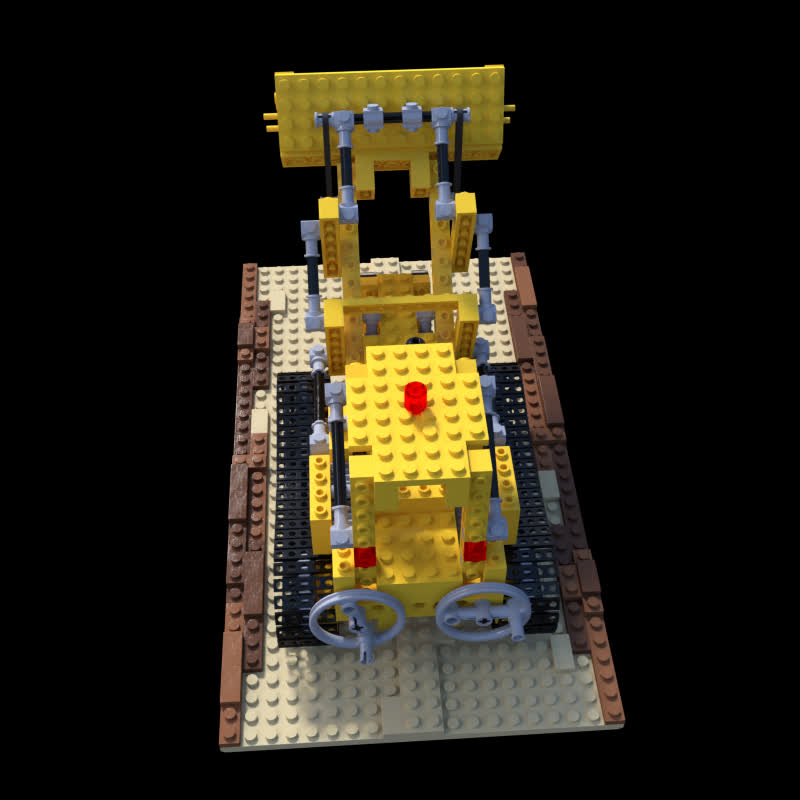} &  
        \includegraphics[height=0.157\textwidth]{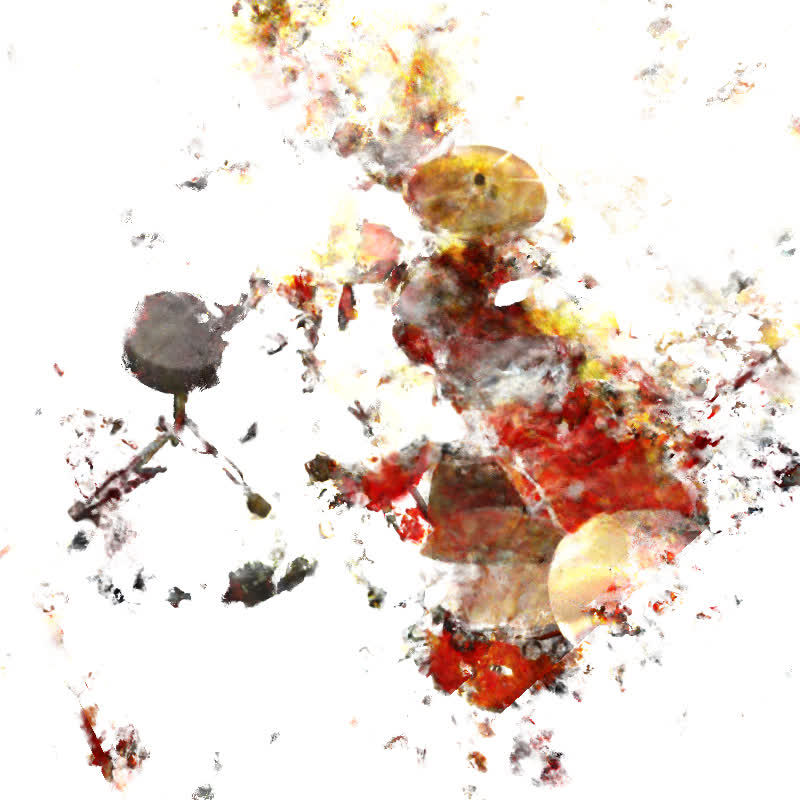} 
        \includegraphics[height=0.157\textwidth]{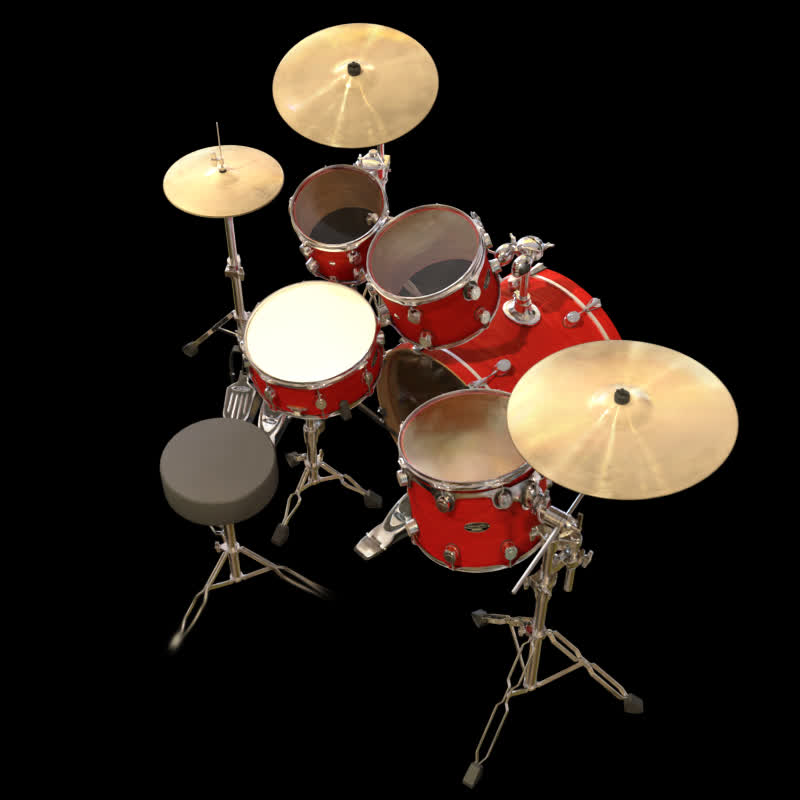} &  
        \includegraphics[height=0.157\textwidth]{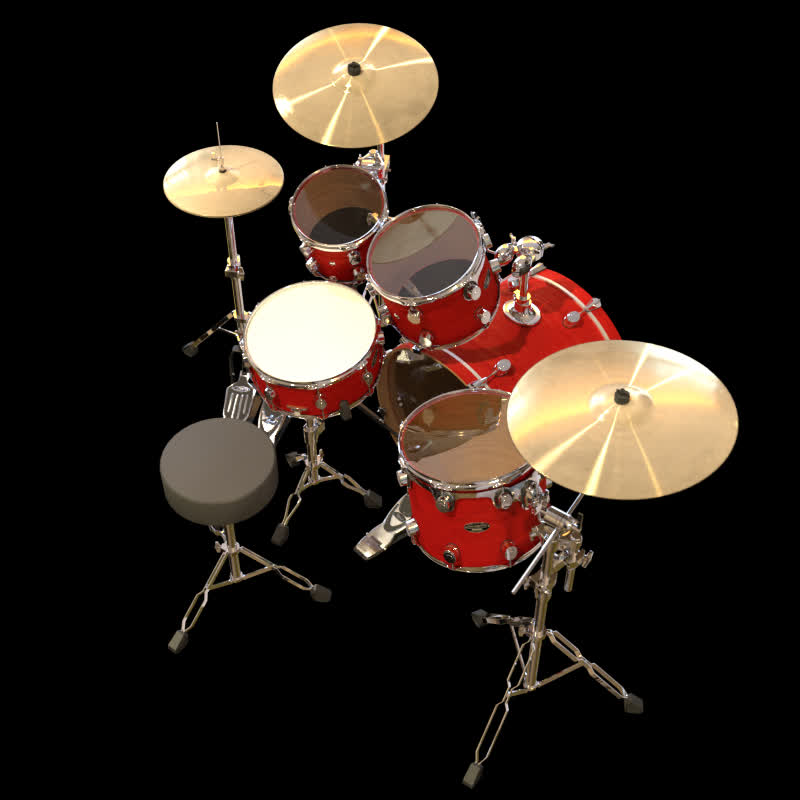} 
        \\
        \multicolumn{3}{c}{(a) Lego} & \multicolumn{3}{c}{(b) Drums}
        \\
        \includegraphics[height=0.157\textwidth]{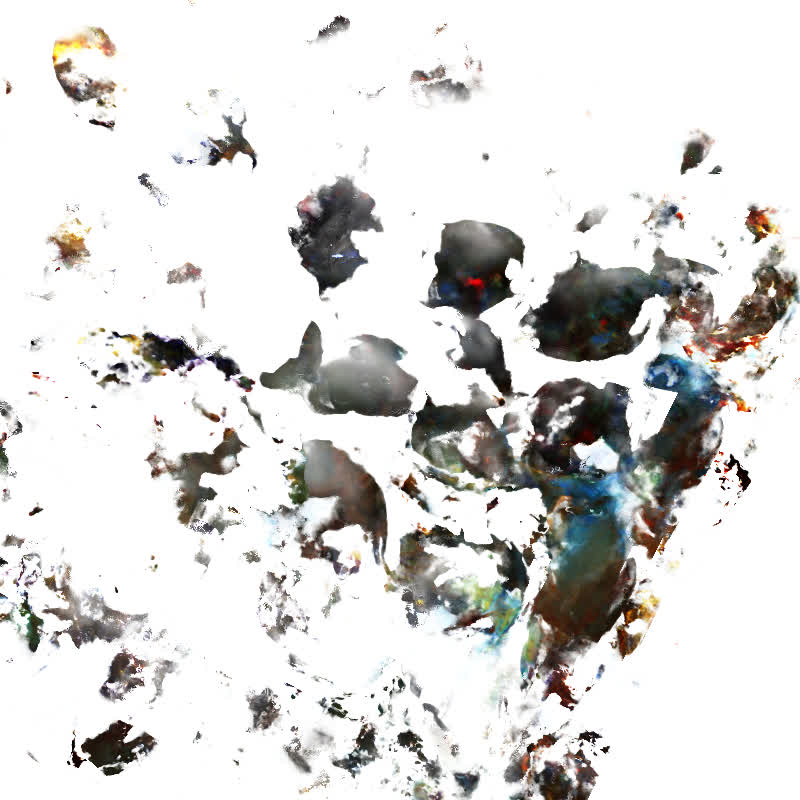} &  
        \includegraphics[height=0.157\textwidth]{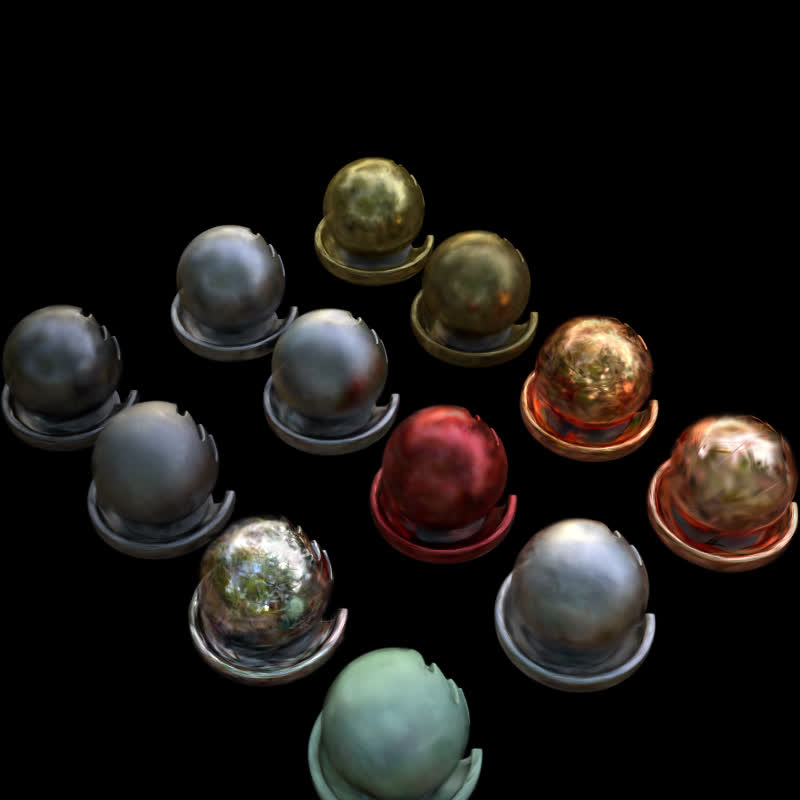} &  
        \includegraphics[height=0.157\textwidth]{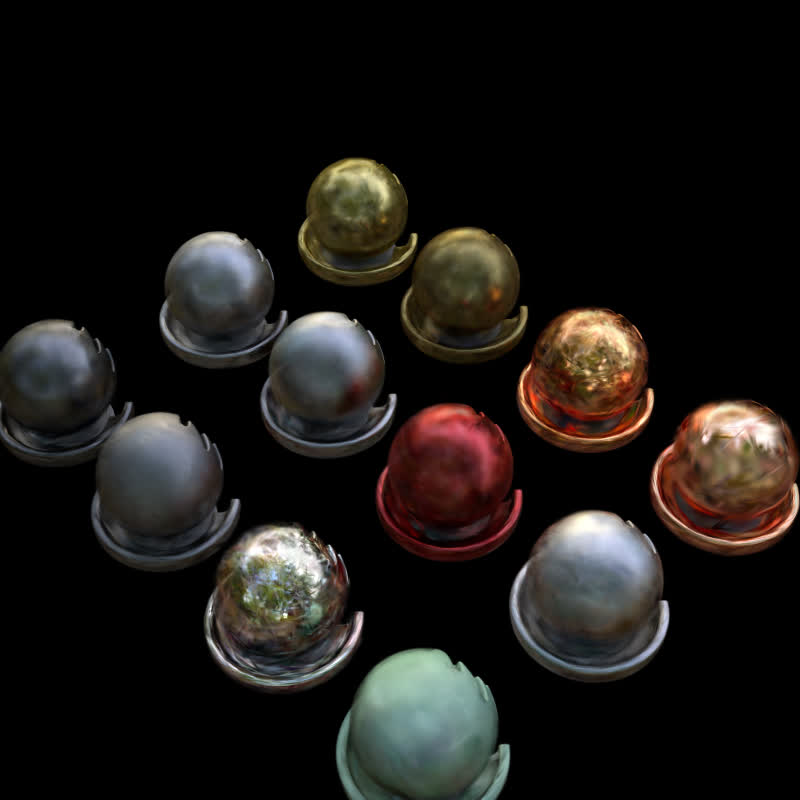} &  
        \includegraphics[height=0.157\textwidth]{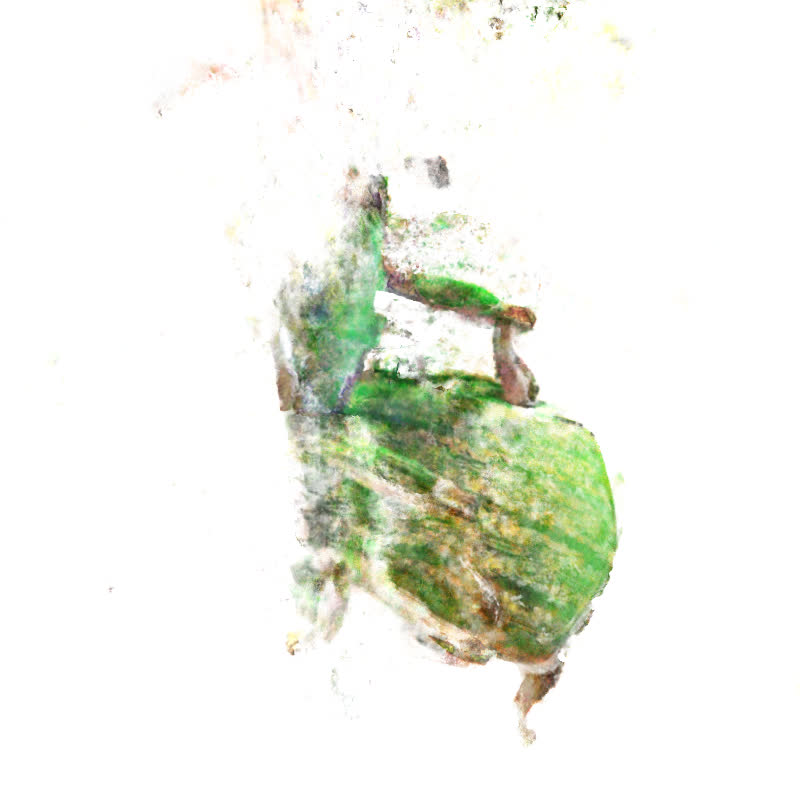} 
        \includegraphics[height=0.157\textwidth]{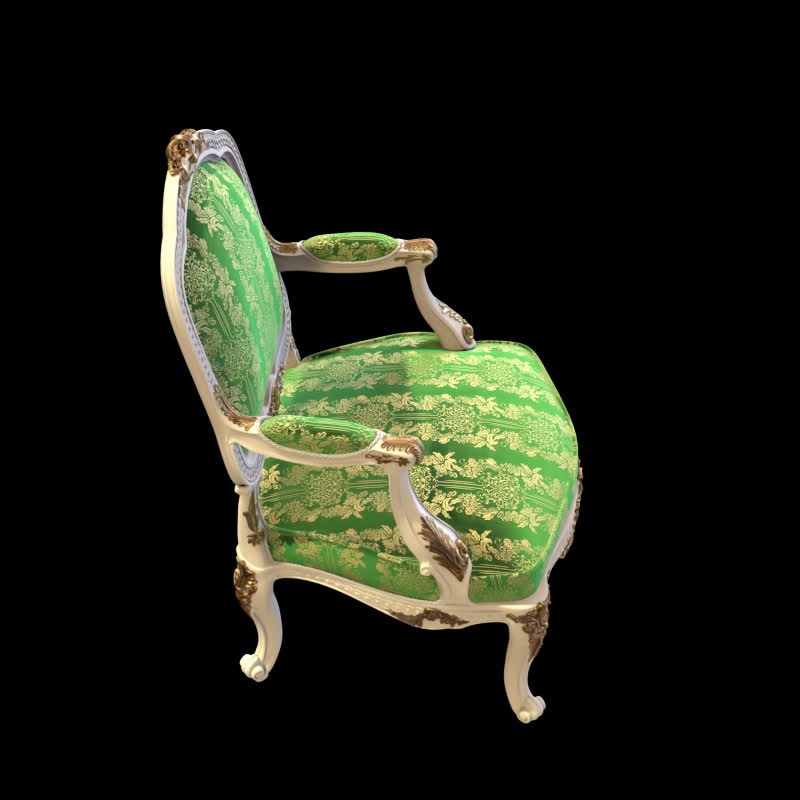} &  
        \includegraphics[height=0.157\textwidth]{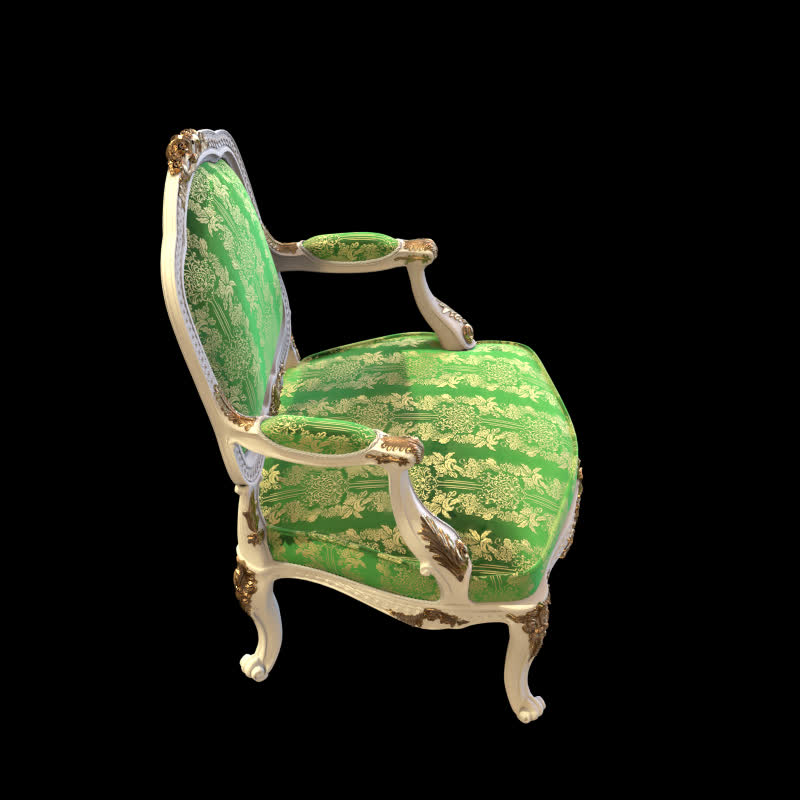} 
        \\
        \multicolumn{3}{c}{(c) Material} & \multicolumn{3}{c}{(d) Chair}
        \\
        &&&&&\\
    \end{tabular}
     \caption{We carry qualitative comparison with CamP at noise level 0.15. Each scene show CamP, our method, and the ground truth, from left to right. Our method is able to produce sharp renderings at this noise level, where CamP fails.}
    \label{fig:qualitative-perturb}
\end{figure*}

\subsection{Joint Distortion and Pose Optimization}
\label{sec:2.2}
Our approach supports efficient optimization of camera parameters, either independently (with perspective images) or in combination, as shown in~\cref{fig:pose_distortion}, with distortion modeling. This ensures that our pipeline remains robust even when both distortion and camera parameters are inaccurate.

To evaluate our model's ability to jointly optimize lens distortion and other camera parameters in real-world scenes, we introduce a setting where both camera extrinsics and intrinsics are perturbed in the FisheyeNeRF dataset. 
Specifically, we add Gaussian noise with a standard deviation of $s=0.15$ to each image's camera extrinsic and intrinsic parameters.
Despite this additional noise, our method successfully recovers the lens distortions while generating high-quality novel-view synthesis renderings (\cref{fig:fisheyenerf_Perturb}).

\begin{figure*}[th]
    \centering
    \setlength{\tabcolsep}{1pt} % Adjust space between columns if needed
    \begin{tabular}{cccc} % 4 columns (Vertical Caption | Image Set 1 | Image Set 2 | Image Set 3)

        % First row: Chairs
        \raisebox{1\height}{\hspace{-0.5cm}\rotatebox{90}{{Chairs}}} &
        \includegraphics[width=0.3\textwidth]{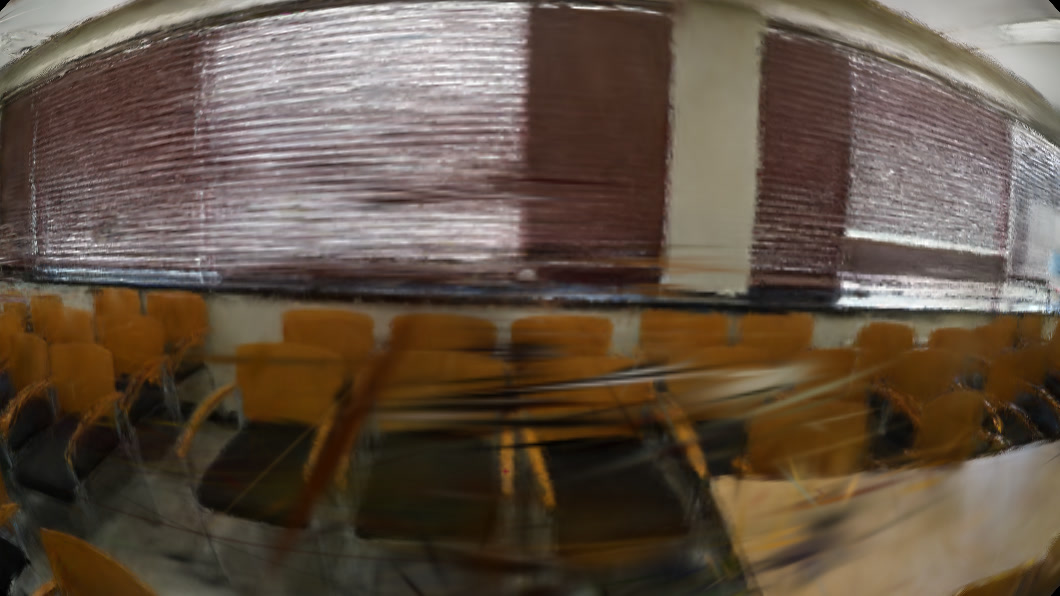} &
        \includegraphics[width=0.3\textwidth]{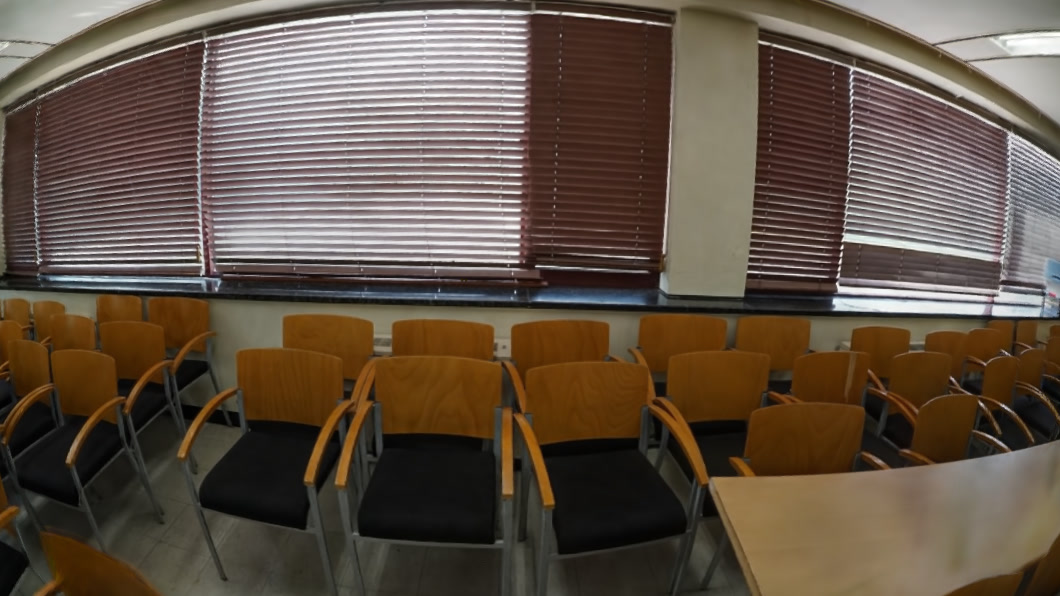} &
        \includegraphics[width=0.3\textwidth]{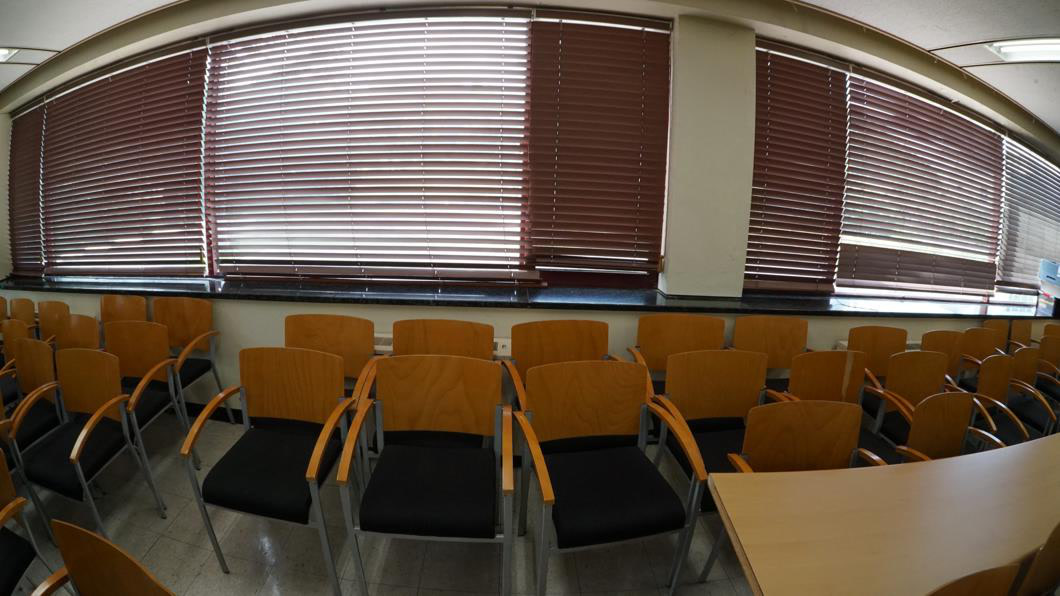} \\
        
        % Second row: Cube
        \raisebox{1\height}{\hspace{-0.5cm}\rotatebox{90}{{Cube}}} &
        \includegraphics[width=0.3\textwidth]{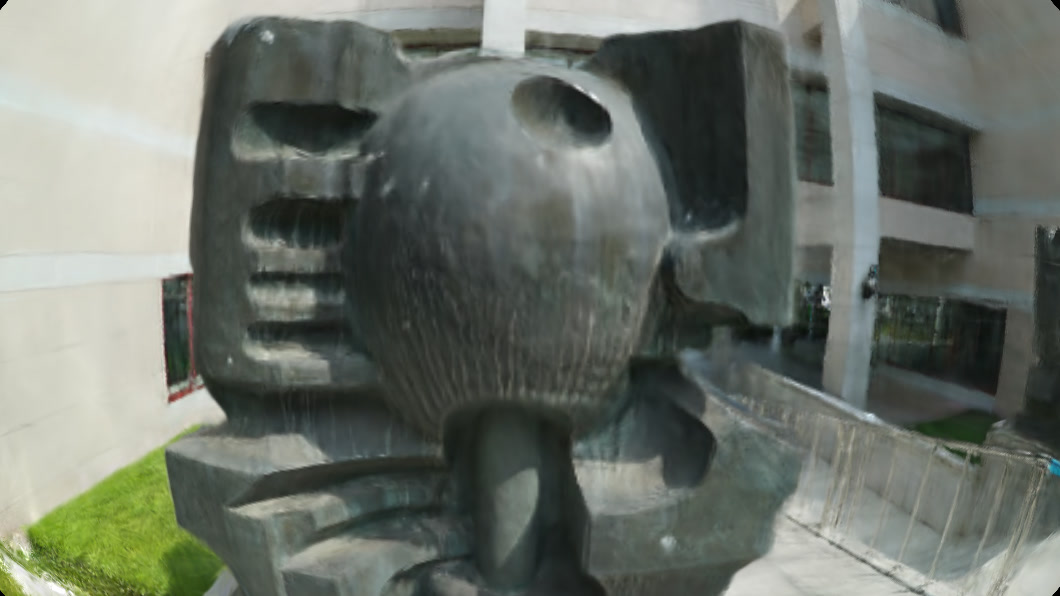} &
        \includegraphics[width=0.3\textwidth]{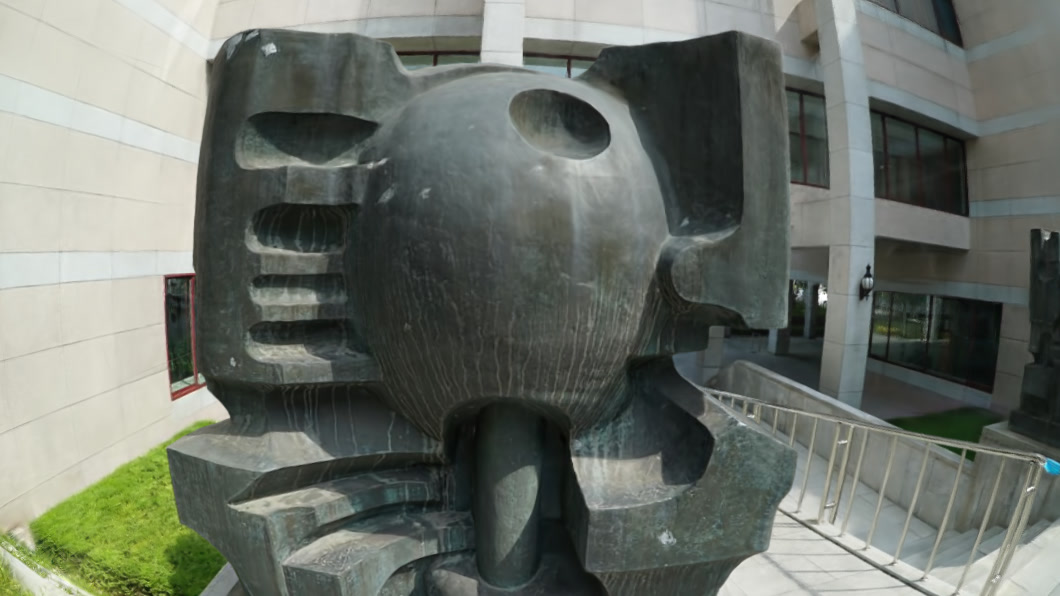} &
        \includegraphics[width=0.3\textwidth]{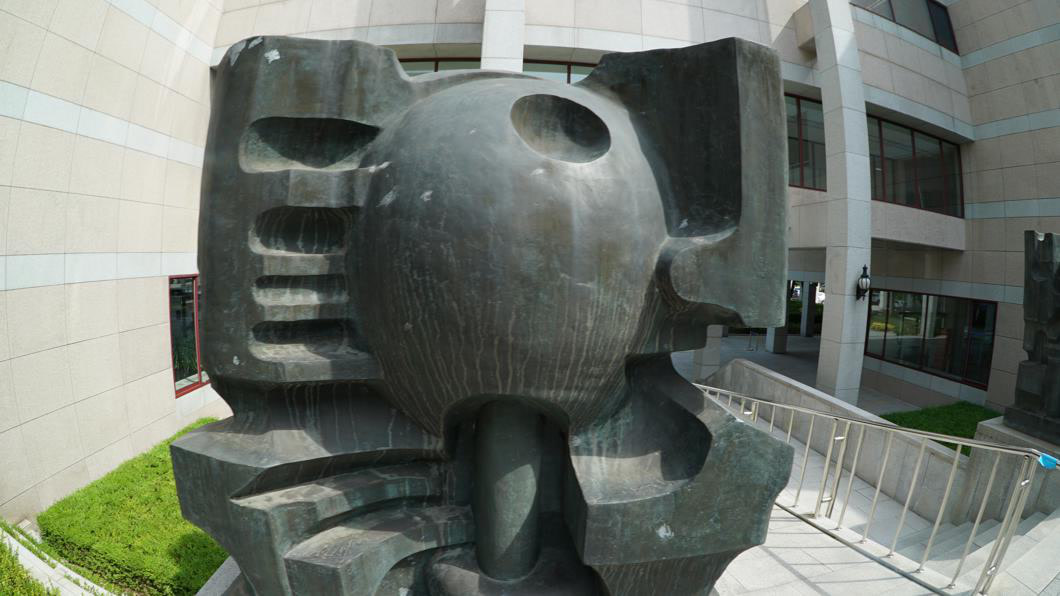} \\
        
        % Third row: Flowers
        \raisebox{1\height}{\hspace{-0.5cm}\rotatebox{90}{{Flowers}}} &
        \includegraphics[width=0.3\textwidth]{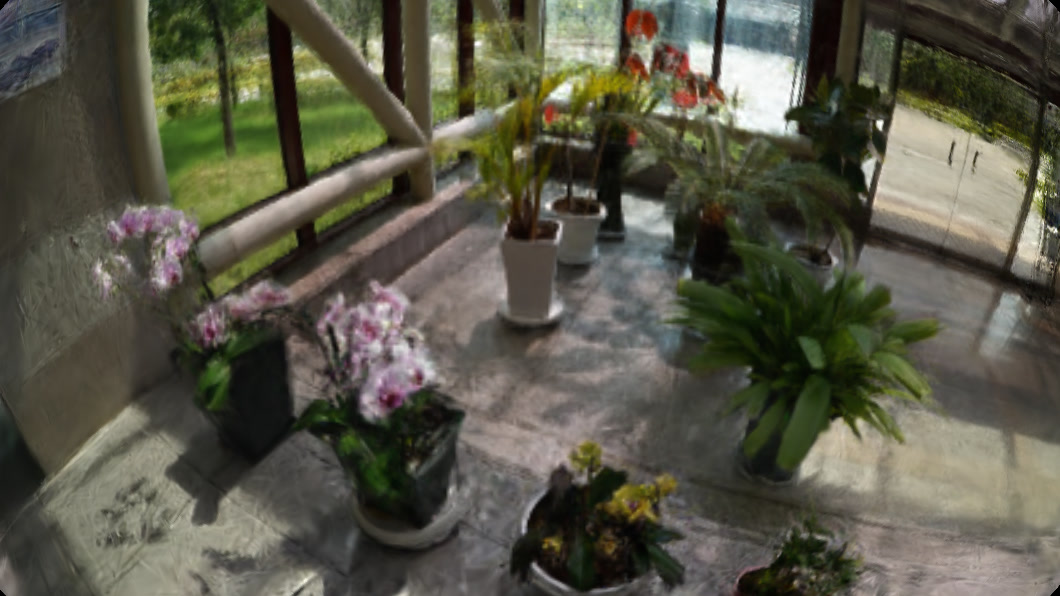} &
        \includegraphics[width=0.3\textwidth]{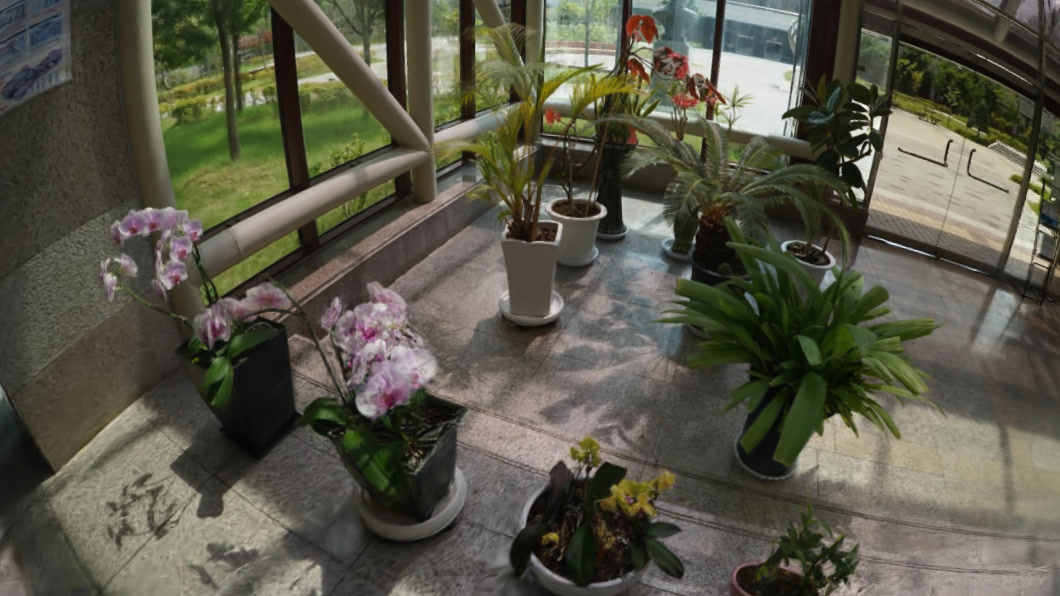} &
        \includegraphics[width=0.3\textwidth]{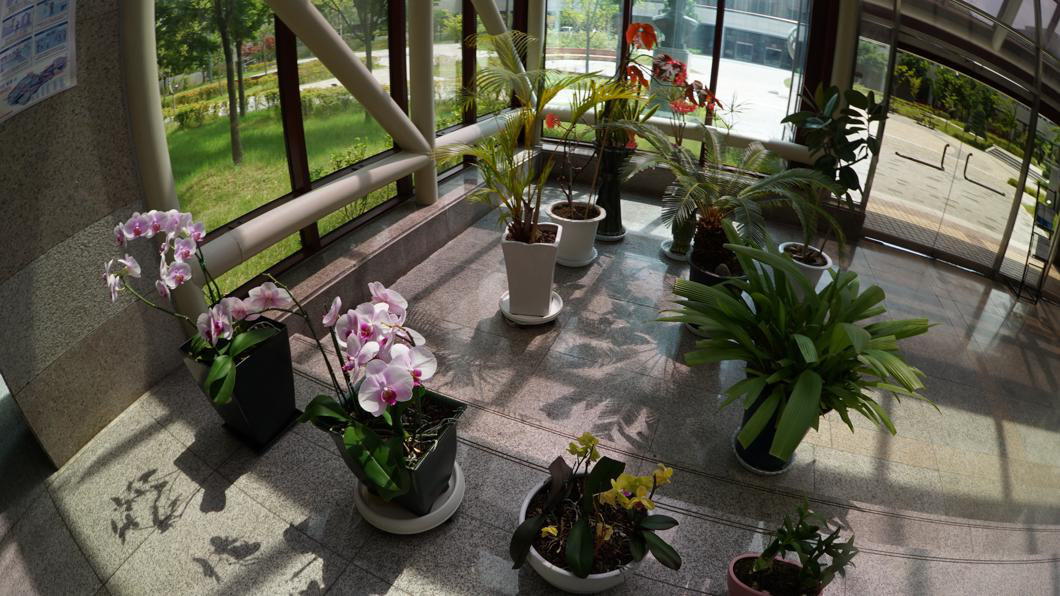} \\

        % % Third row: Flowers
        % \raisebox{1\height}{\hspace{-0.5cm}\rotatebox{90}{{Globe}}} &
        % \includegraphics[width=0.3\textwidth]{images/ablation_iresnet/woopt_rock.png} &
        % \includegraphics[width=0.3\textwidth]{images/ablation_iresnet/opt_rock.png} &
        % \includegraphics[width=0.3\textwidth]{images/ablation_iresnet/gt_rock.png} \\

        \multicolumn{1}{c}{} & \multicolumn{1}{c}{(a) Freeze Intialization} & \multicolumn{1}{c}{(b) Ours} & \multicolumn{1}{c}{(c) GT}
        \\
    \end{tabular}

    \caption{\textbf{Comparison on Invertible ResNet Optimization}. This version shows the comparison transposed, grouping by scenes instead of optimization stages.}
    \label{fig:transposed_opti_iresnet}
\end{figure*}

One key observation is that COLMAP~\cite{schoenberger2016sfm} can robustly estimate camera extrinsics but struggles with intrinsic parameters and distortion. When we enable extrinsic optimization during reconstruction, the camera poses are refined only slightly, indicating that the initial poses are already quite reliable. Regarding lens distortion, as demonstrated in \cref{fig:transposed_opti_iresnet} and discussed in \textbf{Hybrid Field} of Sec.4.4 and\cref{sec:inaccurate_colmap_distortion}, COLMAP's distortion estimation lacks accuracy, highlighting the necessity of our hybrid distortion field for improved expressiveness and precision.

We hypothesize that this limitation stems from the structure-from-motion (SfM) pipeline~\cite{schoenberger2016sfm} in COLMAP. COLMAP primarily utilizes the central region of raw images (corresponding to a small FOV), where conventional distortion models perform well. Consequently, lens distortion has minimal influence on extrinsic estimation, as COLMAP can still rely on the image center to solve linear equations. However, when attempting to leverage the full FOV of raw images for reconstruction, the limitations of a fixed distortion model and a single-plane projection become pronounced. 

To validate the self-calibration capability of our pipeline, we manually introduce noise into the extrinsics produced by COLMAP and jointly optimize extrinsics, intrinsics, and distortion. As shown in \cref{fig:fisheyenerf_Perturb}, our approach effectively refines all camera parameters and distortion simultaneously. Without extrinsic and intrinsic optimization, the hybrid field can only predict coarse distortion, while misaligned poses contribute to the blurry reconstruction seen in~\cref{fig:fisheyenerf_Perturb} (a). Even with significant perturbations in poses and intrinsics, our method robustly recovers accurate camera parameters and the distortion field after training, as illustrated in~\cref{fig:fisheyenerf_Perturb} (b).

\begin{table}[t]
  \centering
  \scalebox{0.85}{
  \begin{tabular}{cccccc}
    \toprule
    \multirow{2}{*}{Methods} & \multicolumn{3}{c}{Image Metrics} & \multicolumn{2}{c}{Camera Metrics} \\
    \cmidrule(lr){2-4} \cmidrule(lr){5-6}
    & PSNR & SSIM & \multicolumn{1}{c}{LPIPS} & Position & Orientation \\
    \midrule
    % Zip-NeRF & \xmark & 33.10 & 0.971 & 0.031 & - & - \\
    % GS & \xmark & 33.31 & 0.969 & 0.030 & - & - \\
    % \midrule
    3DGS  & 16.54 & 0.733 & 0.273 & 0.2911 & 5.015 \\
    CamP  & 19.07 & 0.840 & 0.289 & 0.1879 & 5.619  \\
    Ours  & \textbf{32.84} & \textbf{0.964} & \textbf{0.034} & \textbf{0.0082} & \textbf{0.919} \\
    \bottomrule
  \end{tabular}
  }
  
    \caption{Comparison with CamP~\cite{park2023camp} and 3DGS~\cite{kerbl20233d} in the NeRF-Synthetic dataset.
  We report average camera orientation errors in degrees, and position error in world units.
  \label{tab:quantitative_perturb_synthetic}
  }
\end{table}
\begin{table}[htbp]
  \centering
  \scalebox{0.85}{
    \begin{tabular}{c|cccccc}
    \toprule
    \textbf{Scenes} & \textbf{Metric} & \textbf{0} & \textbf{0.1} & \textbf{0.15} & \textbf{0.2} & \textbf{0.25} \\
    \midrule
    \multirow{3}{*}{Chair} & SSIM & 0.987 & 0.988 & 0.988 & 0.987 & 0.980 \\
                           & PSNR & 35.81 & 36.10 & 35.99 & 35.83 & 34.35 \\
                           & LPIPS & 0.012 & 0.012 & 0.012 & 0.012 & 0.019 \\
    \midrule
    \multirow{3}{*}{Lego}  & SSIM & 0.983 & 0.974 & 0.972 & 0.969 & 0.965 \\
                           & PSNR & 35.77 & 34.78 & 34.29 & 33.70 & 33.15 \\
                           & LPIPS & 0.015 & 0.021 & 0.023 & 0.025 & 0.029 \\
    \midrule
    \multirow{3}{*}{Drums} & SSIM & 0.955 & 0.954 & 0.953 & 0.953 & 0.946 \\
                           & PSNR & 26.17 & 26.15 & 26.04 & 26.01 & 25.30 \\
                           & LPIPS & 0.037 & 0.038 & 0.039 & 0.040 & 0.045 \\
    \midrule
    \multirow{3}{*}{Materials} & SSIM & 0.960 & 0.960 & 0.951 & 0.942 & 0.843 \\
                               & PSNR & 29.99 & 29.93 & 28.91 & 27.95 & 15.17 \\
                               & LPIPS & 0.034 & 0.036 & 0.044 & 0.052 & 0.158 \\
    \midrule
    \multirow{3}{*}{Mic} & SSIM  & 0.991 & 0.989 & 0.987 & 0.974 & 0.923 \\
                         & PSNR  & 35.34 & 34.58 & 33.65 & 30.07 & 18.78 \\
                         & LPIPS & 0.006 & 0.008 & 0.010 & 0.019 & 0.087 \\
    \midrule
    \multirow{3}{*}{Ship} & SSIM  & 0.907 & 0.873 & 0.776 & 0.719 & 0.700 \\
                          & PSNR  & 30.91 & 28.66 & 20.96 & 16.80 & 15.10 \\
                          & LPIPS & 0.106 & 0.126 & 0.203 & 0.262 & 0.294 \\
    \midrule
    \multirow{3}{*}{Ficus} & SSIM  & 0.987 & 0.987 & 0.984 & 0.955 & 0.859 \\
                           & PSNR  & 34.85 & 34.84 & 33.99 & 28.08 & 18.54 \\
                           & LPIPS & 0.012 & 0.012 & 0.014 & 0.039 & 0.125 \\
    \midrule
    \multirow{3}{*}{Hotdog} & SSIM  & 0.985 & 0.985 & 0.985 & 0.983 & 0.982 \\
                            & PSNR  & 37.67 & 37.65 & 37.63 & 37.05 & 36.53 \\
                            & LPIPS & 0.020 & 0.020 & 0.020 & 0.022 & 0.025 \\
    \bottomrule
    \end{tabular}%
  }
  \caption{Quantitative Comparison on the Perturbed Synthetic Dataset}
\label{tab:perturb-syn}%
\end{table}%

{\begin{table*}[ht]

  \centering
  \scalebox{0.65}{
    \begin{tabular}{ccccccccccccccccccc}
    \toprule
    \multicolumn{1}{c}{\multirow{2}[2]{*}{Method}} & \multicolumn{3}{c}{Chairs} & \multicolumn{3}{c}{Cube} & \multicolumn{3}{c}{Flowers} & \multicolumn{3}{c}{Globe} & \multicolumn{3}{c}{Heart} & \multicolumn{3}{c}{Rock}\\
    \cmidrule(lr){2-4}
    \cmidrule(lr){5-7}
    \cmidrule(lr){8-10}
    \cmidrule(lr){11-13}
    \cmidrule(lr){14-16}
    \cmidrule(lr){17-19}
    \multicolumn{1}{c}{} & SSIM  & PSNR  & LPIPS & SSIM  & PSNR & LPIPS & SSIM  & PSNR & LPIPS & SSIM  & PSNR & LPIPS & SSIM  & PSNR  & LPIPS & SSIM  & PSNR  & LPIPS \\
    \midrule
    3DGS~\cite{kerbl20233d}     &0.431&14.06&0.547&0.507&15.21&0.533&0.281&12.91&0.609&0.502&15.09& 0.530&0.505&15.19&0.549&0.297&12.70&0.595\\

    Ours (Freeze Init) &0.583&18.28&0.290&0.637&21.64&0.296&0.443&18.09&0.379&0.580&19.63&0.327&0.660&20.87&0.282&0.511&20.24&0.280\\

    Ours &\textbf{0.832}&\textbf{23.45}&\textbf{0.106}&\textbf{0.786}&\textbf{24.63}&\textbf{0.162}&\textbf{0.693}&\textbf{22.01}&\textbf{0.172}&\textbf{0.790}&\textbf{23.63}&\textbf{0.126}&\textbf{0.775}&\textbf{23.42}&\textbf{0.195}&\textbf{0.787}&\textbf{24.88}&\textbf{0.145}\\
    \bottomrule
    \end{tabular}%
  } 
  \caption{\textbf{Ablation on Invertible ResNet Optimization}. We compare our final optimized hybrid field with a fixed hybrid field reconstruction on FisheyeNeRF~\cite{jeong2021self}. While the distortion estimated from COLMAP~\cite{schoenberger2016sfm} significantly improves quality compared to vanilla 3DGS~\cite{kerbl20233d}, optimizing the invertible ResNet further enhances performance.}
  \label{tab:colmap_init_opt}%
\end{table*}%
}

\subsection{Pure Camera Optimization} 
\label{sec:2.3}
We use the NeRF-Synthetic dataset~\cite{mildenhall2021nerf}, which includes known ground truth camera poses. 
The dataset contains 100 viewpoints for training and 200 additional viewpoints for computing test error metrics.
We first add noise to perturb the rotation angles of camera extrinsics, the positions of the camera centers, and the focal lengths. 
These noisy cameras are used to train both the baselines and our method.
We compare our method with CamP~\cite{park2023camp}, implemented on ZipNeRF~\cite{barron2023zip}, a state-of-the-art method for joint optimization of 3D scenes and camera extrinsics and intrinsics. In addition to CamP, we also report the performance of vanilla Gaussian Splatting without pose optimization.

The models are evaluated on two criteria, following the protocol in CamP~\cite{park2023camp}.
First, we measure the accuracy of the estimated camera poses in the training views after accounting for a global rigid transformation. 
Second, we measure the quality of rendered images at the held-out camera poses after a test-time optimization to adjust for a potential global offset. 
\cref{tab:quantitative_perturb_synthetic} shows that our method outperforms both vanilla 3DGS and CamP in both image and camera metrics by a large margin.

We further evaluate our model by perturbing the camera poses with varying levels of noise. Specifically, we add Gaussian noise with a standard deviation ranging from 0 to 0.3 to the camera poses. For reference, we report the increased noise levels for each scene in~\cref{tab:perturb-syn}. Handling larger noise in camera poses presents a significant challenge. As shown in \cref{fig:robust_to_noise} and \cref{fig:qualitative-perturb}, CamP’s performance drops significantly due to its tendency to fall into local minima when the initial camera poses contain substantial errors. In contrast, our method exhibits much slower degradation in novel-view synthesis performance.

We also include the video ``pose\_opt.mp4" to visualize the optimization process.

\section{Distortion Estimation from COLMAP}
\label{sec:inaccurate_colmap_distortion}
In practice, the distortion estimated from the SfM~\cite{schoenberger2016sfm} pipeline can be used as an initialization for our hybrid field, stabilizing training and accelerating convergence. However, these parameters are inaccurate when derived from highly distorted images. We verify the necessity of optimizing our hybrid field during reconstruction both quantitatively and qualitatively.

To assess this, we fit the COLMAP~\cite{schoenberger2016sfm} distortion parameters within our hybrid field and freeze the network. As shown in~\cref{tab:colmap_init_opt}, the distortion initialization from COLMAP improves upon Vanilla 3DGS~\cite{kerbl20233d} but remains far from accurate in modeling distortion compared to our approach. While the hybrid field produces a roughly correct distortion field, it results in blurry reconstructions, as shown in~\cref{fig:transposed_opti_iresnet}. This degradation is particularly noticeable in scenes with many straight lines, such as the jalousie windows in the chairs scene. The photometric evaluation in~\cref{tab:colmap_init_opt} demonstrates that our fully optimized hybrid field significantly outperforms the estimated distortion parameters. These results highlight the importance of optimizing our hybrid field for achieving more accurate reconstruction and distortion modeling.

\section{Computational Efficiency}
\paragraph{Training Time.}
To verify the hypothesis that our hybrid method achieves a better balance between expressiveness and efficiency, we perform an ablation study on the FisheyeNeRF dataset~\cite{jeong2021self}. 
Specifically, we analyze the grid resolution of $\textbf{P}_c$, which determines the number of evaluations required for the most computationally expensive part of the pipeline—the invertible ResNet. 
\cref{tab:ablation_control_pts} reports the PSNR and training time for three different grid resolutions, as well as for Fisheye-GS~\cite{liao2024fisheye} and vanilla 3DGS~\cite{kerbl20233d}. 
We observe that increasing the resolution of $\textbf{P}_c$ leads to better performance but longer training time. Further reducing the grid resolution does not significantly shorten computation time, as other operations, such as gradient computation for camera parameters, begin to dominate. Fortunately, all hybrid solutions significantly outperform vanilla 3DGS and Fisheye-GS. The lowest-resolution setting we tested introduces only a $7$-minute training time overhead compared to 3DGS, in exchange for a $>8$ dB boost in PSNR.

{\begin{figure}[t]
    \centering
    \setlength{\tabcolsep}{1pt} % Adjust space between columns if needed
    \scalebox{0.95}{
    \begin{tabular}{cc} % 4 columns (Vertical Caption | Image Set 1 | Image Set 2 | Image Set 3)
    
        \includegraphics[width=0.23\textwidth]{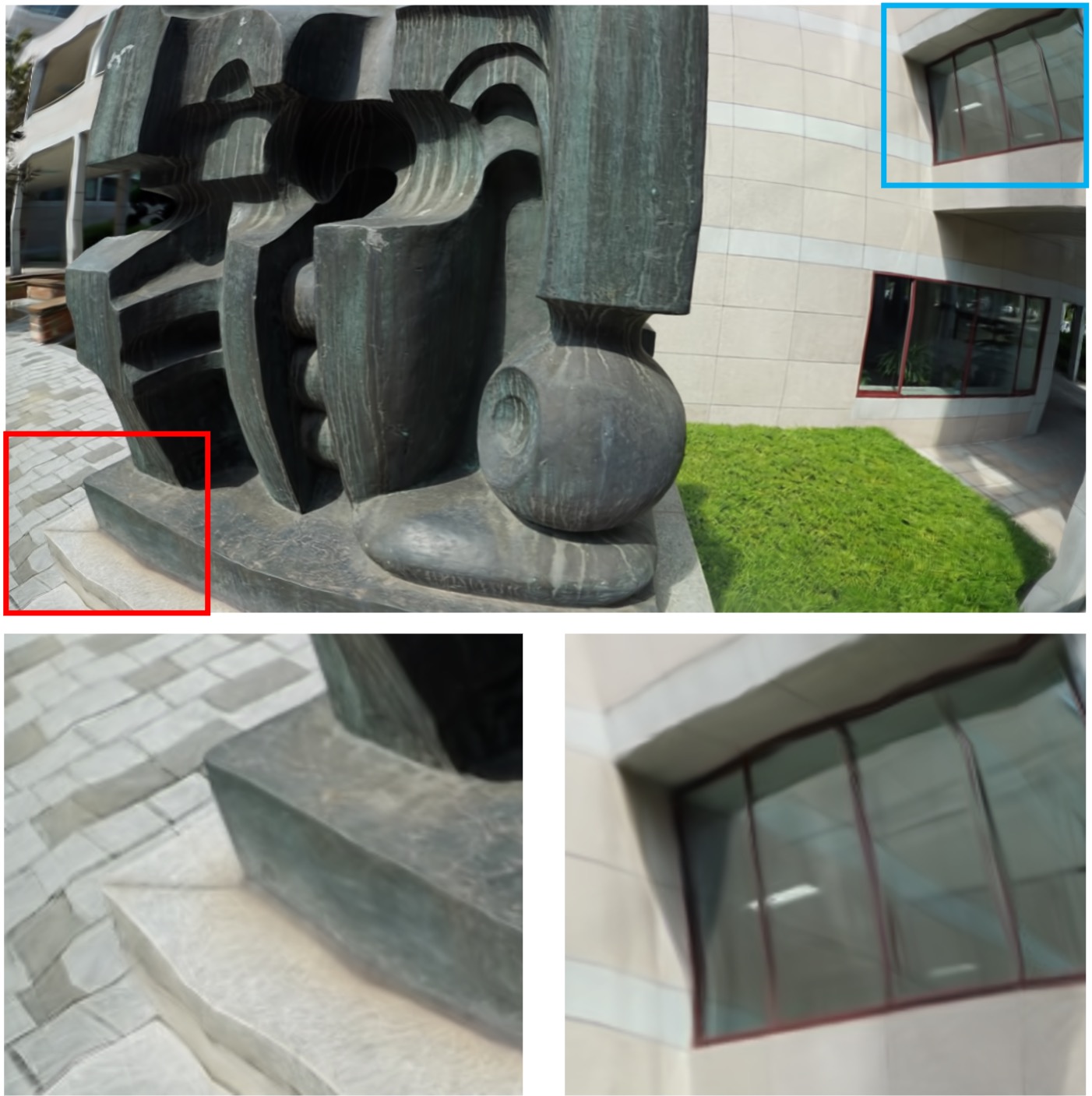} &
        \includegraphics[width=0.23\textwidth]{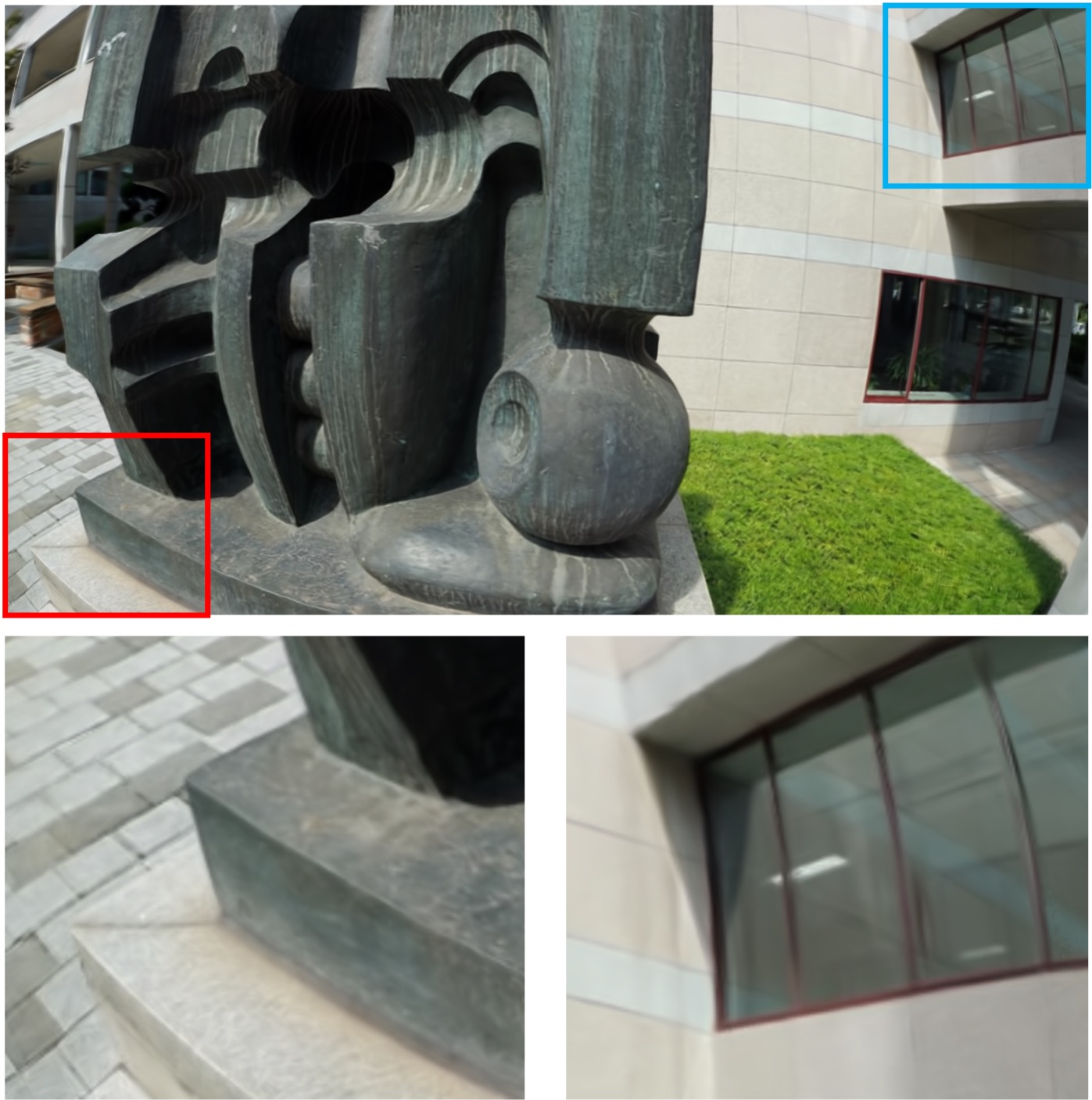}
        \\
        
        \multicolumn{1}{c}{(a) Low Resolution} & \multicolumn{1}{c}{(b) High Resolution}
        \\
    \end{tabular}
    }
    \caption{\textbf{Resolution of Control Grid}. When the resolution of the control grid is decreased, the central region retains decent quality due to minimal distortion. However, as highlighted by the red and blue boxes in the corners of the image, a sparse control grid for the hybrid field results in noticeably distorted renderings.}
    \label{fig:control_pts_resolution}
\end{figure}
}
% \begin{table}[htbp]
\begin{table}[t]
% \begin{wraptable}{R}{0.5\textwidth}
\centering
% of our algorithm 
% \todo{report the impact of sampled grid resolution on performance and training time.}
% }
\begin{tabular}{ccc}
\toprule
$\textbf{P}_c$ Resolution & PSNR~($\uparrow$)  & Time (mins) \\
\midrule
% $\downarrow 4 \times$ 
265 $\times$ 149 & 23.67 & 55 \\
% 265 $\times$ 149 & 23.21 & +57 \\
% $\downarrow 8 \times$ 
132 $\times$ 74 & 22.99 & 36 \\
% 132 $\times$ 74 & 22.99 & +17 \\
% $\downarrow 16 \times$ 
        66 $\times$ 37 & 22.44 & 25 \\
% 66 $\times$ 37 & 22.44 & +6 \\
% 33 $\times$ 18    & 21.68 & 25  \\
% 33 $\times$ 18    & 21.68 & +6  \\
\midrule
Explicit Grid & 14.93 & 22 \\
Fisheye-GS~\cite{liao2024fisheye} & 21.84 & 46 \\
3DGS~\cite{kerbl20233d} & 14.19 & 18 \\
% 32    & 21.68 & 25 mins \\
% 64    & 21.15 & 25 mins \\
\bottomrule
\end{tabular}%
\caption{\textbf{Ablation Study on Control Grid Resolution}. Parameter study on different control point resolutions, showing that our method has favorable cost/performance trade-off.}
\label{tab:ablation_control_pts}%
\end{table}%
% \end{wraptable}%

\paragraph{Control Grid Resolution.}
A higher-resolution control grid results in a smoother distortion field representation but slower training. To better illustrate the effect of control grid resolution, we visualize two types of resolutions (\textit{i.e.,} 265 $\times$ 149 and 66 $\times$ 37). While there are no significant differences in the central region, the distortion at the edges is better recovered with a higher-resolution grid. Since the distortion field becomes smoother, a high-resolution control grid produces more accurate distorted lines, as shown in the red and blue boxes of~\cref{fig:control_pts_resolution}.

\section{Extra Experiments}
{
\begin{table}[t]
  \centering
  \scalebox{0.75}{
    \begin{tabular}{ccccccccc}
    \toprule
    \multicolumn{1}{c}{\multirow{2}[0]{*}{Method}} & \multicolumn{3}{c}{Garden} & \multicolumn{3}{c}{Studio}\\
    \cmidrule(lr){2-4}
    \cmidrule(lr){5-7}
    \multicolumn{1}{c}{} & SSIM  & PSNR  & LPIPS & SSIM  & PSNR & LPIPS\\
    \midrule
    Fisheye-GS~\cite{liao2024fisheye} & 0.530 & 14.94 & 0.542 & 0.536 & 12.24 & 0.549\\
    Ours & \textbf{0.882} & \textbf{27.85} & \textbf{0.144} & \textbf{0.965} & \textbf{33.86} & \textbf{0.044}\\

    \bottomrule
    \end{tabular}%
  } 
    \caption{\textbf{Evaluation on Real-World Wide-Angle Captures}. We evaluate Fisheye-GS~\cite{liao2024fisheye} and our method on two real-world scenes captured with large FOV fisheye cameras. Our method outperforms the baseline by a significant margin.}
    \label{tab:quantitative_large_small_netflix}%
\end{table}%

\begin{table}[t]
  \centering
  \scalebox{0.8}{
    \begin{tabular}{cccccc}
        \toprule
        Method & Test View & Num & SSIM & PSNR & LPIPS \\
        \midrule
        Fisheye-GS~\cite{liao2024fisheye} & \multirow{2}{*}{Fisheye} & 100 & 0.630 & 15.94 & 0.463 \\
        % \multirow{4}{*}{Ours} 
        % && 10 & 0.770 & 22.93 & 0.230 \\
        % && 25 & 0.830 & 26.54 & 0.167 \\
        % && 50 & 0.851 & 28.14 & 0.152 \\
        Ours&& 100 & \textbf{0.886} & \textbf{30.72} & \textbf{0.146} \\
        \bottomrule
    \end{tabular}
  }
  \vspace{-2mm}
    \caption{\textbf{Evaluation on Mitsuba Scenes}. The comparison between our method and Fisheye-GS~\cite{liao2024fisheye} illustrates the expressiveness of our hybrid field.}
    \label{tab:quantitative_mitsuba}%
  \vspace{-6mm}
\end{table}
}
\subsection{Quantitative Comparisons with Fisheye-GS}
In addition to Fig.~5 in the main paper, we also provide a quantitative evaluation of our method compared with the baseline Fisheye-GS~\cite{liao2024fisheye} in~\cref{tab:quantitative_large_small_netflix} and~\cref{tab:quantitative_mitsuba}. 
The performance degradation observed in the baseline method is primarily due to the limitations of the conventional camera distortion model used during reconstruction, which struggles at the edges of large FOVs. 
As a result, there is no geometric consistency in the peripheral regions to produce uniform gradients for optimizing the Gaussians.
When rendering novel views, these Gaussians appear as large floaters that occlude the camera view. We further visualize this phenomenon in the failure case video reconstructed by Fisheye-GS~\cite{liao2024fisheye}. The center area is revealed as we gradually decrease the scale of the Gaussians.
Please refer to our supplementary video ``fisheye-gs\_failure.mp4" for a comparison with the parametric distortion model.

{\begin{figure}[t]
    \centering
    \setlength{\tabcolsep}{1pt} % Adjust space between columns if needed
    \scalebox{0.95}{
    \begin{tabular}{ccc} % 4 columns (Vertical Caption | Image Set 1 | Image Set 2 | Image Set 3)

        \includegraphics[width=0.158\textwidth]{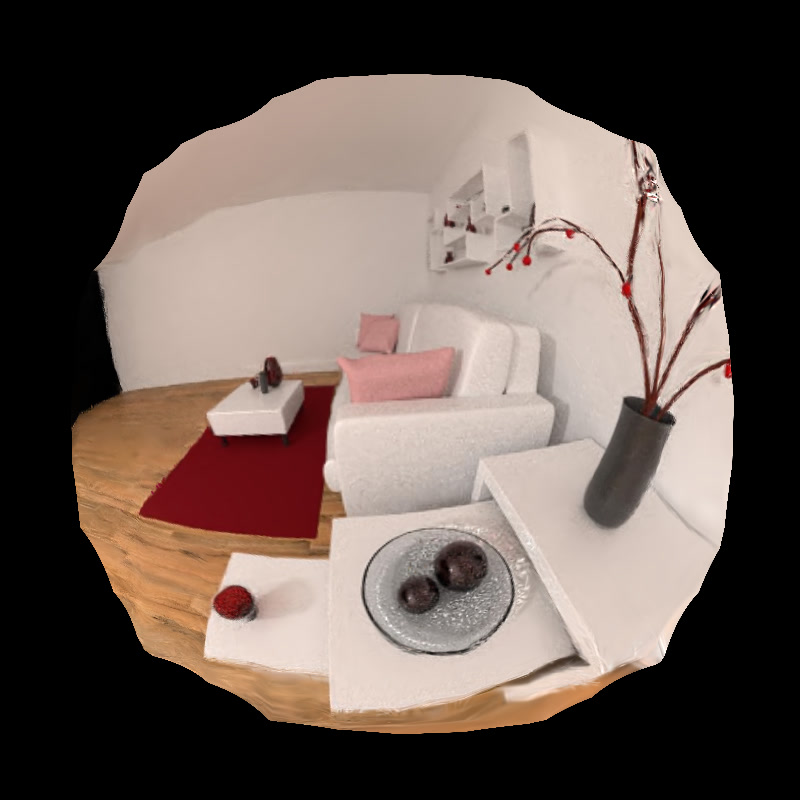} &
        \includegraphics[width=0.158\textwidth]{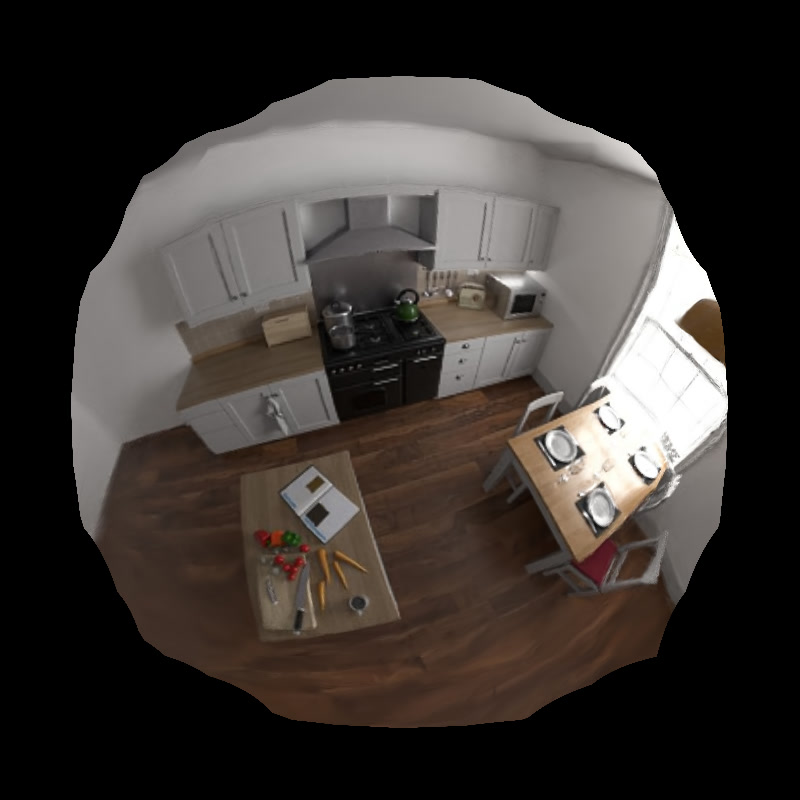} &
        \includegraphics[width=0.158\textwidth]{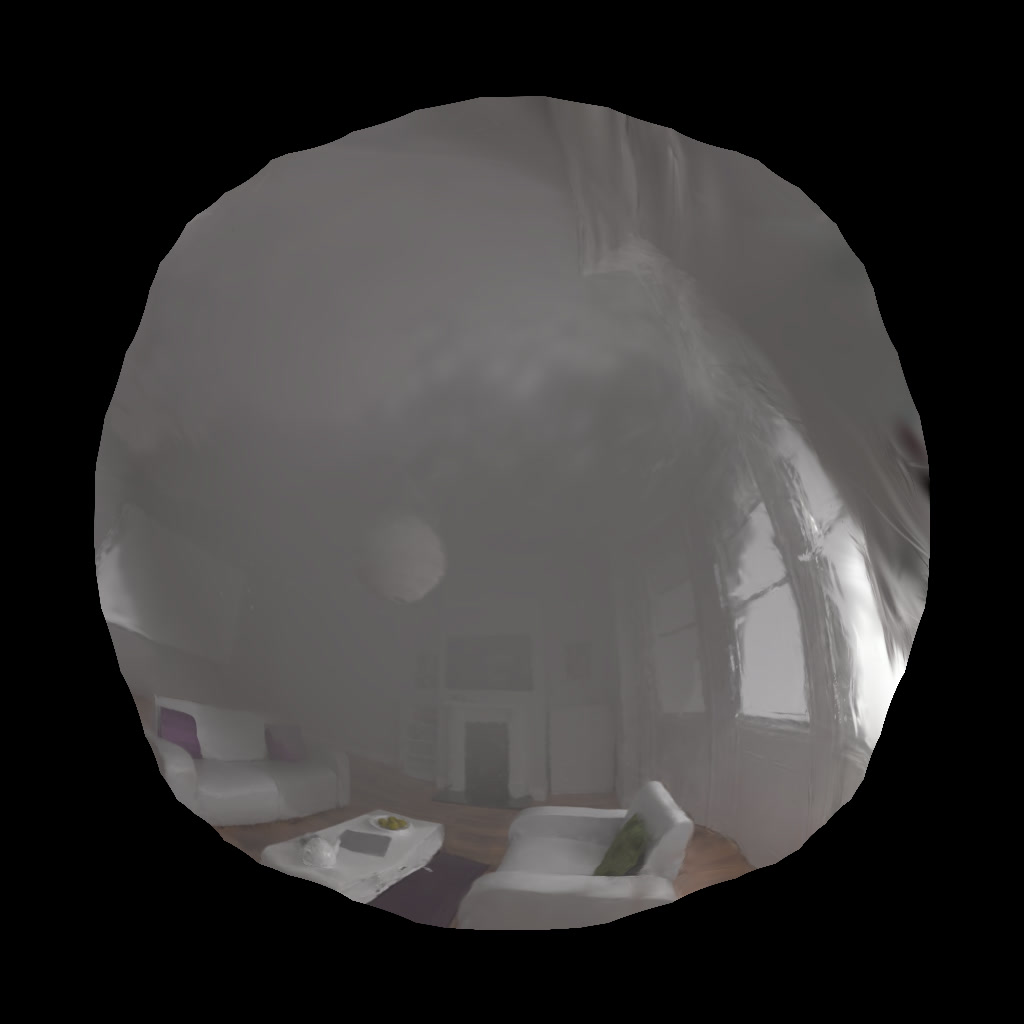}
        \\

        \includegraphics[width=0.158\textwidth]{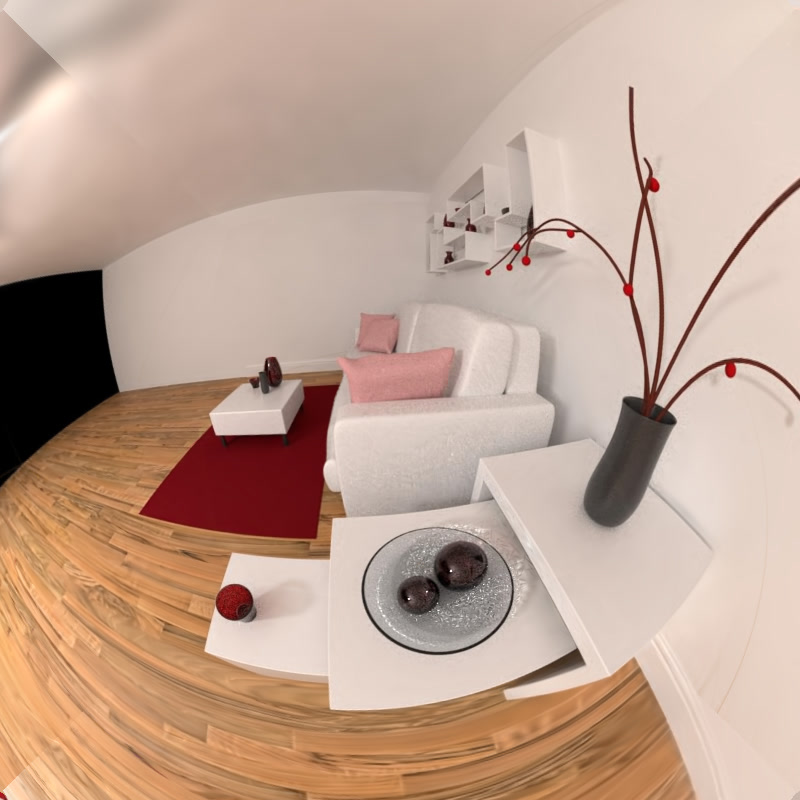} &
        \includegraphics[width=0.158\textwidth]{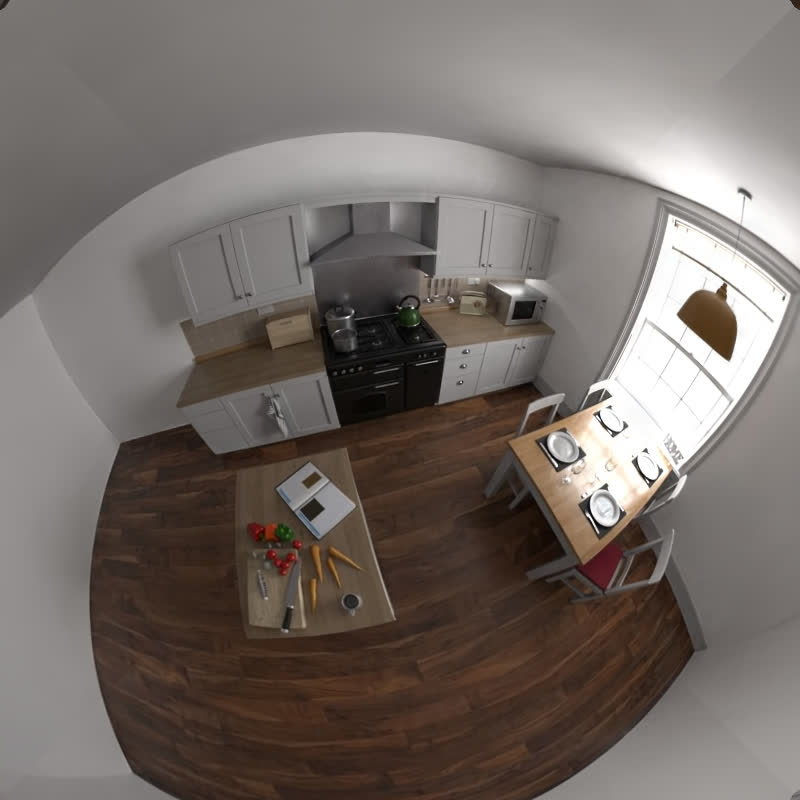} &
        \includegraphics[width=0.158\textwidth]{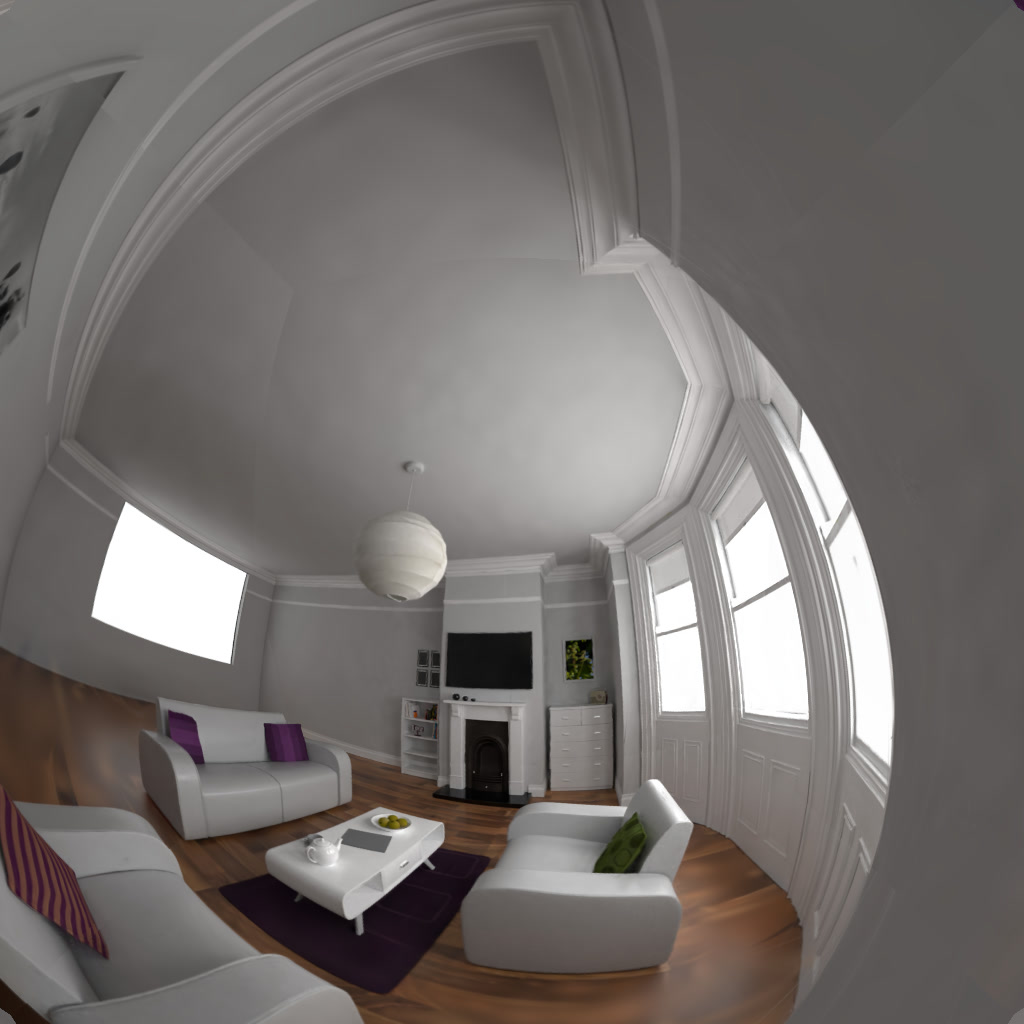}
        \\
        \multicolumn{1}{c}{(a) Living} & \multicolumn{1}{c}{(b) Kitchen} & \multicolumn{1}{c}{(c) Room}
        \\
    \end{tabular}
    }
    \caption{\textbf{Single Planar Projection with Hybrid Field}. Our hybrid field can be directly applied to a single plane during rasterization. However, the limitation of single planar projection is that it cannot cover the full FOV of the raw images, leading to partial loss of information in the peripheral regions.}
    \label{fig:ablation_cubemap}
\end{figure}
}

\subsection{Qualitative Results of Cubemap Ablation}
As illustrated in Fig.~2 of the main paper, we apply a cubemap to overcome the limitations of perspective projection. Even with our hybrid distortion field, we can only utilize the central region of raw fisheye images, as shown in the bottom row of~\cref{fig:ablation_cubemap}. In contrast, rendering a cubemap enables us to achieve a larger FOV, allowing us to compute the photometric loss with reference raw images for optimization, as demonstrated in Fig.~7 of the main paper. We also provide a quantitative evaluation in Tab.~4 of the main paper, showing that the final reconstruction quality is compromised when using a single planar projection.

Additionally, the boundary of the final distorted rendering is irregular, primarily because the distortion information at the boundary heavily relies on extra regions that are not covered by single planar projections.

\subsection{Qualitative Comparisons with Different Numbers of Input Views}
We report quantitative results in Tab.~2 of the main paper, where our method outperforms the baseline even with significantly fewer input views, down to 10-15\%. 

\begin{figure*}[t]
    \centering
    \setlength{\tabcolsep}{1pt}
    \scalebox{1}{
    \begin{tabular}{cccc}
        \includegraphics[width=0.23\textwidth]{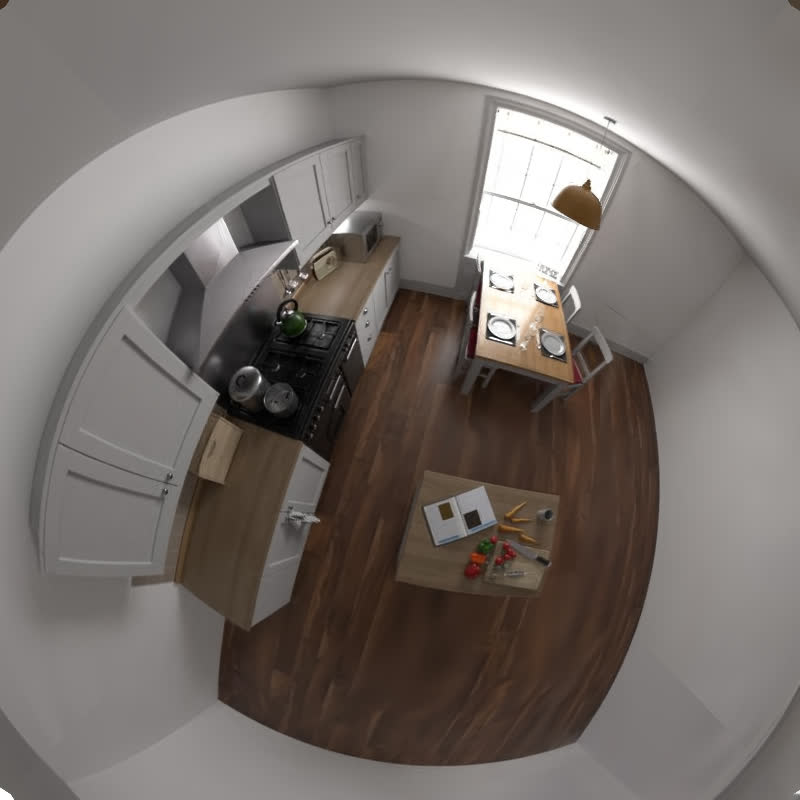} &
        \includegraphics[width=0.23\textwidth]{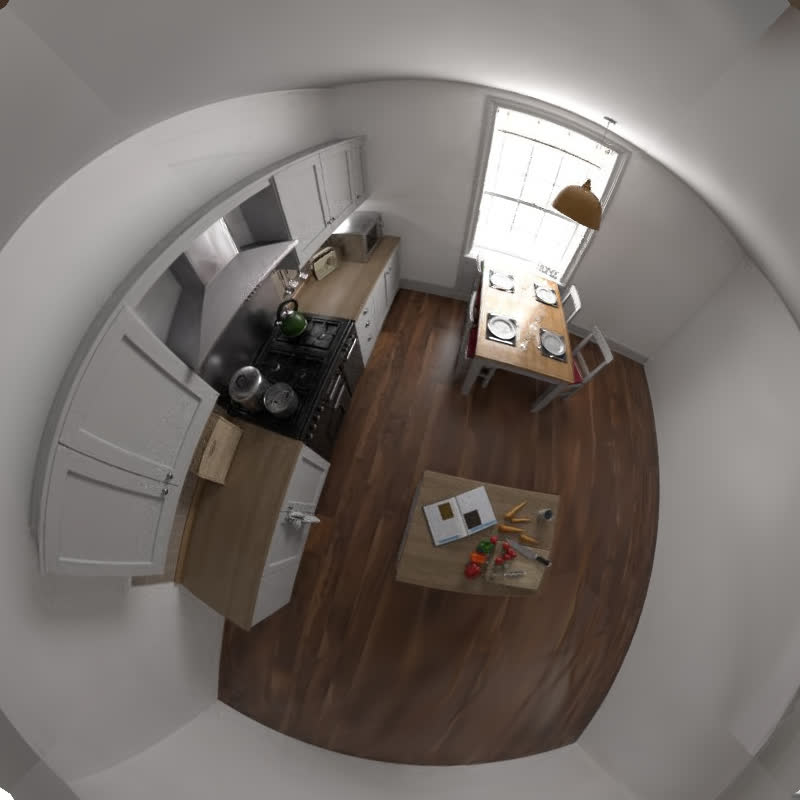} &
        \includegraphics[width=0.23\textwidth]{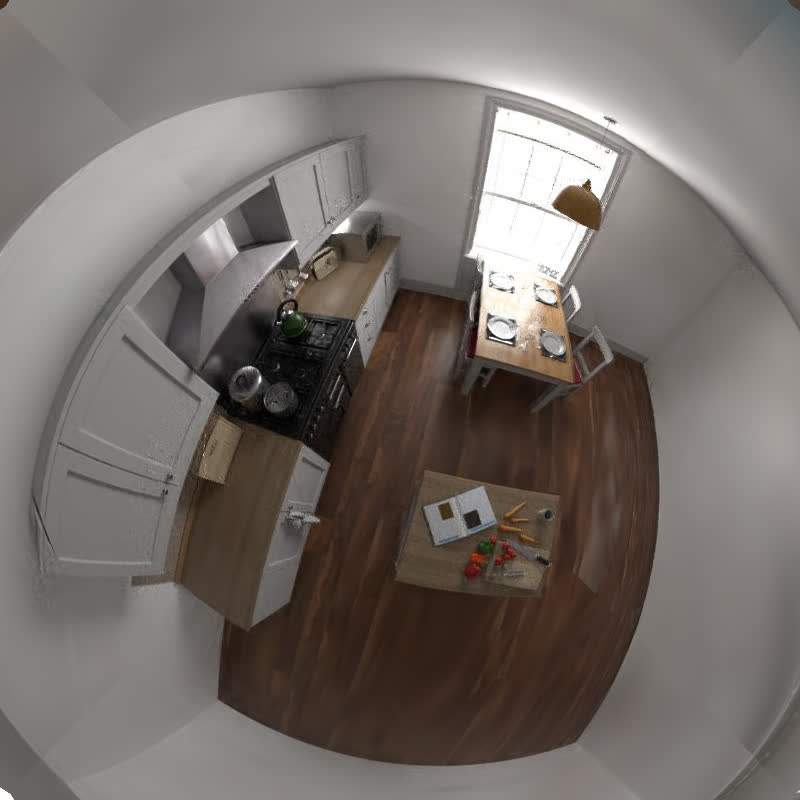} &
        \includegraphics[width=0.23\textwidth]{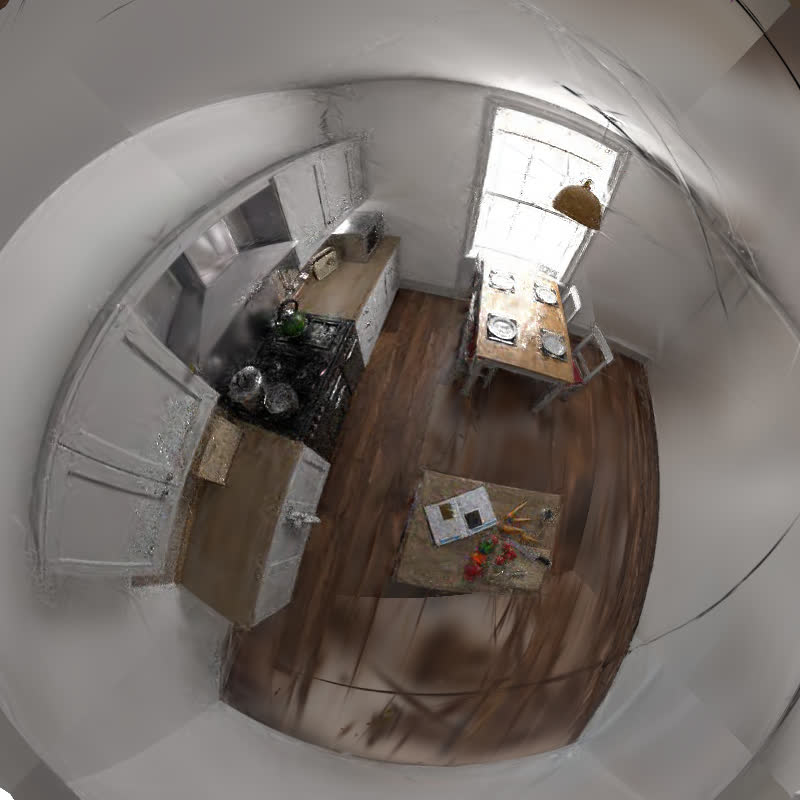}
        \\

        \includegraphics[width=0.23\textwidth]{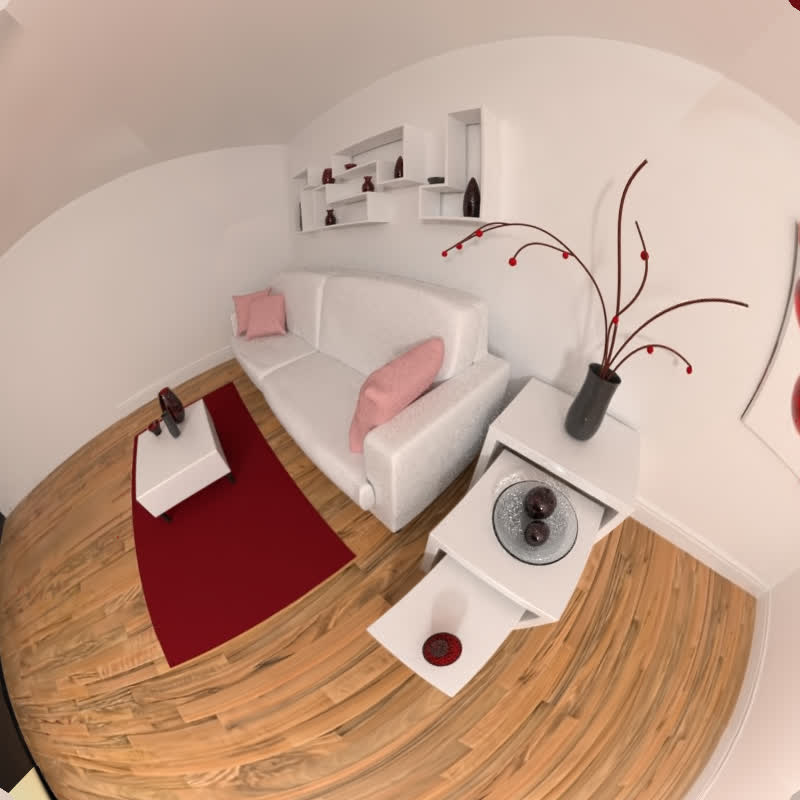} &
        \includegraphics[width=0.23\textwidth]{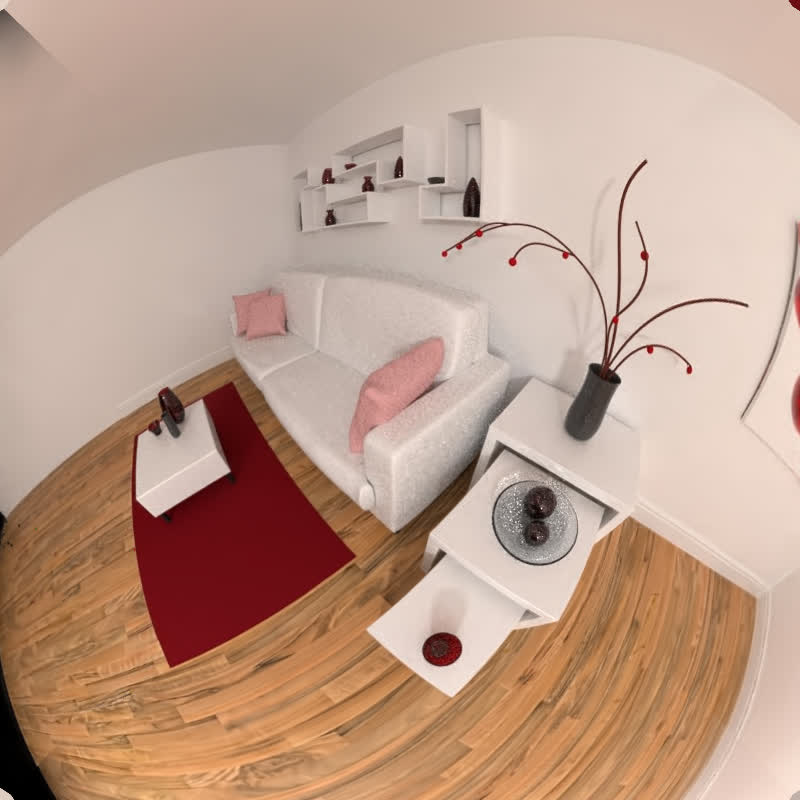} &
        \includegraphics[width=0.23\textwidth]{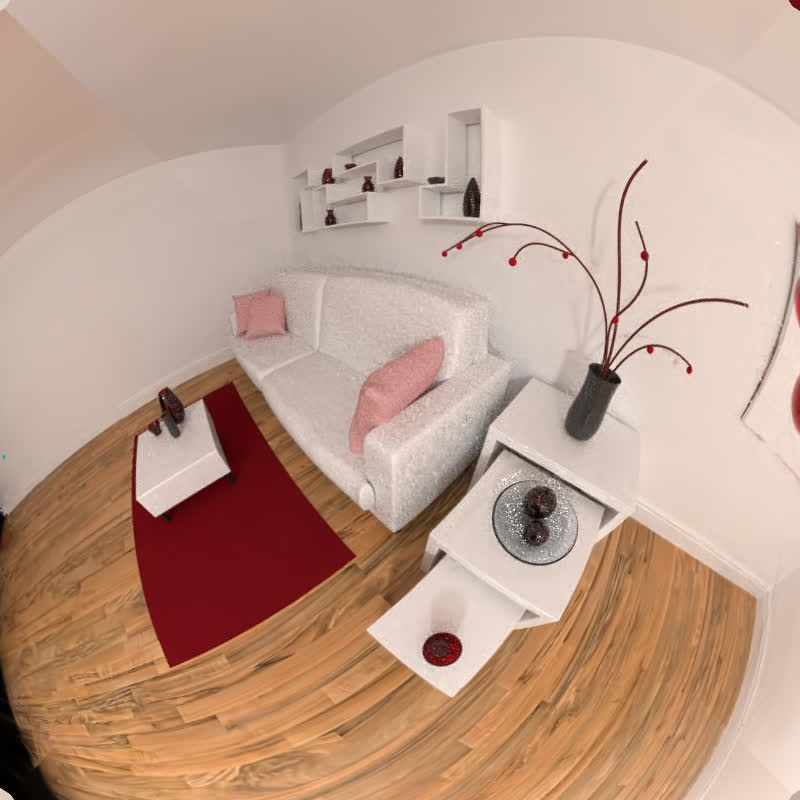} &
        \includegraphics[width=0.23\textwidth]{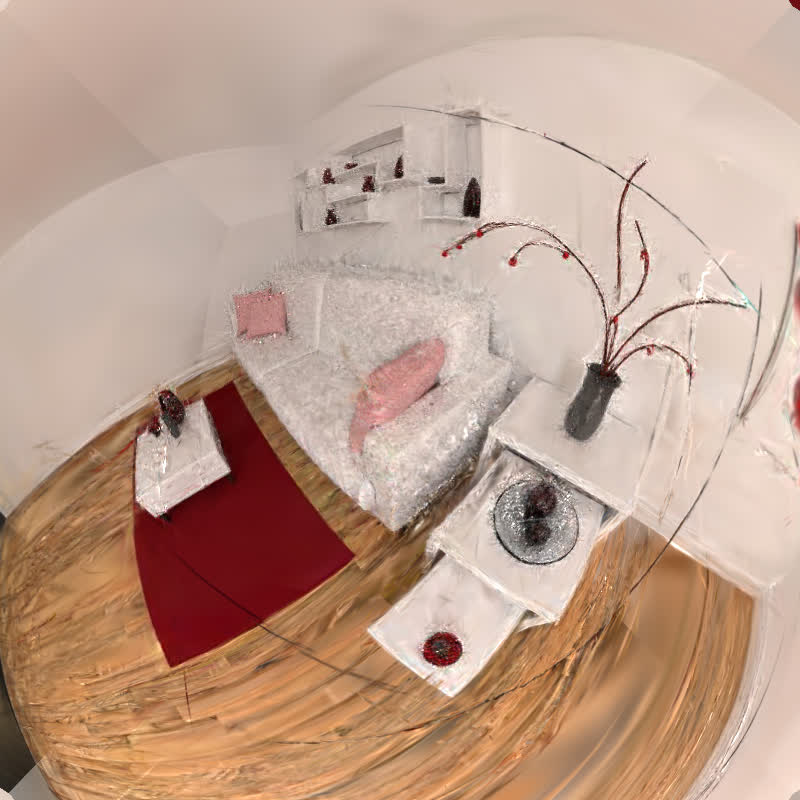}
        \\
        \multicolumn{1}{c}{(a) 100 Views} & \multicolumn{1}{c}{(b) 50 Views} & \multicolumn{1}{c}{(c) 25 Views} & \multicolumn{1}{c}{(d) 10 Views}
        \\
    \end{tabular}
    }
    \caption{\textbf{Qualitative Evaluation of Reconstruction with Varying Numbers of Large FOV Inputs}. Our method achieves high-quality reconstruction even with a relatively small number of input images, thanks to our hybrid distortion representation and cubemap resampling.}
    \label{fig:view_num}
\end{figure*}

To illustrate the impact of varying input view counts, we visualize the rendering quality as the number of input views decreases. Even with as few as 25 input views, our method still achieves reasonable performance, as shown in~\cref{fig:view_num}. This demonstrates our method's ability to effectively utilize large FOV cameras and achieve comprehensive scene coverage during reconstruction.

\subsection{Comparisons with Regular FOV Cameras}
\paragraph{Synthetic Mitsuba Scene Captures.} To carefully control the experimental settings, we customized a camera module in the Mitsuba ray tracer~\cite{jakob2010mitsuba} using camera parameters derived from DSLR lenses, as profiled in the open-source Lensfun~\cite{lensfun}. 
We also utilize 3D assets, including geometry, materials, and lighting, from~\cite{resources16} to produce renderings.
Our synthetic dataset contains three large indoor scenes and four object-centric scenes. 
All indoor scenes follow a Sigma 180\si{\degree} circular fisheye camera. 
Two of the object scenes are rendered with a 120\si{\degree} fisheye lens, while the others are rendered with classic radial distortion. 

To capture detailed perspectives of the scene, cameras are placed close to the objects at varying distances. For Room2, since objects are uniformly distributed in the space, we placed a set of camera centers along a Hilbert curve, then oriented each fixed camera center to cover the surroundings. The number of images captured at each point is reduced for 180\si{\degree} images compared to those with a 90\si{\degree} FOV, as shown in the number of captures in~\cref{tab:decrease_number_each_scene}.

\begin{table}[t]
  \centering
  \scalebox{0.53}{
    \begin{tabular}{ccccccccccc}
        \toprule
        \multicolumn{1}{c}{\multirow{2}[0]{*}{Method}} & \multicolumn{1}{c}{\multirow{2}[0]{*}{Num}} & \multicolumn{3}{c}{Mitsuba Kitchen} & \multicolumn{3}{c}{Mitsuba Room1} & \multicolumn{3}{c}{Mitsuba Room2}\\
        \cmidrule(lr){3-5}
        \cmidrule(lr){6-8}
        \cmidrule(lr){9-11}
        \multicolumn{1}{c}{} & & SSIM  & PSNR  & LPIPS & SSIM  & PSNR & LPIPS & SSIM  & PSNR & LPIPS\\
        \midrule
        3DGS~\cite{kerbl20233d} & 200 & 0.470 & 11.09 & 0.435 & 0.595 & 15.27 & 0.406 & 0.897 & 30.88 & 0.155 \\
        \multirow{4}{*}{Ours} & 10 & 0.644 & 23.03 & 0.346 & 0.556 & 22.26 & 0.377 & 0.644 & 23.03 & 0.346 \\
        & 25 & 0.689 & 24.75 & 0.303 & 0.613 & 24.21 & 0.318 & 0.825 & 24.81 & 0.254 \\
        & 50 & 0.708 & 25.44 & 0.285 & 0.640 & 25.07 & 0.301 & 0.856 & 28.26 & 0.214 \\
        & 100 & \textbf{0.794} & \textbf{27.56} & \textbf{0.272} & \textbf{0.686} & \textbf{26.27} & \textbf{0.314} & \textbf{0.920} & \textbf{33.21} & \textbf{0.107} \\
        \midrule
        Fisheye-GS~\cite{liao2024fisheye} & 100 & 0.601 & 14.30 & 0.485 & 0.549 & 15.54 & 0.548 & 0.739 & 17.97 & 0.355\\
        \multirow{4}{*}{Ours} & 10 & 0.801 & 26.04 & 0.202 & 0.703 & 23.69 & 0.232 & 0.806 & 19.06 & 0.256 \\
        & 25 & 0.855 & 28.62 & 0.155 & 0.783 & 26.71 & 0.180 & 0.853 & 24.28 & 0.167\\
        & 50 & 0.870 & 29.55 & 0.151 & 0.801 & 27.57 & 0.172 & 0.883 & 27.29 & 0.133\\
        & 100 & \textbf{0.886} & \textbf{30.72} & \textbf{0.146} & \textbf{0.842} & \textbf{28.46} & \textbf{0.180} & \textbf{0.929} & \textbf{31.16} & \textbf{0.095}\\
        \bottomrule
    \end{tabular}%
  } 
    \caption{\textbf{Evaluation on Mitsuba Synthetic Scenes}. We compare our method with vanilla 3DGS~\cite{kerbl20233d} and Fisheye-GS~\cite{liao2024fisheye} on a set of held-out captures. Since vanilla 3DGS does not support fisheye rendering, we render several perspective images at the same locations and look-at directions for comparison. We directly compare the fisheye rendering results with both Fisheye-GS and our method.}
    \label{tab:decrease_number_each_scene}%
\end{table}%

{\begin{figure}[t]
    \centering
    \setlength{\tabcolsep}{1pt} % Adjust space between columns if needed
    \scalebox{0.95}{
    \begin{tabular}{cccccc} % 4 columns (Vertical Caption | Image Set 1 | Image Set 2 | Image Set 3)

        \raisebox{0.7\height}{\hspace{-0.5cm}\rotatebox{90}{{Room1}}} &
        \includegraphics[width=0.15\textwidth]{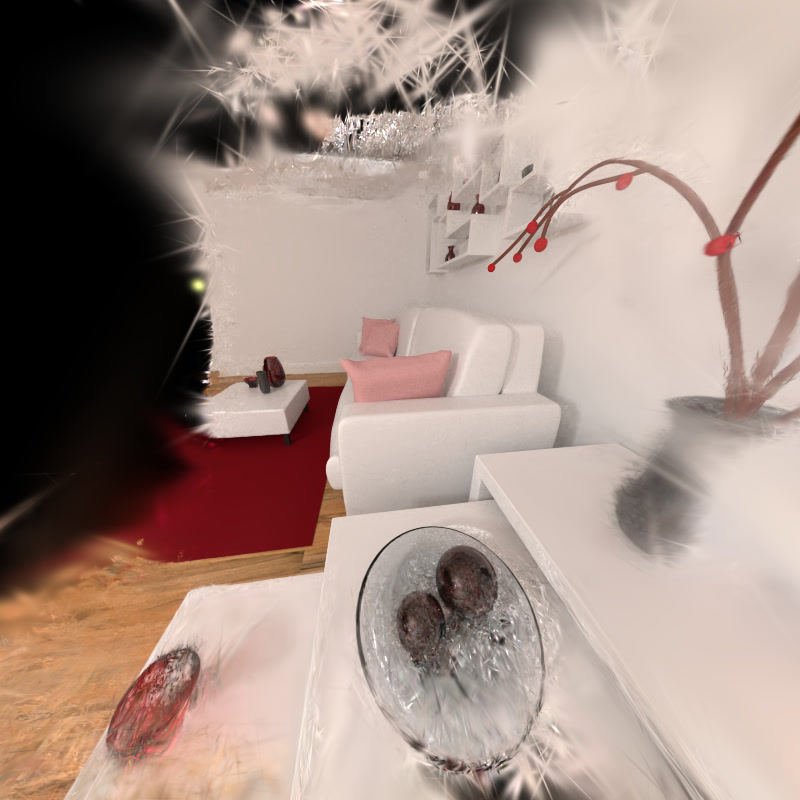} &
        \includegraphics[width=0.15\textwidth]{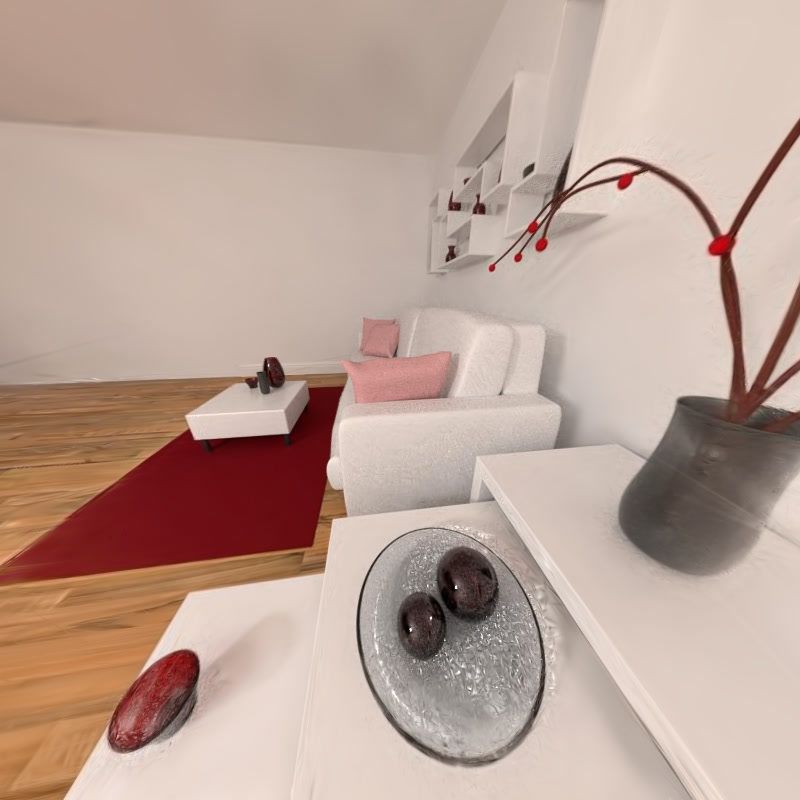} &
        \includegraphics[width=0.15\textwidth]{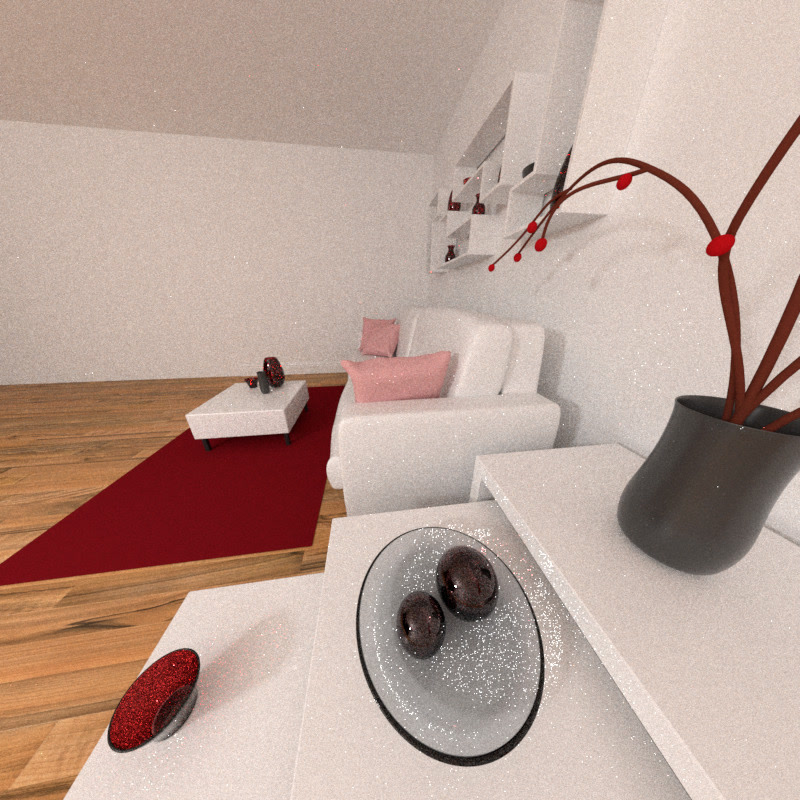}
        \\

        \raisebox{0.5\height}{\hspace{-0.5cm}\rotatebox{90}{{Kitchen}}} &
        \includegraphics[width=0.15\textwidth]{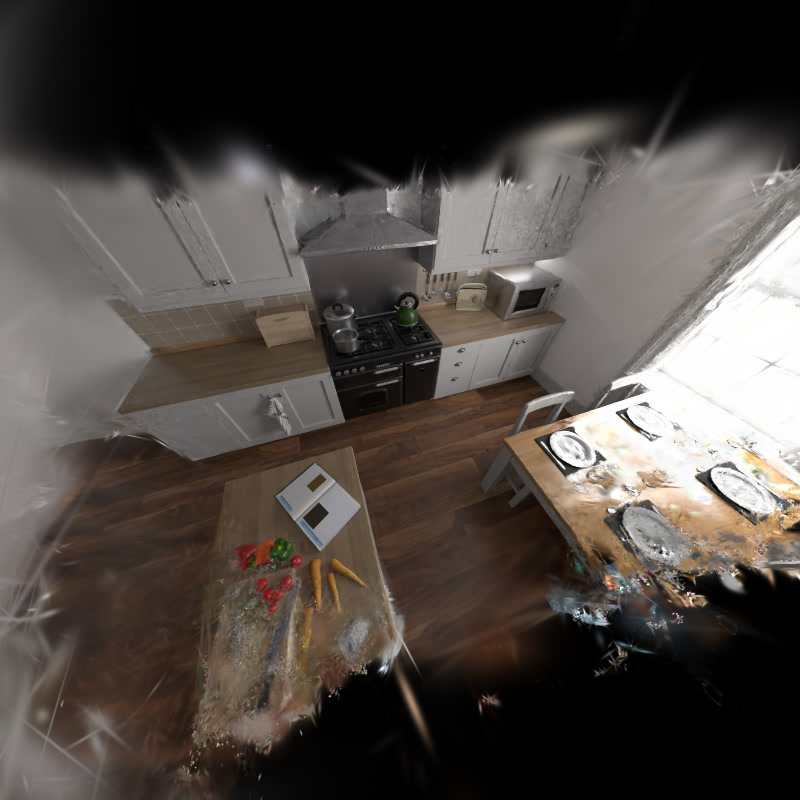} &
        \includegraphics[width=0.15\textwidth]{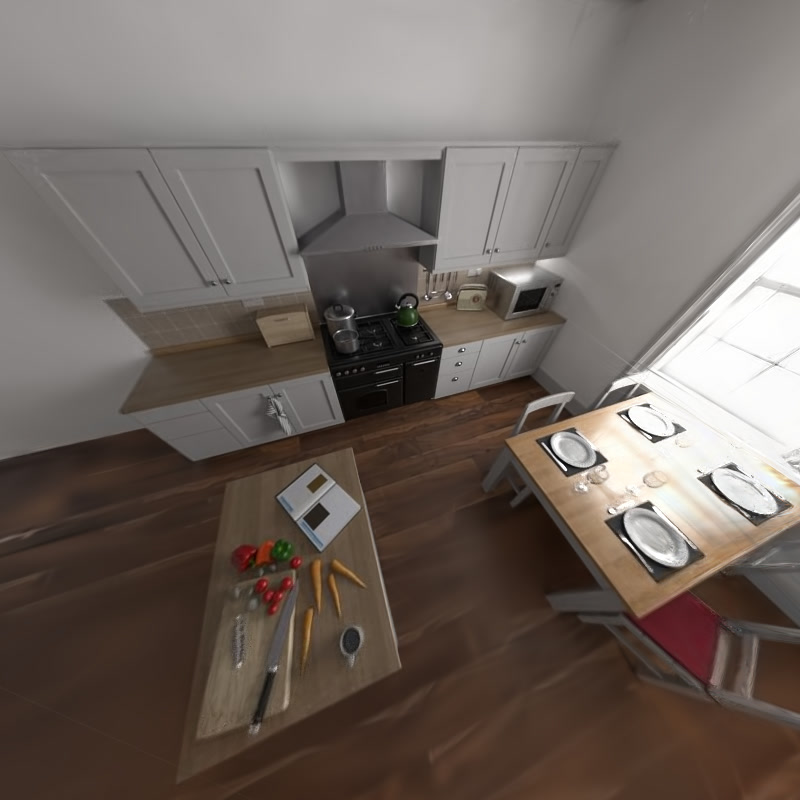} &
        \includegraphics[width=0.15\textwidth]{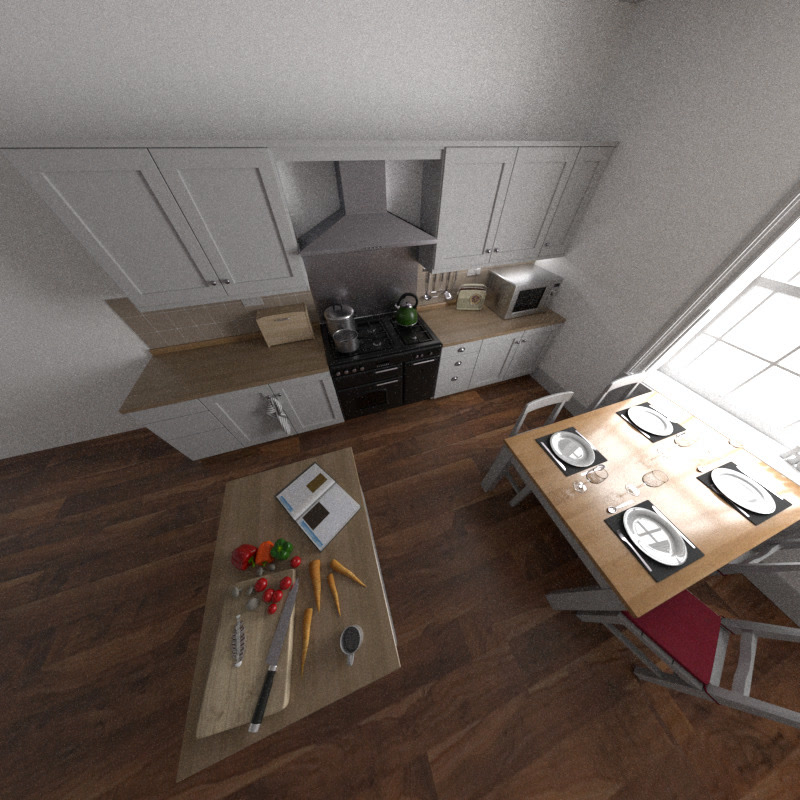}
        \\
                
        \raisebox{0.8\height}{\hspace{-0.5cm}\rotatebox{90}{{Room2}}} &
        \includegraphics[width=0.15\textwidth]{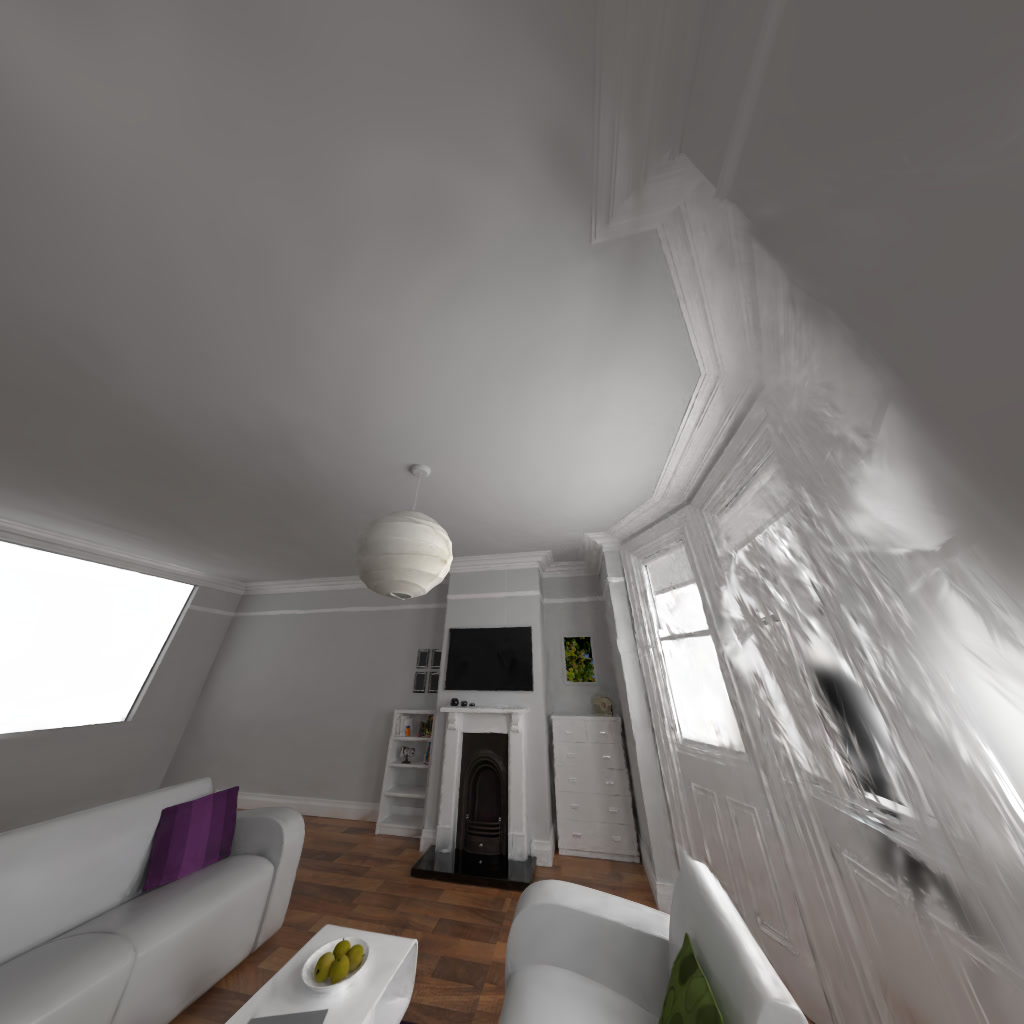} &
        \includegraphics[width=0.15\textwidth]{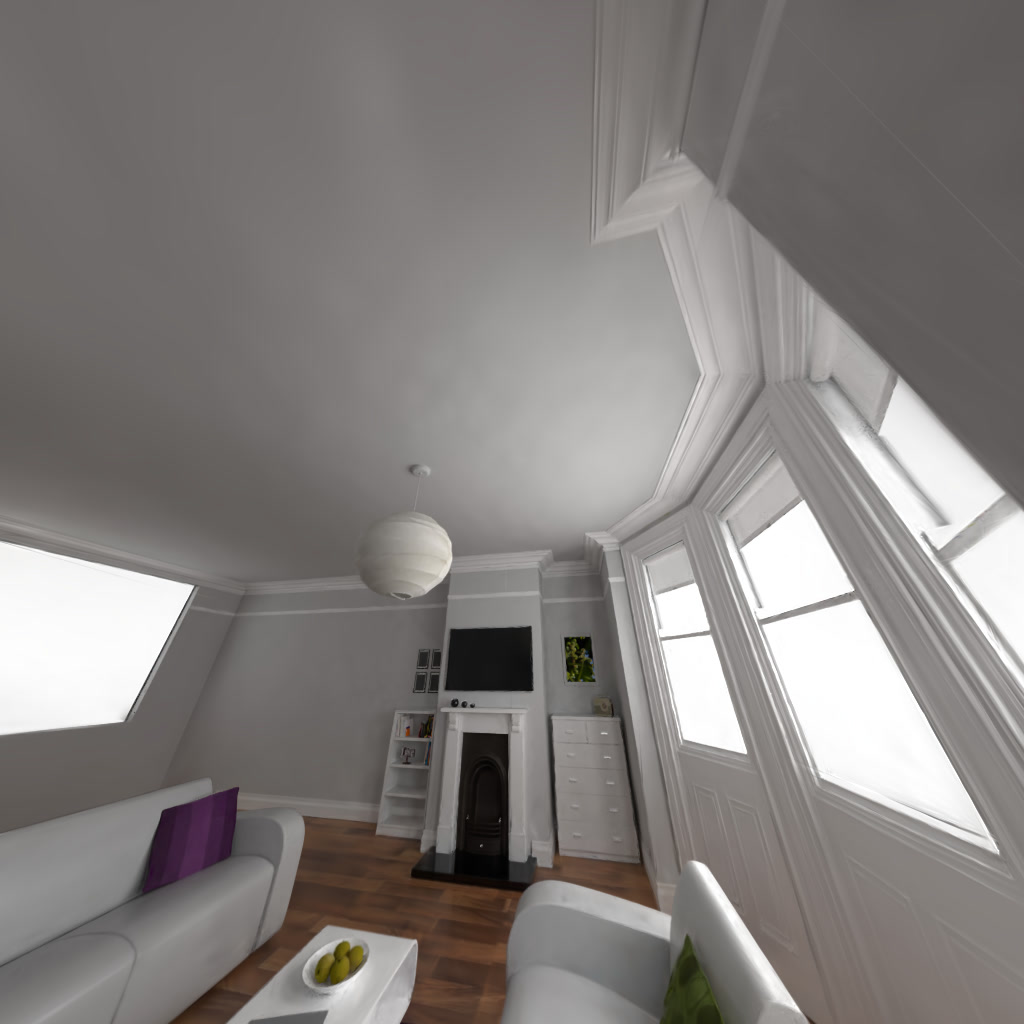} &
        \includegraphics[width=0.15\textwidth]{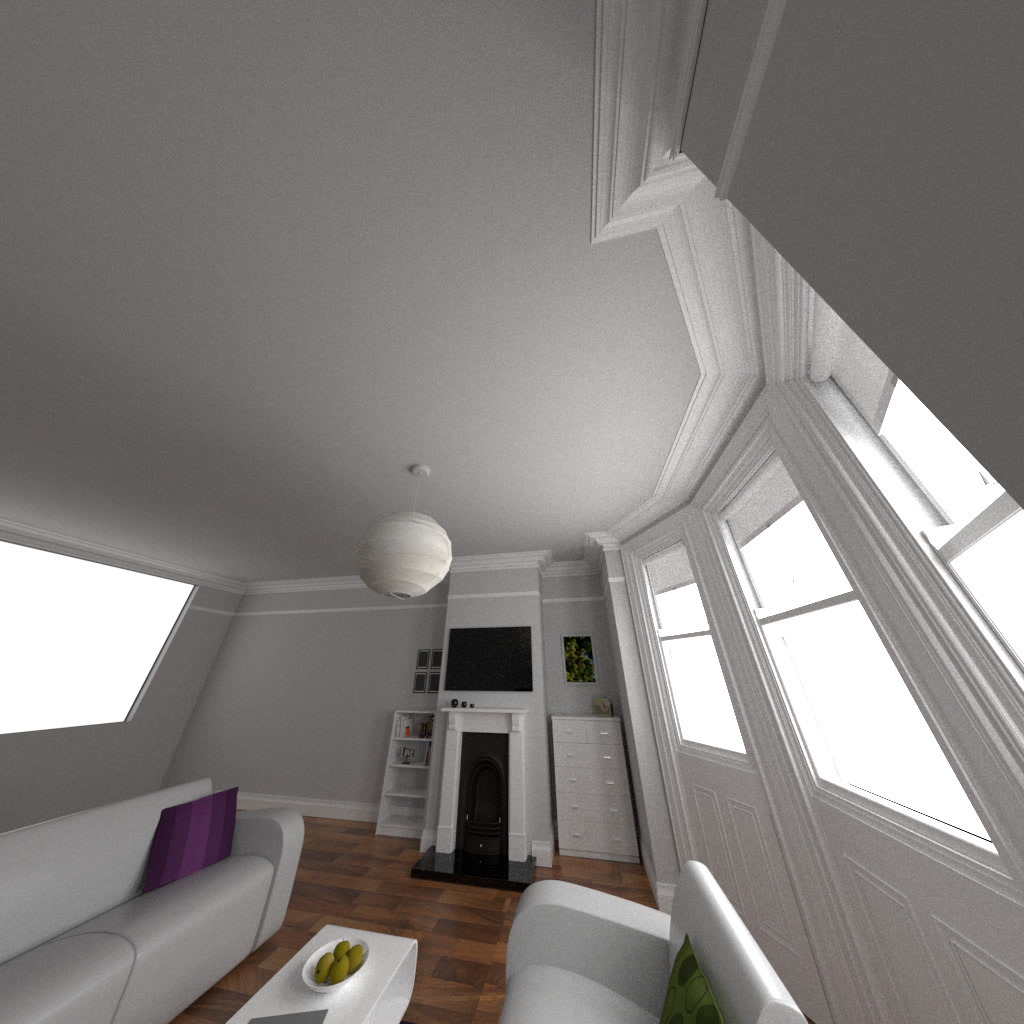}
        \\
        
        \multicolumn{1}{c}{} & \multicolumn{1}{c}{(a) Small FOV} & \multicolumn{1}{c}{(b) Ours} & \multicolumn{1}{c}{(b) GT}
        \\
    \end{tabular}
    }
    \caption{\textbf{Evaluation of Perspective Rendering}. After reconstruction, our method can render perspective views with arbitrary FOV. We compare the perspective renderings produced by our method with those rendered from small-FOV reconstructions using 3DGS~\cite{kerbl20233d}.}
    \label{fig:largefov_vs_smallfov_mitsuba_in_perspective}

\end{figure}
}

{
\begin{figure*}[t]
    \centering
    \setlength{\tabcolsep}{1pt} % Adjust space between columns if needed
    \begin{tabular}{ccc} % 5 columns
        \includegraphics[height=0.3\textwidth]{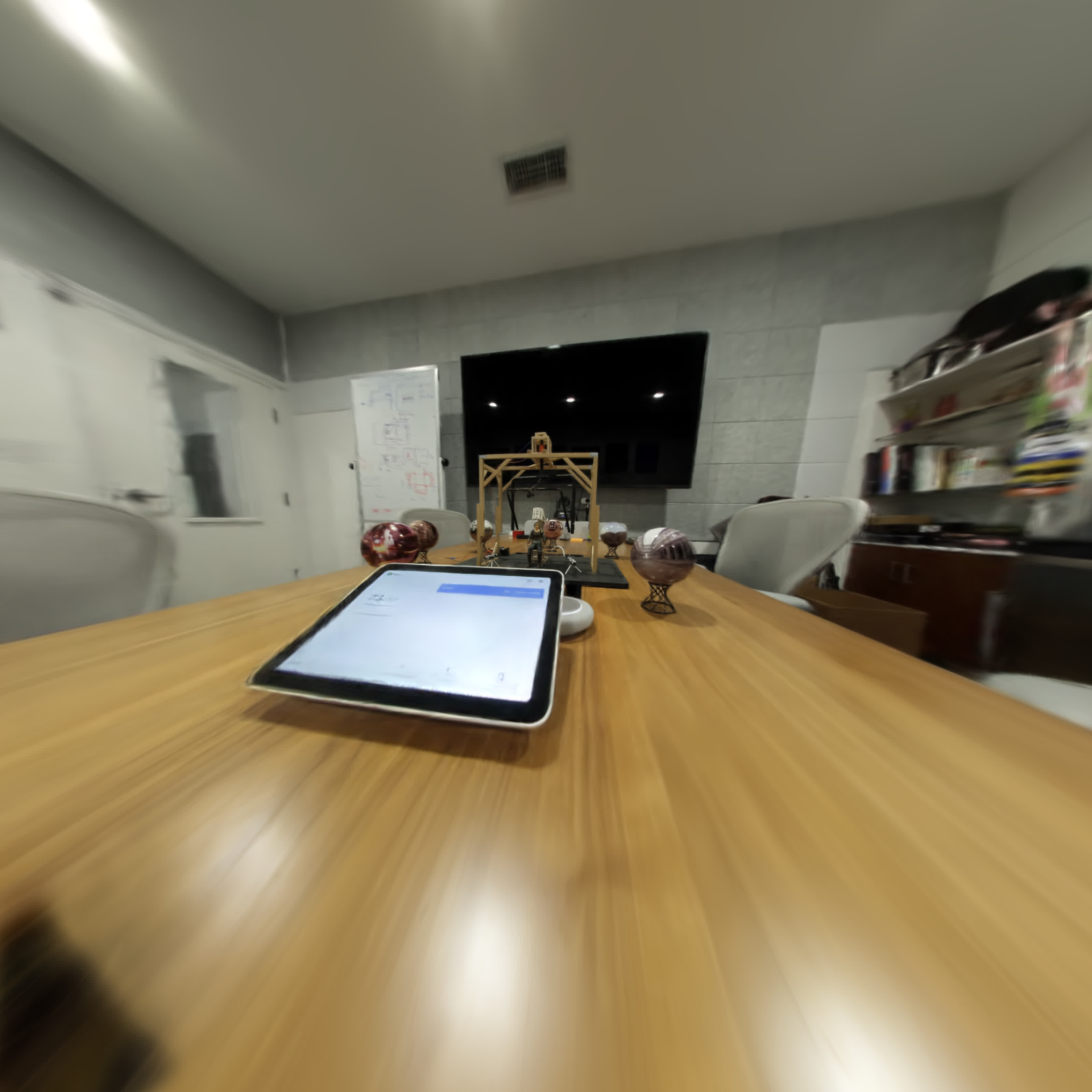} &
         \includegraphics[height=0.3\textwidth]{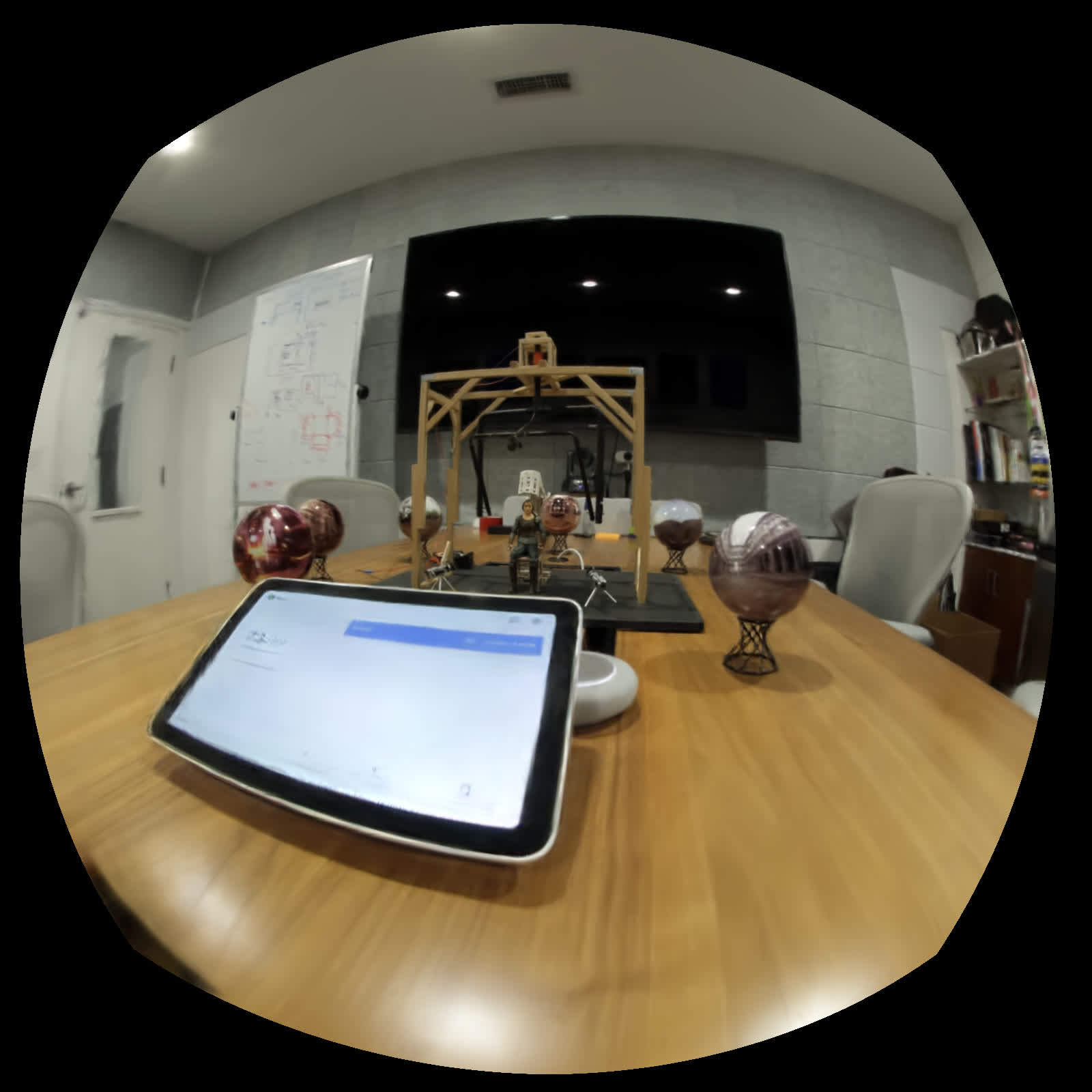} &
        \includegraphics[height=0.3\textwidth]{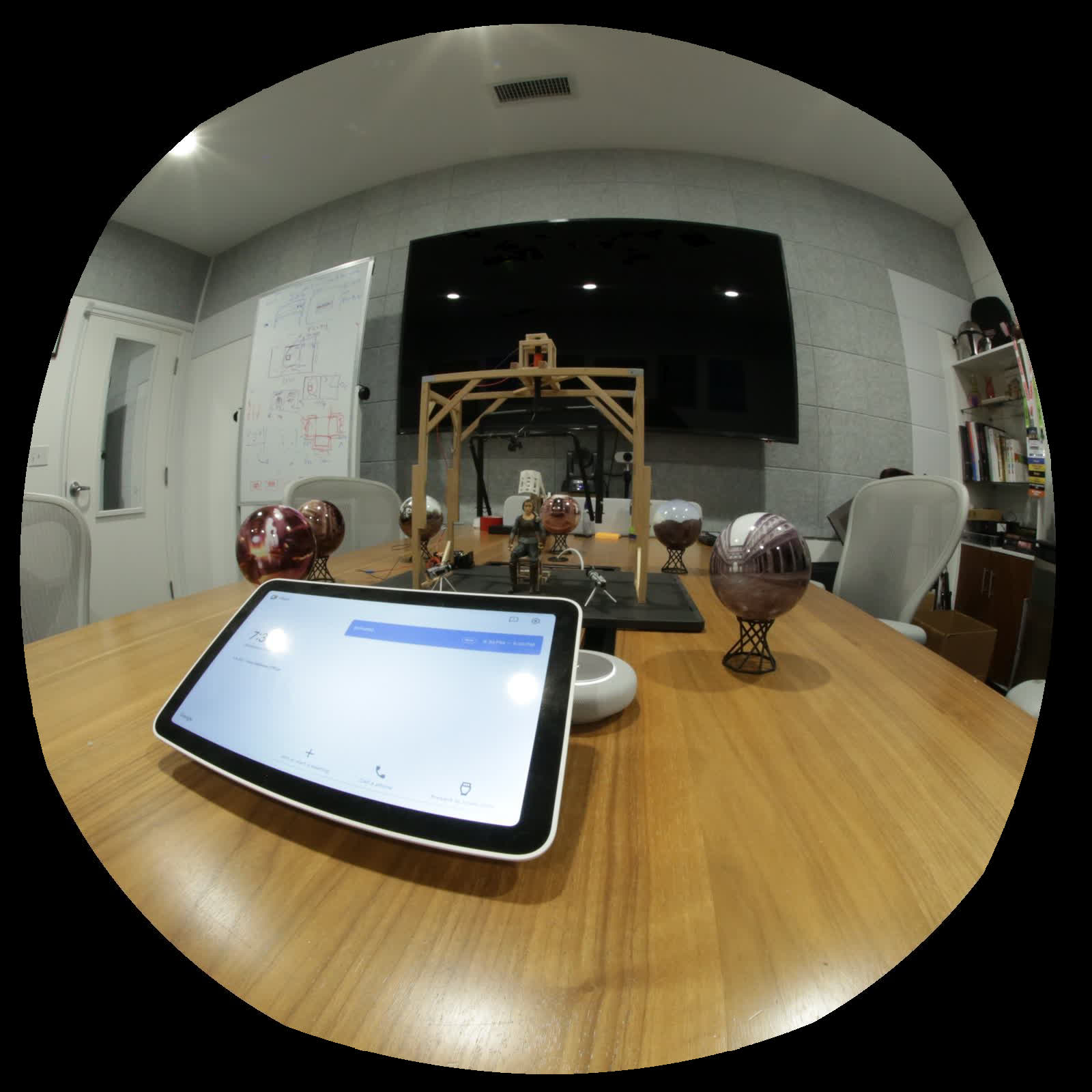} \\

        \includegraphics[height=0.3\textwidth]{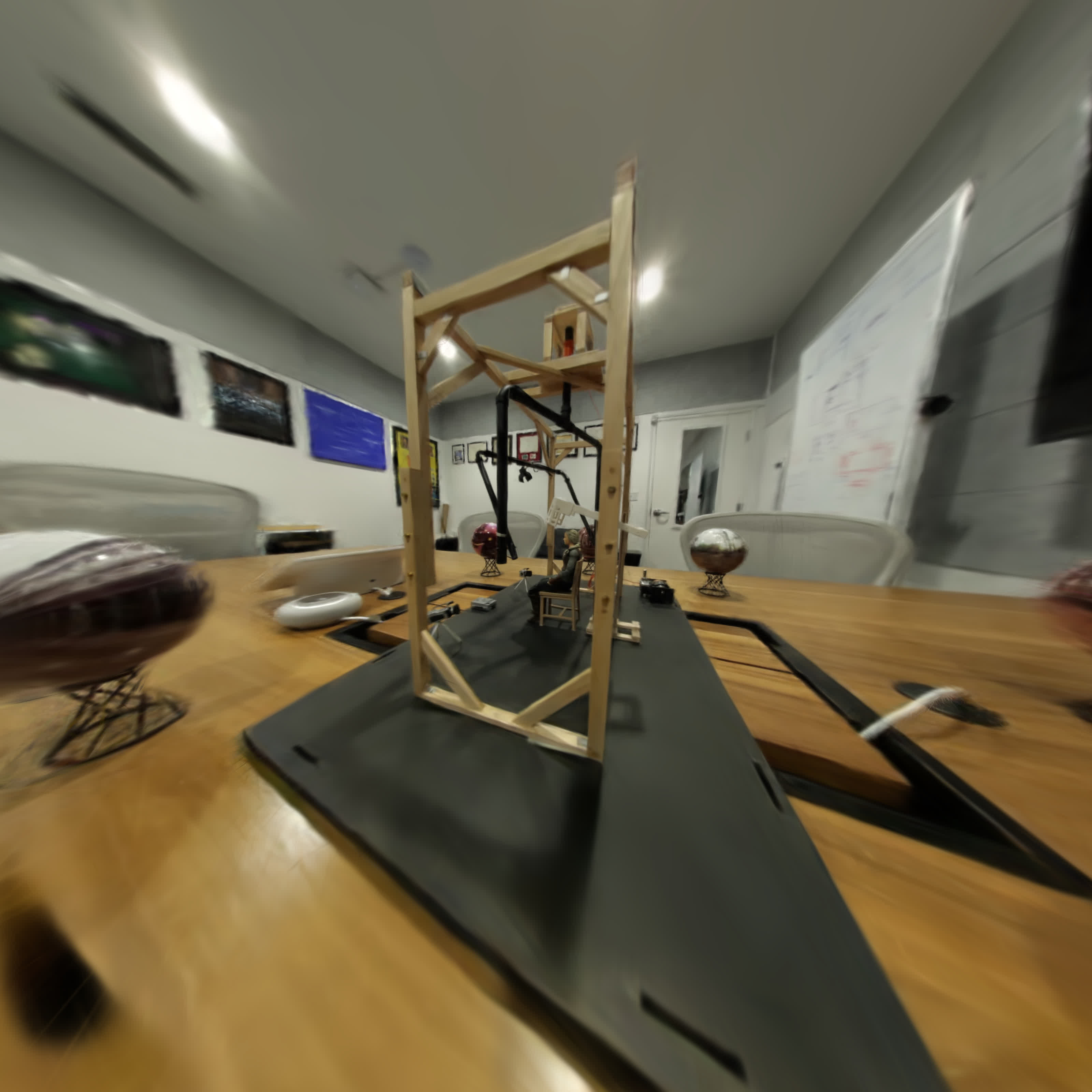} &
         \includegraphics[height=0.3\textwidth]{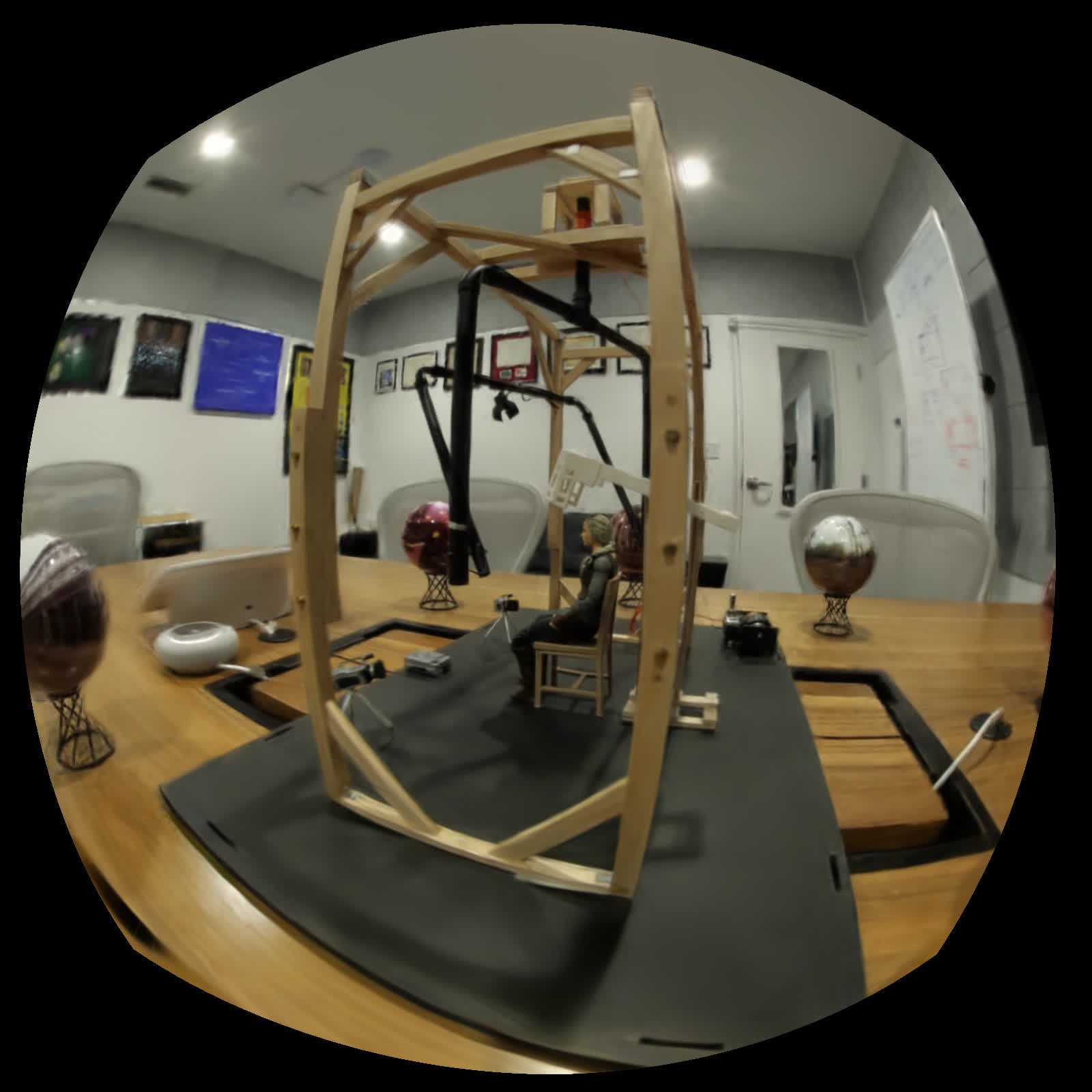} &
        \includegraphics[height=0.3\textwidth]{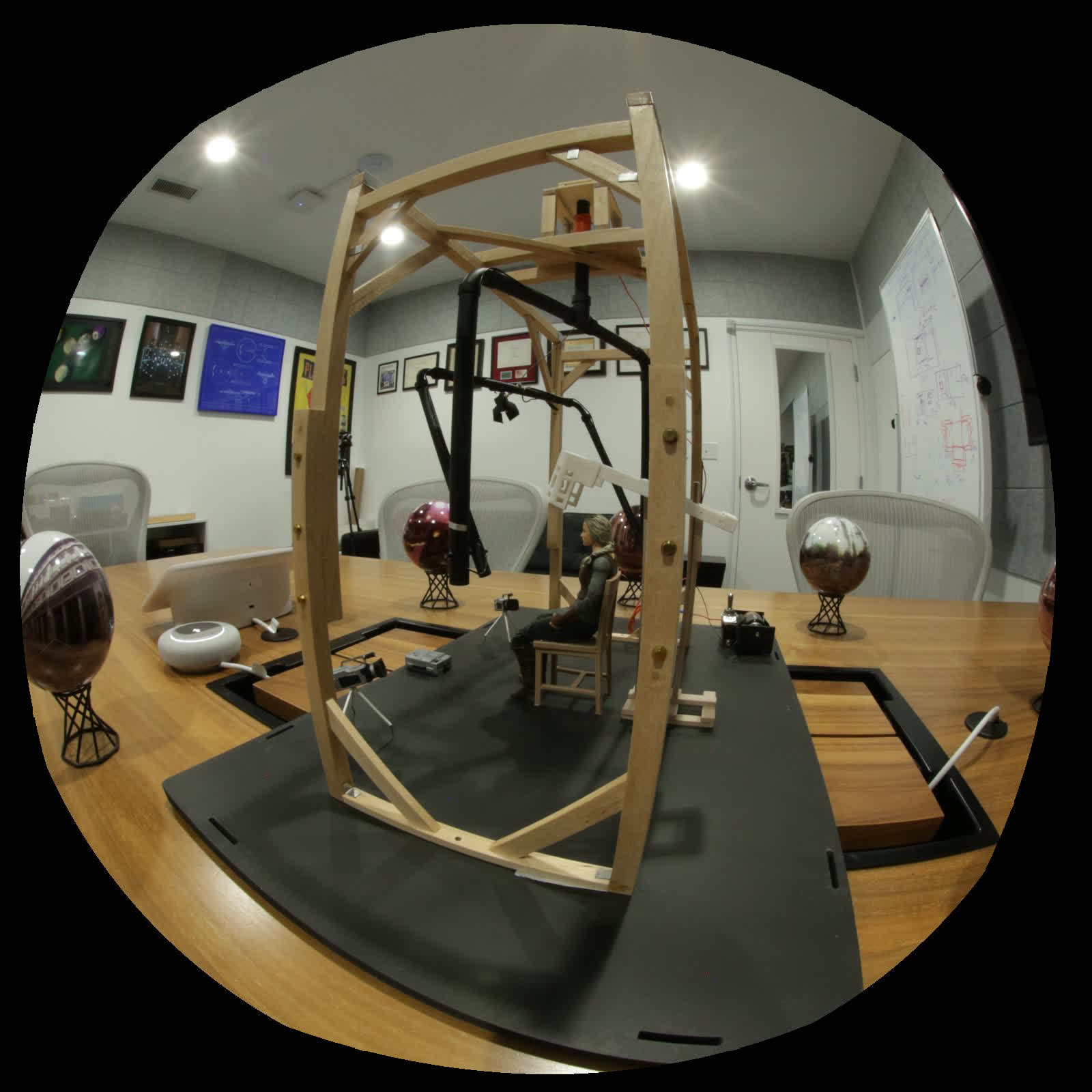} \\

        \includegraphics[height=0.3\textwidth]{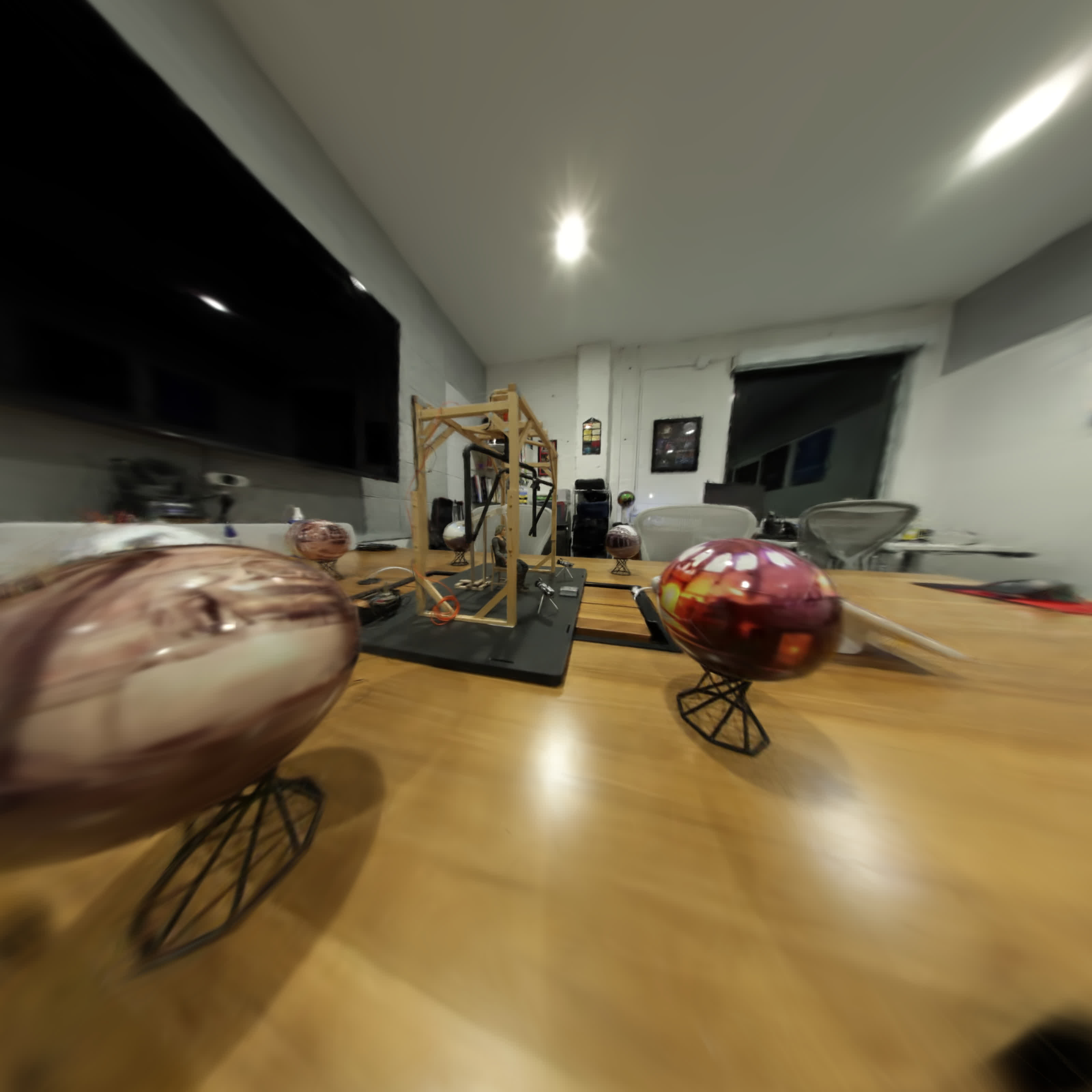} &
         \includegraphics[height=0.3\textwidth]{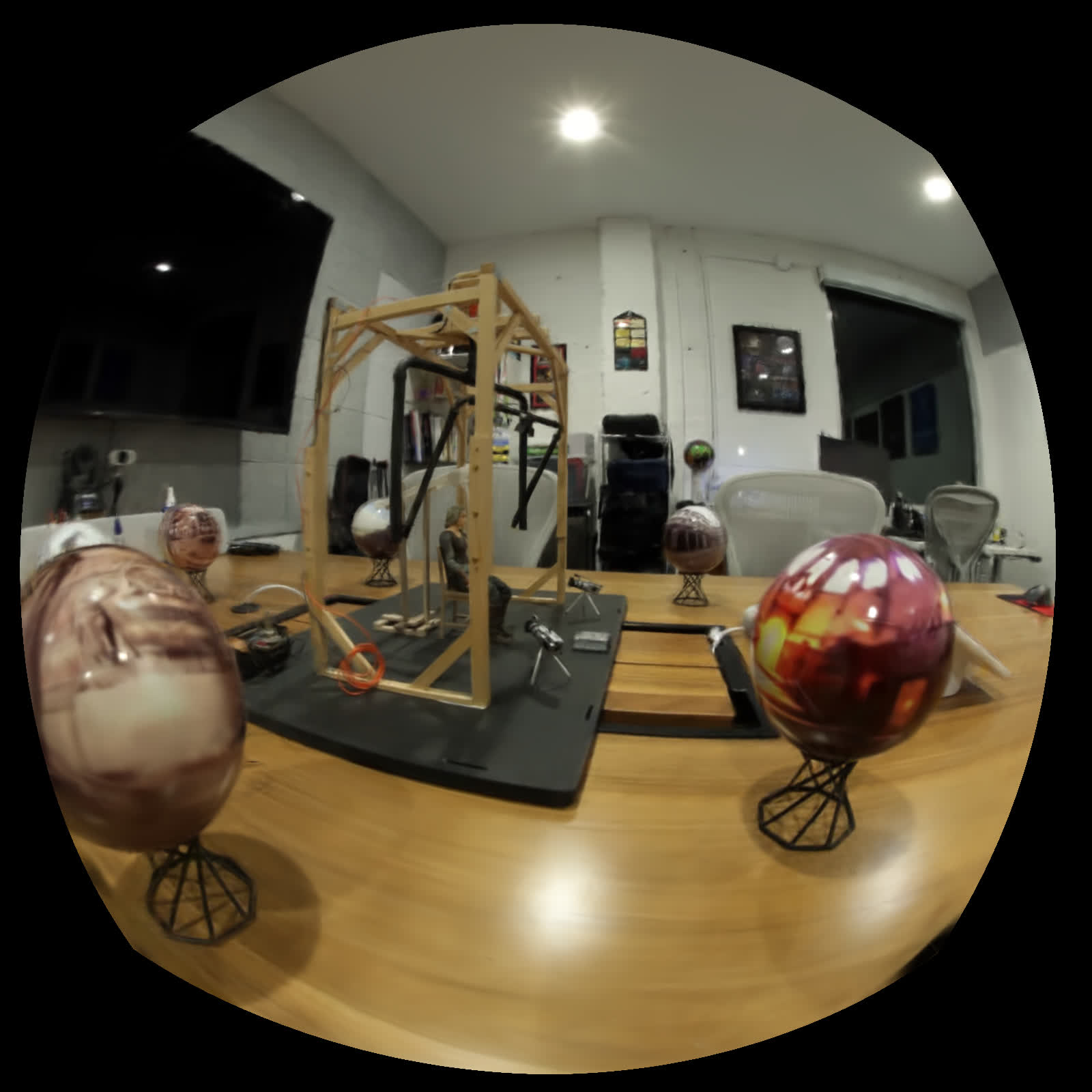} &
        \includegraphics[height=0.3\textwidth]{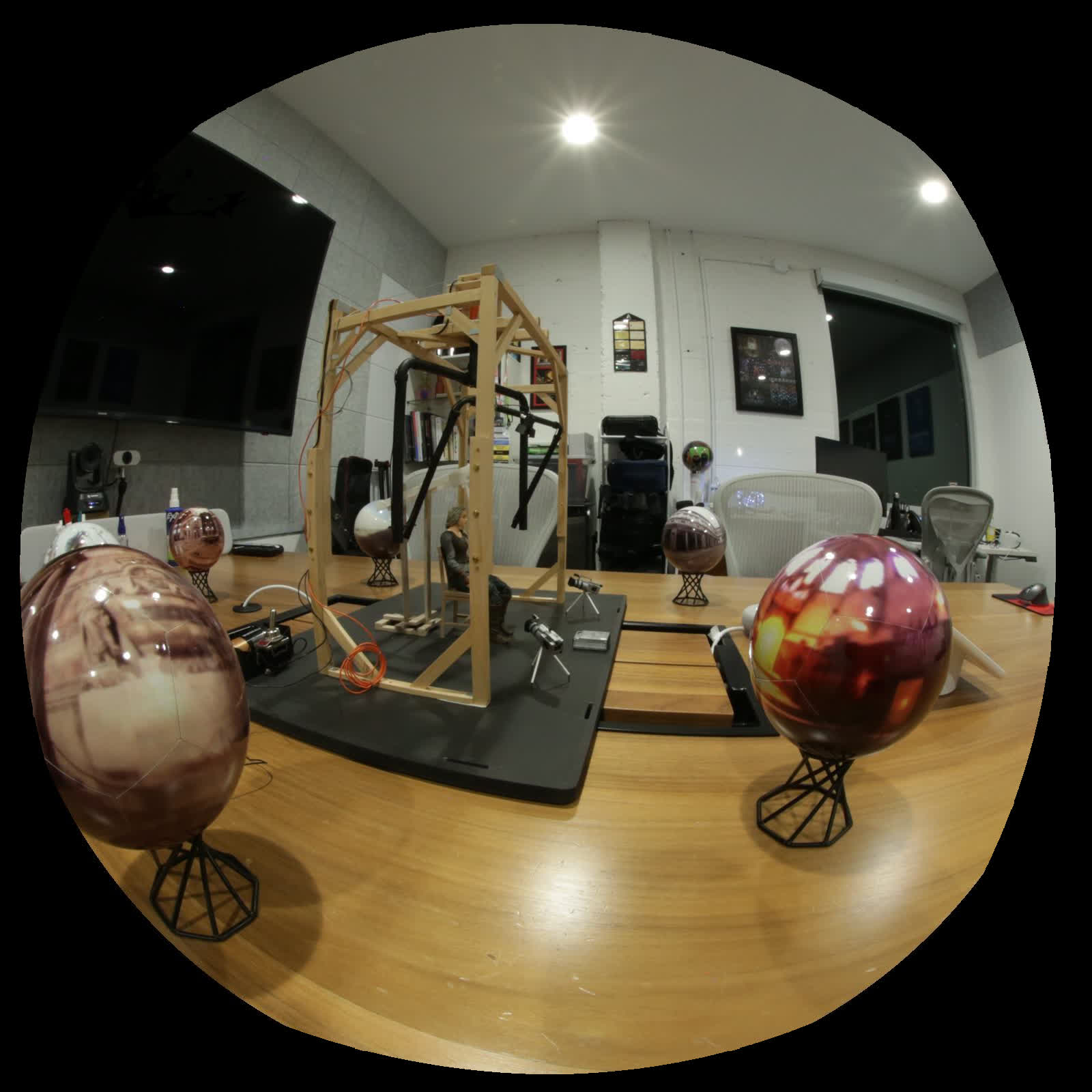}\\

        \multicolumn{1}{c}{(a) Ours (Perspective)} & \multicolumn{1}{c}{(b) Ours (Fisheye)} & \multicolumn{1}{c}{(c) GT}
    \end{tabular}

    \caption{\textbf{Reconstruction from 180\si{\degree} FOV Fisheye Captures.} Using a Canon fisheye camera, we capture the scene and reconstruct the office with our method. Both perspective and fisheye views are rendered to demonstrate the quality of our reconstruction.}

    \label{fig:supp_paul_office}

\end{figure*}
}
{
\begin{figure*}[t]
    \centering
    \setlength{\tabcolsep}{1pt} % Adjust space between columns if needed
    \begin{tabular}{ccc} % 5 columns
        \includegraphics[height=0.19\textwidth]{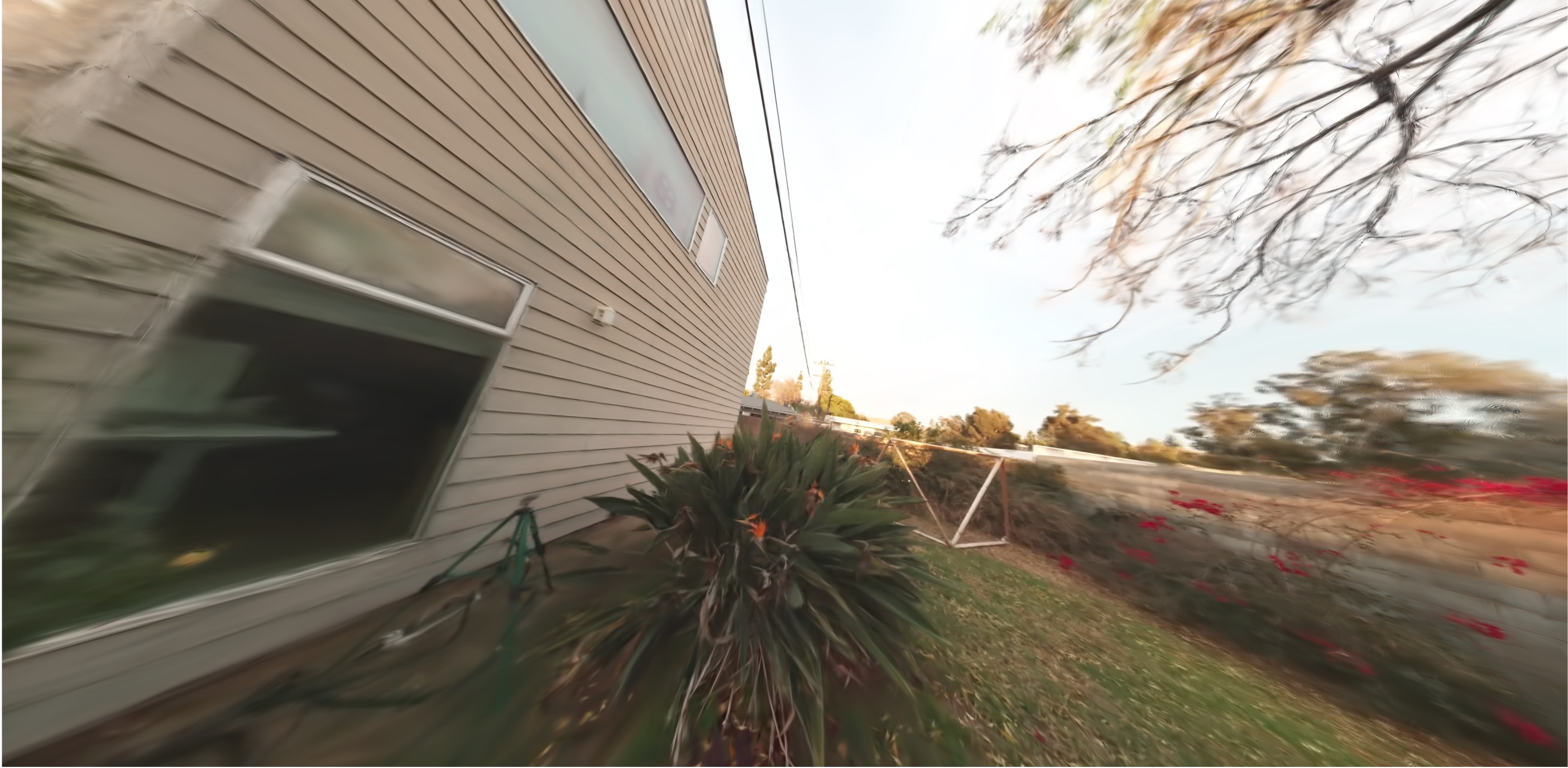} &
         \includegraphics[height=0.19\textwidth]{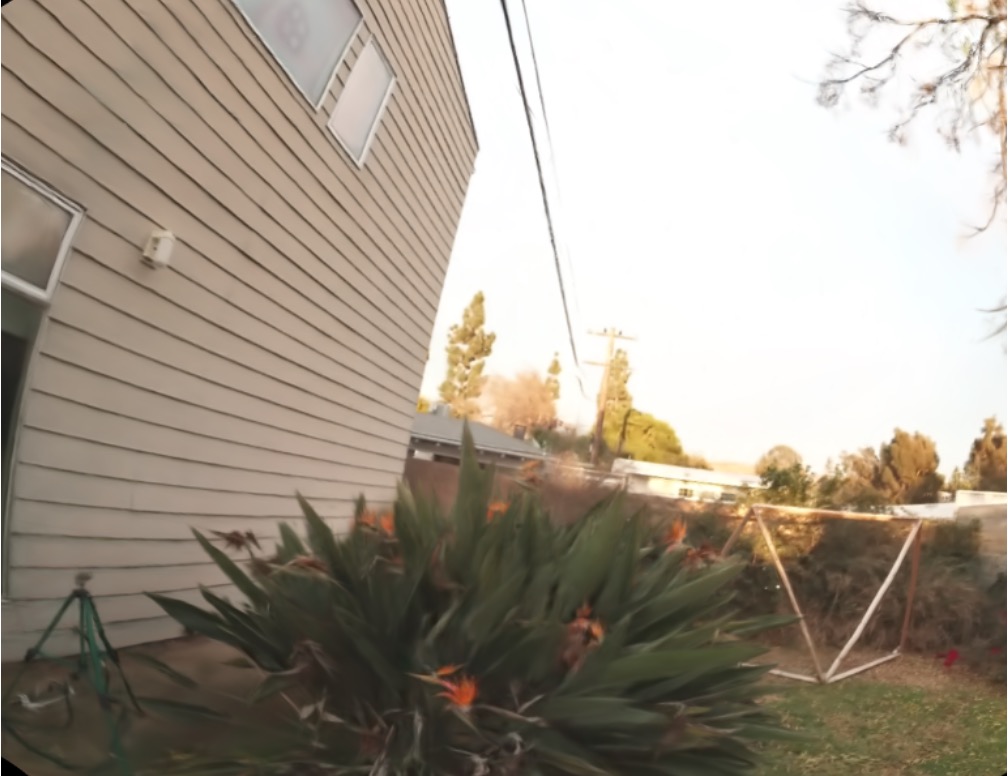} &
        \includegraphics[height=0.19\textwidth]{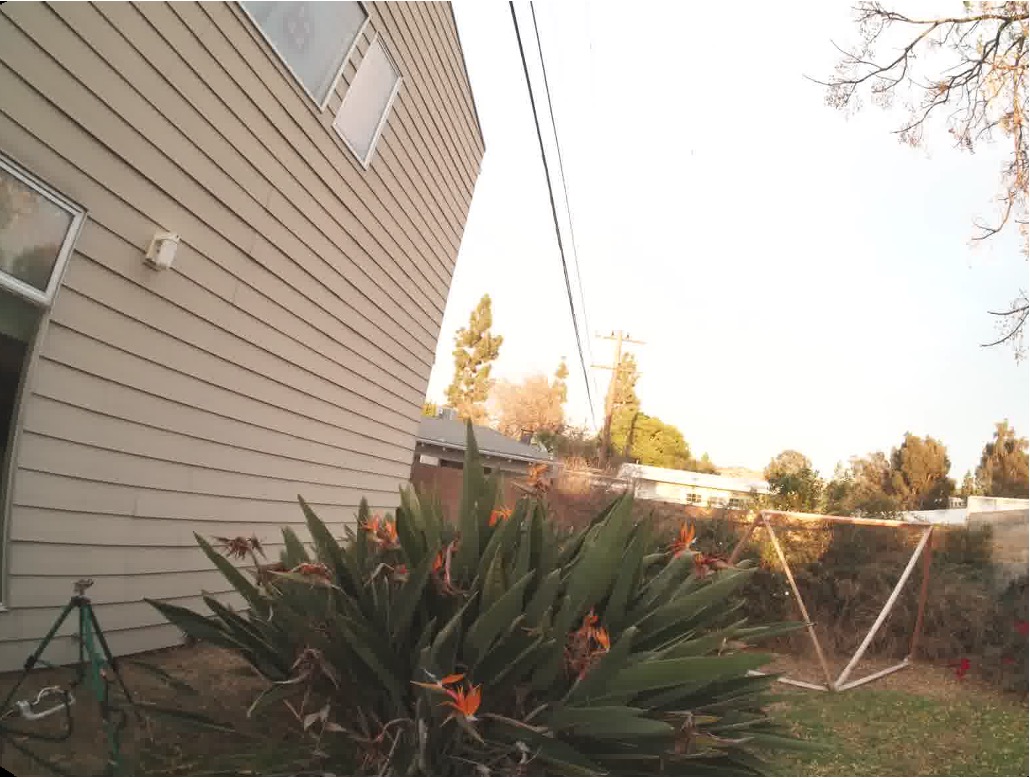} \\

        \includegraphics[height=0.19\textwidth]{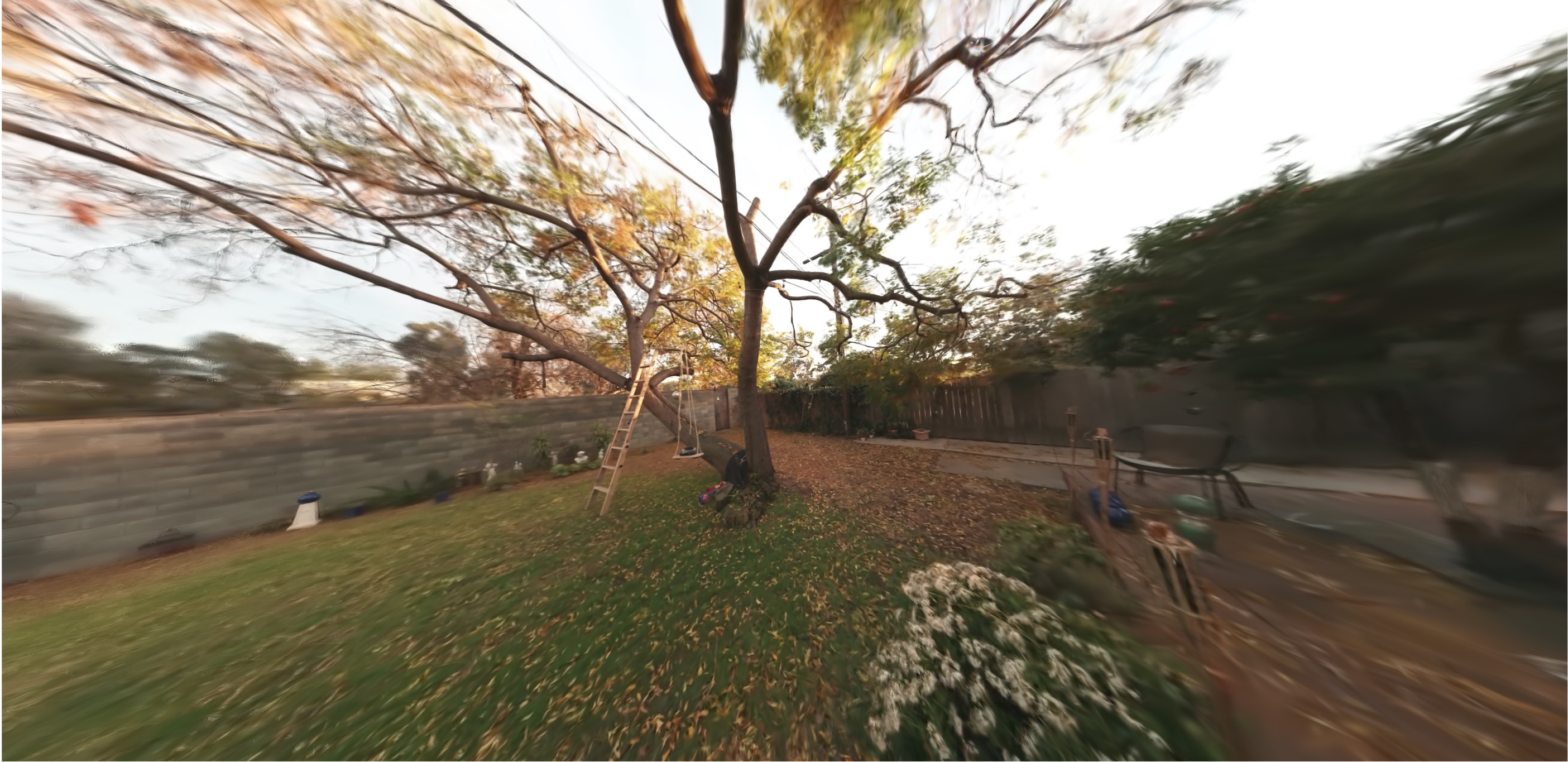} &
         \includegraphics[height=0.19\textwidth]{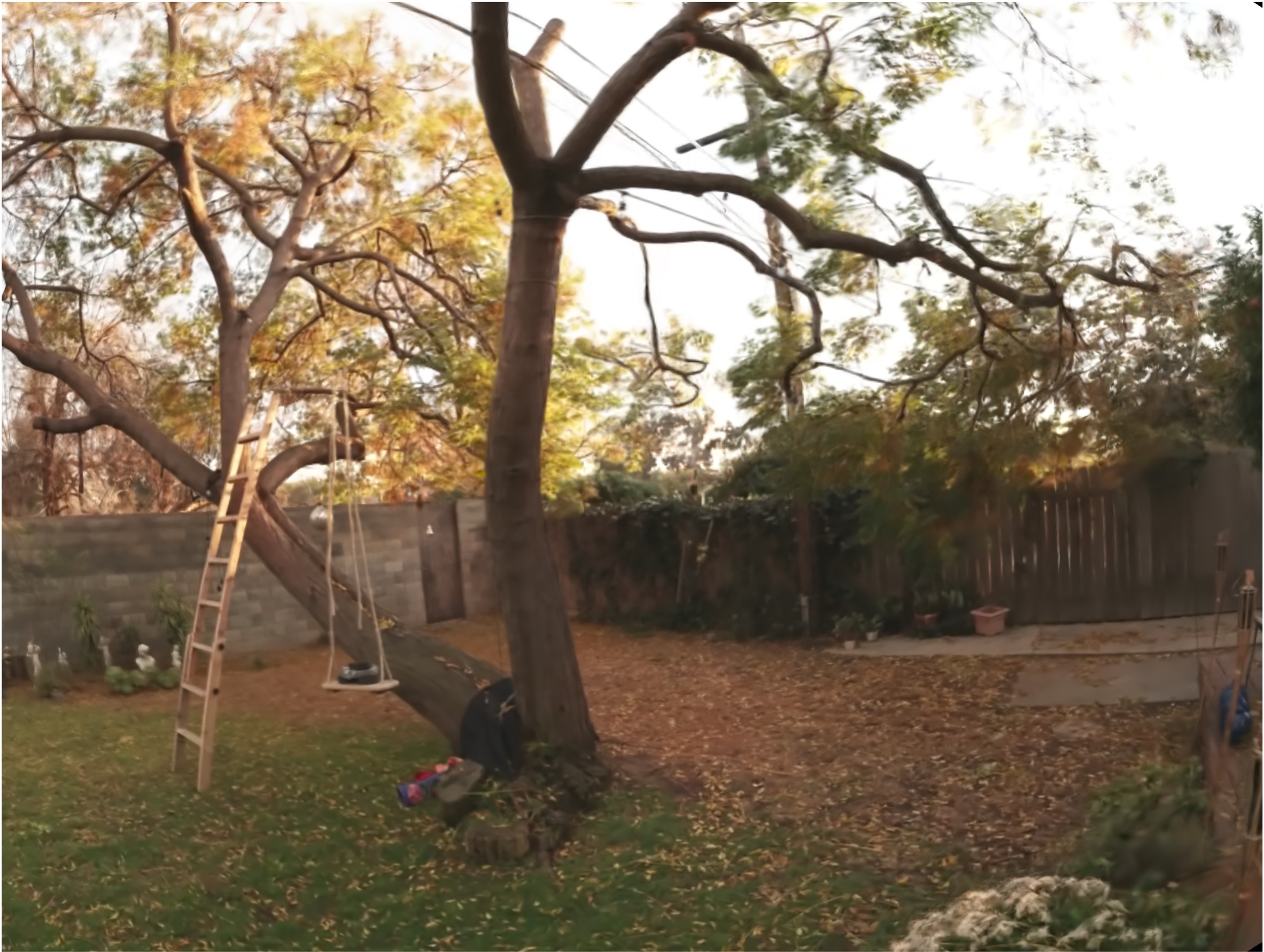} &
        \includegraphics[height=0.19\textwidth]{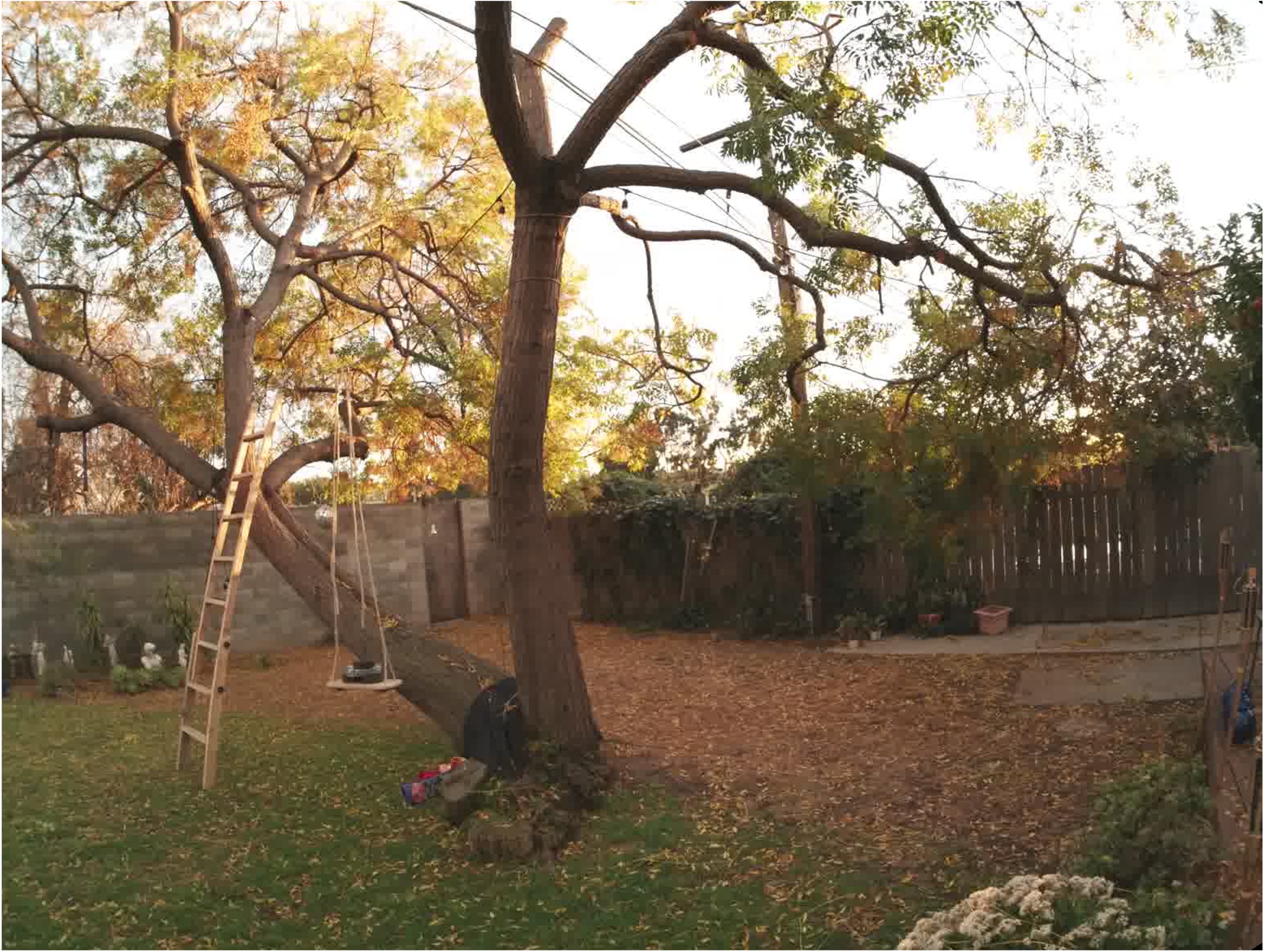} \\

        \includegraphics[height=0.19\textwidth]{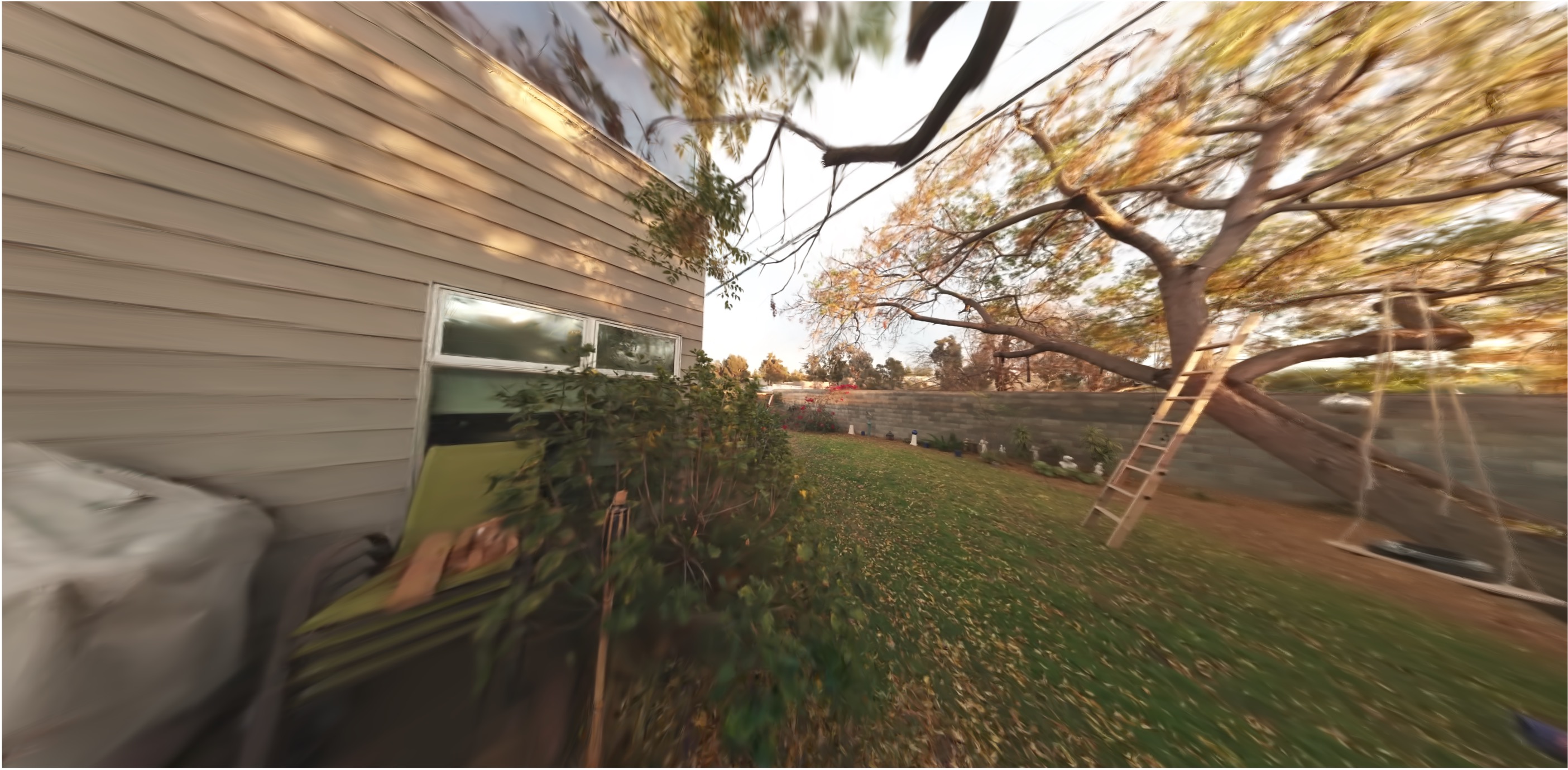} &
         \includegraphics[height=0.19\textwidth]{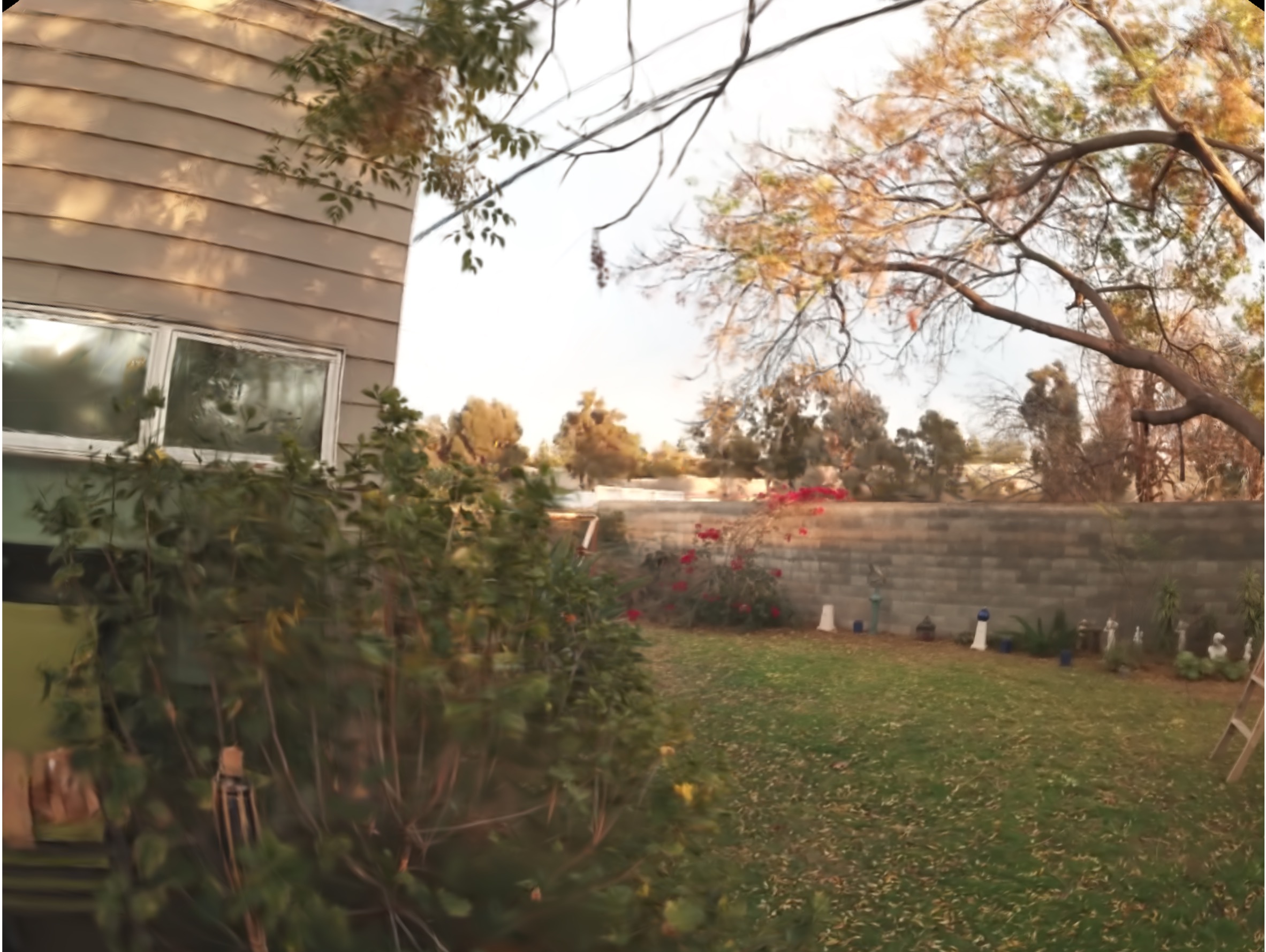} &
        \includegraphics[height=0.19\textwidth]{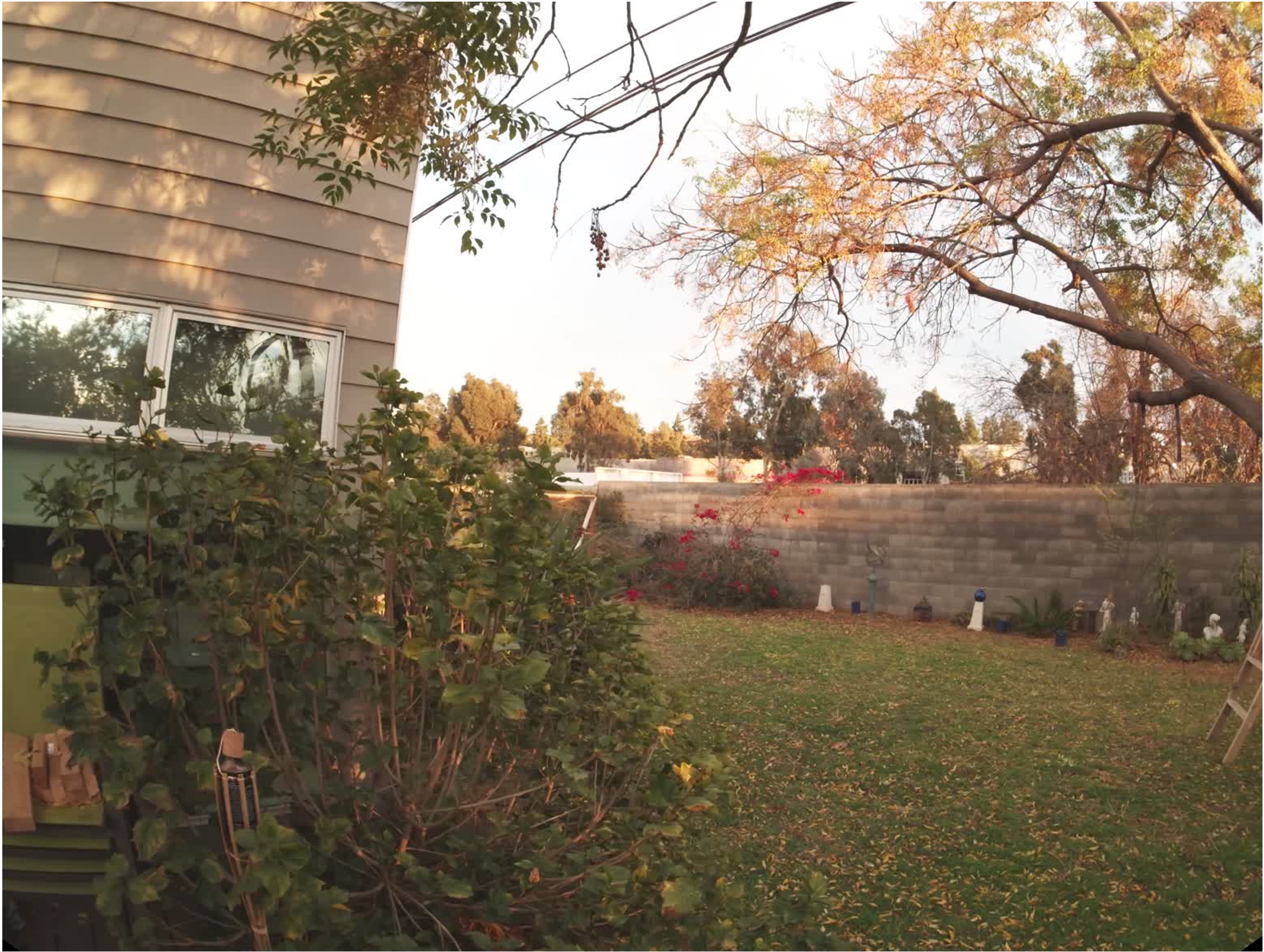} \\

        \multicolumn{1}{c}{(a) Ours (Perspective)} & \multicolumn{1}{c}{(b) Ours (Fisheye)} & \multicolumn{1}{c}{(c) GT}
    \end{tabular}

    \caption{\textbf{Large FOV Reconstruction from a Customized Fisheye Rig.} We reconstruct a backyard from images captured using a fisheye rig. Our method achieves accurate geometric corrections, such as straightening the lines on the wall and the edges of the house.}

    \label{fig:supp_garden}

\end{figure*}
}

\paragraph{Qualitative Results.}
To verify the accuracy of self-calibration, we generate a set of hold-out cameras that share the same distribution as the training set. For each validation camera, we render paired perspective and 180\si{\degree} fisheye images. Unlike real-world datasets such as Garden and Studio, our synthetic dataset allows direct comparison with ground-truth perspective views. As shown in~\cref{fig:largefov_vs_smallfov_mitsuba_in_perspective}, we render an additional 20\si{\degree} for the hold-out cameras to highlight the difference in coverage between our method and conventional capture approaches.

It is worth noting that in the comparison between our method and 3DGS~\cite{kerbl20233d}, we evaluate in perspective views since 3DGS~\cite{kerbl20233d} does not support fisheye rendering, whereas our method can generate perspective views after training. In contrast, we directly compare our method with Fisheye-GS~\cite{liao2024fisheye}, as both methods natively support fisheye rendering.

\subsection{More Real-world Scenes}
\paragraph{Scene Captures.}
We use a Canon 5D Mk III DSLR camera with a Canon 8mm-15mm fisheye lens, zoomed out to 8mm with a 180\si{\degree} FOV, to capture a complex indoor office scene, where we place models and spheres on a table. Images are taken close to the table to capture the details of the various models and spheres. We also use a Meike 3.5mm f/2.8 ultra-wide-angle circular fisheye lens to capture the same office. 

Additionally, we mount two fisheye cameras on a rig configured such that the cameras are perpendicular to each other. This camera rig is used to capture a backyard scene by walking clockwise and counterclockwise twice to record videos. The benefit of using this rig is that the relative pose between the two fisheye cameras is fixed, simplifying the SfM~\cite{schoenberger2016sfm} pipeline for estimating accurate poses.

\paragraph{Qualitative Results.}
Qualitative results are shown in~\cref{fig:supp_paul_office}. Our method effectively recovers details in the central region while accurately modeling lens distortion for background elements, such as painting frames and lines on the white wall.

As shown in~\cref{fig:supp_garden}, our method converges to an accurate calibration, ensuring that lines on the house's surface and the ladder leaning against the tree remain straight when rendered in perspective views.

\subsection{Adaptability to Different Lens Distortions}

\begin{table}[t]
  \centering
  \scalebox{1}{
    \begin{tabular}{cccc} 
    \toprule
    Method & PSNR & SSIM & LPIPS \\ 
    \midrule
    Vanilla-GS~\cite{kerbl20233d}  & 25.37 & 0.923 & 0.121 \\
    Ours                           & \textbf{28.00} & \textbf{0.932} & \textbf{0.093} \\
    \bottomrule
    \end{tabular}
  }
  \caption{\textbf{Evaluation on Radial Distortion of Mitsuba Synthetic Scenes}. We compare our method with 3DGS~\cite{kerbl20233d} in object-centric scenes with slight radial distortion, where our method still produces better reconstructions.}
  \label{tab:table_radial}
\end{table}

As mentioned in the last part of Sec.~4.2 of the main paper and also shown in Fig.~6, we introduced synthetic distortions, including both radial and tangential components, to images from the LLFF dataset~\cite{mildenhall2019llff}.

We also apply moderate radial distortion to our synthetic dataset and reconstruct several object scenes. After training, we can render undistorted images. The perspective rendering increases the FOV while maintaining the same camera extrinsics. As shown in~\cref{fig:ours_obj_radial}, distorted edge lines, such as those on the Lego and Car objects, are correctly recovered into straight lines, demonstrating the capability of our hybrid field to model radial distortion effectively. We also report quantitative evaluations in~\cref{tab:table_radial}. Note that the radial distortion is relatively subtle, so the improvement is not as pronounced compared to other types of lenses.

{
\begin{figure*}[t]
    \centering
    \setlength{\tabcolsep}{1pt} % Adjust space between columns if needed
    \begin{tabular}{cccc} % 5 columns
         \includegraphics[height=0.22\textwidth]{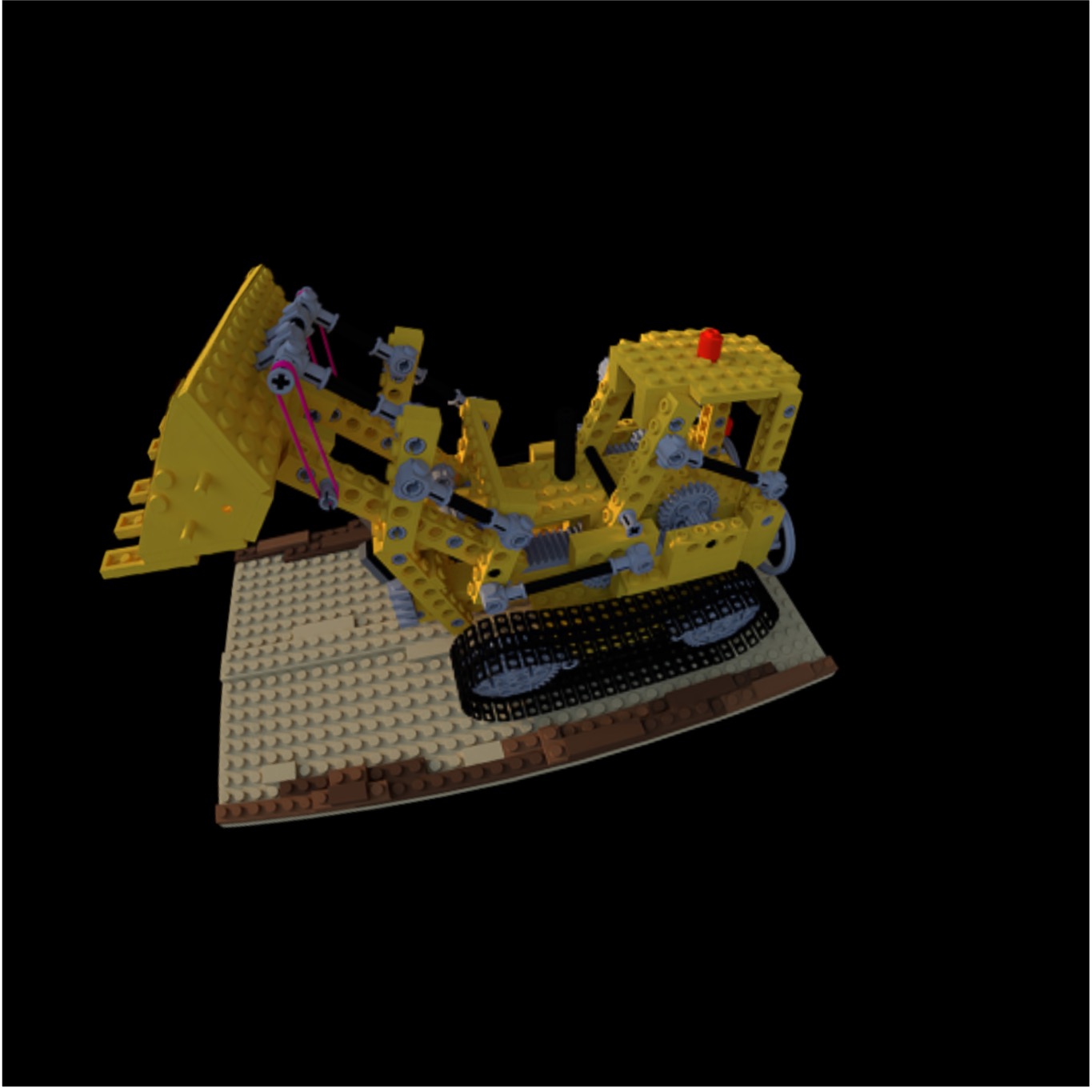} &
        \includegraphics[height=0.22\textwidth]{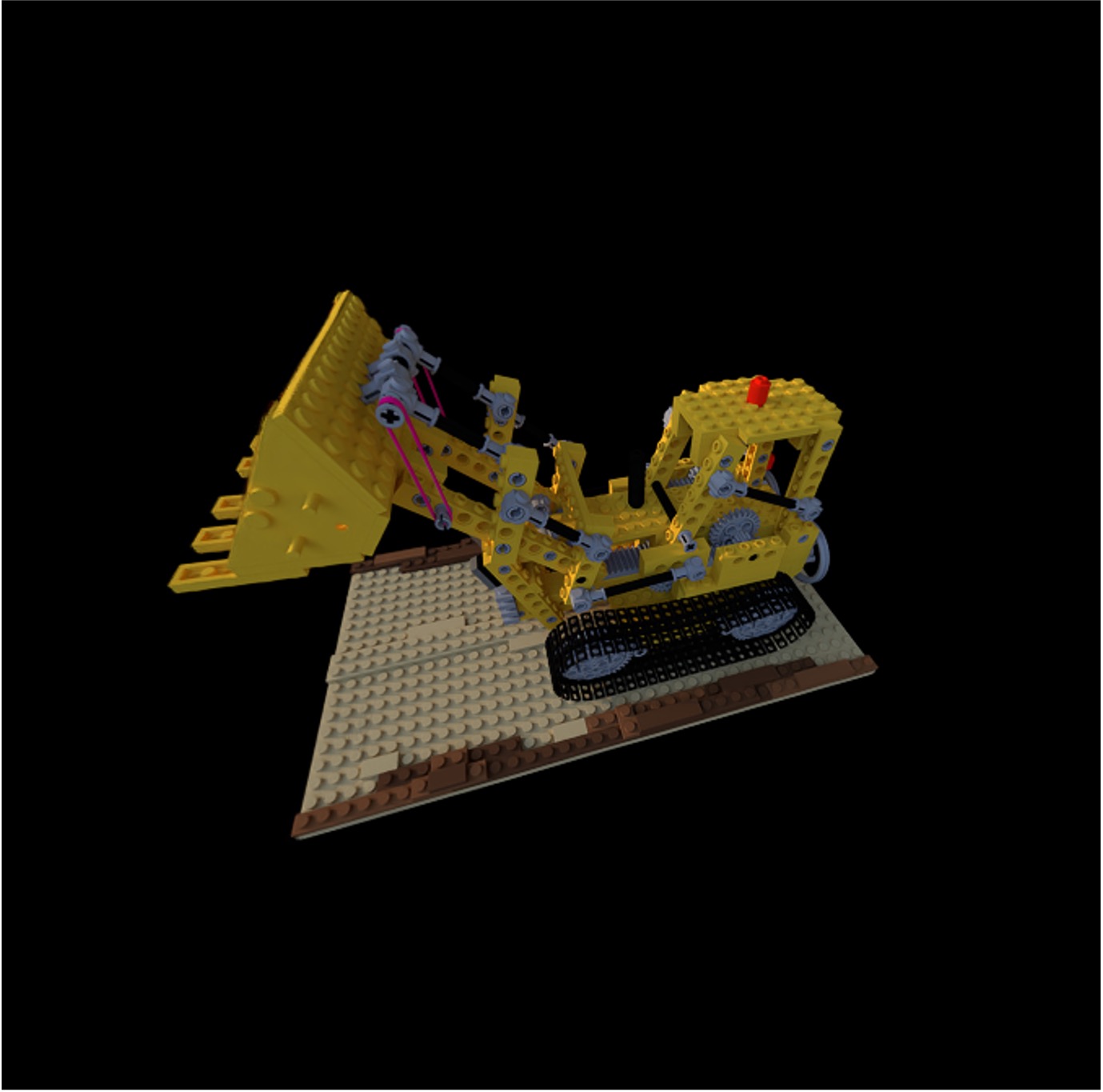} &
        \includegraphics[height=0.22\textwidth]{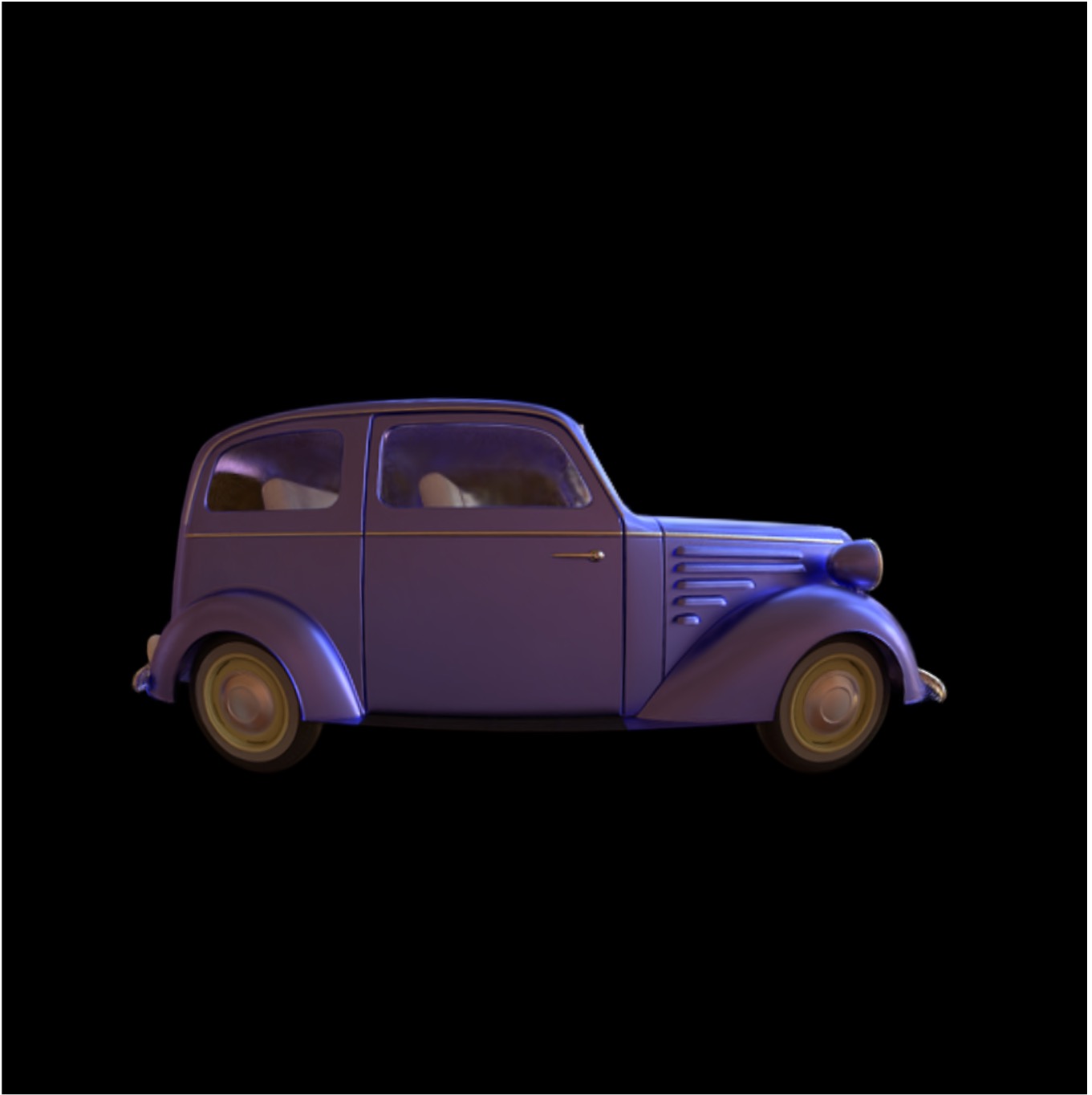} &
        \includegraphics[height=0.22\textwidth]{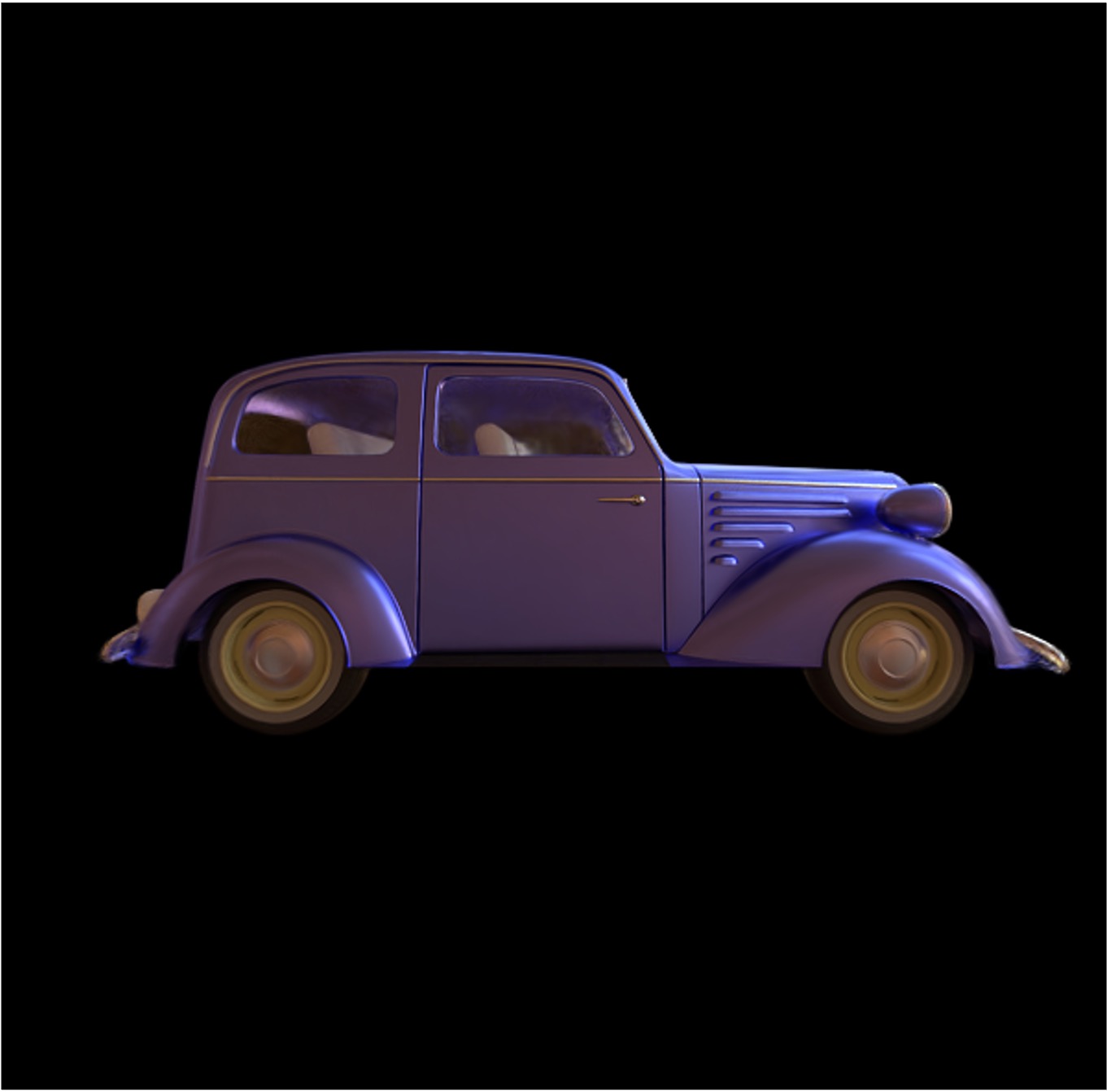}\\
        \multicolumn{2}{c}{(a) Lego} & \multicolumn{2}{c}{(b) Car}
        \vspace{-0mm}
        \\
        \includegraphics[height=0.22\textwidth]{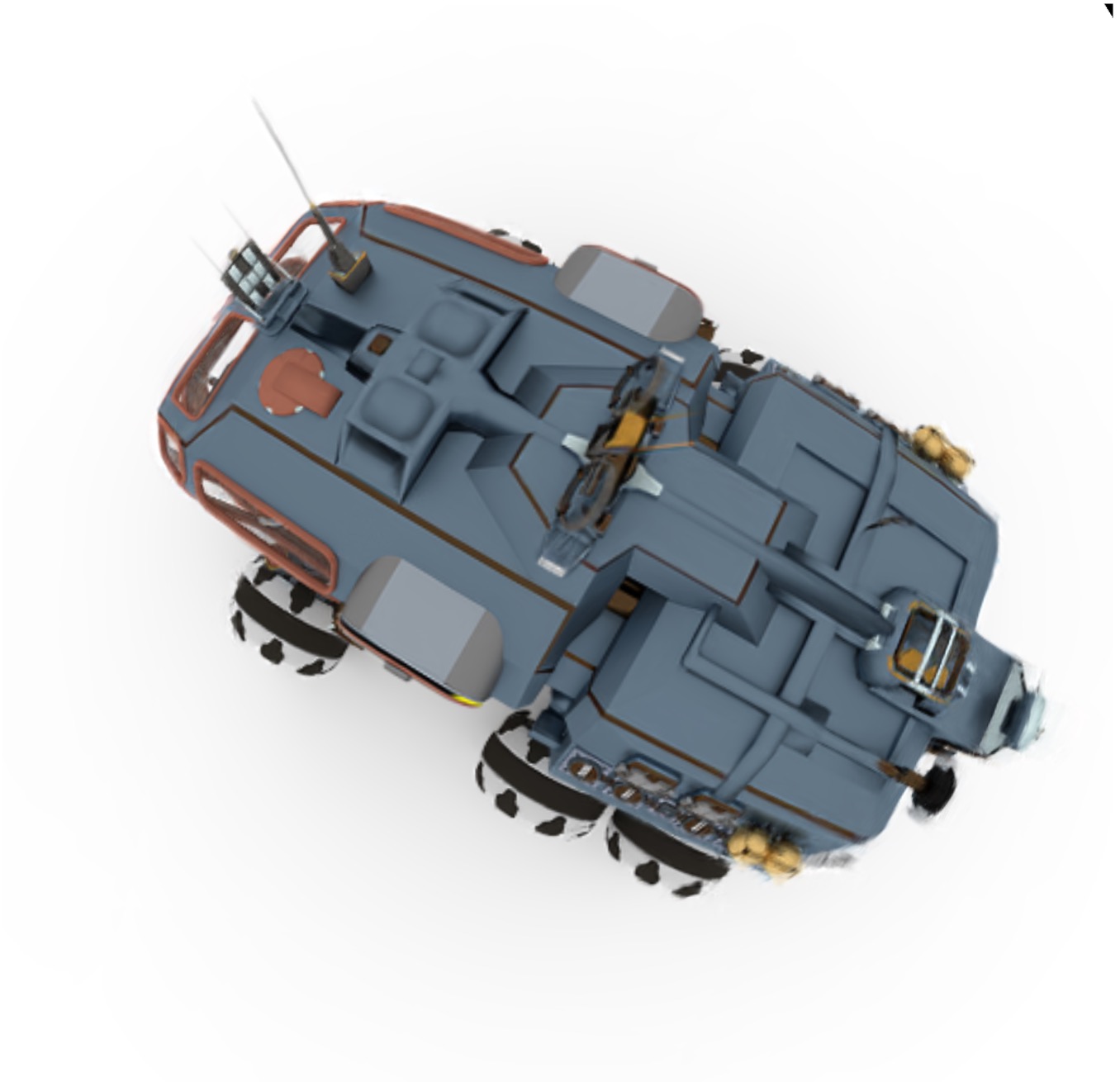} &
        \includegraphics[height=0.22\textwidth]{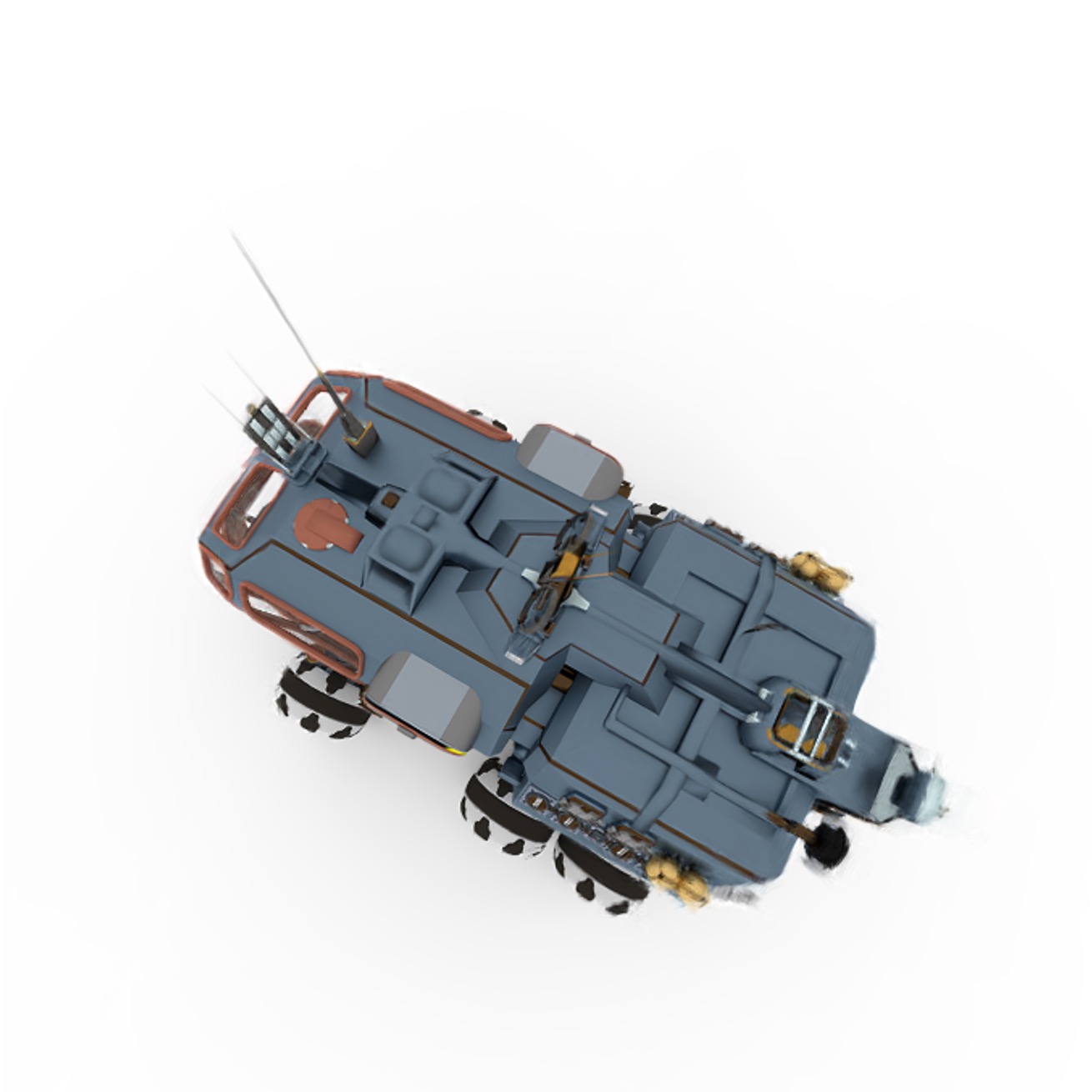} &
        \includegraphics[height=0.22\textwidth]{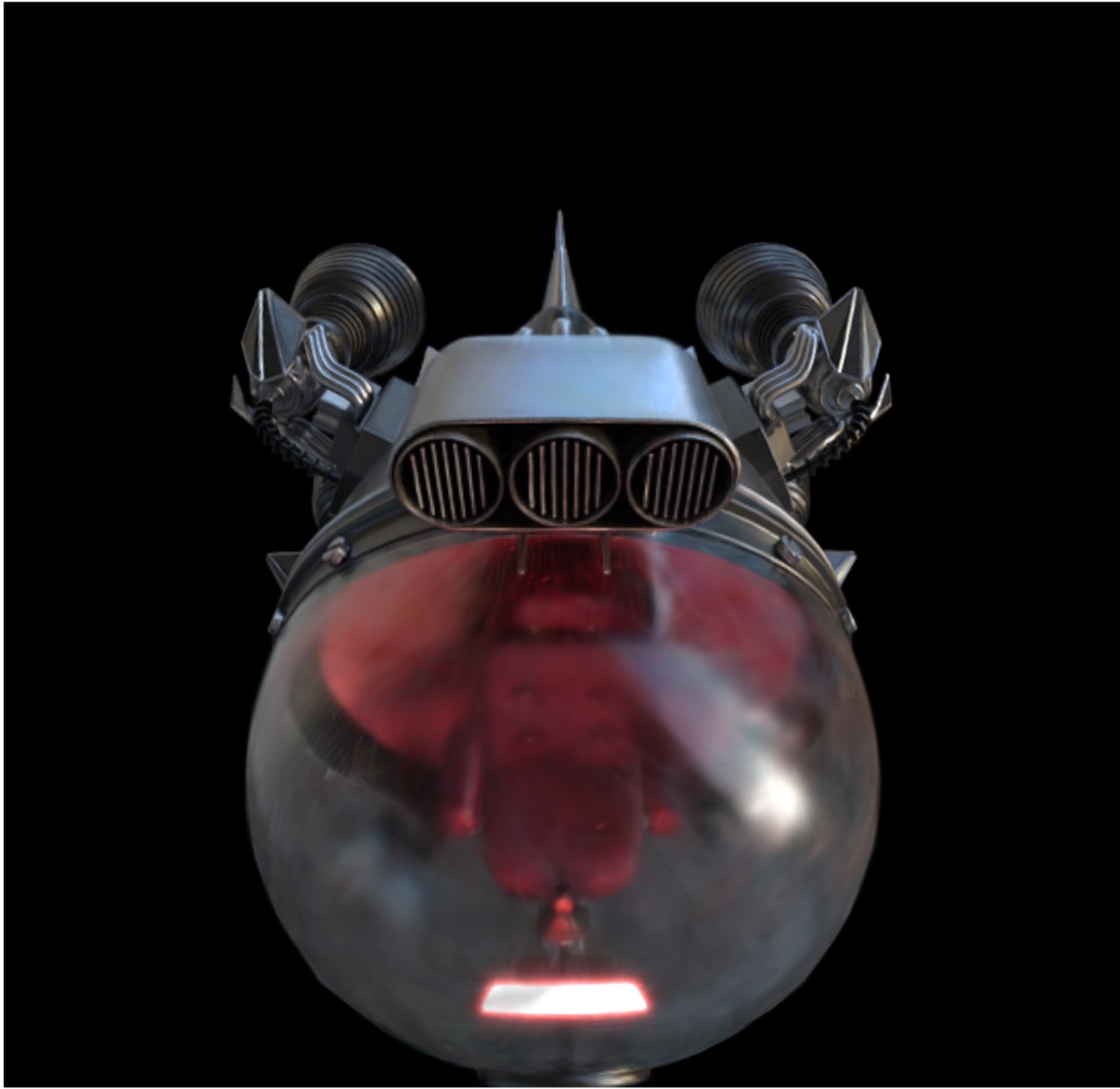} &
        \includegraphics[height=0.22\textwidth]{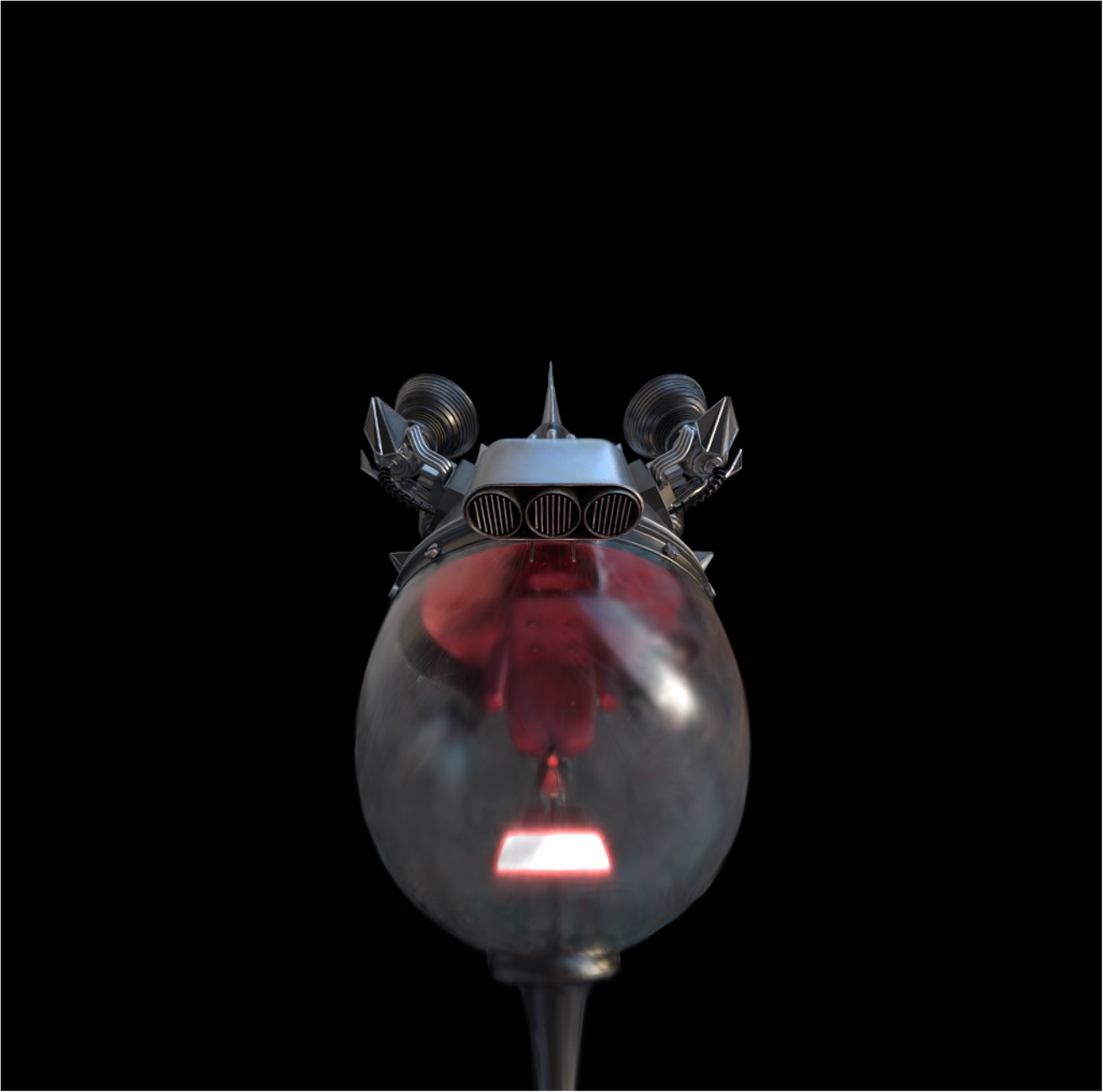}\\
        \multicolumn{2}{c}{(c) Rover} & \multicolumn{2}{c}{(d) Spaceship}
    \end{tabular}

    \caption{\textbf{Radial and Perspective Rendering}. We evaluate our method on a synthetic radial distortion dataset. Our approach successfully recovers slight radial distortion during reconstruction and enables perspective rendering upon completion of training.}

    \label{fig:ours_obj_radial}

\end{figure*}
}

\begin{figure}[t] % Position at the top of the page
    \centering
    \scalebox{0.3}{
        \includegraphics[]{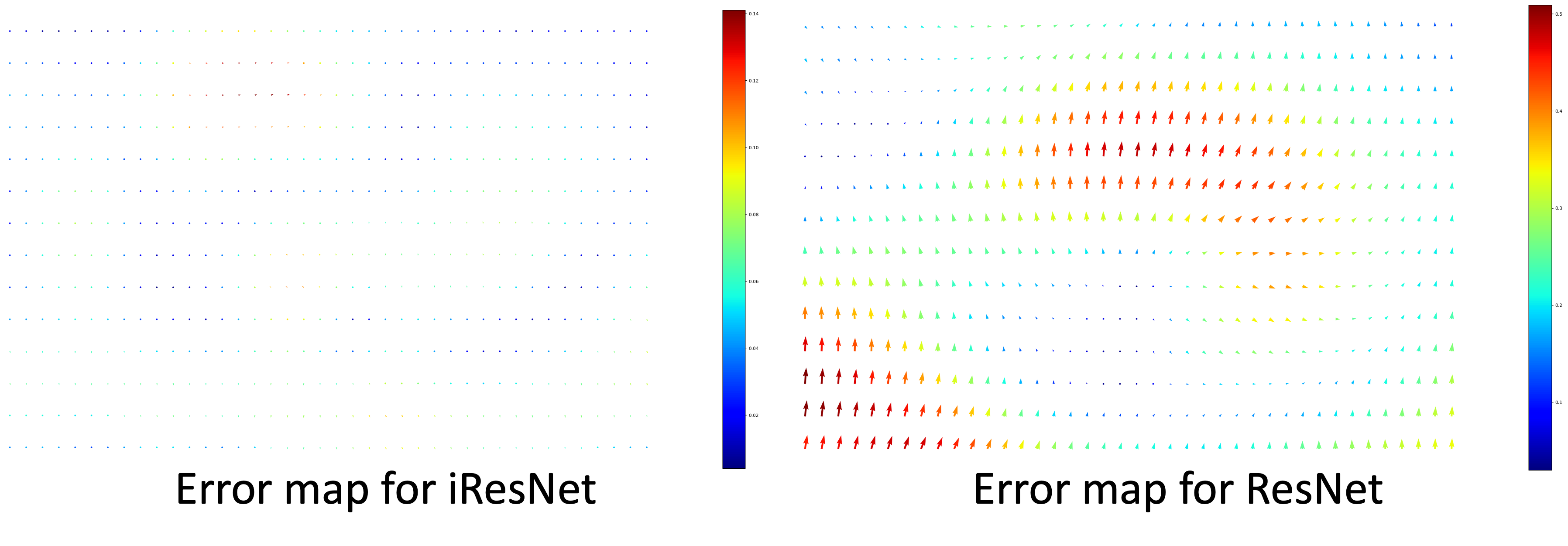} % Scale image width  
    }
    \caption{\textbf{Distortion Error Map}. We visualize the error map between the predicted distortion and the ground truth distortion from Mitsuba synthetic scenes.} % Add a caption
    \label{fig:error_map} % Label for referencing
\end{figure}

\subsection{Comparison of Regular and Invertible ResNet}
Neural networks can model complex non-linear fields, but the key advantage of iResNet is its effective regularization. Light rays passing through the lens are strictly bijective and invertible. iResNet, using fixed-point iteration, enforces this property at minimal cost. We visualize the error map compared to the distortion of the GT lens in synthetic scenes in~\cref{fig:error_map}. We show that iResNet predicts smooth distortion with low error, whereas ResNet produces a highly asymmetric field with large errors. The large error produced by ResNet is largely due to the lack of regularization. The displacement predicted by ResNet can be arbitrary and does not follow the two properties that real-world light rays hold.

\subsection{Qualitative Results in Undistorted Rendering}
{
\begin{figure*}[t]
    \centering
    \setlength{\tabcolsep}{1pt} % Adjust space between columns if needed
    \begin{tabular}{cccc} % 5 columns
        
        \includegraphics[height=0.12\textwidth]{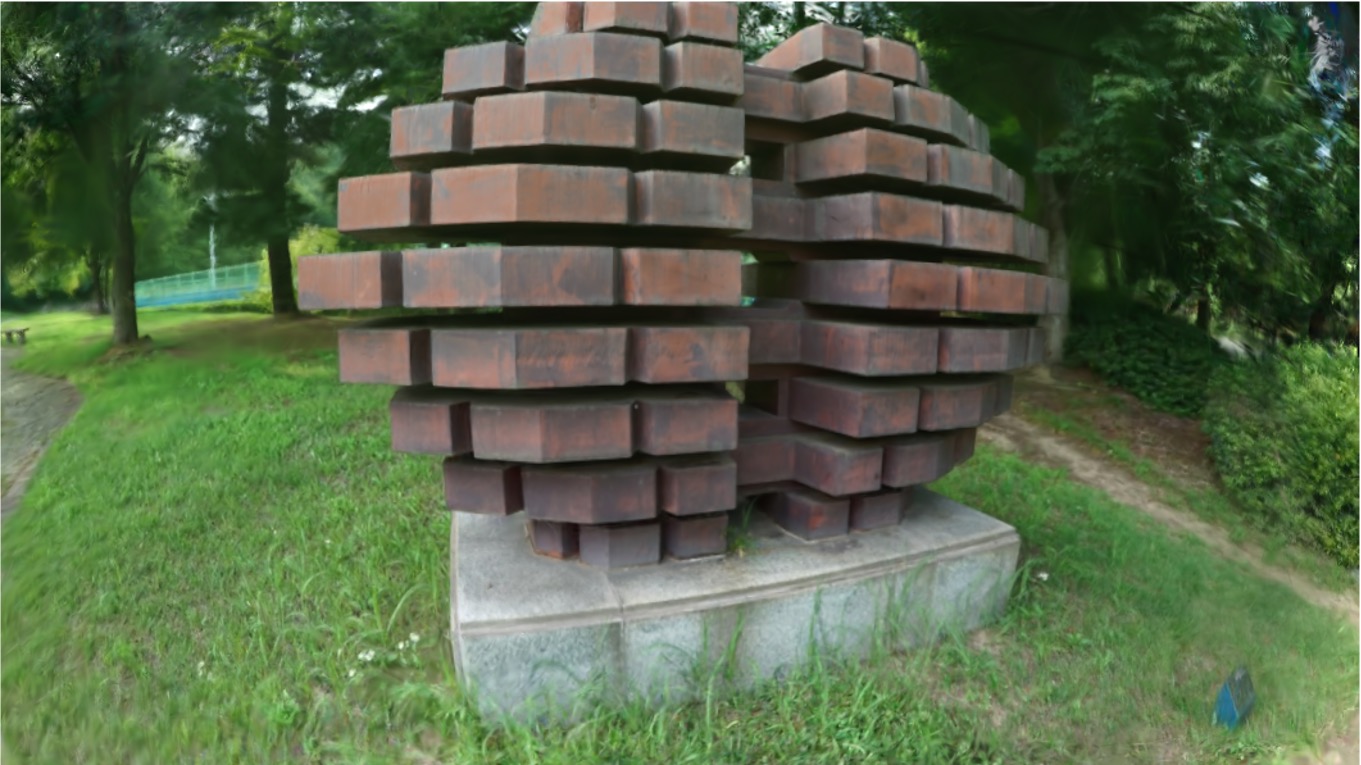} &
        \includegraphics[height=0.12\textwidth]{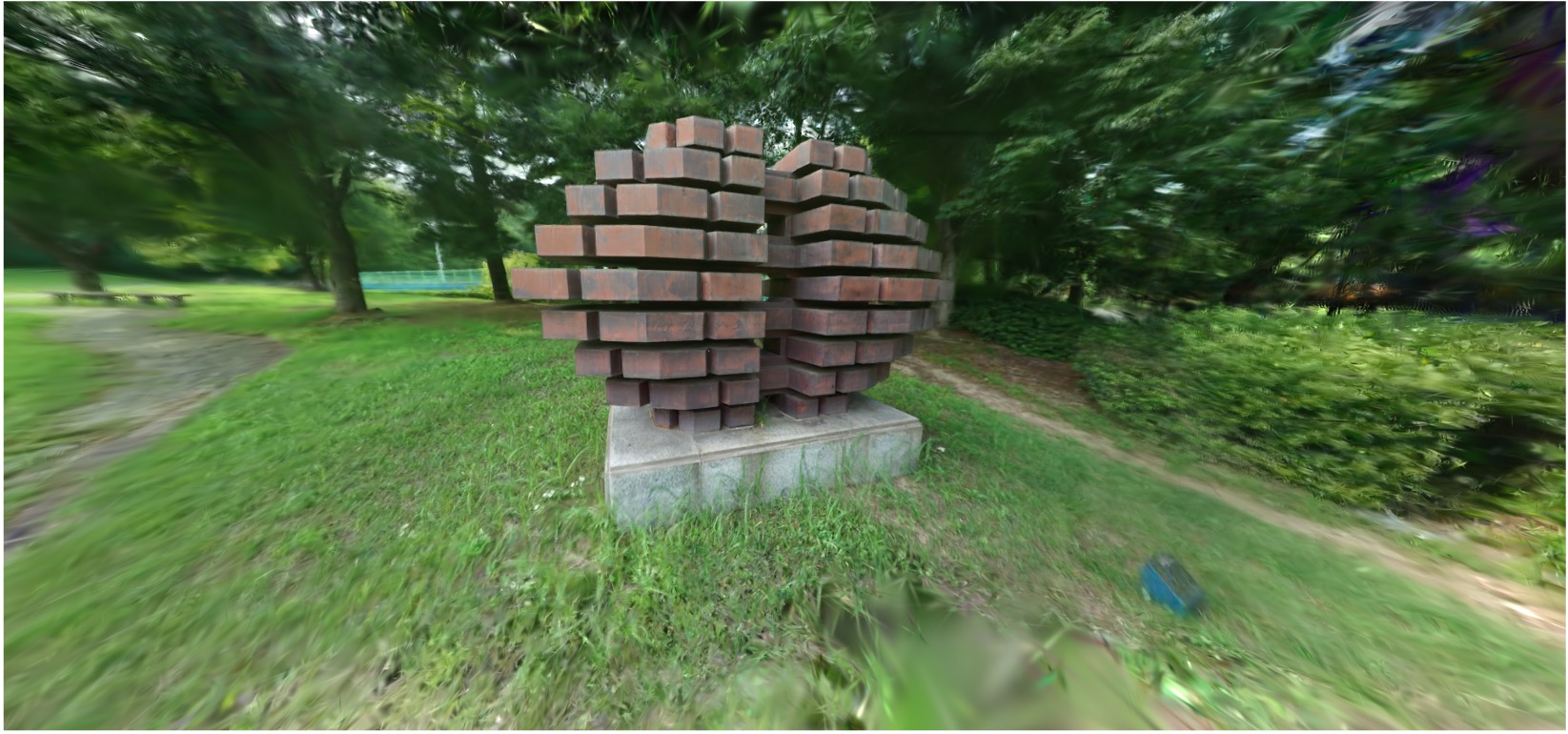} & 
        \includegraphics[height=0.12\textwidth]{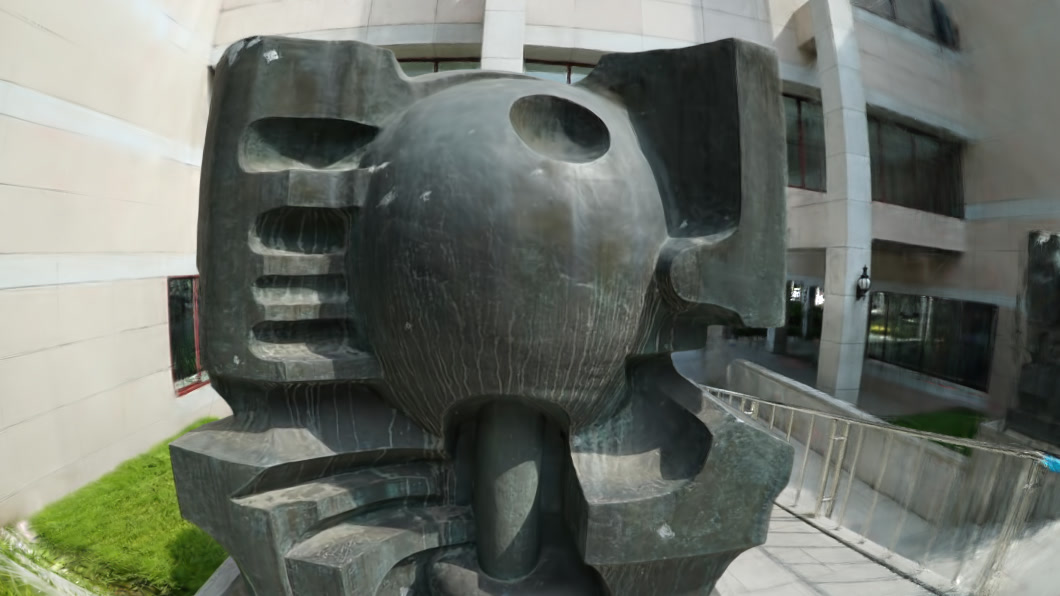} &
        \includegraphics[height=0.12\textwidth]{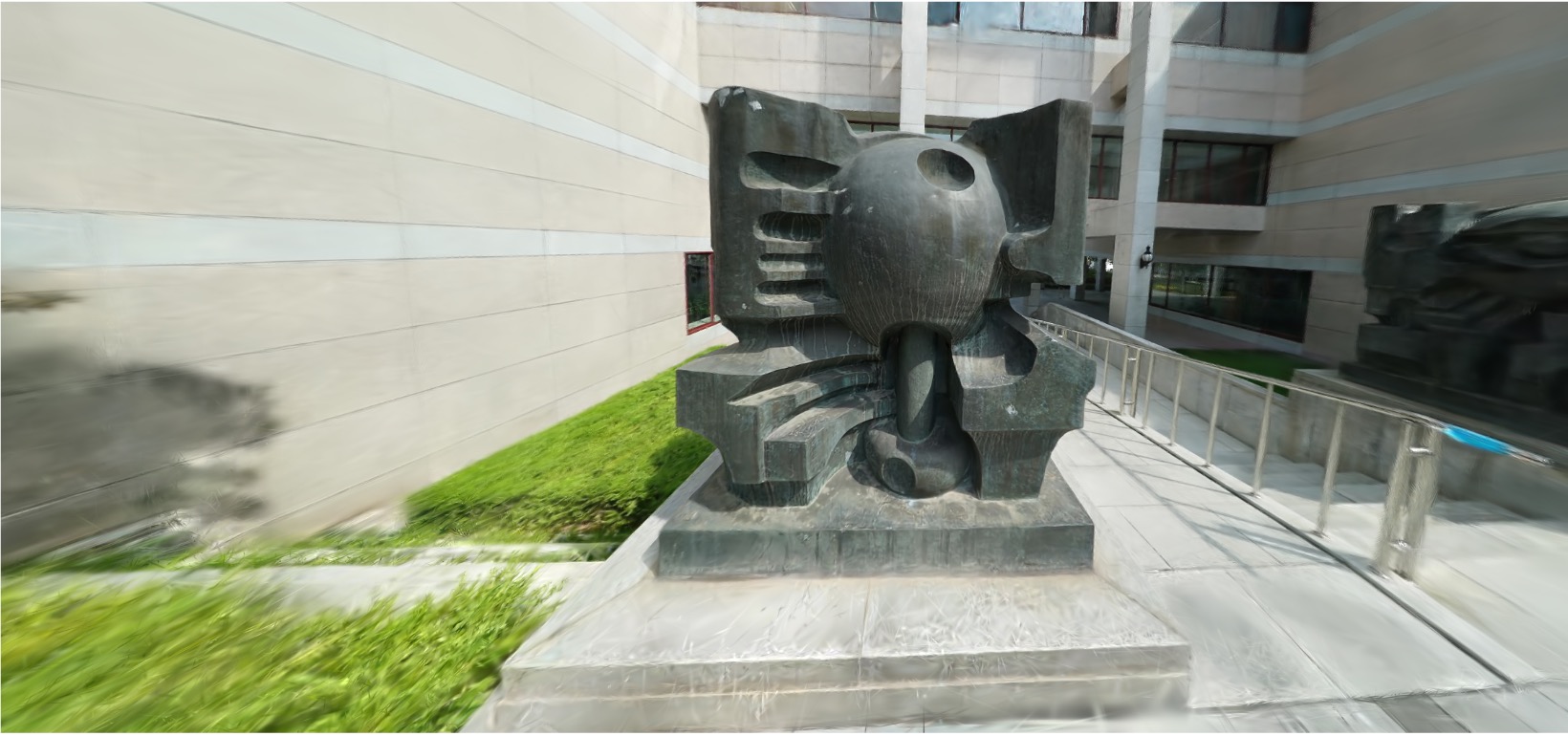} \\
        \multicolumn{2}{c}{(a) Heart} & \multicolumn{2}{c}{(b) Cube}
        \\
        \includegraphics[height=0.12\textwidth]{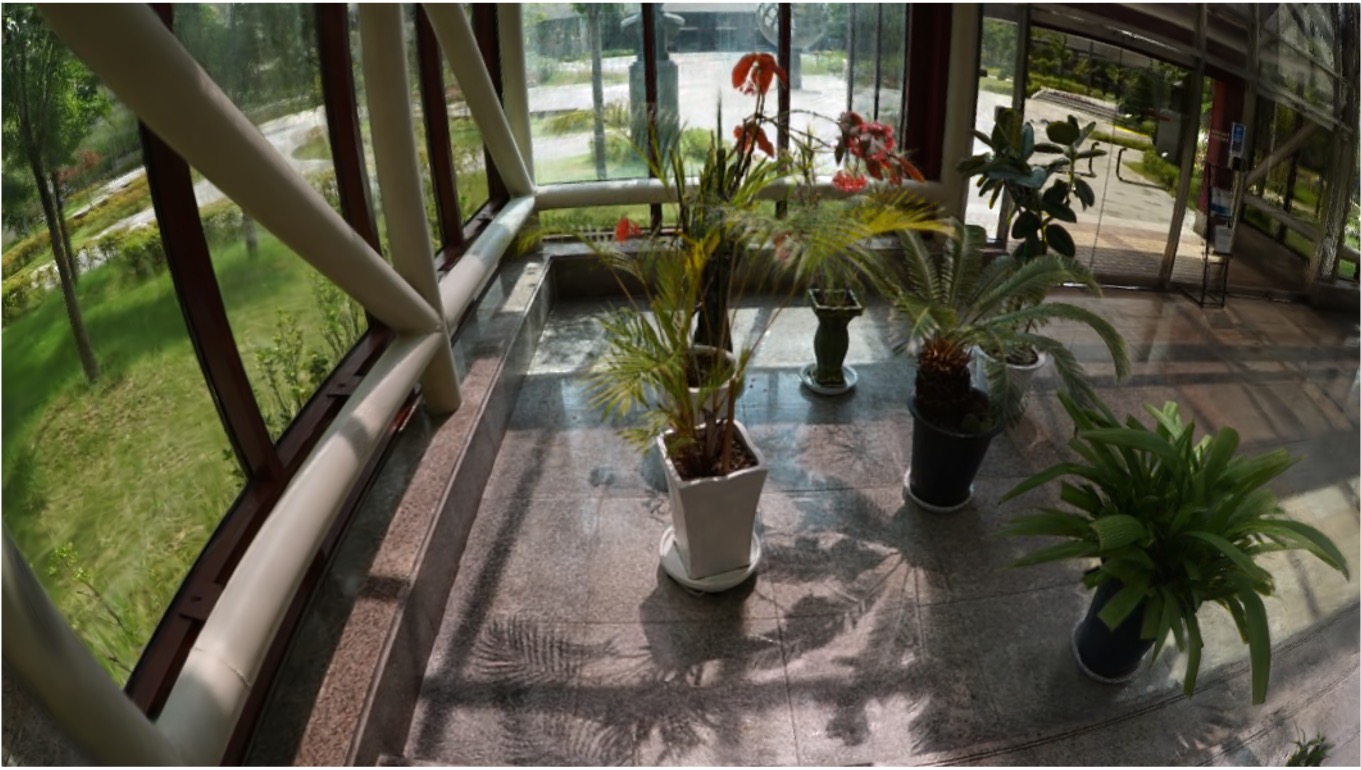} &
        \includegraphics[height=0.12\textwidth]{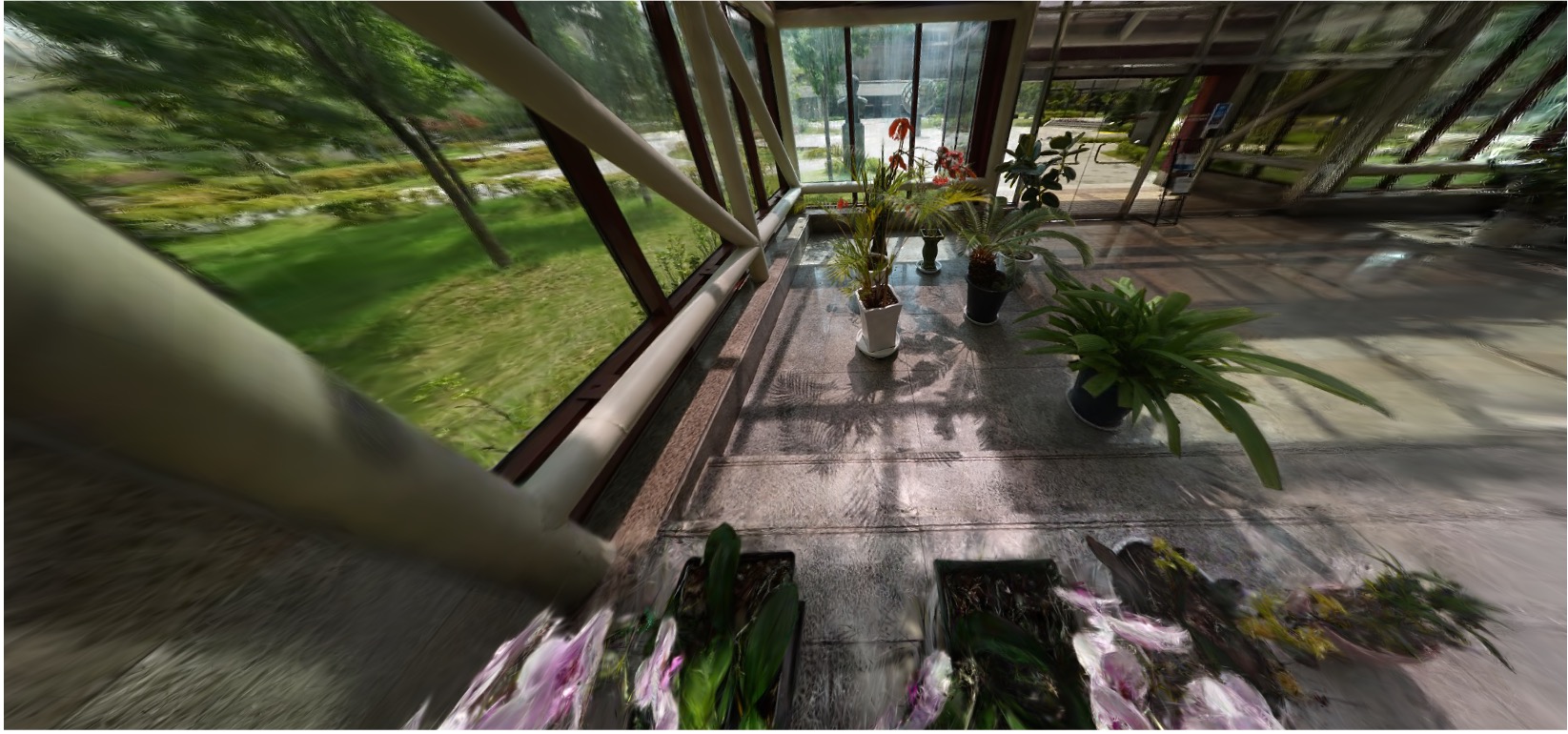} &
        \includegraphics[height=0.12\textwidth]{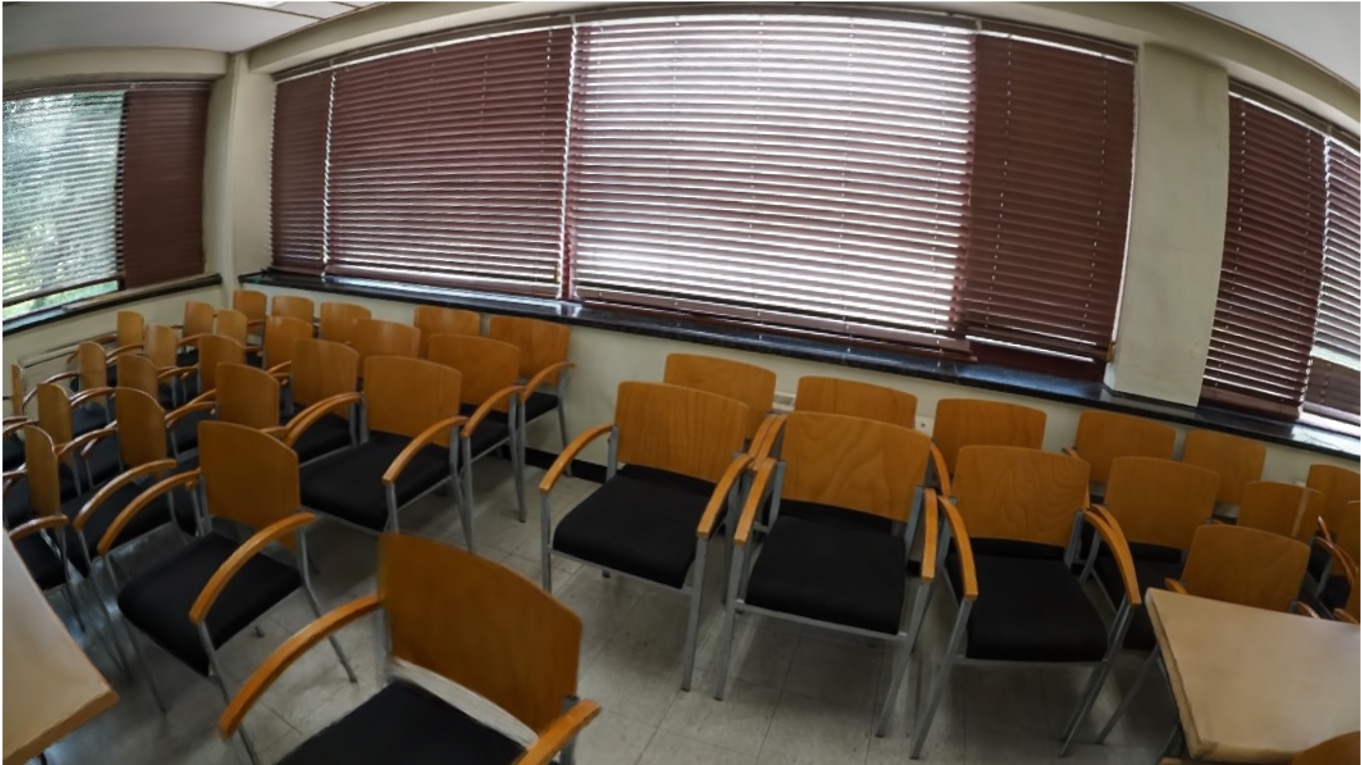} &
        \includegraphics[height=0.12\textwidth]{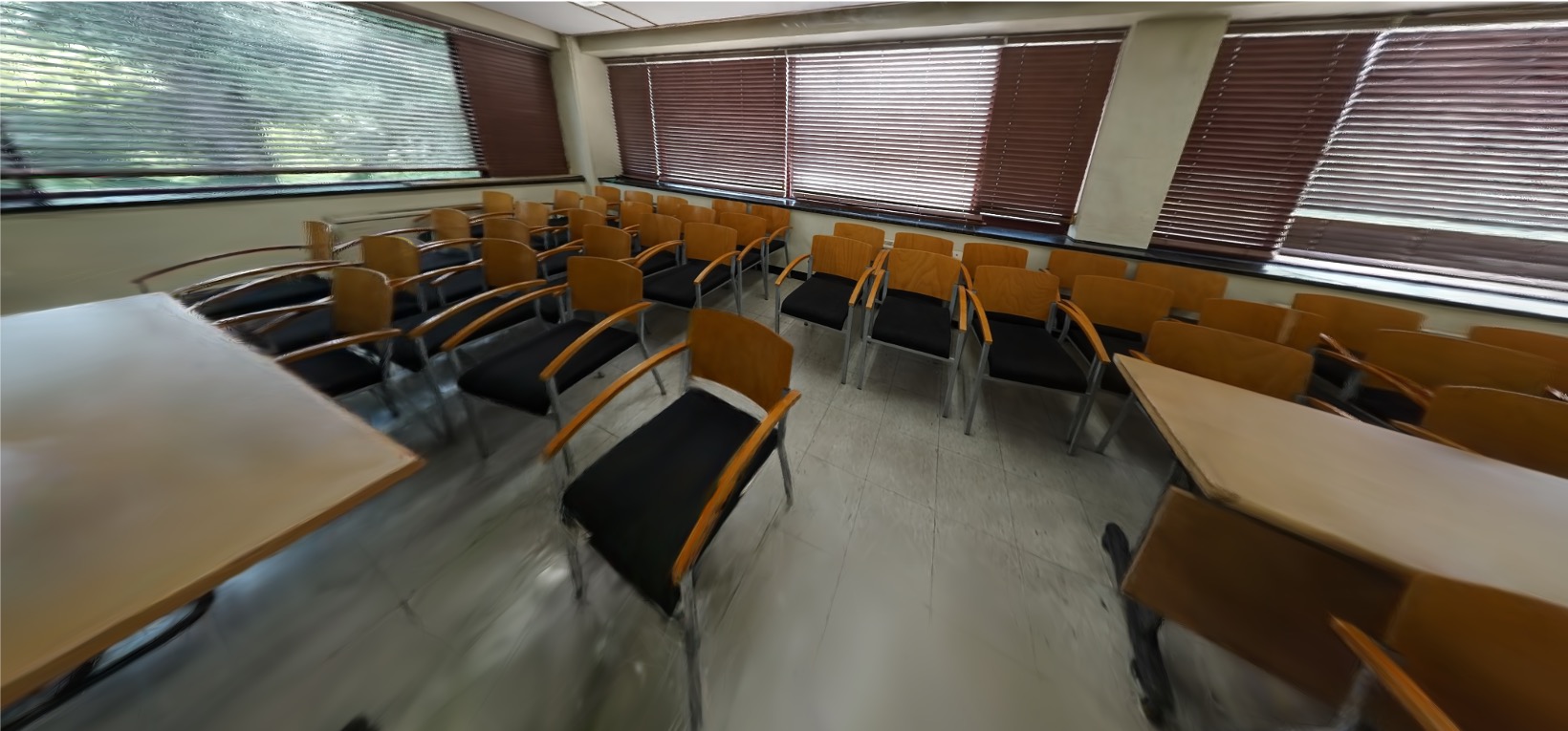}\\
        \multicolumn{2}{c}{(c) Flowers} & \multicolumn{2}{c}{(d) Chairs}
    \end{tabular}

    \caption{\textbf{Fisheye and Perspective Rendering}. After optimization, our method allows rendering in either fisheye or perspective views. Perspective rendering can be achieved by simply removing the hybrid field.}

    \label{fig:ours_undistort}

\end{figure*}
}
We provide additional rendering results in this section. We render both fisheye and perspective views on the FisheyeNeRF dataset~\cite{jeong2021self}. In~\cref{fig:ours_undistort}, we fix the view direction and camera location for fisheye rendering and extend the FOV for perspective rendering. In scenes such as Cube, Chairs, and Flowers, we observe that straight lines are accurately recovered during reconstruction. The lines on the wall behind the Cube and the window frames serve as strong evidence that our self-calibration system precisely models lens distortion.

\section{Implementation Details}
Our implementation is based on the codebase from Gaussian Splatting~\cite{kerbl20233d} and gsplat~\cite{ye2024gsplatopensourcelibrarygaussian}. We use the same loss function as 3DGS for training~\cite{kerbl20233d}. 
The invertible ResNet is constructed using FrEIA~\cite{freia}. 
We follow \citet{kerbl20233d} to select hyperparameters for optimizing 3D Gaussians. 
We also adopt the implementation of MCMC densification~\cite{kheradmand20243d}. 
Compared with vanilla densification, MCMC helps remove floaters by using opacity thresholding to relocate dead Gaussians. 
While the final quantitative results on the test set remain largely unchanged, applying the MCMC technique reduces visual floaters in novel viewpoints. 
For high-resolution scene captures such as Backyard and Office, we also use bilateral grids and anti-aliasing~\cite{yu2024mip} for improved quality.

As explained in~\cref{fig:transposed_opti_iresnet}, optimizing our hybrid field is essential for successful self-calibration. 
We use Adam~\cite{kingma2014adam} to train the invertible ResNet. 
The initial learning rate for the invertible ResNet is set to 1e-5 and gradually decreases to 1e-7 for FisheyeNeRF~\cite{jeong2021self}. 
The final learning rate for real-world captures is 1e-8, including Studio, Garden, and Backyard in \Cref{fig:supp_garden}, as well as more complex real-world captures such as Office~\cref{fig:supp_paul_office}. 
The learning rate for Mitsuba indoor synthetic scenes is set to 1e-8, while for object-centric scenes, it is 1e-7.
% After estimating distortion parameters from COLMAP, we uniformly sample points following the estimated distortion parameters to initialize the invertible ResNet, which typically takes approximately 1 minute to complete. 
All experiments are conducted on a single NVIDIA GeForce RTX 3090.

\section{Failure Cases and Limitations}
Real-world outdoor captures often include the sky.
Reconstructing the sky poses challenges due to moving clouds and the large uniform regions of blue and white without textures. 
The 3DGS~\cite{kerbl20233d} method tends to assign large Gaussians to the sky, resulting in artifacts when rendering novel views. 
Occasionally, some large Gaussians leak into the scene's center, appearing as a thin film in front of the camera. 
Similarly, for indoor scenes, regions with uniform textures, such as colored walls, present challenges. 
These textureless walls are often represented by Gaussians with large covariance matrices, causing similar rendering artifacts as observed with the sky.

The Gaussian sorting we propose alleviates the intensity discontinuities at the boundaries of cubemap faces caused by the multiple projections of a single Gaussian. 
However, since the projection of 3D covariance follows the equation in Sec.~3.1 of the main paper, identical 3D Gaussians can still result in different 2D covariances on different faces. 
This issue can be addressed by implementing smoother transitions between projection faces, such as spherical projection.

Finally, we do not account for the entrance pupil shift phenomenon commonly observed in fisheye lenses.
% , where rays near the center of the field of view converge at a deeper point within the lens than rays at the periphery. 
This effect is distinct from the lens distortion we are currently modeling. 
As a result, our method still struggles with such cameras, as shown in \Cref{fig:supp_paul_office}. 
While entrance pupil shift is negligible for distant scenes, it can cause splat misalignments in near-field scenes (\textit{e.g.,} the blurry sphere surface in the Office scene shown in the video), as the shift can reach up to half a centimeter for full-frame lenses.
It remains an exciting direction to study how to model such lens effects to further improve reconstruction quality.

\end{document}